**A Major Project Report**

**On**

# DESIGN AND FABRICATION OF SOLAR POWERED REMOTE CONTROLLED ALL TERRAIN SPRAYER AND MOWER ROBOT

Submitted in partial fulfillment of the requirements for the degree of

**Bachelor of Technology**

In

**MECHANICAL ENGINEERING**

Submitted by

**A.SRI TARUN (17261A0361)**
**K.E. SHARAN KUMAR (17261A0380)**
**N. MANICHANDRA (17261A0395)**
**T.S. SRIKANTH (17261A03A6)**

**Under the guidance of**
**Mrs. K UDAYANI**
**Assistant Professor**
**Department of Mechanical Engineering**

**DEPARTMENT OF MECHANICAL ENGINEERING**

**MAHATMA GANDHI INSTITUTE OF TECHNOLOGY**

**(Affiliated to JNTU, Hyderabad; Accredited by NBA, AICTE-New Delhi)**
**Kokapet (Vill), Rajendra Nagar (Mandal), Ranga Reddy (Dist.)**
**Chaitanya Bharathi P.O., Gandipet, Hyderabad-500075.**

**June-2021**

MAHATMA GANDHI INSTITUTE OF TECHNOLOGY
(Estd. in 1997 by Chaitanya Bharathi Educational Society)
(Permanent Affiliation to JNTUH; Six UG Programs Accredited by NBA in 2009 & 2014)
Kokapet, Rajendra Nagar (Mandal), Ranga Reddy (Dist.)
Chaitanya Bharathi P.O., Hyderabad - 500 075, TELANGANA, INDIA.

# MAHATMA GANDHI INSTITUTE OF TECHNOLOGY

(Accredited by NAAC with _A‘ Grade & Accredited by NBA, New Delhi,
Affiliated to Jawaharlal Nehru Technological University Hyderabad)
Gandipet, Hyderabad – 500075, Telangana.
Website: www.mgit.ac.in

## CERTIFICATE

This is to certify that the Project Report entitled **"DESIGN AND FABRICATION OF SOLAR POWERED REMOTE CONTROLLED ALL TERRAIN SPRAYER AND MOWER ROBOT"** is a record of bonafide work carried out by the following students during the academic year 2020-2021 in partial fulfillment of requirement of the award of Degree of Bachelor of Technology in Mechanical Engineering.

**A. SRI TARUN (17261A0361)**

**K.E. SHARAN KUMAR (17261A0380)**

**N. MANI CHANDRA (17261A0395)**

**T.S. SRIKANTH (17261A03A6)**

**Guide**
**Mrs. K Udayani**
Asst. Professor

**Dr. K. Sudhakar Reddy**
Head of the Department

**Internal Examiner**

**External Examiner**

# ACKNOWLEDGEMENT

The completion of the project brings with it a sense of satisfaction, but it is never complete without thanking those people who made it possible and whose constant support has crowned our efforts with Success.

With deep sense of gratitude, we acknowledge the esteemed guidance, help and active co-operation rendered by our guide **Mrs. K UDAYANI**, Asst. Professor of Mechanical Engineering. Her inspiring guidance has sustained the effort that has led to successful completion of this project.

We are also thankful to **Dr. K. SUDHAKAR REDDY**, Head of the Department of Mechanical Engineering for his valuable guidance and support.

Our sincere thanks to **Prof. K. JAYA SHANKAR**, Principal, Mahatma Gandhi Institute of Technology for the encouragement and support.

Finally, yet importantly, we would like to express heartfelt thanks to the entire teaching and non-teaching staff of the Department of Mechanical Engineering for their continuous support during our coursework and project.

**A.SRI TARUN (17261A0361)**

**K.E. SHARAN KUMAR (17261A0380)**

**N. MANI CHANDRA (17261A0395)**

**T.S. SRIKANTH (17261A03A6)**

# ABSTRACT

Manual spraying of pesticides and herbicides to crops and weed inhibitors onto the field are quite laborious work to humans. Manual trimming of selected unwanted plants or harvested crops from the field is also difficult.

Our project proposes a multipurpose solar powered, flexible, Remote Controlled, semi-automated spraying robot with 4 Degrees of Freedom (DoF) in spatial movement, with an additional plant mowing equipment.

The robot is designed to spray pesticide/insecticide directly onto individual lesions minimizing wastage or excess chemical spraying, hence making the system cost effective and also environment friendly. It is designed to cut down undesired plants selectively by remotely controlling the start and stop of the mowing system.

Alternatively, it also serves the purpose of maintaining lawns and sports field made of grass. The same system can be used for water spraying and mowing the grass to desired levels, leading to proper maintenance of the field.

The robot is designed to move at 1.4m/s, with an effective spraying area of 0.98 sq. m. by the nozzle and an effective cutting area of 0.3 sq. m. by the mower, when stationary. The prototype has a battery back-up of 7.2hrs under minimum load conditions.

# MOTIVATION

- Use of chemical sprays is of major concern as it is hazardous to both human health and the environment. In general, most of the agriculturists use human operated knapsack sprayers that require more time to cover a given area. Motorized sprayers can cover much larger area in a stipulated time. During this process, though the farmers take lot of precautionary measures like wearing gloves, masks and outfits, sprayed chemicals will adversely affect their health. Therefore, use of autonomous robots provide a safe environment for farming, along with increase in efficient production of agri-products due to increased level of monitoring and control of agricultural fields.

- Sports fields for professional sports need to be specifically maintained to allow convenient and non-obstructive gameplay. The grass on the playing field need to be perfectly trimmed and sprayed regularly but use of sprinklers in the middle of the field can be obstructive. Manual lawn mowing can be time consuming repetitive and stressful, so automation of the process can mitigate all these constraints.

# CONTENTS

# LIST OF FIGURES

# LIST OF TABLES

# 1. INTRODUCTION

India is agrarian economies and most of rural populations depend on agriculture to earn their livelihood. The farming methods at present are manual or semi-automatic with high involvement of labourers. In the recent years, the number of labour availability is reducing continuously along with increase in their wages. There is a requirement of higher productivity. Hence the device is to be designed which helps farmers to overcome the stated problem. Automated Robots can provide us the solution.

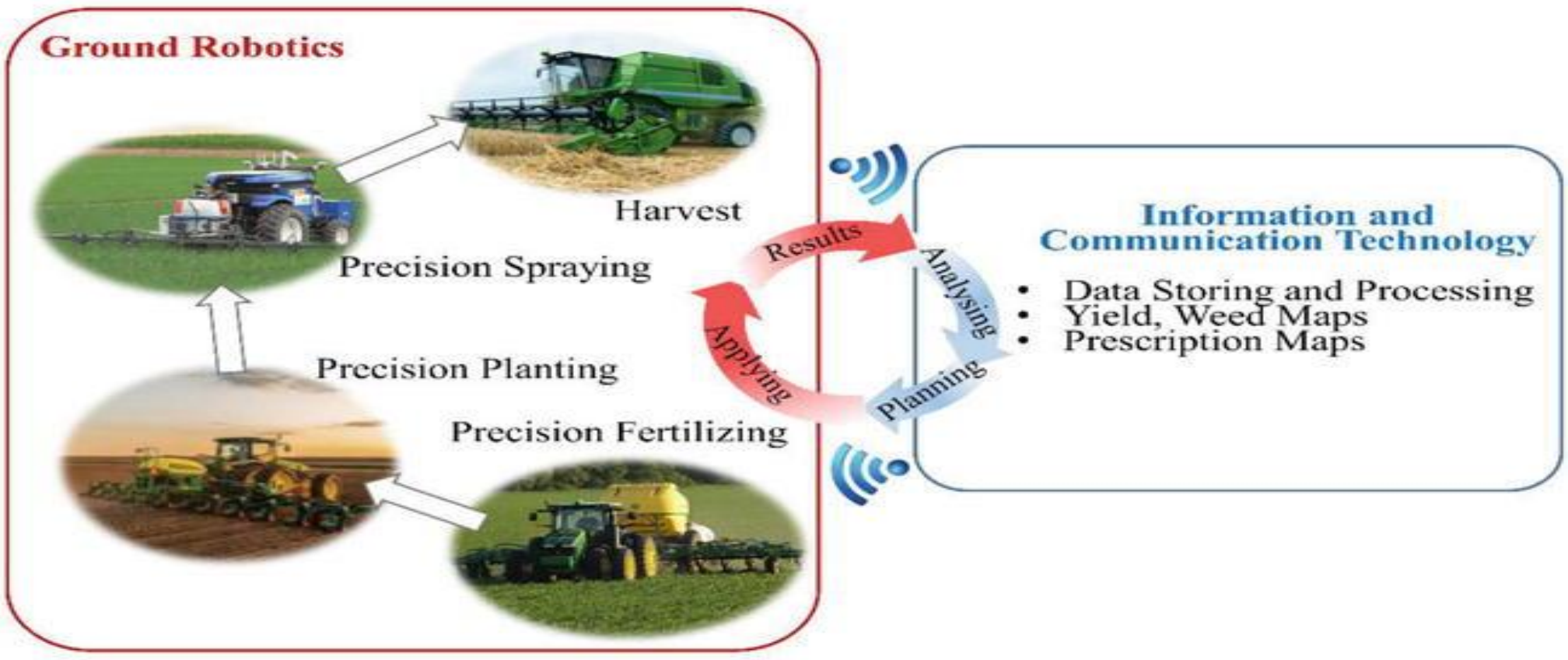

*Figure 1.1: Automation in agriculture*

The main application of robots in the commercial sector has been concerned with the substitution of manual human labour by robots or mechanized systems to make the work more time efficient, accurate, uniform and less costly. One may argue the social implications of such developments, for example, the effects on employment through loss of blue-collar jobs to the more efficient robotic counterpart; there are also ethical considerations that may be argued. Whilst there may well be some validity to the argument in some cases, this current project is unique in the number of stakeholders that are affected in a positive sense. The farmers benefits are found in more efficient maintenance of the crops and either less work for themselves or a decreased need for the employment of others (arguably, an expensive process). Increased demand on growers has begun to be met with increased specific automation in many fields, as producers believe that automation is a viable and sometimes necessary method to ensure maximum profits with minimum costs. Indeed, Hopkins argues that automation enables the expansion of a farm without having to invest more financial resources on labour. Merchants may benefit from increased sales due to a lower cost product; the consumers will benefit, likewise, from a lower cost product of comparable quality. The stakeholders that benefit most, at least from an ethical or social perspective, however, are the farm workers. This project presents the design and construction of an autonomous robot that seeks to address some of the human health concerns associated with farms. This robot is designed as a base for developing systems to enable the automation of farming processes such as the spraying of pesticides, picking of fruit and the caring for diseased plants. The system is designed to be as modular as possible, enabling the development and/or modification of any of the individual tasks.

### *1.1 Pesticide spraying*

The pesticides have a vital influence of the agribusiness. Nearly 35% of crops have been safeguarded from the insects using pesticides. The pesticides are needed for agriculture field to increase the efficiency but they are also injurious to human and also to the environment. In the current methods, the farmers use the backpack sprayer which is manually operated by the human along the crop fields. They used to spray the pesticides in the targeted way manually. Here the sprayer is connected to the back of the tractor and this tractor was driven by the human. The pesticides were sprayed to the crops along the field. This method does not uses the selective spraying and the pesticides are spread to the field.

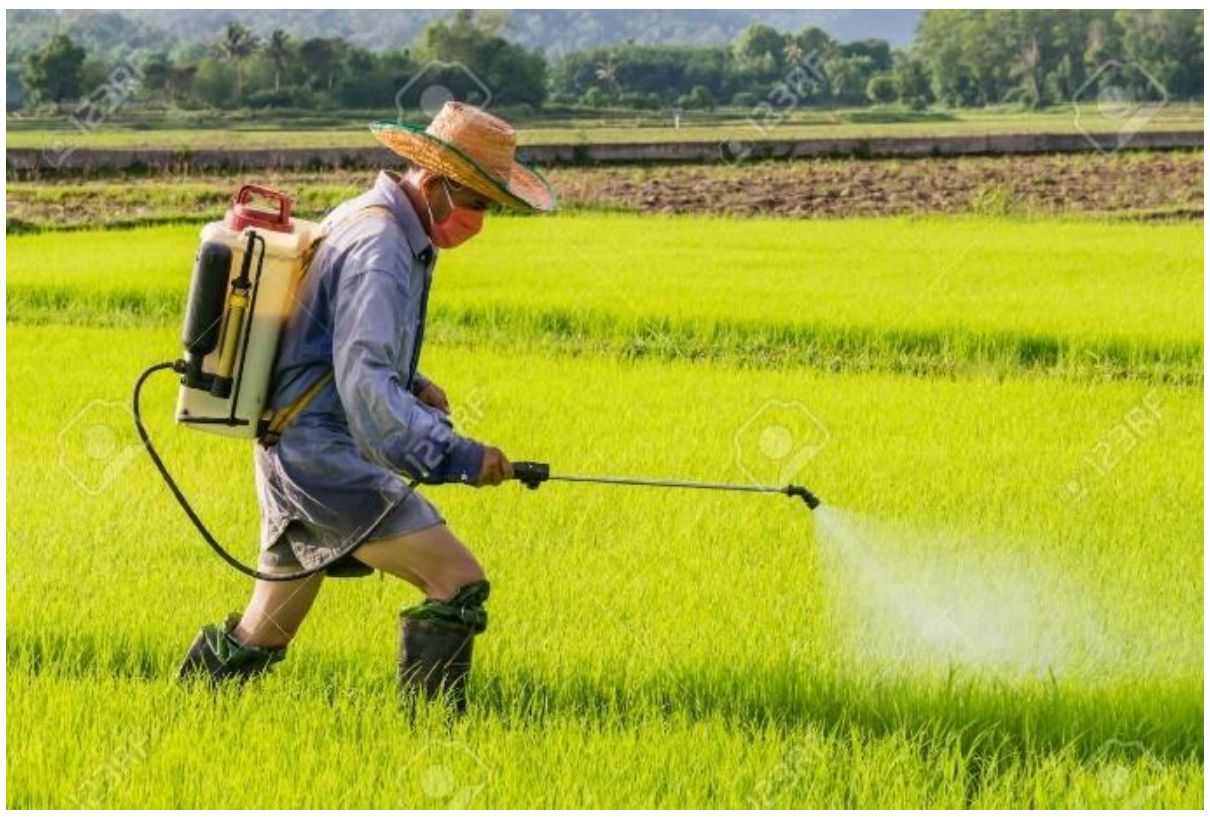

*Figure 1.2: Manual Pesticide spraying*

In spite of the utilization of pesticide assurance gear (individual head veil and focal filtration framework for the manual and automated spraying strategies, separately) the human is as yet presented to unsafe pesticides that can cause negative medical problems.

Other than wellbeing concerns, automated and manual spraying strategies have different downsides. The motorized spraying isn't target explicit and is intended to splash a harvest strip with rearranged stature (e.g., for spraying only the grape bunches the rancher will show the shower spouts to shower a strip 0.5 m wide with no thought of the natural product area). Moreover, manual spraying is repetitive work, moderate, and restricted because of the absence of laborers horticulture.

### *1.2 Significance of solar energy*

Every day, the sun radiates an enormous amount of energy called solar energy. It radiates more energy in one day than the world uses in one year. This energy comes from within the sun itself. Like most stars, the sun is a big gas ball made up mostly of hydrogen and helium gas. The sun makes energy in its inner core in a process called nuclear fusion. It takes the sun's energy just a little over eight minutes to travel the 93 million miles to Earth. Solar energy travels at the speed of light, or 186,000 miles per second, or 3.0 x 10^8 meters per second. Only a small part of the visible radiant energy (light) that the sun emits into space ever reaches the Earth, but that is more than enough to supply all our energy needs. According to the national air space association (NASA), there is a 1.361 $kW/m^2$ of solar irradiance received at the top of Earth's atmosphere.

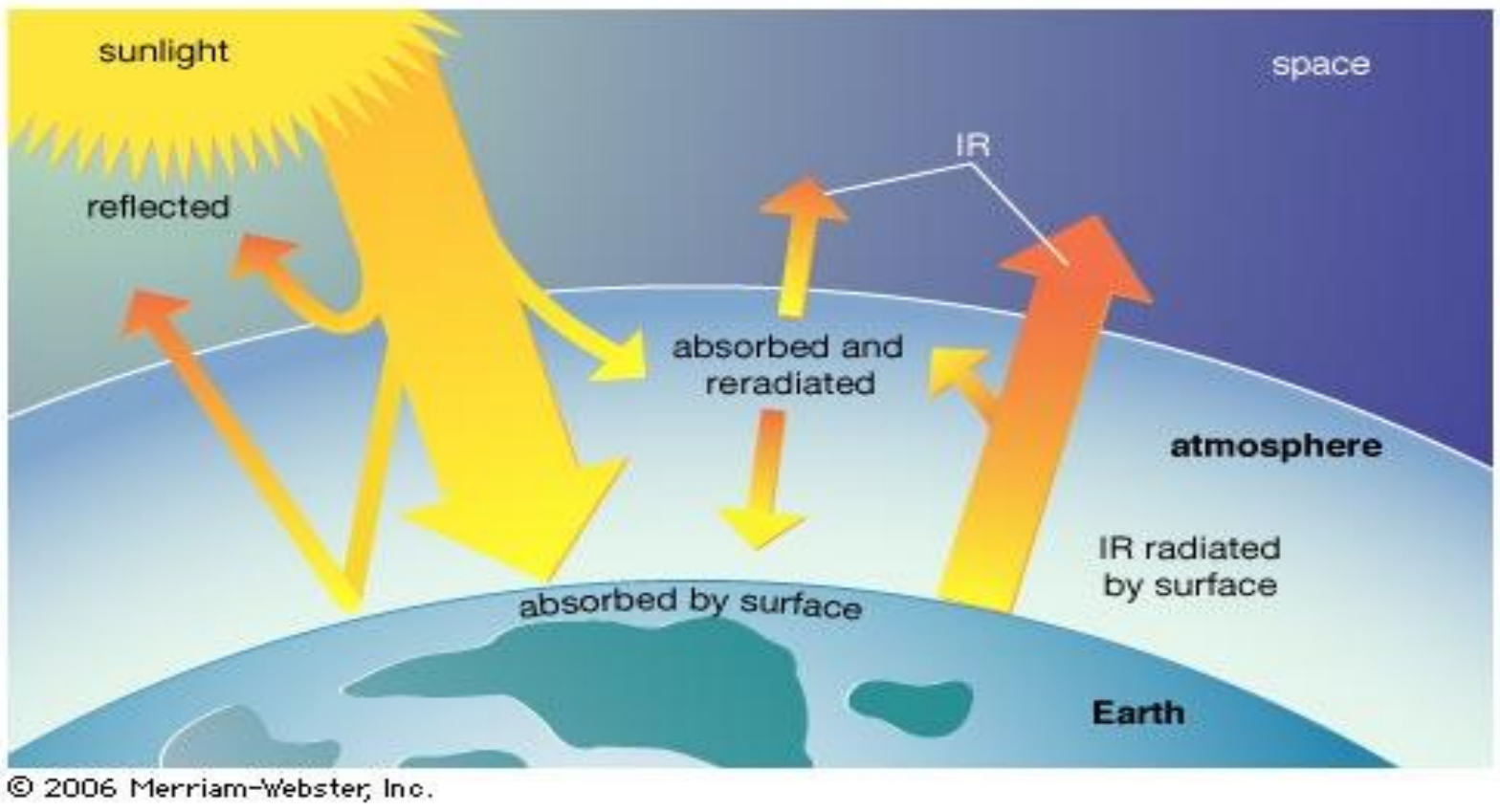

*Figure 1.3: Solar energy incident on earth*

Approximately 1.8/10MW amount of power from the sun has been interrupted by the planet Earth, which is thousands of times larger than the present global consumption rate of the energy. This has motivated the governments, researchers and power industries to increase their investments in the renewable energy industry aiming to utilize more this clean energy and relief the global warming. Every hour enough solar energy reaches the Earth to supply our nation's energy needs for a year solar energy is considered a renewable energy source due to this fact. Today, people use solar energy to heat buildings and water and to generate electricity. Solar energy accounts for a very small percentage of U.S. energy less than one percent. Solar energy is mostly used by residences and to generate electricity.

### *1.3 Mower*

The conventional grass cutters have been widely used recently by workers in the gardening and agricultural industries. However, the manual handled grass cutters are consuming a lot of energy and producing air pollution which can directly affect the workers' health. The conventional grass cutters are also creating a high level of noise and vibration which can cause serious health issues such as grip strength, decreased hand sensation and dexterity, finger blanching or white fingers and carpal tunnel.

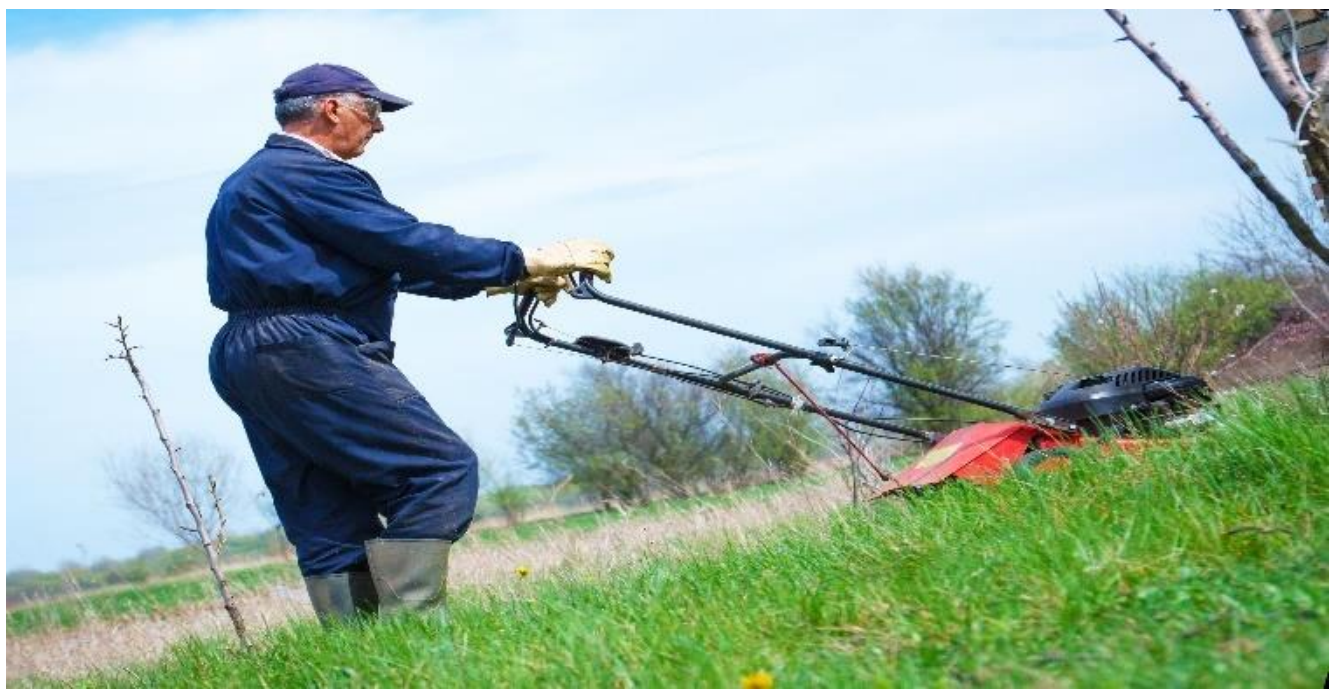

*Figure 1.4: Manual Mowing*

Hence, by evaluating the above 3 parameters we had found a way to integrate these 3 functions in a single robot, which would increase the productivity and reduce idle time for farmers. This motivated us to design and fabricate a model that utilizes solar energy for spraying pesticides and Mowing. Semi-automatic solar pesticide sprayer and Mower which consists of solar panel, a battery, motor, pump, container, cutting blades and microcontroller is a 4 wheeled vehicle which is operated by a wireless remote which runs on power source as a DC battery. So, with this background, design and construction of solar powered sprayer and Mower system was made. The control of the vehicle is achieved using an inbuilt microcontroller unit which is programmed to respond to a wireless device via Bluetooth.

# 2. LITERATURE SURVEY

## *2.1 Pesticide spraying*

With flourishing technology that is introduced in this 21st century, there is numerous types of robots been used in agricultural activity starting from the cultivation process to the production process. The autonomous robot had been introduced in various application such is in underwater, rescue, line following robot based on metal detection. In agriculture field, the usage of robotics in agriculture operation able to help to increase the production process and improve efficiency. One of the types of the robot used in agriculture is for the purpose of pesticide spraying with the ability to navigate in the farm, recognize the target and regulate the spraying mechanism.

The use of autonomous robot pesticide sprayer as the substitution of the worker who used conventional pesticide sprayer can be applicable. Besides, the demand for the agriculture robot also stimulates the consciousness of how important its role in the current and future generations. The survey conducted shows that the demand for robots and drones in agriculture will be expected to be rose from 2018 to 2038. Hence, the usage of the autonomous robot is assumed to rise thus replacing the current labor worker. This granular20 years market forecast covers all the aspect of the agricultural robots and drones for 16 market categories with the expectation by the end of 2038, market of the robots and drones in these categories is predicted will close to 35 billion with the viable technology and ongoing market demand by considering its technology and application.

Nevertheless, the common problem with an autonomous robot use in agricultural activity is the navigation method used to able the robot fully-operated with decision making capability. In order to navigate through all the field, there are some research has been done. It can be done through infrastructure ready or to be without infrastructure. Some research on RFID based navigation are conducted to be implemented as navigation tools.

As artificial intelligence (AI) starts to emerge, the current robot should be able to navigate the next movement by the adaptation of the surrounding environment and decide which path it will take. The typical method used in the detection is based on the targeted object orientation or repelled signal emits from the sensor itself then calculates the distance in between it. Other than that, there is also the robot that uses the vision observation then accumulates all the acquired data to generate the data fusion that enables the robot to navigate itself through the farm.

The second problem with the agricultural robot is due to the dissemination of the pesticide to the crops. Unregulated spraying during the disposition of the pesticide to the crop can lead to the low rate of coverage on leaves, wastage of pesticide and hazardous exposure to workers due to disperse pesticide to the desired target.

With regulated spraying by the pump, the higher coverage of dissemination to the crops can be achieved whereby the positioning of each crop was varied from one another in the farm. Furthermore, instead of hiring the workers to do miscellaneous work on the farm which can affect themselves, it can be done by an autonomous agriculture robot thus save the expenses on the labor worker.

Lastly, the designed robot used in agriculture having the difference performance index depends on the variable they want to achieve. Certain researchers may focus on UAV based pesticide spraying, localization and motion control of agriculture mobile robot, pest image identification and else. This also same goes to the type of the plant being used as the target which differs from one another in terms of size, leaves density and height.

Hence, it would be difficult to decide which designed robot was most successful at the time being. In this research, the aim of this study is to design and develop an autonomous pesticide sprayer for the chili fertigation system. Then, this study intends to implement a flexible sprayer arm to spray the pesticide under the crop's leaves, respectively. This study involves the development of unmanned pesticide sprayer that can be mobilized autonomously. It is because the pesticide is a hazardous component that can be affected human health in the future if it exposed during manual spraying method especially in a closed area such as in the greenhouse.

The flexible sprayer boom also can be flexibly controlled in the greenhouse and outdoor environment such as open space farms. It is expected to have a successful pesticide management system in the fertigation-based farm by using the autonomous pesticide sprayer robot.

Besides, the proposed autonomous pesticide sprayer also can be used for various types of crops such as rock melon, tomato, papaya. pineapples, vegetables and etc.

The development of the autonomous pesticide sprayer prototype consists of two parts where the navigation system and the spraying system. The interconnection between the selected components in the designed robot is crucial and plays a major role to make sure the robot function as desired.

Misconnection between the electronic components can lead to malfunction of the designed system thus deviated the operation from achieving the project objective.

### *2.2 Solar cell*

A solar cell or photovoltaic cell is a wide area electronic device that converts solar energy into electricity by the photovoltaic effect. Photovoltaic is the field of technology and research related to the application of solar cells as solar energy. Sometimes the term solar cell is reserved for devices intended specifically to capture energy from sunlight, while the term photovoltaic cell is used when the source is unspecified. Assemblies of cells are used to make solar modules, or photovoltaic arrays.

*2.2.1 First generation:* First generation cells consist of large-area, high quality and single junction devices. First generation technologies involve high energy and labor inputs which prevent any significant progress in reducing production costs. Single junction silicon devices are approaching the theoretical limiting efficiency of 33% and achieve cost parity with fossil fuel energy generation after a payback period of 5-7 years.

*2.2.2 Second generation:* Second generation materials have been developed to address energy requirements and production costs of solar cells. Alternative manufacturing techniques such as vapor deposition and electroplating are advantageous as they reduce high temperature processing significantly. It is commonly accepted that as manufacturing techniques evolve production costs will be dominated by constituent material requirements, weather this be a silicon substrate, or glass cover. Such processes can bring costs down to a little under but because of the defects inherent in the lower quality processing methods, have much reduced efficiencies compared to First Generation. The most successful secondgeneration materials have been cadmium telluride (CdTe), copper indium gallium solenoid, as glass or ceramics reducing material mass and therefore costs. These technologies do hold promise of higher conversion efficiencies, particularly CIGS-CIS, DSC and CdTe offers significantly cheaper production costs. In CdTe production represented 4.7% of total market share, thin-film silicon 5.2% and CIGS 0.5%.

*2.2.3 Third generation:* Third generation technologies aim to enhance poor electrical performance of second generation (thin-film technologies) while maintaining very low production costs. Current research is targeting conversion efficiencies of 30-60% while retaining low-cost materials and manufacturing techniques. They can exceed the theoretical solar conversion efficiency limit for a single energy threshold material; which was calculated in 1961 by Shockley and Queasier as 31% under 1 sun illumination and 40.8% under maximal concentration of sunlight (46,200 suns, which makes the latter limit more difficult to approach than the former).

### *2.3 Mower*

India is the second largest producer of wheat and rice, the world's major food staples. India ranked within the world's five largest producers of over 80% of agricultural produce items, including many cash crops such as coffee and cotton, in 2010. India is also one of the world's five largest producers of livestock and poultry meat, with one of the fastest growth rates, as of 2011 and this process of harvesting is conventional that consumes mote efforts and cost of crop cutting. The lawn mower is an aid in the mundane task of grass cutting and tending to lawns. Due to the revolution of green movement in the present scenario the industries with major campus areas are changing the percentage of greenery in the campuses and increased greenery causes increased effort and money to tend to.

Due to increased availability of system on chips, the lawn mower can be automated very easily and also the reduced size and cost of Dc motors causes the system to be independent of fossil fuels to be able to tap into renewable energies. The presence of Ultrasonic sensors and light dependent resistors in a smaller and cheaper packaging cause the bot to be more aware of its surroundings. Due to the presence of Arduino in the system causes and increase in the module that can be added. Traditional design of lawn mowers had motored powered engines which required regular maintenance such as engine oil and greasing.

They also created a lot of noise pollution and air pollution. In the cold and harsh environment, the fossil fuel powered motors tend to freeze and not run. These problems are solved by using electric motors. They are also much greener because they use solar panel. The mower uses battery chorded system causes a range as alimentation and damage to the chords. There are several types of mowers, each suited to a particular scale and purpose. The smallest types, non-powered push mowers, are suitable for small residential lawns and gardens. Electrical or piston engine-powered push-mowers are used for larger residential lawns (although there is some overlap). Riding mowers, which sometimes resemble small tractors, are larger than push mowers and are suitable for large lawns, although commercial riding lawn mowers (such as zero-turn mowers) can be "stand-on" types, and often bear little resemblance to residential lawn tractors, being designed to mow large areas at high speed in the shortest time possible. The largest multi-gang (multi-blade) mowers are mounted on tractors and are designed for large expanses of grass such as golf courses and municipal parks, although they are ill-suited for complex terrain.

Dipin and Chandrasekhar explained that the Solar Powered Vision Based Robotic Lawn Mower is an autonomous lawn mower that will allow the user to the ability to cut their grass with minimal effort. Unlike other robotic lawn mowers on the market, this design requires no perimeter wires to maintain the robot within the lawn and also with less human effort in the manual mode operation. They studied that there is some preset pattern installed in the robot, in the automatic mode operation no human effort needed for the operation and helps to cut different patterns in the lawn very easily with less time. Through an array of sensors safety takes major consideration in the device, this robot will not only stay on the lawn, it will avoid and detect objects and humans. And also, it detects the land boundaries and start mowing upon the predefine pattern with the help of installed camera and MATLAB programming.

## *2.4 Literature review*

| S. No. | Author name | Research Paper Title | Conclusion | Publish date & Year | Gaps in Literature |
|---|---|---|---|---|---|
| 1. | Julian Se nchez-Hermosilla, Francisco Rodriguez Ramon Gonzalez, Jose Luis Guzman2and Manuel Berenguel | A mechatronic description of an autonomous mobile robot for agricultural tasks in greenhouses | The mechanical design of the mobile robot has been carried out using CAD/CAE technologies in which the main features of greenhouses, the electronic components have been considered | 1 march 2010 | Gives only a description of processes carried out in green houses. |
| | Tony E. Grift | Development of Autonomous Robots for Agricultural Applications | The flexibility of the robot was not truly employed, since simple front wheel steering proved sufficient for between-row guidance. The SICK laser unit provided | 5 June 2003 | Only used for scouting of row crop operations. Very bulky and expensive setup. |
| 3. | Vijaykumar N Chalwa, Shilpa S Gundagi | Mechatronics Based Remote Controlled Agricultural Robot | The robot for agricultural purpose an Agrobot is a concept for the near the performance and cost of the product once optimized, will prove to be work through in the agricultural spraying operations. | 7 JULY 2014 | The Sprayer does not have any degrees of freedom. Completely dependent on grid power. The setup height is too low. The cost is high. |
| 4. | Yan Li, Chunlei Xia, Jangmyung Lee | Vision based pest detection and automatic spray of greenhouse plant | In this paper, a depth measurement method for detecting pests on leaves is proposed. And it provides pests position for automatically spraying pesticide on the leaves where the pest model pasted on. | 5 August 2009 | The process of pest detection is time consuming and this setup can only be used in greenhouse plants. |
| 5. | Philip J. Sammons, Tomonari Furukawa and Andrew Bulgin | Autonomous Pesticide Spraying Robot for use in a Greenhouse | The results showed the robot was able to successfully the physical specifications outlined by NCGH so as to be able to function within their greenhouses. The robot was able to drive up and | 9 September 2005 | The robots motion sis only limited to the tracks arranged around the green house. |
| | | | back along the tracks in the greenhouse. | | |
| 6. | Harshit Jain,Nikunj Gangrade, Sumit Paul, Harshal Gangrade, Jishnu Ghosh | Design and fabrication of Solar Pesticide sprayer. | It is observed that, this model of solar powered pesticide sprayer is more cost effective and gives the effective results in spraying operation. As it runs on the non-conventional energy source i.e. solar energy, it is widely available at free of cost. | March - 2018 | The chasis setup is not automated so it has to be moved manually. |
| 7. | Binod Poudel, Ritesh Sapkota, Ravi Bikram Shah, Navaraj Subedi, Anantha Krishna G.L. | “Design and Fabrication of Solar Powered Semi Automatic Pesticide Sprayer | This project demonstrates the implementation of robotics and mechatronics in the field of agriculture. This being a test model the robustness of the vehicle is not very high. In addition the safety and long term health of the farmers is ensured by eliminating human labor completely from this process. It does not compromise the performance of a petrol based pesticide sprayer. | July -2017 | The system has a two wheel drive which compromises the robustness and the model can only be used for pesticide spraying. |
| 8. | Kiran Kumar B M,M S Indira, S Nagaraja Rao Pranupa S | Design and Development of Three DoF Solar Powered Smart Spraying Agricultural Robot | The design and development of a solar powered, remotely operated three degrees of freedom pesticide spraying robot for use in agriculture is presented in this paper. The prototype gave a fairly good rate of area coverage with a reasonably low operating cost. | March - April 2020 | The actuating mechanisms for providing degrees of freedom are quite slow and unstable. This setup can only be used for pesticide spraying. |

### *2.5 Our approach*

- With the proposed design of the robot in this project, the above-mentioned gaps can be eliminated completely. This project integrates two of the major activities in agriculture which are Pesticide spraying and Crop Cutting (or Weed Removal).

- Robots are used for various agricultural tasks such as harvesting, cultivation of crop, weeding, and spraying pesticides. The features of any agricultural robot are classified as: Guidance, Detection, Action and Mapping. Guidance refers to robo-navigation; Detection assimilates surrounding information; Mapping extracts field features; and Action ensures the completion of the intended task. All four features of an agriculture robot are independent in nature.

- The controller and the navigation enabling motor units are powered by a solar photovoltaic system. Hence our project focusses on green, cost effective in terms of labour and quantity of pesticides and an efficient spraying system safe to both humans and the environment.

## 3. OBJECTIVES

- The major objectives are the attributes which the device must meet. They are:
- To reduce human effort in the agricultural field with the use of small machine.
- To perform all operations at single time, hence increases production and saves time.
- It should perform all operations on command.
- It should be safe and simple to control.
- It should be reliable.
- It should be durable and economical.
- To reduce human effort within the agricultural field with the employment of small robot.
- To perform all the operations at single time, hence increases production and decrease idle time.
- To complete great amount of labour in less time.
- Farmer can control the robot through remote by sitting at one side and operate easily.
- The usage of solar may be utilized for Battery charging. because the Robot works within the field, the rays of the sun may be used for solar energy generation.
- To reduce the dependence on grid power, the solar energy is employed and therefore a battery is placed to store the energy and use it whenever required.

# 4. COMPONENTS

**HARDWARE COMPONENTS REQUIRED**:

1. HC-05 Bluetooth module
2. Arduino Uno
3. L293D driver
4. Solar panel
5. DC motors
6. Linear Actuators
7. Mobile app
8. Battery
9. Manual Switch
10. Ply wood
11. Clamps
12. Nuts and Bolts
13. Wires
14. Wheels
15. Resister
16. Pump
17. Mower Blade
18. Storage tank
19. Pipes
20. Nozzles
21. Personal Computer

### *4.1 Components in detail*

*4.1.1Bluetooth Module HC-05 And Arduino:* HC05 module is pretty easy to use and Bluetooth Serial Port Protocol (SPP) module is fabricated for transparent wireless serial connection setup. The HC-05 Bluetooth module can be used to communicate between two microcontrollers like Arduino or communicate with any device with Bluetooth functionality like a Phone or Laptop and is shown in the Fig.4. To control the entire system, Bluetooth HC05 is connected to Arduino and to android smartphone wirelessly.

pairing the HC- 05 module with microcontrollers is very easy because it works using the SPP.

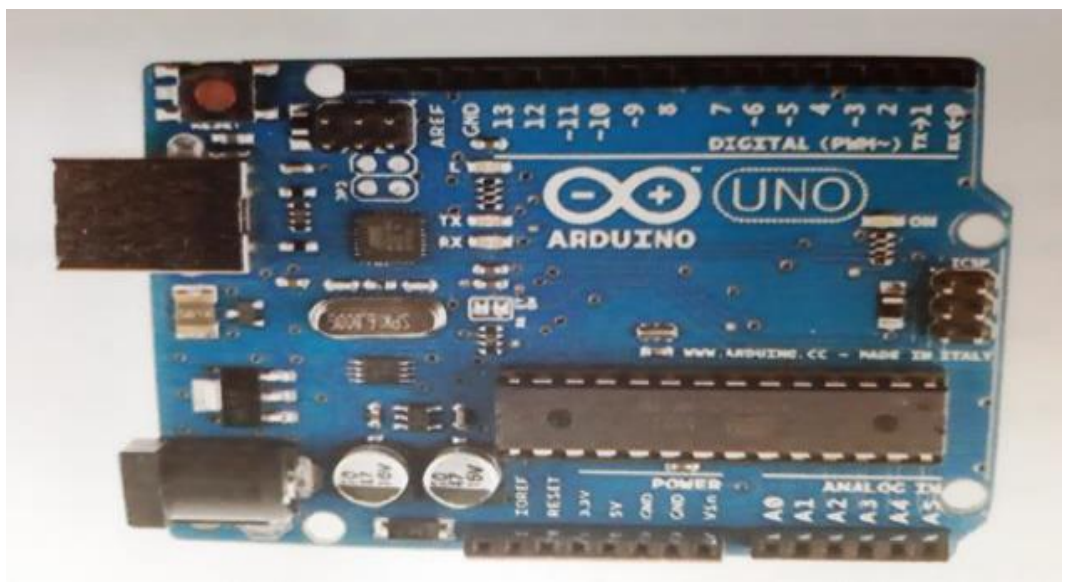

*Figure 4.1: Arduino UNO board*

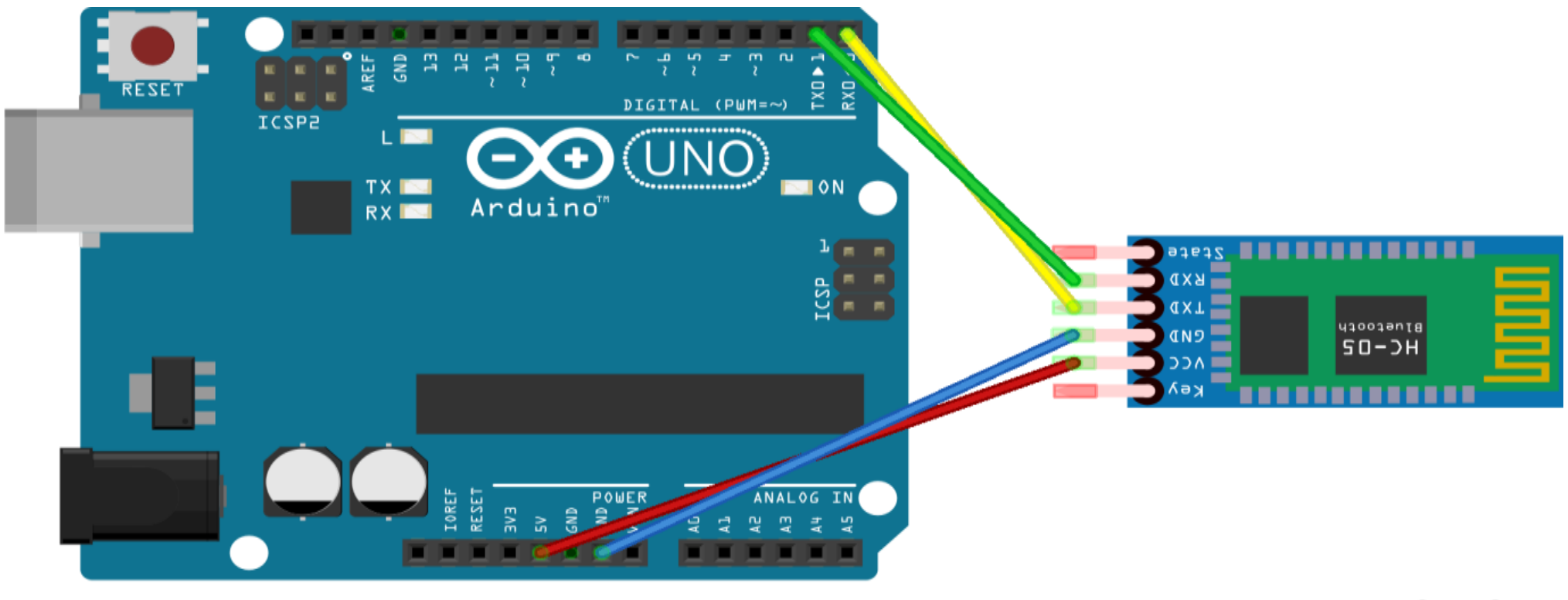

*Figure 4.2: Arduino and Bluetooth connection*

New Features of the board are as follows

Microcontroller: ATmega328

Input Voltage: 5V

Recommended Voltage: 7-12V

Limits of Input Voltage: 6-20V

Digital I/O Pins: 14 of which 6 are for PWM output

Analog Input Pins: 6

DC Current on I/O Pins: 40mA

DC Current on 3.3V Pins: 50mA

Flash Memory: 32KB of which 0.5KB used by boot loader

SRAM: 2KB

EEPROM: 1KB

Clock Speed: 16MHz

**Schematic**: Arduino reference design uses Atmega 8, 168 or 328. Present models use ATmega328. The pin configuration is shown below.

Maximum dimensions of Uno are 2.1x2.7 inches. USB and power jack connectors are extended outside the earlier measurement. The board can be attached to base Four screw holes are provided. Gap between the pins 7 and 8 is 0.16".

*4.1.2 L293D Motor control Shield:* **L293D shield** is a driver board based on **L293** IC, which can drive 4 DC **motors** and 2 stepper or Servo **motors** at the same time. Each channel of this module has the maximum current of 1.2A and doesn't work if the voltage is more than 25v or less than 4.5v.

It can drive:

- 4 bi-directional DC motors with 8-bit speed selection(0-255)
- 2 stepper motors (unipolar or bipolar) with single coil, double coil, interleaved or micro-stepping.
- 2 servo motors

*Power Supply*

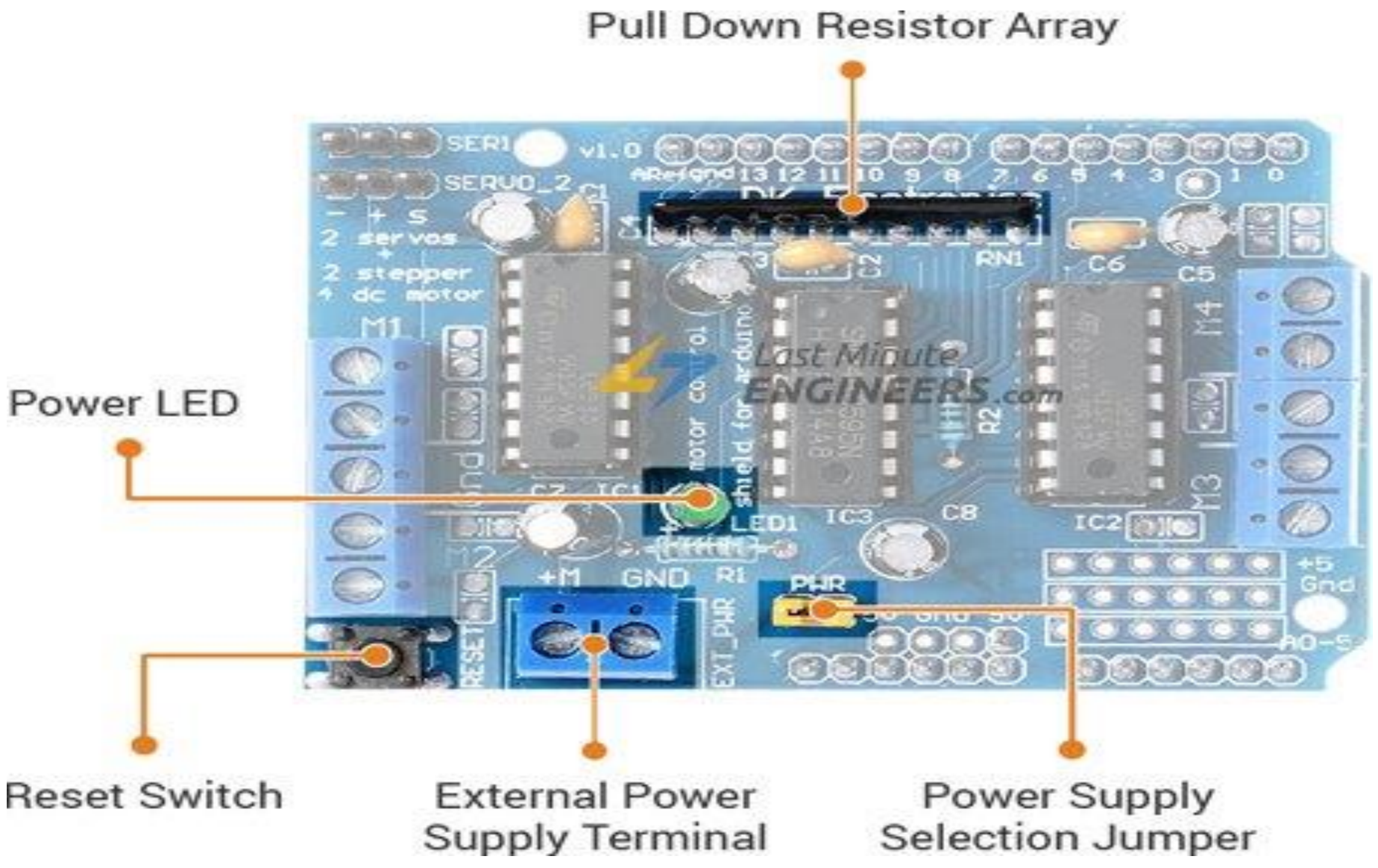

*Figure 4.3: Power supply to Motor shield*

There exist three scenarios when it comes to supplying power for the motors through shield.

- *Single DC power supply for both Arduino and motors:* If you would like to have a single DC power supply for both Arduino and motors, simply plug it into the DC jack on the Arduino or the 2-pin EXT_PWR block on the shield. Place the power jumper on the motor shield. You can employ this method only when motor supply voltage is less than 12V.

- *(Recommended) Arduino powered through USB and motors through a DC power supply:* If you would like to have the Arduino powered off of USB and the motors powered off of a DC power supply, plug in the USB cable. Then connect the motor supply to the EXT_PWR block on the shield. Do not place the jumper on the shield.

- *Two separate DC power supplies for the Arduino and motors:* If you would like to have 2 separate DC power supplies for the Arduino and motors. Plug in the supply for the Arduino into the DC jack, and connect the motor supply to the EXT_PWR block. Make sure the jumper is removed from the motor shield.

*Output Terminals*

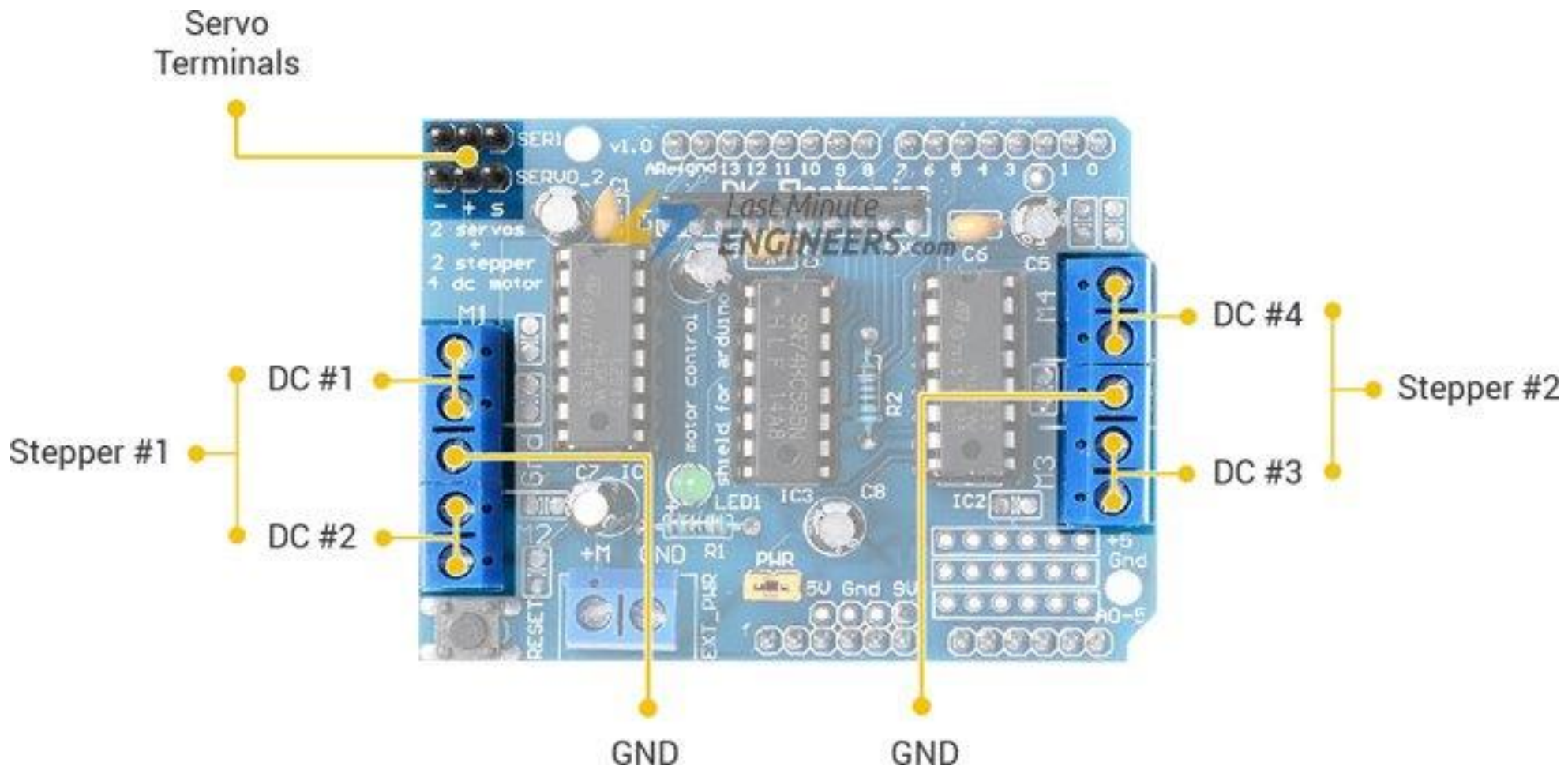

*Figure 4.4: Output terminals*

The output channels of both the L293D chips are broken out to the edge of the shield with two 5-pin screw terminals viz. M1, M2, M3 & M4. You can connect four DC motors having voltages between 4.5 to 25V to these terminals. Each channel on the module can deliver up to 600mA to the DC motor. However, the amount of current supplied to the motor depends on system's power supply.

You can also connect two stepper motors to output terminals. One stepper motor to motor port M1-M2 and other to M3-M4. The GND terminal is also provided if you happen to have a unipolar stepper motor. You can connect the centre taps of both stepper motors to this terminal. The shield brings out the 16bit PWM output lines to two 3-pin headers to which you can connect two servo motors.

*Driving DC Motors with L293D Shield:* Now that we know everything about the shield, we can begin hooking it up to our Arduino!. Start by plugging the shield on the top of the Arduino. Next, connect power supply to the motors. Although you can connect DC motors having voltages between 4.5 to 25V to the shield, in our experiment we are using DC Motors that are rated for 9V. So, we will connect external 9V power supply to the EXT_PWR terminal. Now, connect the motor to either M1, M2, M3 or M4 motor terminals. In our experiment we are connecting it to M4.

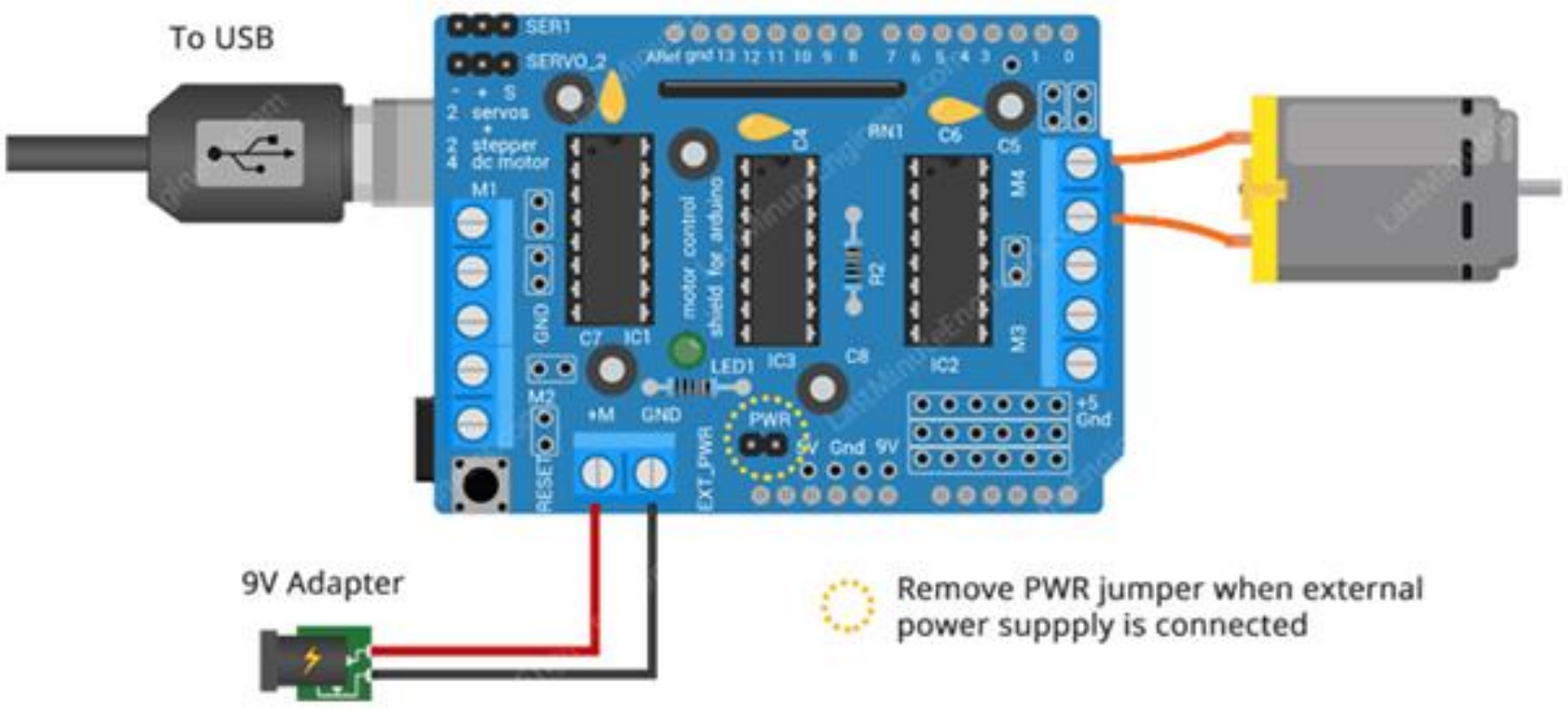

*Figure 4.5: Wiring DC motor*

*Wiring DC Motor to L293D Motor Shield & Arduino:* The following sketch will give you complete understanding on how to control speed and spinning direction of a DC motor with L293D motor driver shield and can serve as the basis for more practical experiments and projects.

*4.1.3 Solar panel:* A solar panel, or photo-voltaic (PV) module, is an assembly of photo-voltaic cells mounted in a framework for installation. Solar panels use sunlight as a source of energy and generate direct current electricity. A collection of PV modules is called a PV panel, and a system of panels is an array. Arrays of a photovoltaic system supply solar electricity to electrical equipment.

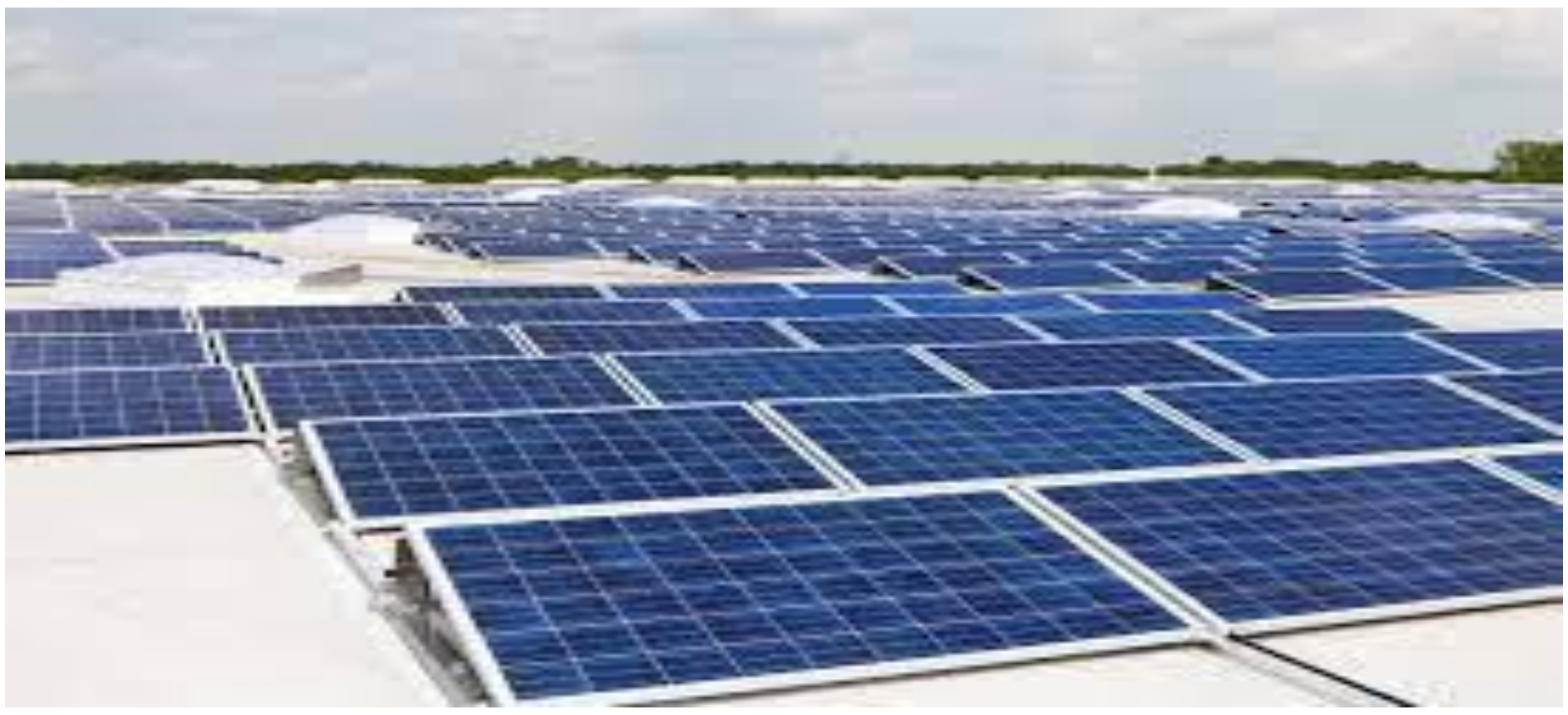

*Fig 4.6. Solar panel*

*4.1.3.1 Theory and constitution:* Photovoltaic modules use light energy (photons) from the Sun to generate electricity through the photovoltaic effect. Most modules use water-based crystalline silicon cells or thin-film cells. The structural (load carrying) member of a module can be either the top layer or the back layer. Cells must be protected from mechanical damage and moisture. Most modules are rigid, but semi-flexible ones based on thin-film cells are also available. The cells are connected electrically in series, one to another to the desired voltage, and then in parallel to increase amperage. The wattage of the module is the mathematical product of the voltage and the amperage of the module. The manufacture specifications on solar panels are obtained under standard condition which is not the real operating condition the solar panels are exposed to on the installation site.

A PV junction box is attached to the back of the solar panel and functions as its output interface. External connections for most photovoltaic modules use MC4 connectors to facilitate easy weatherproof connections to the rest of the system. A USB power interface can also be used.

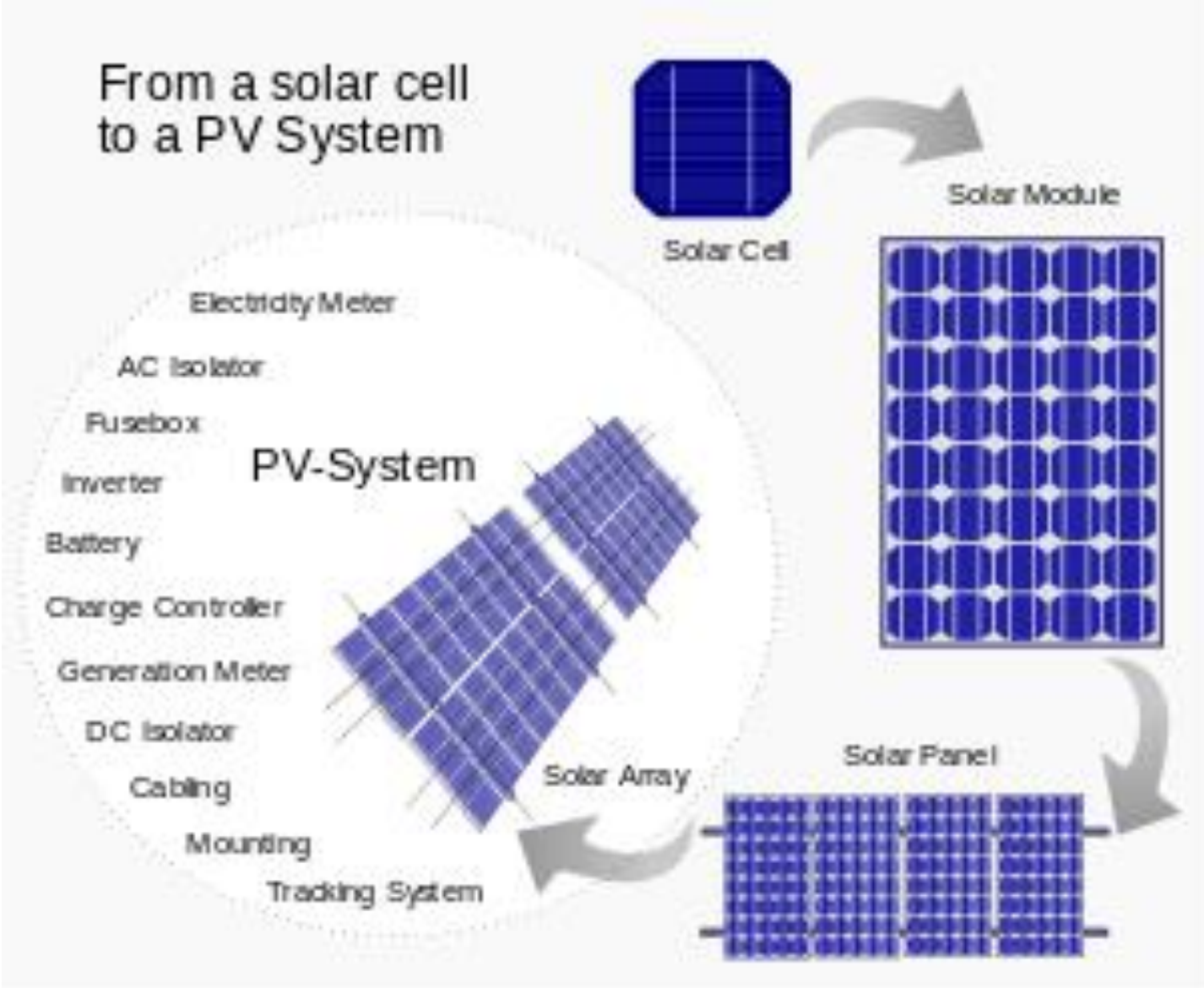

*Fig 4.7. Construction of solar panel*

*4.1.3.2 Solar panel efficiency:* Each module is rated by its DC output power under standard test conditions (STC) and hence the on-field output power might vary. Power typically ranges from 100 to 365 Watts (W). The efficiency of a module determines the area of a module given the same rated output – an 8% efficient 230 W module will have twice the area of a 16% efficient 230 W module. Some commercially available solar modules exceed 24% efficiency. Currently, the best achieved sunlight conversion rate (solar module efficiency) is around 21.5% in new commercial products typically lower than the efficiencies of their cells in isolation. The most efficient mass-produced solar modules have power density values of up to 175 W/m$^2$ (16.22 W/ft$^2$).

Scientists from Spectro lab, a subsidiary of Boeing, have reported development of multi-junction solar cells with an efficiency of more than 40%, a new world record for solar photovoltaic cells. The Spectro lab scientists also predict that concentrator solar cells could achieve efficiencies of more than 45% or even 50% in the future, with theoretical efficiencies being about 58% in cells with more than three junctions.

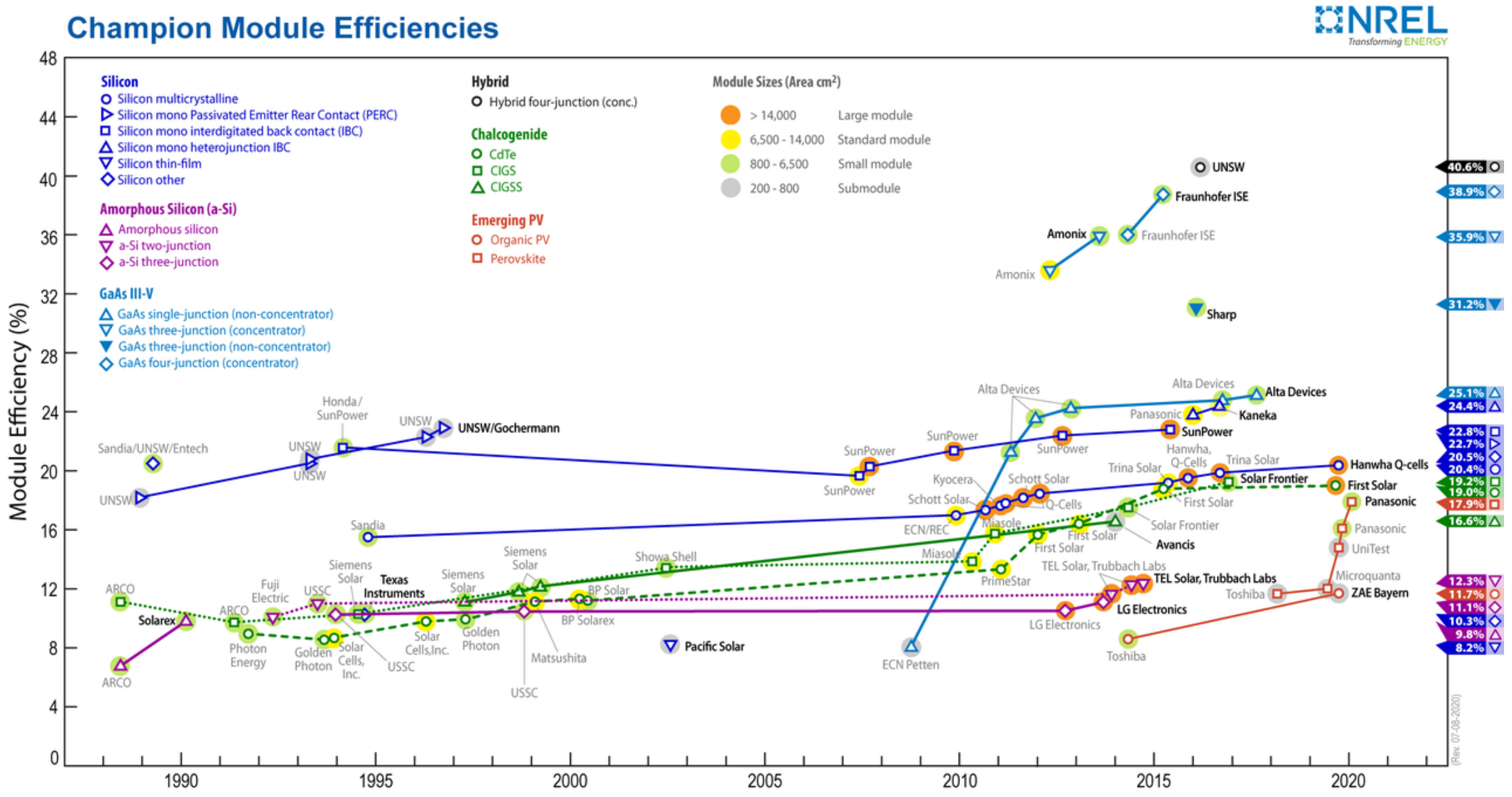

*Fig 4.8. Module efficiencies of a solar panel*

<u>*4.1.3.3 Technological significance:*</u> Most solar modules are currently produced from crystalline silicon (c-Si) solar cells made of multi-crystalline and monocrystalline silicon. In 2013, crystalline silicon accounted for more than 90 percent of worldwide PV production, while the rest of the overall market is made up of thin-film technologies using cadmium telluride, CIGS and amorphous silicon.

Emerging, third generation solar technologies use advanced thin-film cells. They produce a relatively high-efficiency conversion for the low cost compared to other solar technologies. Also, high-cost, high-efficiency, and close-packed rectangular multi-junction (MJ) cells are preferably used in solar panels on spacecraft, as they offer the highest ratio of generated power per kilogram lifted into space. MJ-cells are compound semiconductors and made of gallium arsenide (GaAs) and other semiconductor materials. Another emerging PV technology using MJ-cells is concentrator photovoltaics (CPV).

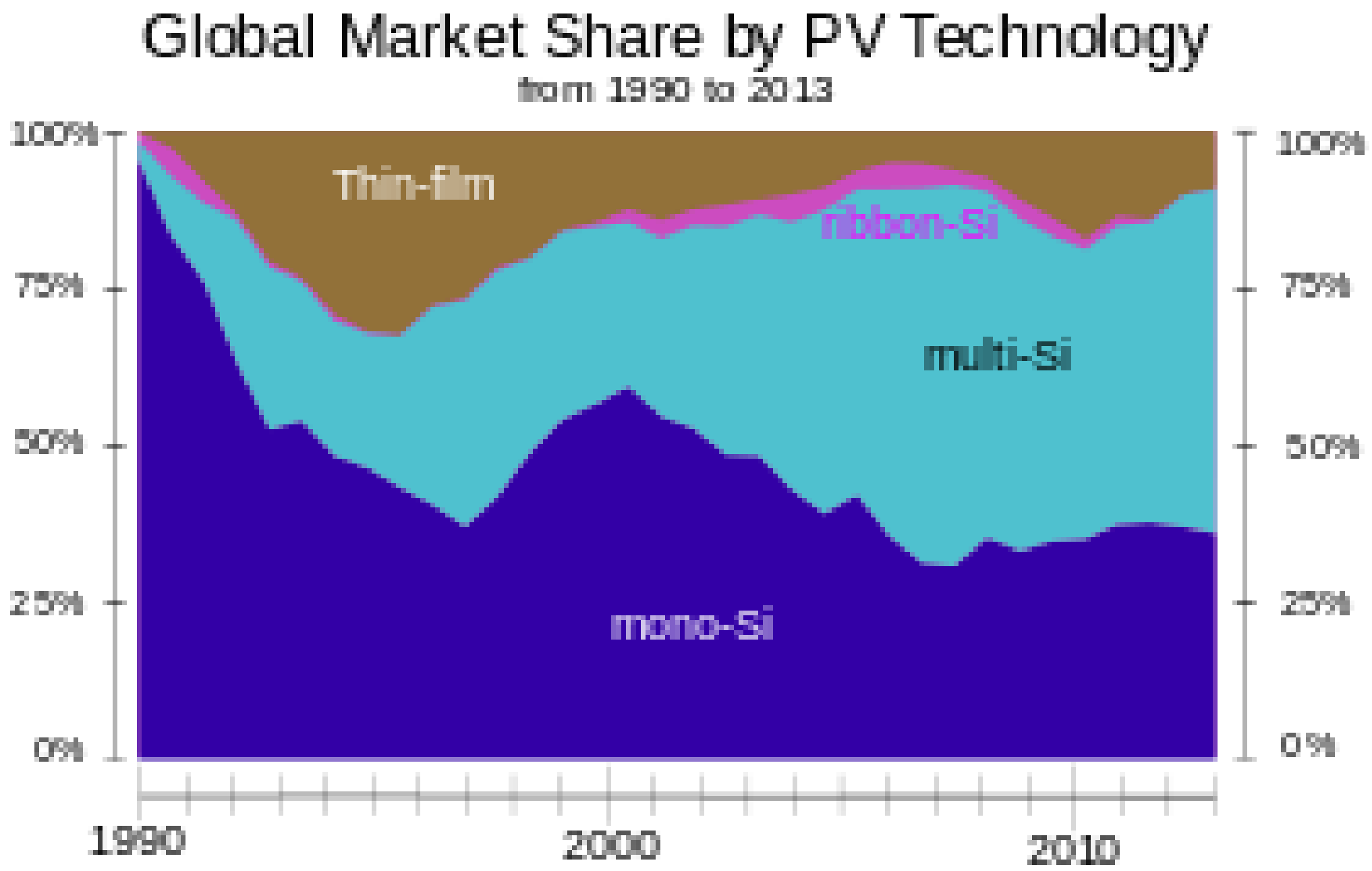

*Fig 4.9. Global market share for solar panels*

*4.1.4 DC Motors:* A DC motor is any of a class of rotary electrical machines that converts direct current electrical energy into mechanical energy. The most common types rely on the forces produced by magnetic fields. Nearly all types of DC motors have some internal mechanism, either electromechanical or electronic; to periodically change the direction of current flow in part of the motor. DC motors were the first type widely used, since they could be powered from existing direct-current lighting power distribution systems. A DC motor's speed can be controlled over a wide range, using either a variable supply voltage or by changing the strength of current in its field windings. Small DC motors are used in tools, toys, and appliances. The universal motor can operate on direct current but is a lightweight brushed motor used for portable power tools and appliances. Larger DC motors are used in propulsion of electric vehicles, elevator and hoists, or in drives for steel rolling mills. The advent of power electronics has made replacement of DC motors with AC motors possible in many applications.

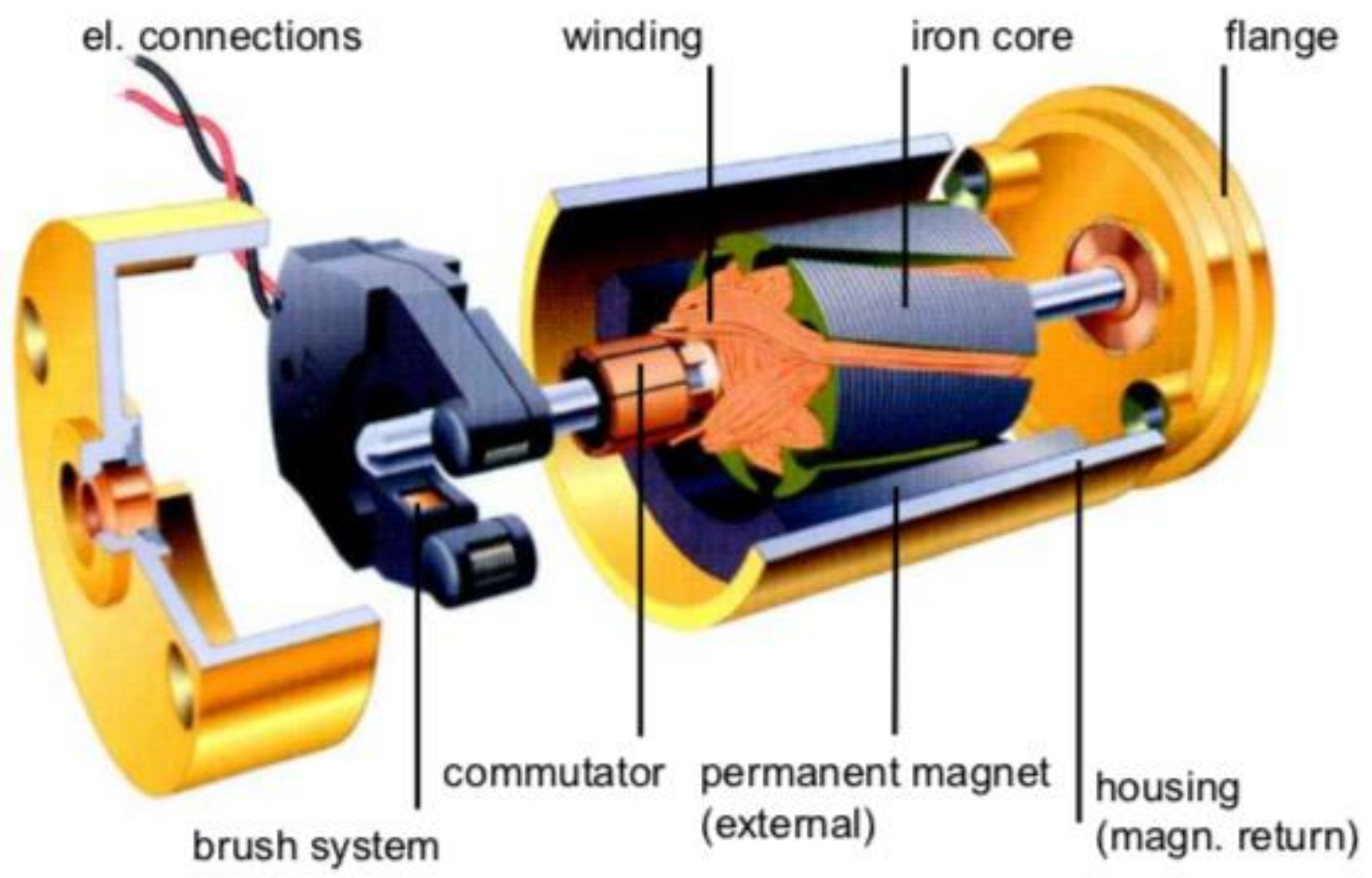

*Fig 4.10. DC motor*

*4.1.5 Battery:* Battery are a collection of one or more cells whose chemical reactions create a flow of electrons in a circuit. All batteries are made up of three basic components: an anode (the '-' side), a cathode (the '+' side), and some kind of electrolyte (a substance that chemically reacts with the anode and cathode). When the anode and cathode of a battery is connected to a circuit, a chemical reaction takes place between the anode and the electrolyte. This reaction causes electrons to flow through the circuit and back into the cathode where another chemical reaction takes place. When the material in the cathode or anode is consumed or no longer able to be used in the reaction, the battery is unable to produce electricity. At that point, your battery is "dead." Batteries that must be thrown away after use are known as primary batteries. Batteries that can be recharged are called secondary batteries.

*4.1.5.1 Types of Batteries*

a.) **NICKEL CADMIUM BATTERIES:** The active components of a rechargeable NiCd battery in the charged state consist of nickel hydroxide (NiOOH) in the positive electrode and cadmium (Cd) in the negative electrode. For the electrolyte, potassium hydroxide (KOH) is normally used. Due to their low internal resistance and the very good current conducting properties, NiCd batteries can supply extremely high currents and can be recharged rapidly. These cells are capable of sustaining temperatures down to -20°C. The selection of the separator (nylon or polypropylene) and the electrolyte (KOH, LiOH,

NaOH) influence the voltage conditions in the case of a high current discharge, the service life and the overcharging capability. In the case of misuse, a very high-pressure may arise quickly. For this reason, cells require as safety valve. NiCd cells generally offer a long service life thereby ensuring a high degree of economy.

b.) **NICKEL METAL HYDRIDE BATTERIES:** The active components of a rechargeable NiMH battery in the charged state consist of nickel hydroxide (NiOOH) in the positive electrode and a hydrogen storing metal alloy (MH) in the negative electrode as well as a potassium hydroxide (KOH)electrolyte. Compared to rechargeable NiCd batteries, NiMH batteries have a higher energy density per volume and weight.

c.) **LITHIUM-ION BATTERIES:** The term lithium-ion battery refers to a rechargeable battery where the negative electrode (anode) and positive electrode (cathode) materials serve as a host for the lithium ion (Li+). Lithium ions move from the anode to the cathode during discharge and are intercalated into (inserted into voids in the crystallographic structure of) the cathode. The ions reverse direction during charging. Since lithium ions are intercalated into host materials during charge or discharge, there is no free lithium metal within a lithium-ion cell. In a lithium-ion cell, alternating layers of anode and cathode are separated by a porous film (separator). An electrolyte composed of an organic solvent and dissolved lithium salt provides the media for lithium-ion transport. For most commercial lithium-ion cells, the voltage range is approximately 3.0 V (discharged, or 0 % state-of charge, SOC) to 4.2 V (fully charged, or 100% SOC).

d.) **SMALL SEALED LEAD ACID BATTERIES:** Rechargeable small sealed lead acid (SSLA)batteries, which are valve regulated lead acid batteries, (VRLA batteries) do not require regular addition of water to the cells, and vent less gas than flooded (wet) lead-acid batteries. SSLA batteries are sometimes referred to as “maintenance free” batteries. The reduced venting is an advantage since they can be used in confined or poorly ventilate spaces.

There are two types of VRLA batteries:  Absorbed glass mat (AGM) battery;  Gel battery (“gel cell”)

An absorbed glass mat battery has the electrolyte absorbed in a fibre-glass mat separator. A gel cell has the electrolyte mixed with silica dust to form an immobilized gel. SSLA batteries include a safety pressure relief valve. As opposed to flooded batteries, a SSLA battery is designed not to spill its electrolyte if it is inverted. In this project we use the type of battery is lead acid battery.

#### *4.1.5.2 Lead Acid Battery*

**Definition:** The battery which uses sponge lead and lead peroxide for the conversion of the chemical energy into electrical power, such type of battery is called a lead acid battery. The lead acid battery is most commonly used in the power stations and substations because it has higher cell voltage and lower cost.

**Construction of Lead Acid Battery:** The various parts of the lead acid battery are shown below. The container and the plates are the main part of the lead acid battery. The container stores chemical energy which is converted into electrical energy by the help of the plates.

**Container** – The container of the lead acid battery is made of glass, lead lined wood, ebonite, the hard rubber of bituminous compound, ceramic materials or moulded plastics and are seated at the top to avoid the discharge of electrolyte. At the bottom of the container, there are four ribs, on two of them rest the positive plate and the others support the negative plates. The prism serves as the support for the plates and at the same time protect them from a short-circuit. The material of which the battery containers are made should be resistant to sulfuric acid, should not deformer porous, or contain impurities which damage the electrolyte.

**Plate –** The plate of the lead-acid cell is of diverse design and they all consist some form of a grid which is made up of lead and the active material. The grid is essential for conducting the electric current and for distributing the current equally on the active material. If the current is not uniformly distributed, then the active material will loosen and fall out.

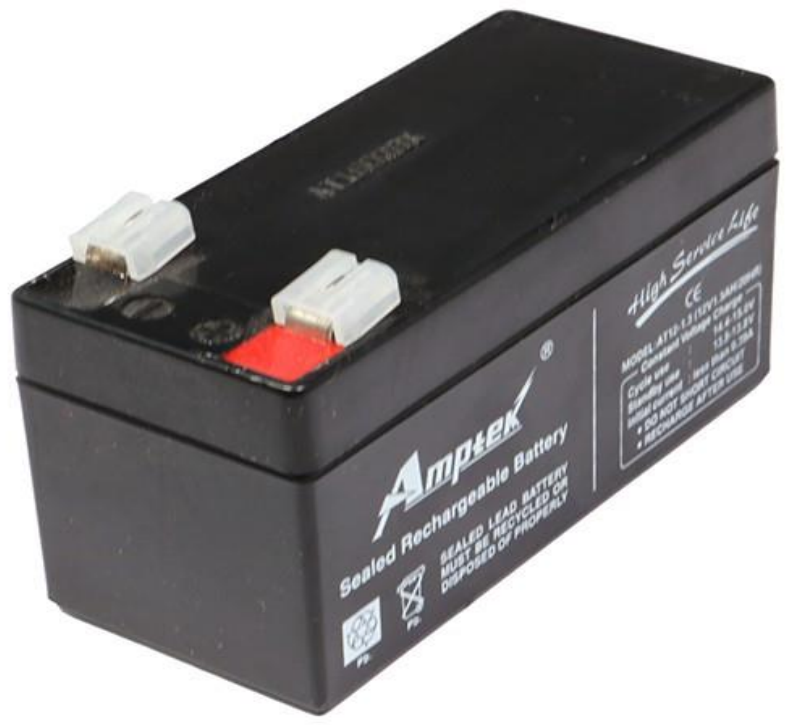

*Fig 4.11: 12V Sealed lead acid battery(1.3Ah)*

The grids are made up of an alloy of lead and antimony. These are usually made with the transverse rib that crosses the places at a right angle or diagonally. The grid for the positive and negative plates are of the same design, but the grids for the negative plates are made lighter because they are not as essential for the uniform conduction of the current. The plates of the battery are of two types. They are the formed plates or Plante plates and pasted or Faure plates. Plante's plates are used largely for stationary batteries as these are heavier in weight and more costly than the pasted plates. But the plates are more durable and less liable to lose active material by rapid charging and discharging. The Plantes plate Haslow capacity weight-ratio. Faure process is much suitable for manufacturing of negative plates rather than positive plates. The negative active material is quite tough, and it undergoes a comparatively low change from charging and discharging.

*4.1.6 Pump:* A pump is a device that moves fluids (liquids or gases), or sometimes slurries, by mechanical action, typically converted from electrical energy into hydraulic energy. Pumps can be classified into three major groups according to the method they use to move the fluid: direct lift, displacement, and gravity pumps. Pump operate by some mechanism (typically reciprocating or rotary), and consume energy to perform mechanical work moving the fluid. Pumps operate via many energy sources, including manual operation, electricity, engines, or wind power, and come in many sizes, from microscopic for use in medical applications, to large industrial pumps.

Mechanical pumps serve in a wide range of applications such as pumping water from wells, aquarium filtering, pond filtering and aeration, in the car industry for water-cooling and fuel injection, in the energy industry for pumping oil and natural gas or for operating cooling towers and other components of heating, ventilation and air conditioning systems. In the medical industry, pumps are used for biochemical processes in developing and manufacturing medicine, and as artificial replacements for body parts, in particular the artificial heart and penile prosthesis.

When a casing contains only one revolving impeller, it is called a single-stage pump. When a casing contains two or more revolving impellers, it is called a double- or multi-stage pump.

*4.1.6.1 Types:* Mechanical pumps may be submerged in the fluid they are pumping or be placed external to the fluid. Pumps can be classified by their method of displacement into positive-displacement pumps, impulse pumps, velocity pumps, gravity pumps, steam pumps and valveless pumps. There are three basic types of pumps: positive-displacement, centrifugal and axial-flow pumps. In centrifugal pumps the direction of flow of the fluid changes by ninety degrees as it flows over impeller, while in axial flow pumps the direction of flow is unchanged.

Centrifugal pumps are used to transport fluids by the conversion of rotational kinetic energy to the hydrodynamic energy of the fluid flow. The rotational energy typically comes from an engine or electric motor. They are a sub-class of dynamic axisymmetric work-absorbing turbomachinery. The fluid enters the pump impeller along or near to the rotating axis and is accelerated by the impeller, flowing radially outward into a diffuser or volute chamber (casing), from which it exits. Common uses include water, sewage, agriculture, petroleum and petrochemical pumping. Centrifugal pumps are often chosen for their high flow rate capabilities, abrasive solution compatibility, mixing potential, as well as their relatively simple engineering.

A centrifugal fan is commonly used to implement an air handling unit or vacuum cleaner. The reverse function of the centrifugal pump is a water turbine converting potential energy of water pressure into mechanical rotational energy.

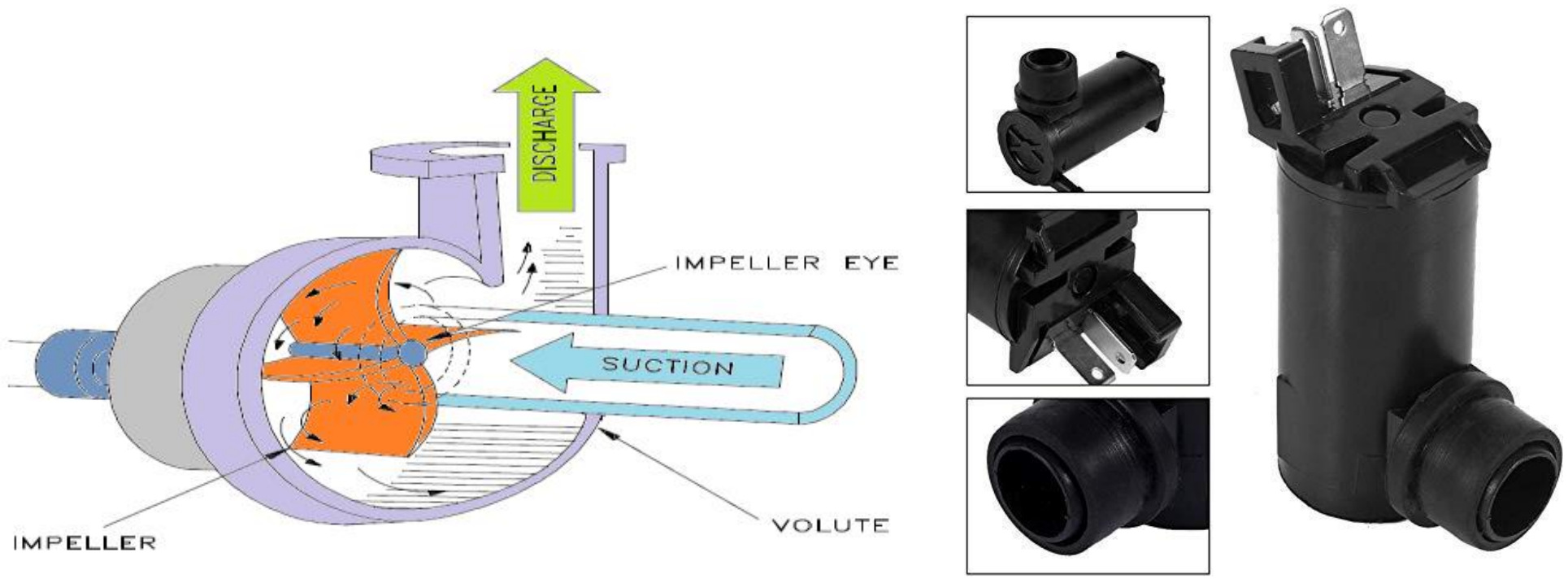

*Fig 4.12: Centrifugal Pump*

*4.1.6.2 Working:* Like most pumps, a centrifugal pump converts rotational energy, often from a motor, to energy in a moving fluid. A portion of the energy goes into kinetic energy of the fluid. Fluid enters axially through eye of the casing, is caught up in the impeller blades, and is whirled tangentially and radially outward until it leaves through all circumferential parts of the impeller into the diffuser part of the casing. The fluid gains both velocity and pressure while passing through the impeller. The doughnut-shaped diffuser, or scroll, section of the casing decelerates the flow and further increases the pressure.

*4.1.7 Linear actuators:* A linear actuator is an actuator that creates motion in a straight line, in contrast to the circular motion of a conventional electric motor. Linear actuators are used in machine tools and industrial machinery, in computer peripherals such as disk drives and printers, in valves and dampers, and in many other places where linear motion is required. Hydraulic or pneumatic cylinders inherently produce linear motion. Many other mechanisms are used to generate linear motion from a rotating motor.

*4.1.7.1 Types*

a.) *Mechanical actuators:* Mechanical linear actuators typically operate by conversion of rotary motion into linear motion. Conversion is commonly made via a few simple types of mechanism:

*Screw*: leadscrew, screw jack, ball screw and roller screw actuators all operate on the principle of the simple machine known as the screw. By rotating the actuator's nut, the screw shaft moves in a line.

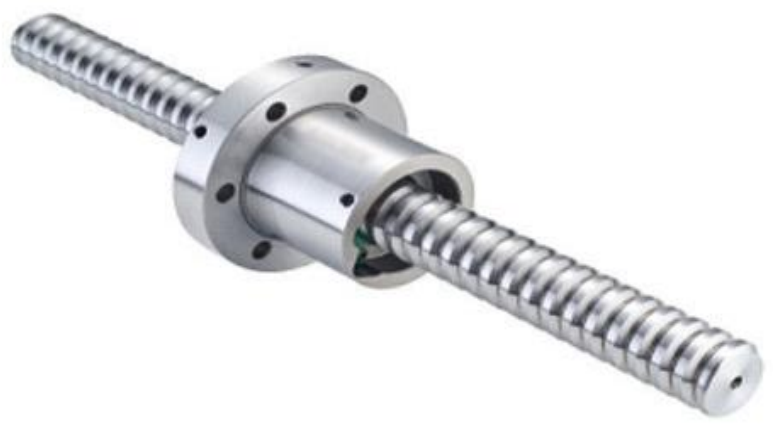

*Fig 4.13: Screw actuator*

*Wheel and axle*: Hoist, winch, rack and pinion, chain drive, belt drive, rigid chain and rigid belt actuators operate on the principle of the wheel and axle. A rotating wheel moves a cable, rack, chain or belt to produce linear motion.[1]

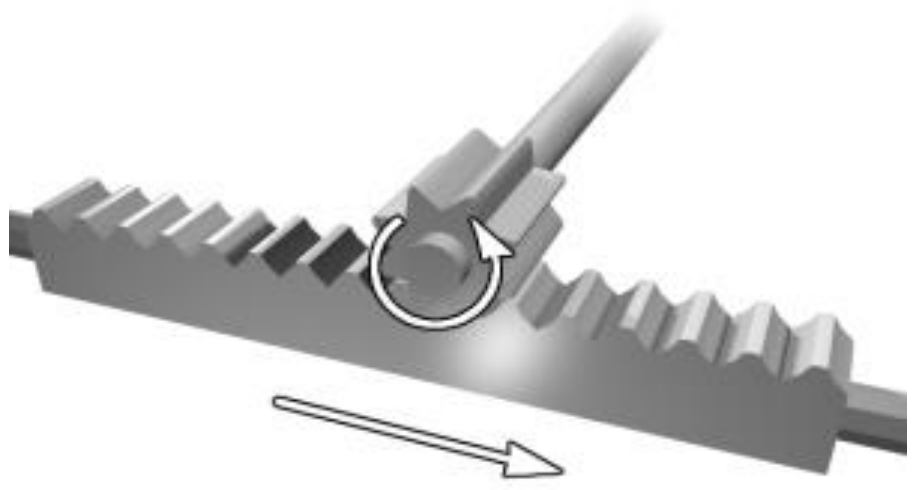

*Fig 4.14: Wheel and axle actuator*

*Cam*: Cam actuators function on a principle similar to that of the wedge, but provide relatively limited travel. As a wheel-like cam rotates, its eccentric shape provides thrust at the base of a shaft.

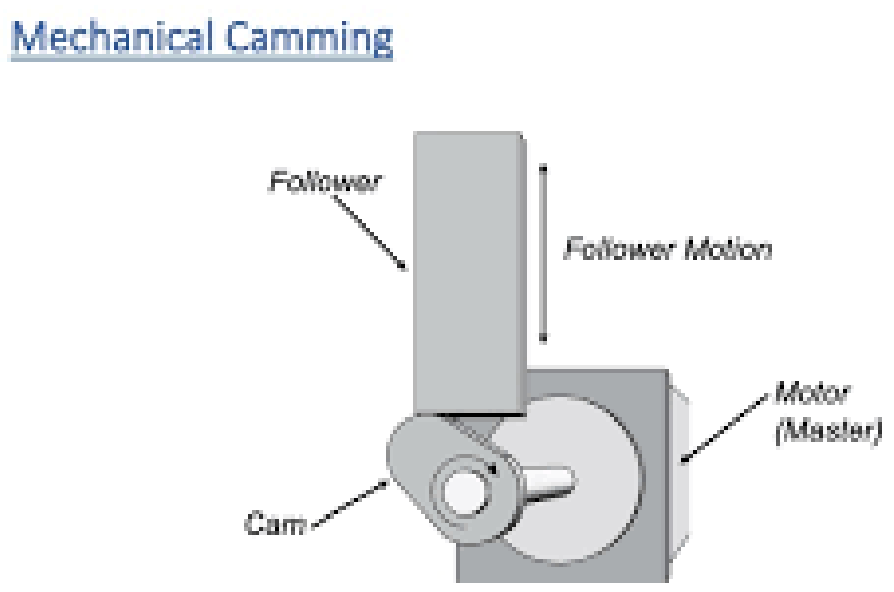

*Fig 4.15: Cam actuation*

Some mechanical linear actuators only pull, such as hoists, chain drive and belt drives. Others only push (such as a cam actuator). Pneumatic and hydraulic cylinders, or lead screws can be designed to generate force in both directions.

Mechanical actuators typically convert rotary motion of a control knob or handle into linear displacement via screws and/or gears to which the knob or handle is attached. A jackscrew or car jack is a familiar mechanical actuator. Another family of actuators are based on the segmented spindle. Rotation of the jack handle is converted mechanically into the linear motion of the jack head. Mechanical actuators are also frequently used in the field of lasers and optics to manipulate the position of linear stages, rotary stages, mirror mounts, goniometers and other positioning instruments. For accurate and repeatable positioning, index marks may be used on control knobs. Some actuators include an encoder and digital position readout. These are similar to the adjustment knobs used on micrometres except their purpose is position adjustment rather than position measurement.

b.) *Electrical actuators:* Electro-mechanical actuators are similar to mechanical actuators except that the control knob or handle is replaced with an electric motor. Rotary motion of the motor is converted to linear displacement. Electromechanical actuators may also be used to power a motor that converts electrical energy into mechanical torque. There are many designs of modern linear actuators and every company that manufactures them tends to have a proprietary method. The following is a generalized description of a very simple electro-mechanical linear actuator.

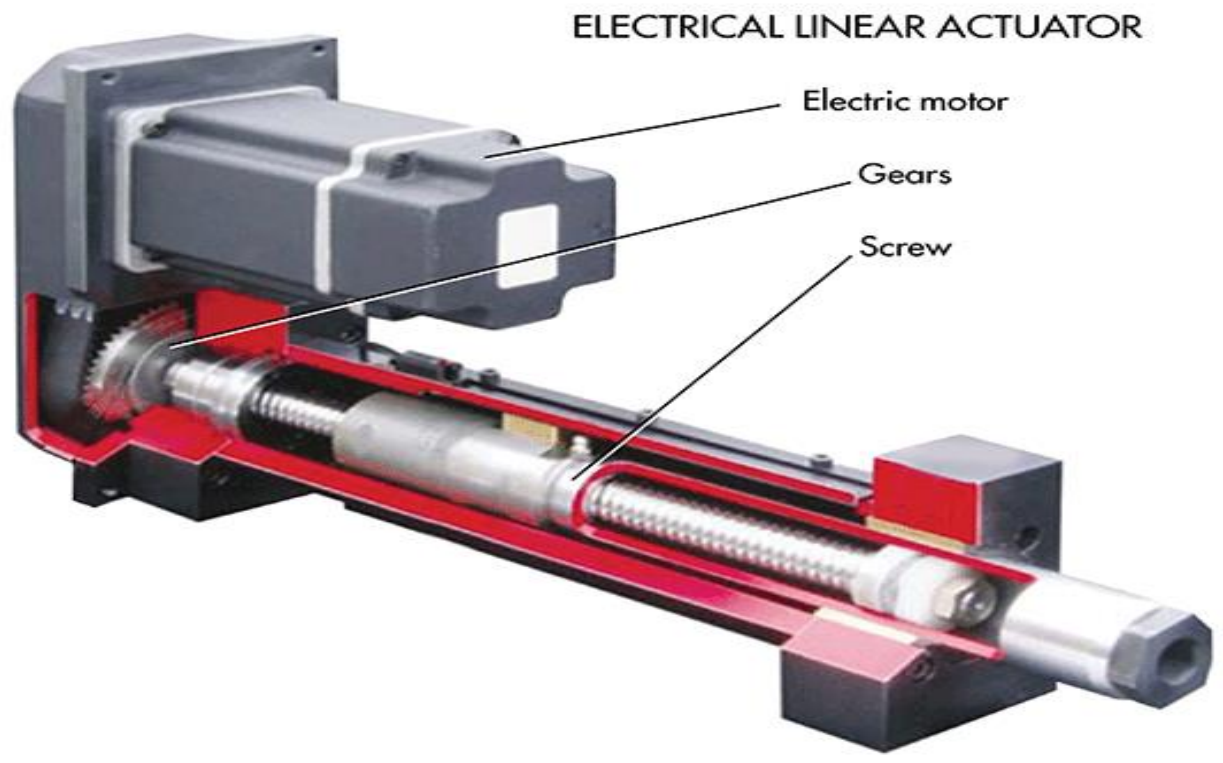

*Fig 4.16: Electrical Actuator*

*4.1.7.1.b.1_Simplified_design:* Typically, an electric motor is mechanically connected to rotate a lead screw. A lead screw has a continuous helical thread machined on its circumference running along the length (similar to the thread on a bolt). Threaded onto the lead screw is a lead nut or ball nut with corresponding helical threads. The nut is prevented from rotating with the lead screw (typically the nut interlocks with a non-rotating part of the actuator body). When the lead screw is rotated, the nut will be driven along the threads. The direction of motion of the nut depends on the direction of rotation of the lead screw. By connecting linkages to the nut, the motion can be converted to usable linear displacement. Most current actuators are built for high speed, high force, or a compromise between the two. When considering an actuator for a particular application, the most important specifications are typically travel, speed, force, accuracy, and lifetime. Most varieties are mounted on dampers or butterfly valves.

There are many types of motors that can be used in a linear actuator system. These include dc brush, dc brushless, stepper, or in some cases, even induction motors. It all depends on the application requirements and the loads the actuator is designed to move. For example, a linear actuator using an integral horsepower AC induction motor driving a lead screw can be used to operate a large valve in a refinery. In this case, accuracy and high movement resolution aren't needed, but high force and speed are. For electromechanical linear actuators used in laboratory instrumentation robotics, optical and laser equipment, or X-Y tables, fine resolution in the micron range and high accuracy may require the use of a fractional horsepower stepper motor linear actuator with a fine pitch lead screw. There are many variations in the electromechanical linear actuator system. It is critical to understand the design requirements and application constraints to know which one would be best.

*4.1.7.2 Comparison of electrical and mechanical actuators*

| Actuator Type | Advantages | Disadvantages |
|---|---|---|
| Mechanical | Cheap, Repeatable, No power source required, Self-contained, Identical behaviour extending or retracting | Manual operation only. No operation. |
| Electro-Mechanical | Cheap, Repeatable, Operation can be automated. Self-contained, Identical behaviour extending or retracting. DC or stepping motors. Position feedback possible. | Many moving parts prone to wear. |

4.1.8 Nozzle: A spray nozzle is a precision device that facilitates dispersion of liquid into a spray. Nozzles are used for three purposes: to distribute a liquid over an area, to increase liquid surface area, and create impact force on a solid surface. A wide variety of spray nozzle applications use a number of spray characteristics to describe the spray.

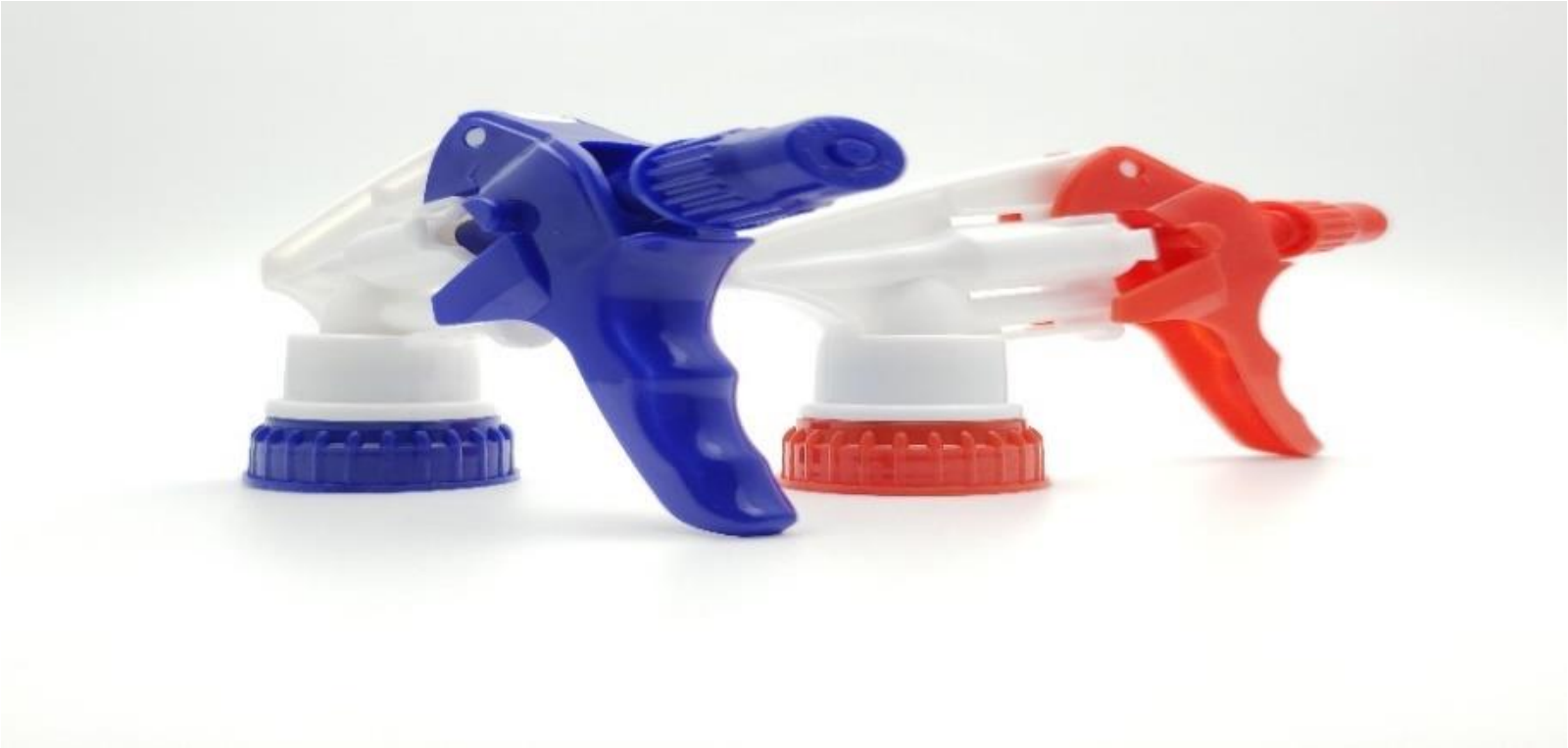

*Fig 4.17: Spray bottle nozzles*

4.1.8.1 *Working of spray bottle nozzle*

a.) *Bottle:* The bottle is the reservoir from which the pump pulls liquid when the user presses the trigger on the spray bottle. The liquid moves from the bottle up the tube. The tube and reservoir are similar in spray bottles and other kinds of pumps, such as pumps on soap dispensers. For thin liquids such as cleaning products, a thinner tube can work, but for thicker liquids such as hand soap, a thicker tube is necessary to pull liquid.

b.) *Pump:* The pump is the main working part of a spray bottle. The pump consists of the trigger mechanism, a piston, a cylinder and a one-way valve. When the user presses the trigger, it forces the piston into the cylinder, which forces the liquid through the nozzle as a concentrated stream of liquid. When the trigger is released, the piston moves back, pulling liquid back into the cylinder. This liquid is forced out of the nozzle the next time the trigger is pressed. A one-way valve at the bottom of the pump only allows liquid to flow up the tube into the pump, not back into the bottle.

c.) *Nozzle:* The nozzle of the spray bottle concentrates the liquid into a stream by forcing it through a very small hole. Spray bottle nozzles also have a one-way valve in them that keeps air from flowing back into the pump and allows for suction within the pump so that liquid can be pulled up the tube. Without this valve, the pump would suck air back into the cylinder, instead of liquid up from the bottle.

# 5. CONCEPTUAL DESIGN

The design was carried out in Fusion 360 3D Modelling software. This software is chosen for this project because it has user friendly interface and it is easy to use. Given below is the step-by-step procedure that was taken by us to create the conceptual design.

### *5.1. The chassis:*

To create the chassis, first create a rectangular block of dimensions (22.9*14.2*9.5 $in^3$). Then sketch two offset rectangles on the two adjacent faces of the block at an offset of 1 inch and use the command "extrude cut" to generate the chassis as shown in the figure below.

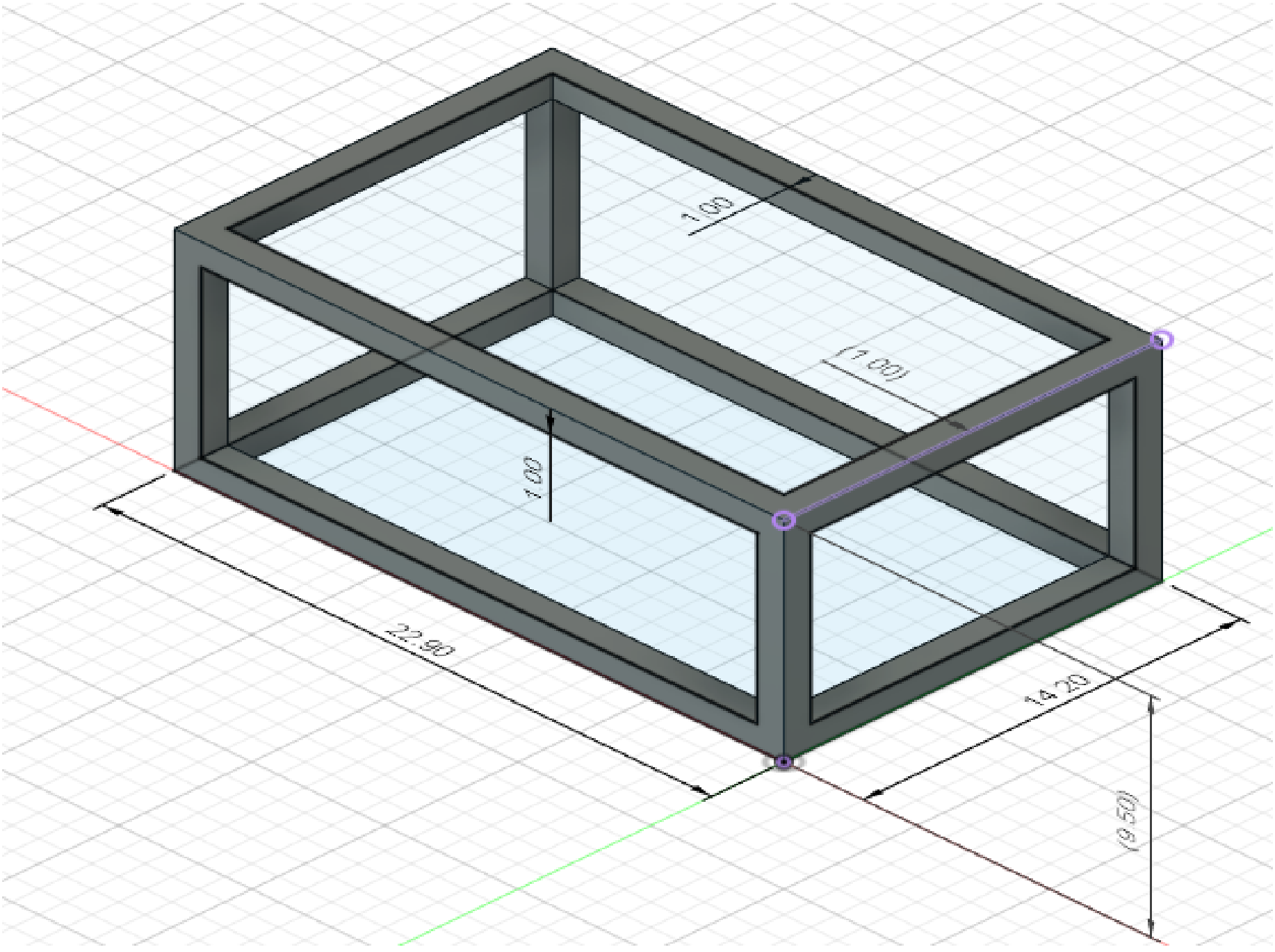

*Fig 5.1: Chassis with dimensions*

Extrude two flat rods from the bottom rectangular frame which divide the frame into 3 equal parts as shown in the figure. This is done to give a support to the base.

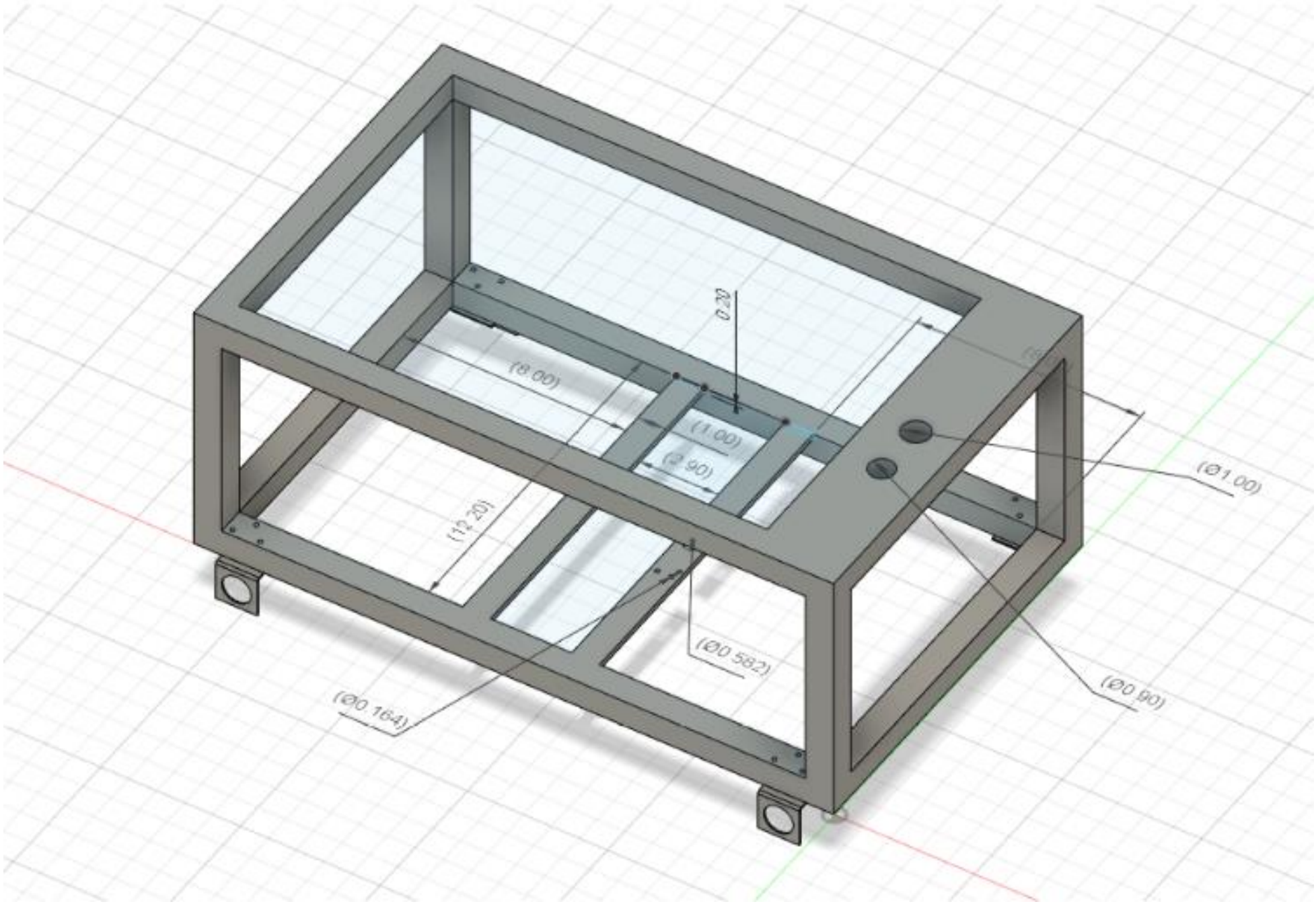

*Fig 5.2: Chassis design with dimensions (Isometric view)*

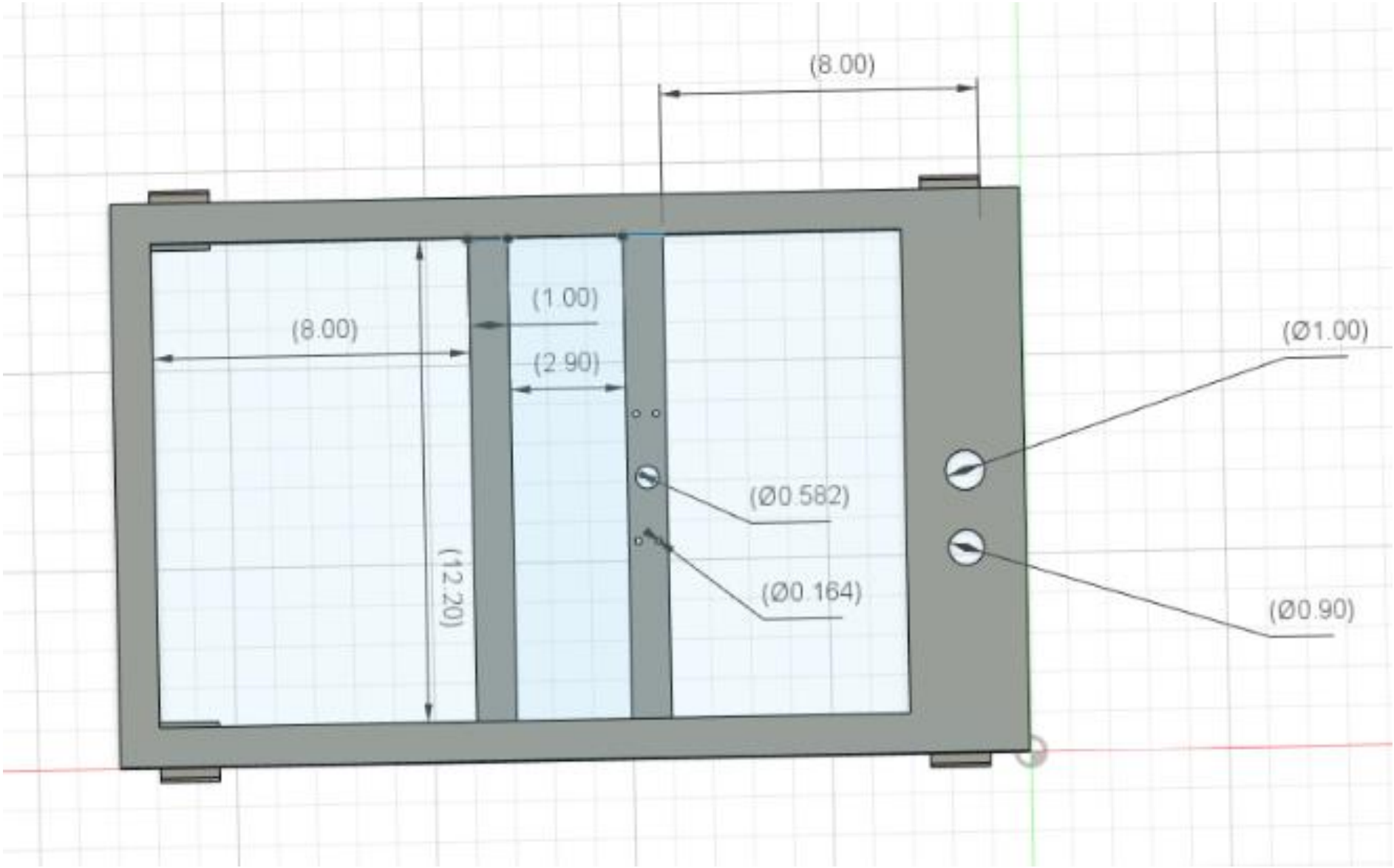

*Fig 5.3: Chassis design with dimensions (Top view)*

<u>*5.1.1 Selection of material for chassis:*</u>

Material selected: Mild Steel EN 1.0301 (Euro Norm)

EN 1.0301 carbon steel contains 0.1% carbon, 0.4% manganese and 0.4 percent silicon. It also contains small amounts of copper (Cu), nickel (Ni), chromium (Cr), aluminum (Al), and molybdenum (Mo).

#### *5.1.1.1 Mechanical properties of EN 1.0301*

| | |
|---|---|
| Material name | 1.0301 |
| US Standard (AISI) | 1011, M1010 |
| Abbreviation | C10 |
| Density (g/cm3) | 7.85 |
| Yield strength (N/mm2) | >=200 |
| Tensile strength (N/mm2) | 350-640 |
| Machinability | Good |
| Weldability | Suitable |

*Table from KIPP.com*

This grade has excellent weldability and is commonly used for extruded, forged, cold headed, and cold-pressed parts and forms. It is primarily used in automotive equipment, furniture, and appliances.

#### *5.1.1.2 Physical Properties of Mild Steel*

High tensile strength.

High impact strength.

Good ductility and weldability.

A magnetic metal due to its ferrite content.

Good malleability with cold-forming possibilities.

Not suitable for heat treatment to improve properties.

### *5.2: Selection of motors for 4-wheel drive*

The DC Motors with high torque and high speed are selected as these motors have to bear the weight of the entire setup and move it according to our command. The motor with Nut is preferred as it can be easily fitted at any desired location.

#### *5.2.1 Specs of Motor*

| | |
|---|---|
| Shaft dia | 0.142 in |
| Shaft length | 1.33 in |
| Speed | 250 rpm |
| Torque | 1.5535 N-mm |

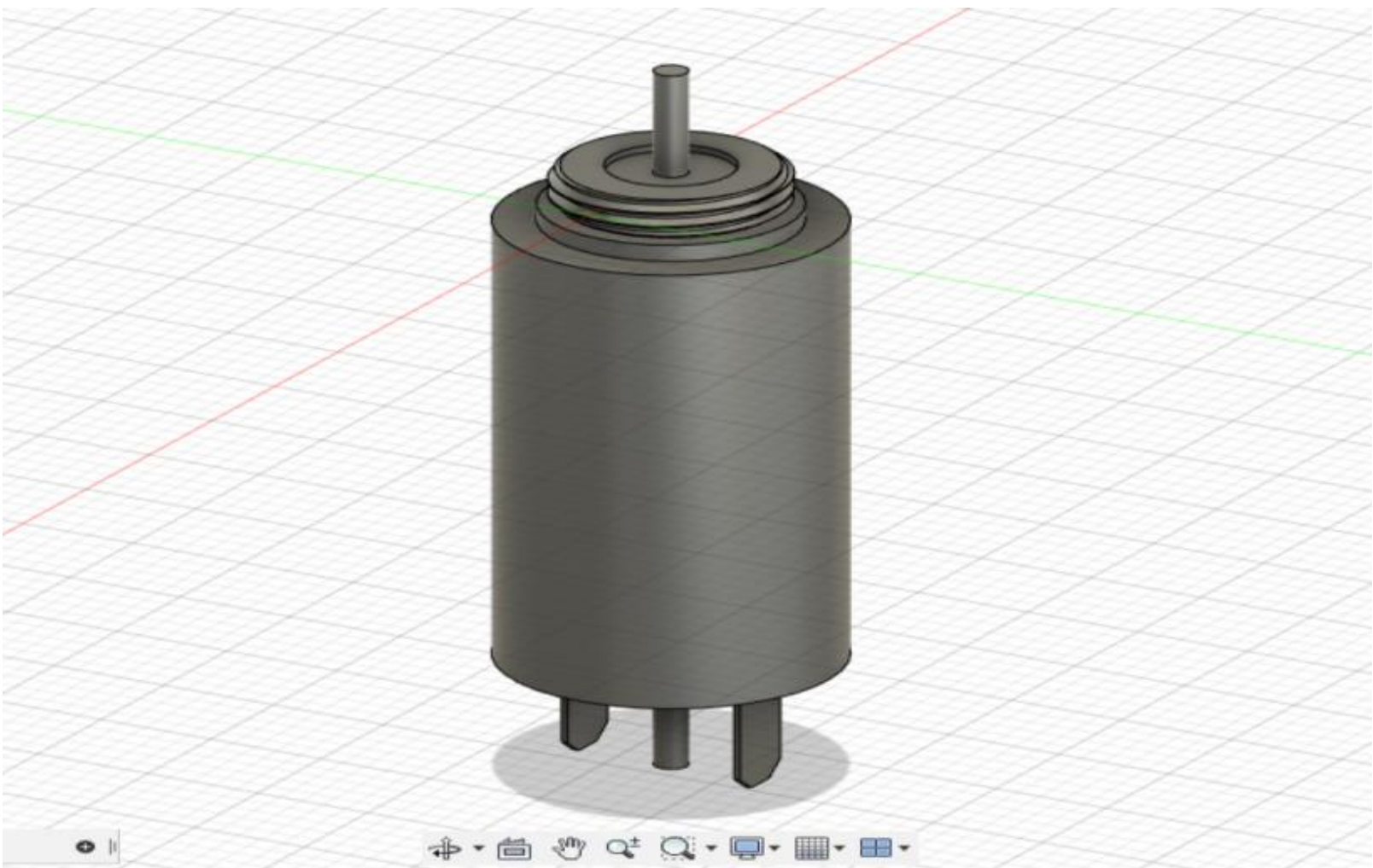

*Fig 5.4:  DC Motor for drive wheels*

### *5.3: Design of brackets to hold the motor*

The Brackets are designed in such a way that the Threads on the motor are on one side of the bracket and the Body of the motor is on the other side. When the nut fastens on the motor threads it would hold the motor tightly with the bracket.  The design of the bracket is as shown in the figure.

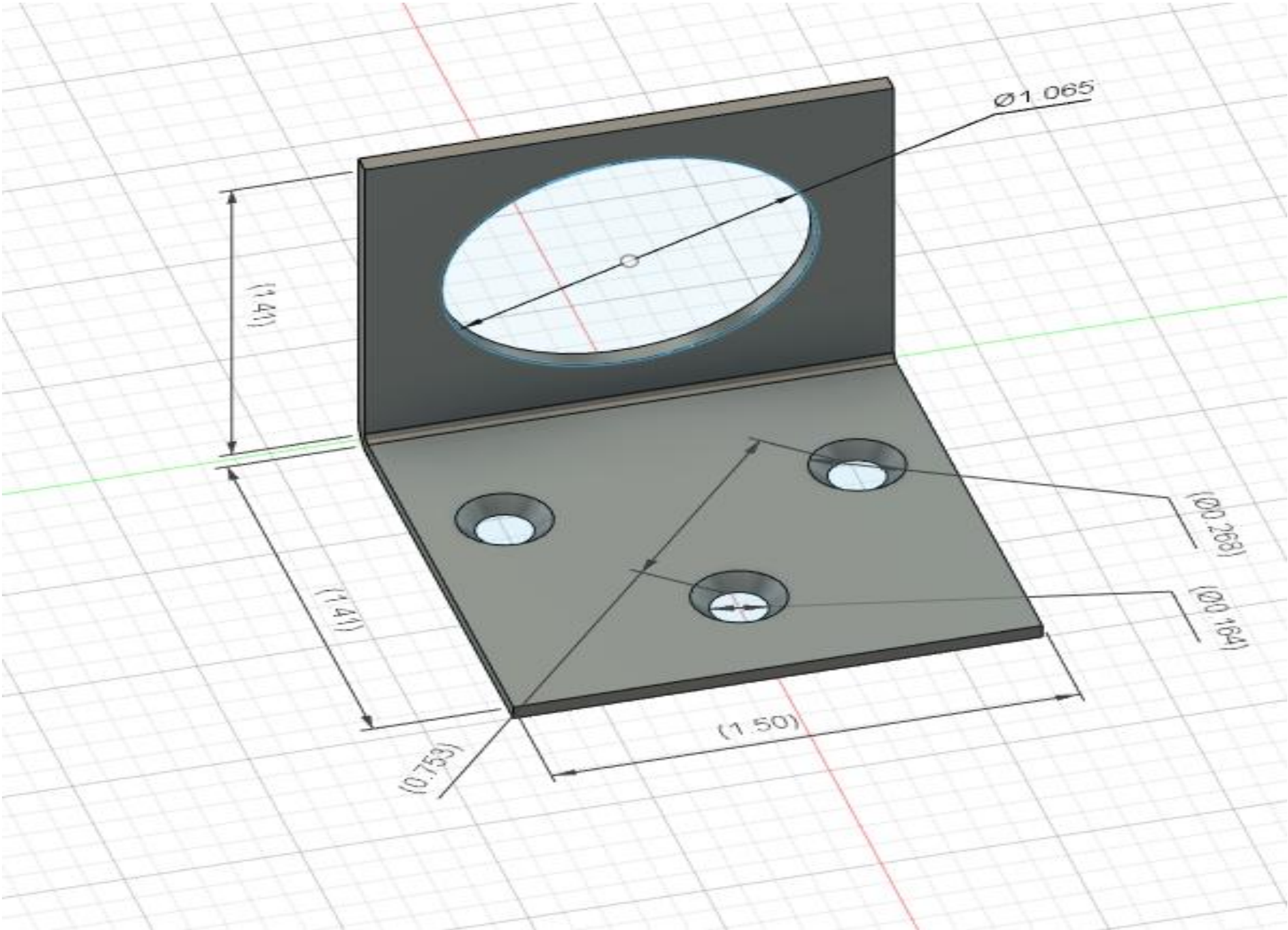

*Fig 5.5: Bracket design with dimensions*

### *5.4: Fixing the brackets to chassis*

The 8mm dia screw and nut combination are chosen to hold the bracket against the chassis. Extrude 3 holes of 8mm at the corners of chassis in such a way that the holes in clamp and holes on chassis are aligned. For each bracket 3 such Screw and Nut combinations are required to make a rigid joint.

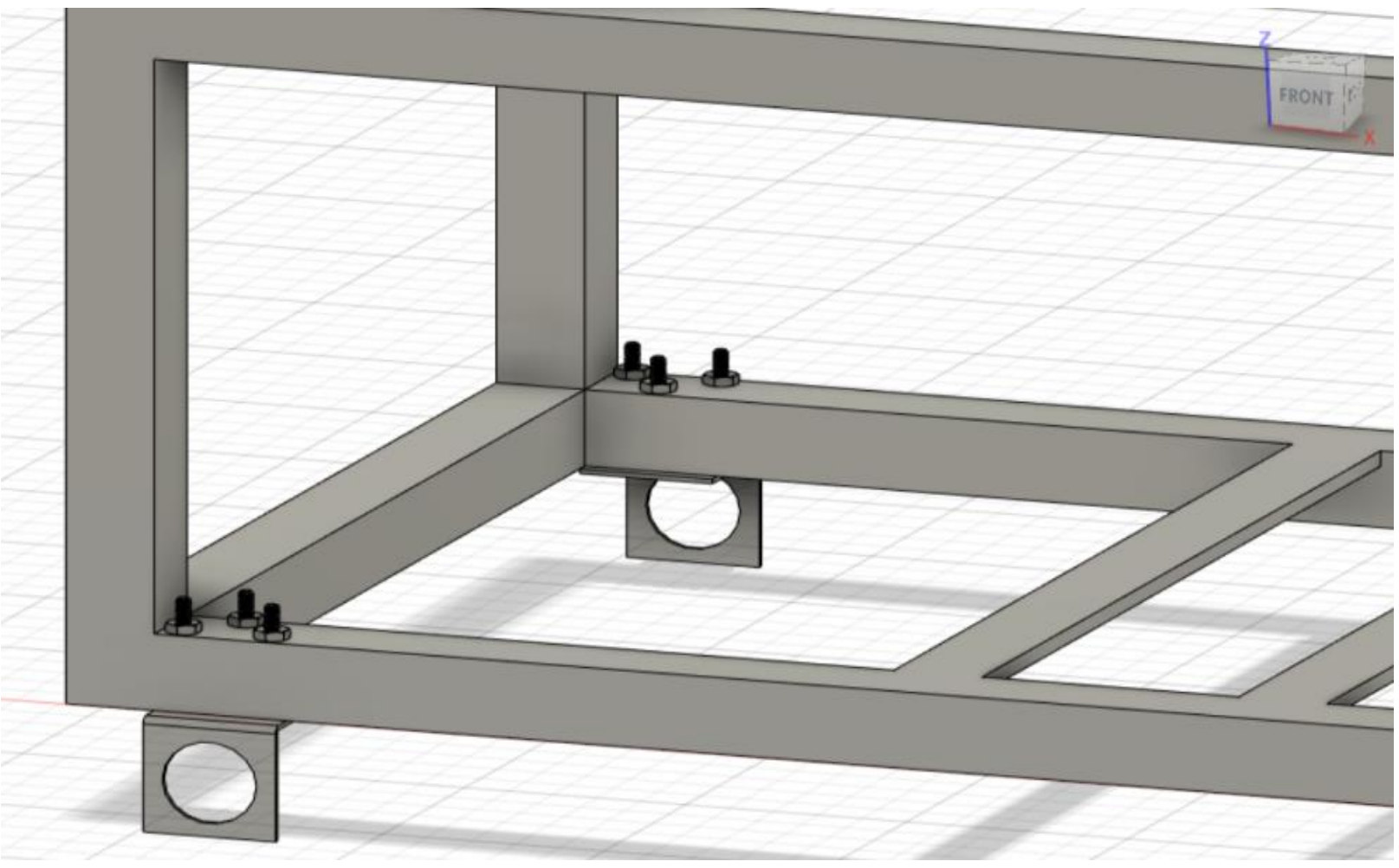

*Fig 5.6: Position of brackets on the chassis*

The same is done for fixing the two forward brackets as well

### *5.5: Fixing motor to the brackets*

The motor is slid into the motor hole provided in the bracket and the nut is fastened. The same is repeated for 4 corners of the chassis.

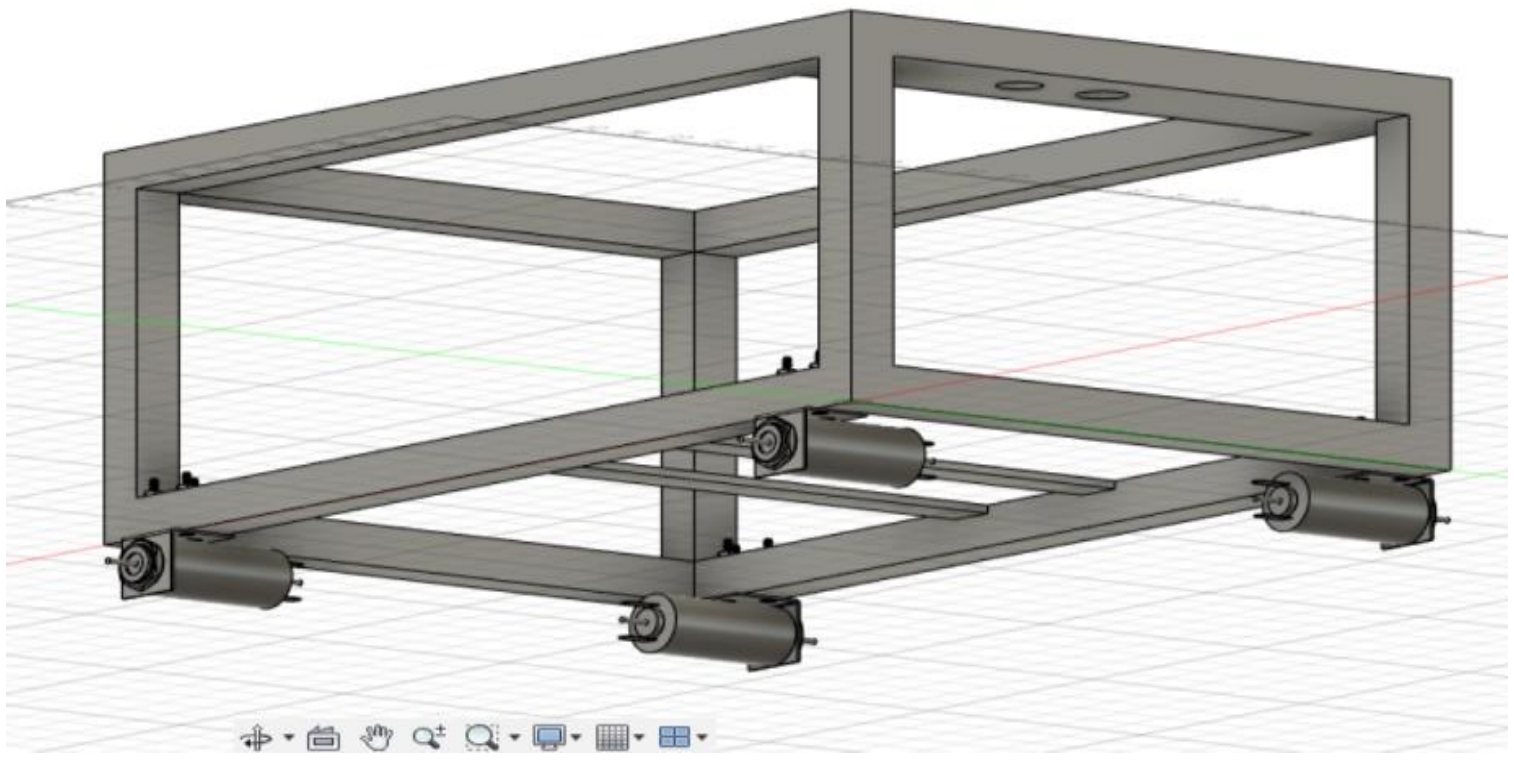

*Fig 5.7: Fixing Motors to the brackets on the chassis*

### *5.6: Fixing the wheel to the motor axle*

Since the wheel has to bear the weight of the entire setup, and the model has to traverse through adverse terrains, the wheel has to be sturdy, gripping and of moderate size. Considering the above aspects, the selection of wheel has been done.

*5.6.1: Specs of wheel:*

| | |
|---|---|
| Wheel dia | 15 in |
| Wheel width | 4.5 in |
| For Axle dia | 0.143 in |
| Style | Tubeless |
| Pressure capacity of Wheel | 4.06 MPa |

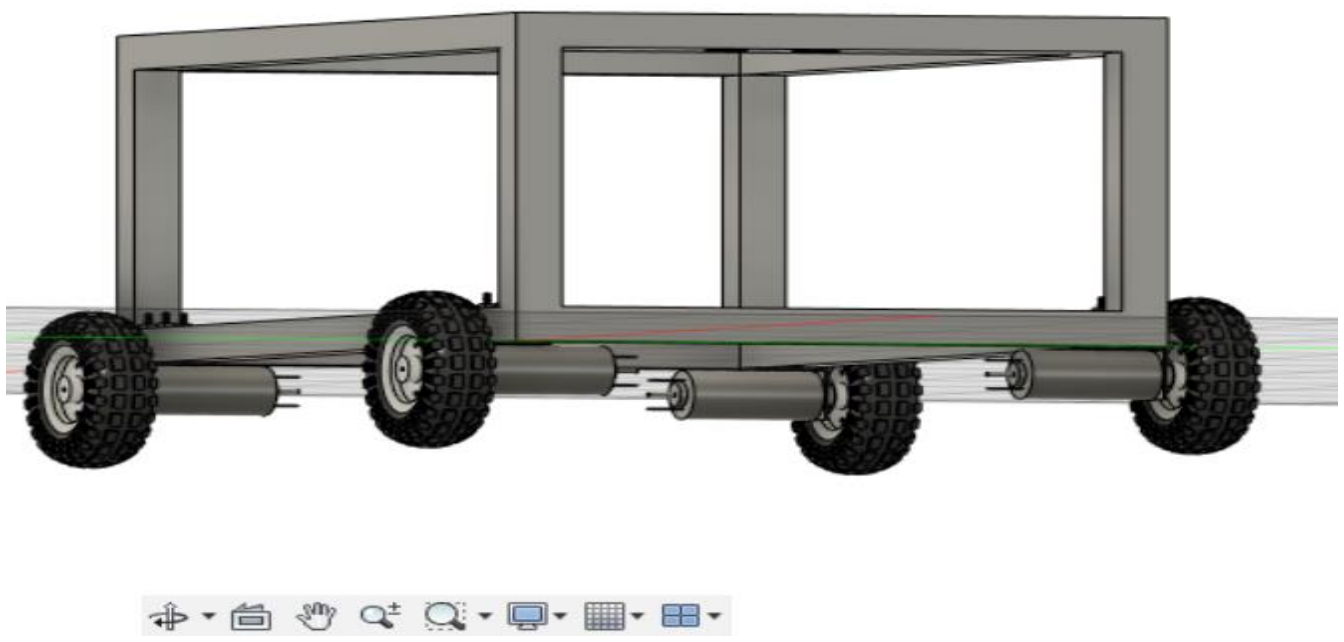

*Fig 5.8: Fixing the wheel to the motor axle*

### *5.7: Base*

A base slab made of wood is to be fitted inside the bottom rectangular space of the chassis. This is made to place all the components of the model such as pesticide storage tank, Battery, Pump etc.  The base is as shown in the figure. A battery holder setup is also made on the base for better stability of battery.

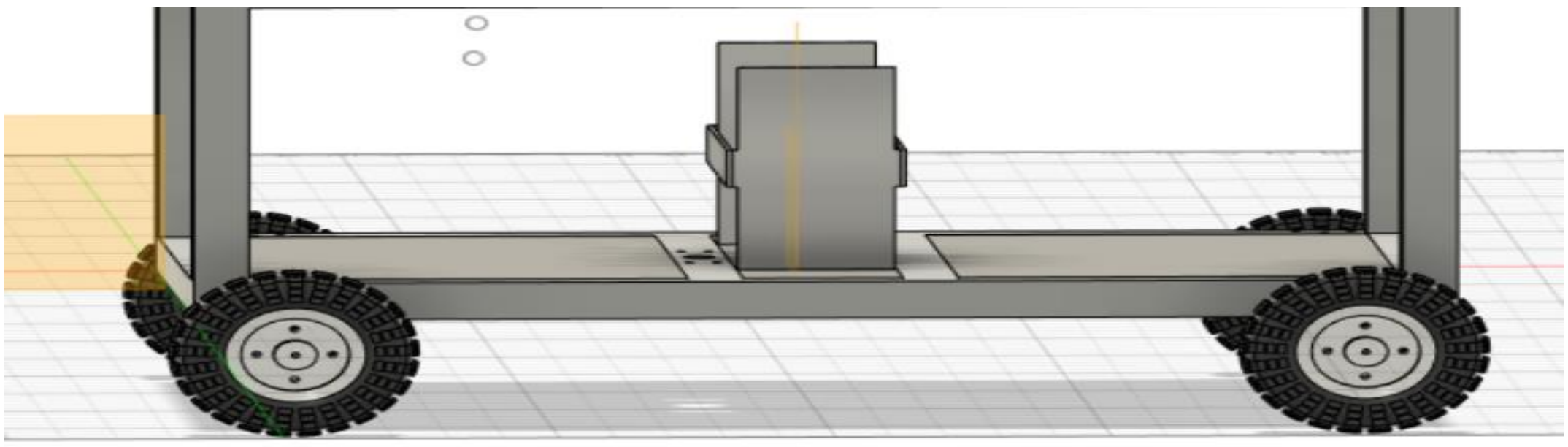

*Fig 5.9: Base with battery holder setup*

### *5.8: Mower*

The selection of motor for mowing has to be based on the speed of the motor. The mower requires the blade to rotate at very high velocities so as to trim the weed or crops effectively and effortlessly.

*5.8.1 Specs of mower motor*

| | |
|---|---|
| Shaft dia | 0.142 in |
| Shaft length | 1.33 in |
| Speed | 1000 rpm |
| Torque | 0.995 N-mm |

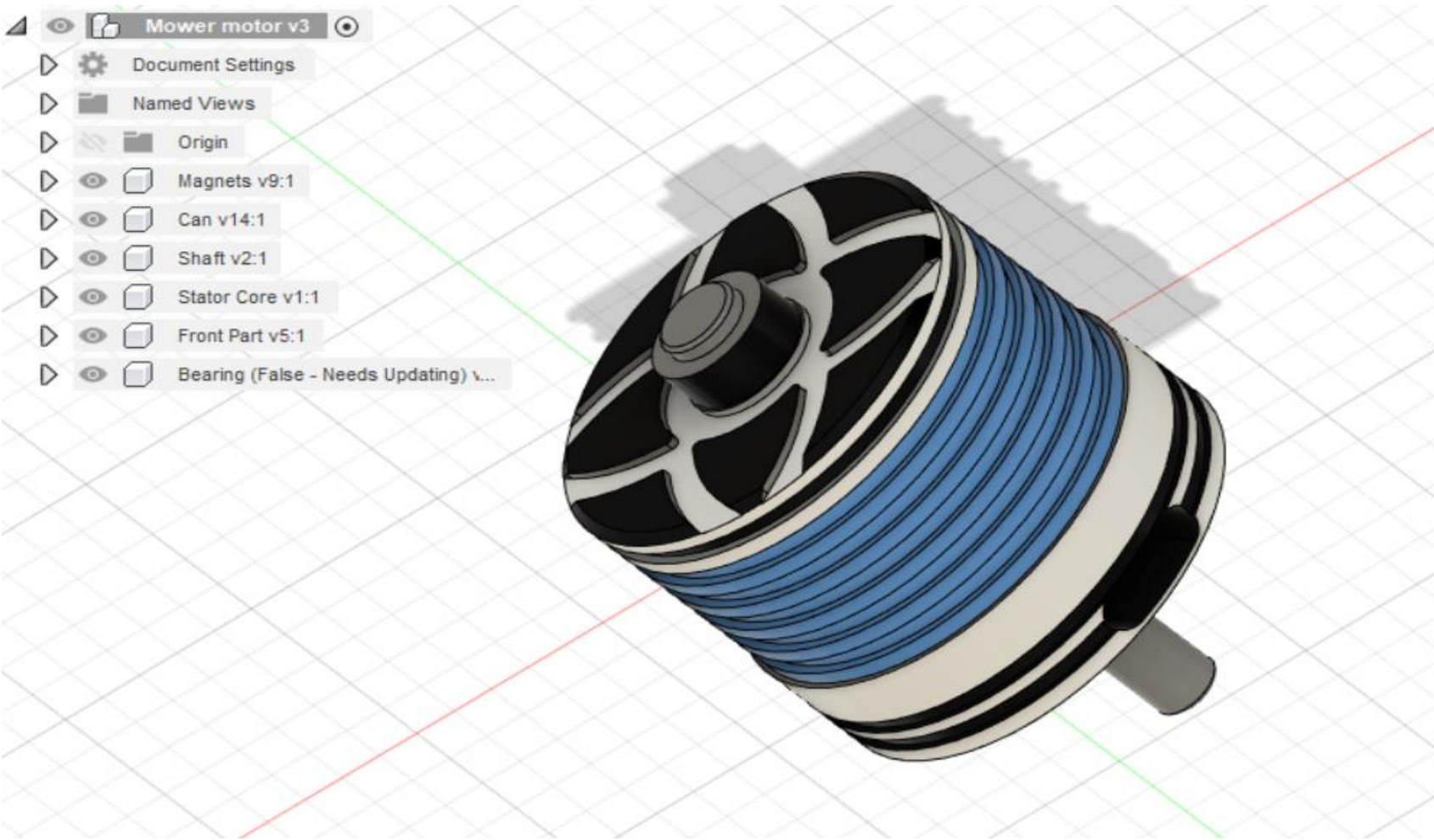

*Fig 5.10: Mower motor*

*5.8.2 Arrangement of mower*

The blade selected for mower is as shown in the figure below. It is selected as this design of cuts higher volumes of the grass at a very good rate. The blade is also simple, compact and easily available.

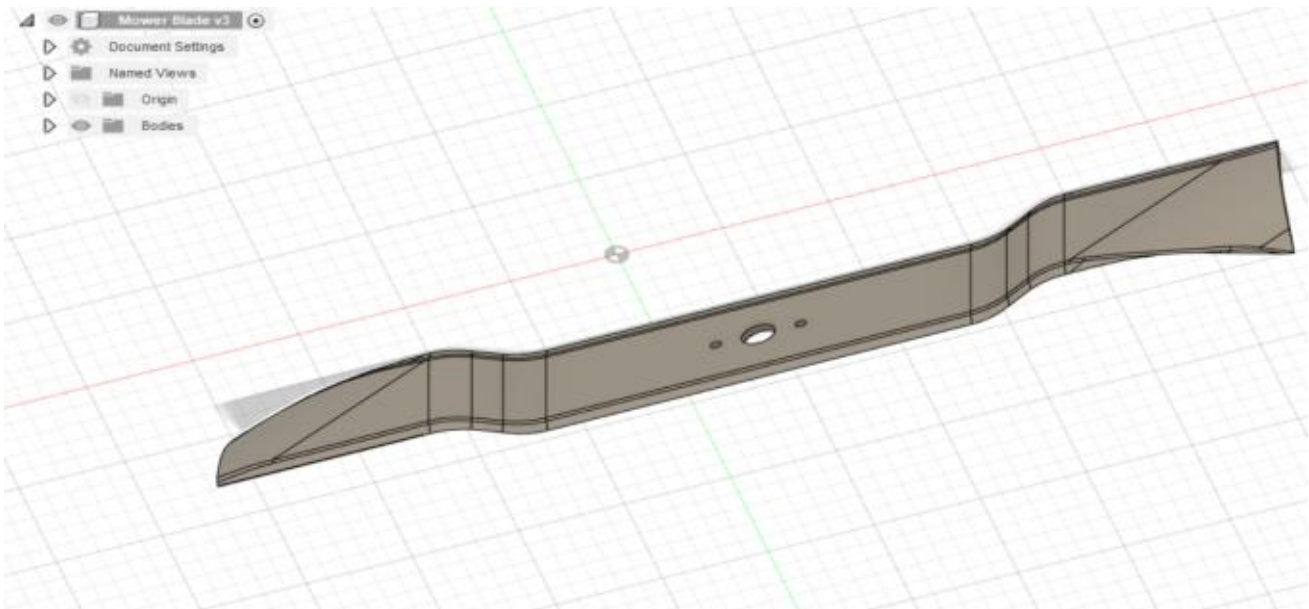

*Fig 5.11: Mower blade*

A hole is made in the center of the mower blade such that the blade slides over the whole length of the mower motor axle. Two rubber stoppers are used to adjust the position of the blade.  The above setup (i.e., the combination of mower motor and blade is referred as Mower.)  gives us flexibility in the aspect of level of the grass field to be maintained. This Mower now has to be fitted to the chassis.  For this, a U shaped bracket has to be designed in such way the it holds the Mower setup tightly to the chassis.

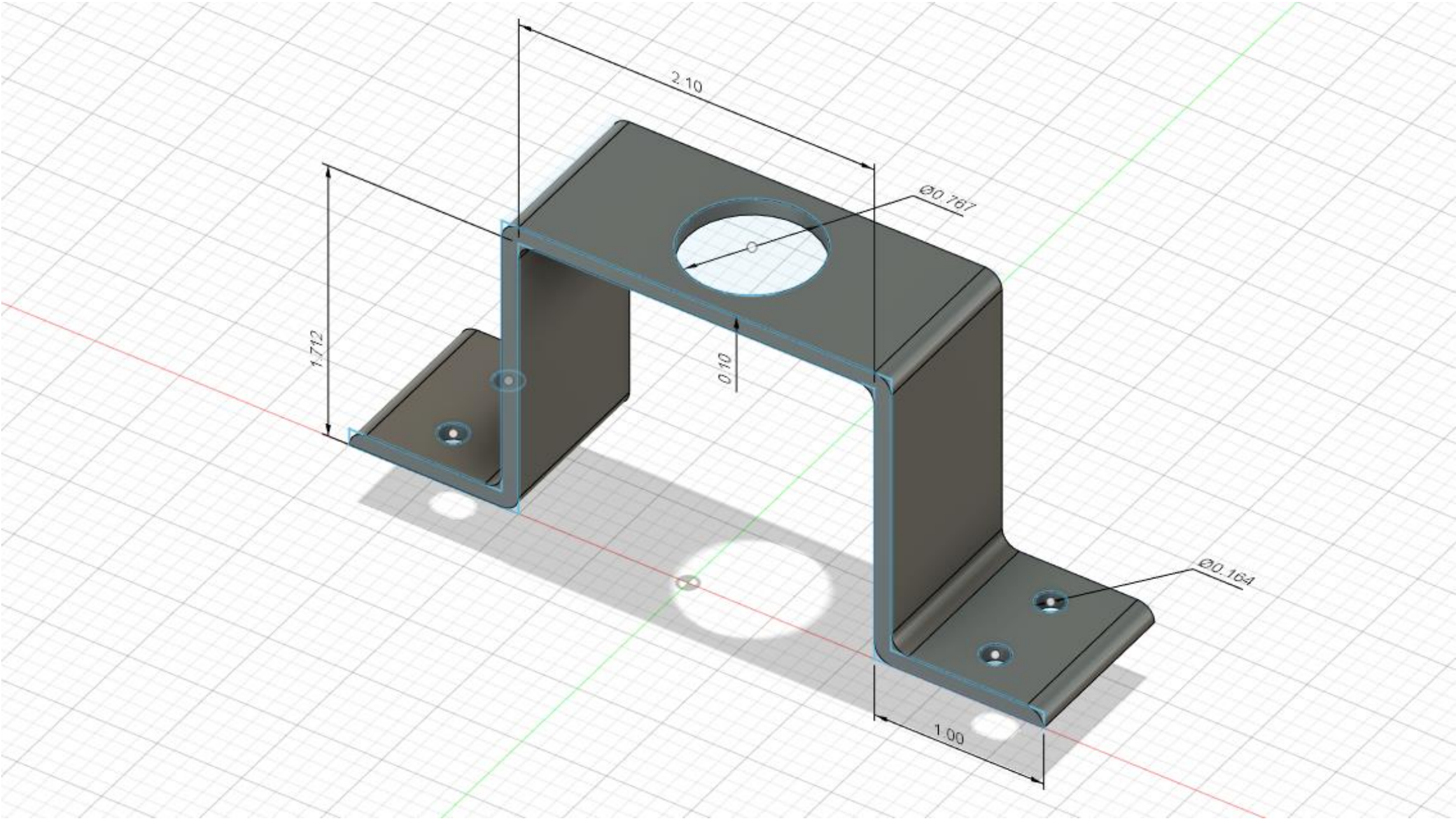

*Fig 5.12: U Shaped bracket*

For this arrangement, first holes have to be extruded in the front support rod of the chassis in such way that the holes in clamp and holes on the rod are aligned. The bigger hole is for the Mower motor's head. Power supply wires are connected to the mower by this hole. The smaller holes are for the screws.

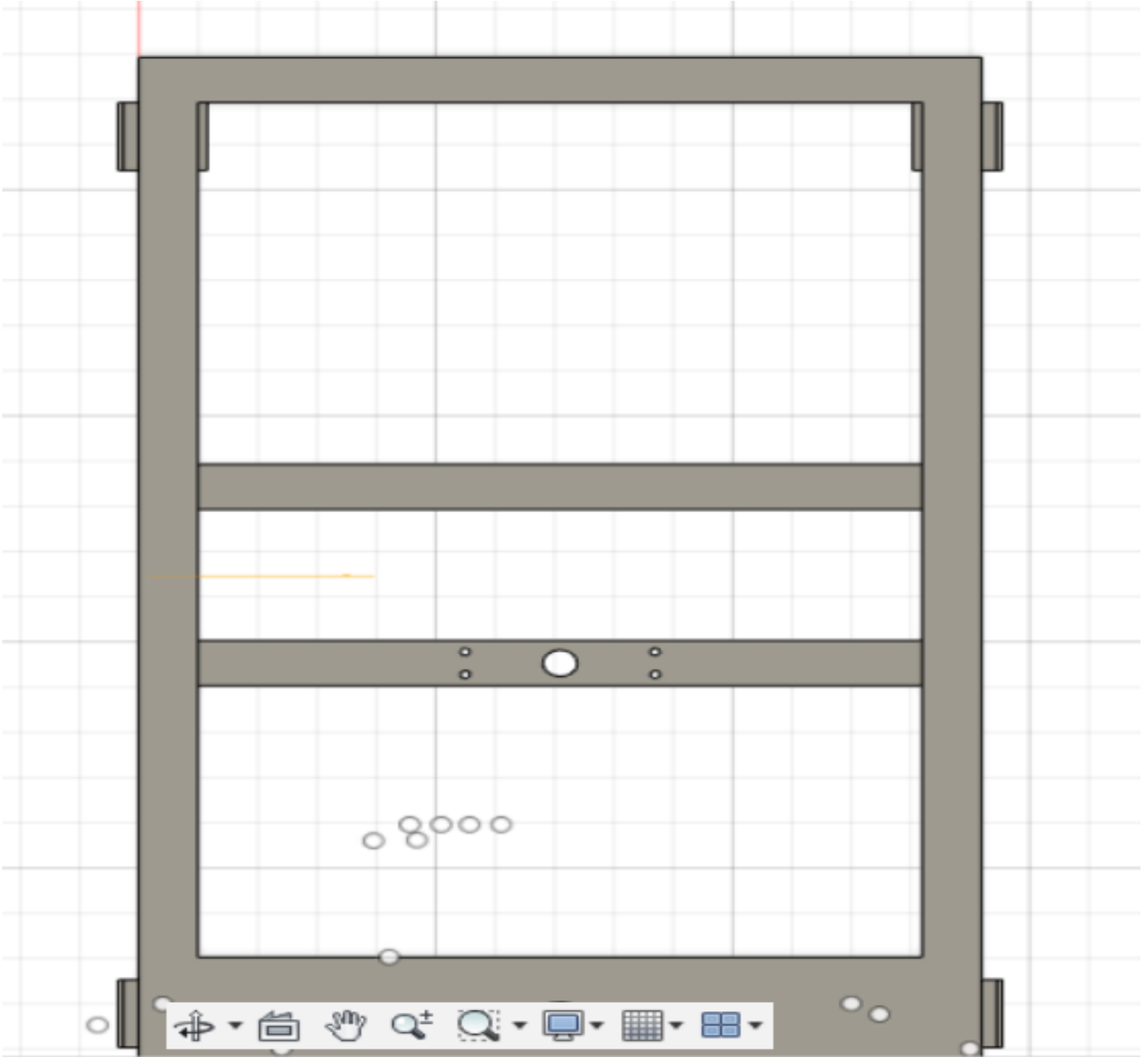

*Fig 5.13: Holes on the support rod to hold Mower*

Then 8mm screw nut combinations on either side of the U bracket are fastened to one of the front support rod of the chassis as shown in the figure.

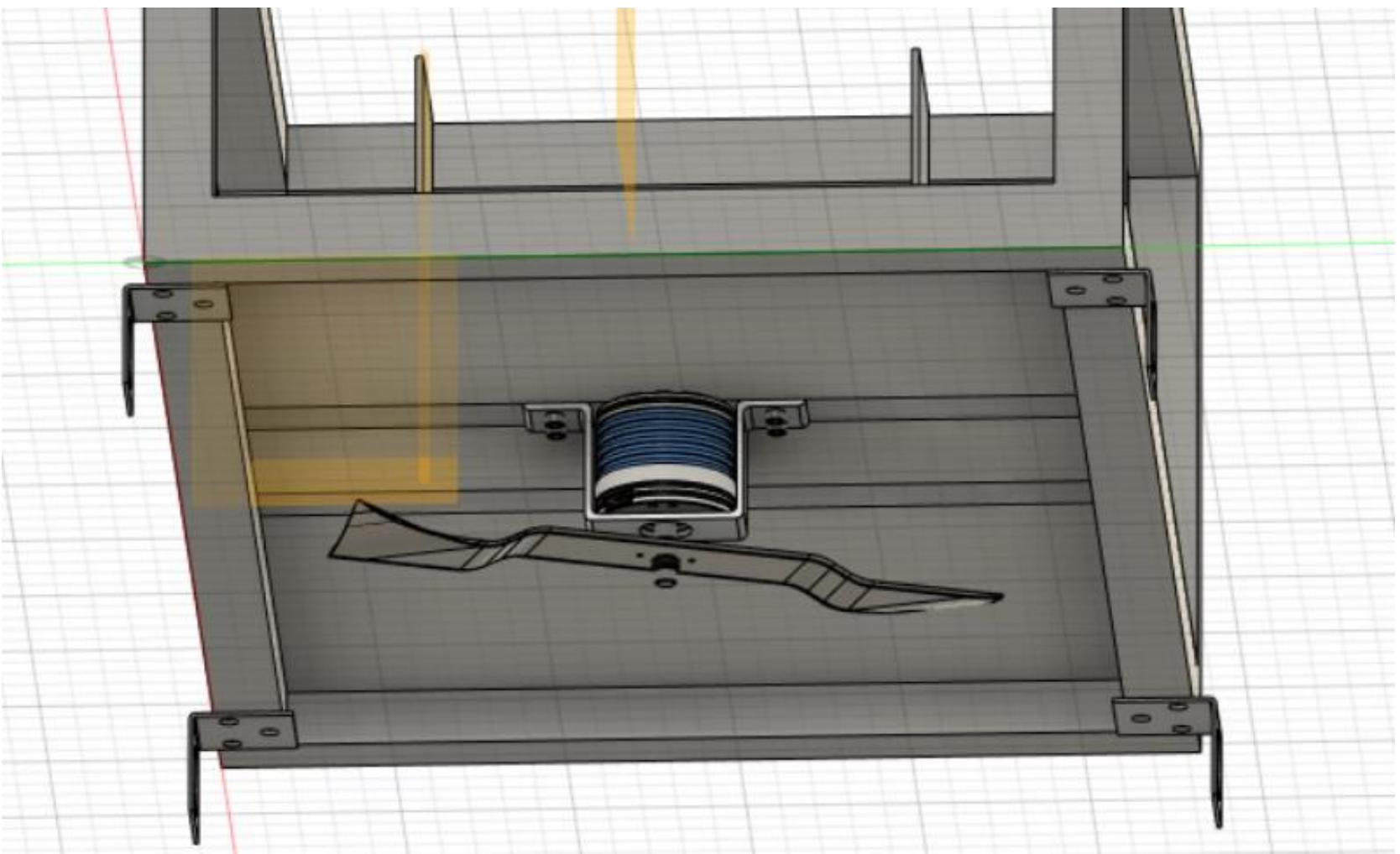

*Fig 5.14: Mower fixed to the support rod*

***5.9: Horizontal actuators***

The horizontal actuators with 100mm actuation in each are positioned as shown in the figure.

### *5.9.1 Specs of actuator*

100:1 12V PLC/RC Control
Peak Power Point: 23 N @ 6 mm/s
Max Speed (no load): 12 mm/s
Input Voltage: 12V
Stall Current: 450 mA at 5 V & 6 V, 200 mA at 12 V
Stroke: 100mm

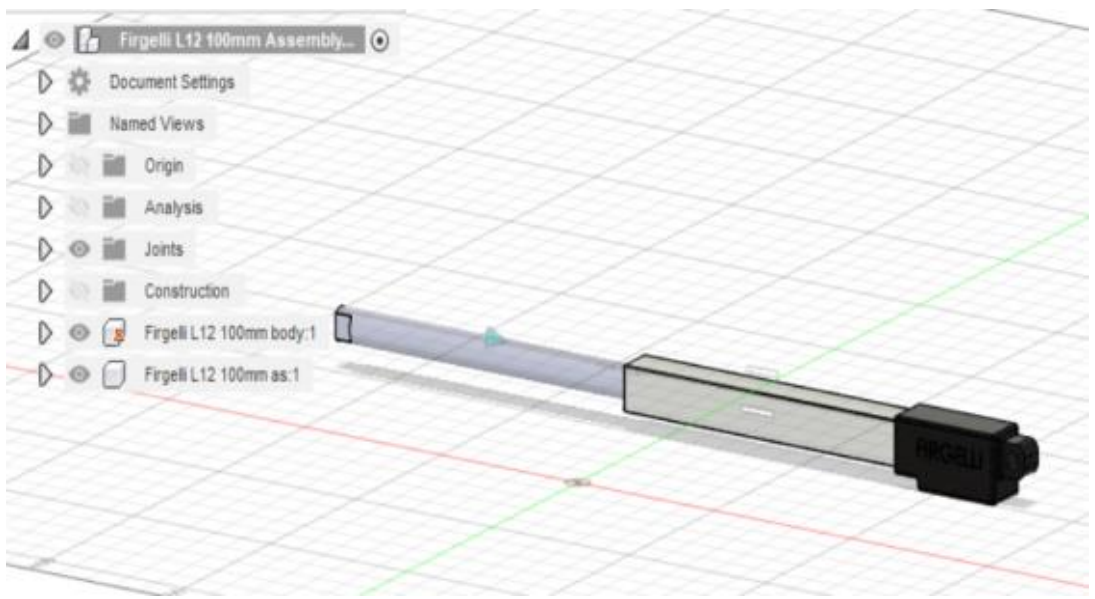

*Fig 5.15: Fully stretched actuator (100mm)*

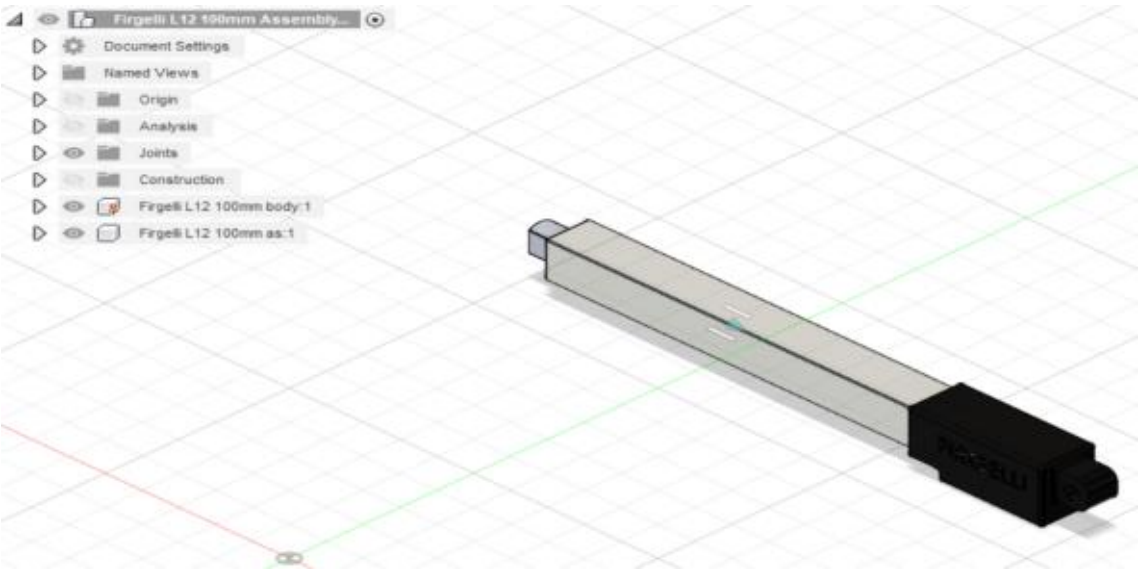

*Fig 5.16: Zero position of the actuator*

The setup for holding the actuators at an angle of 180 degrees to each other is designed as shown in the figure.

### *5.9.2 arrangement of actuator*

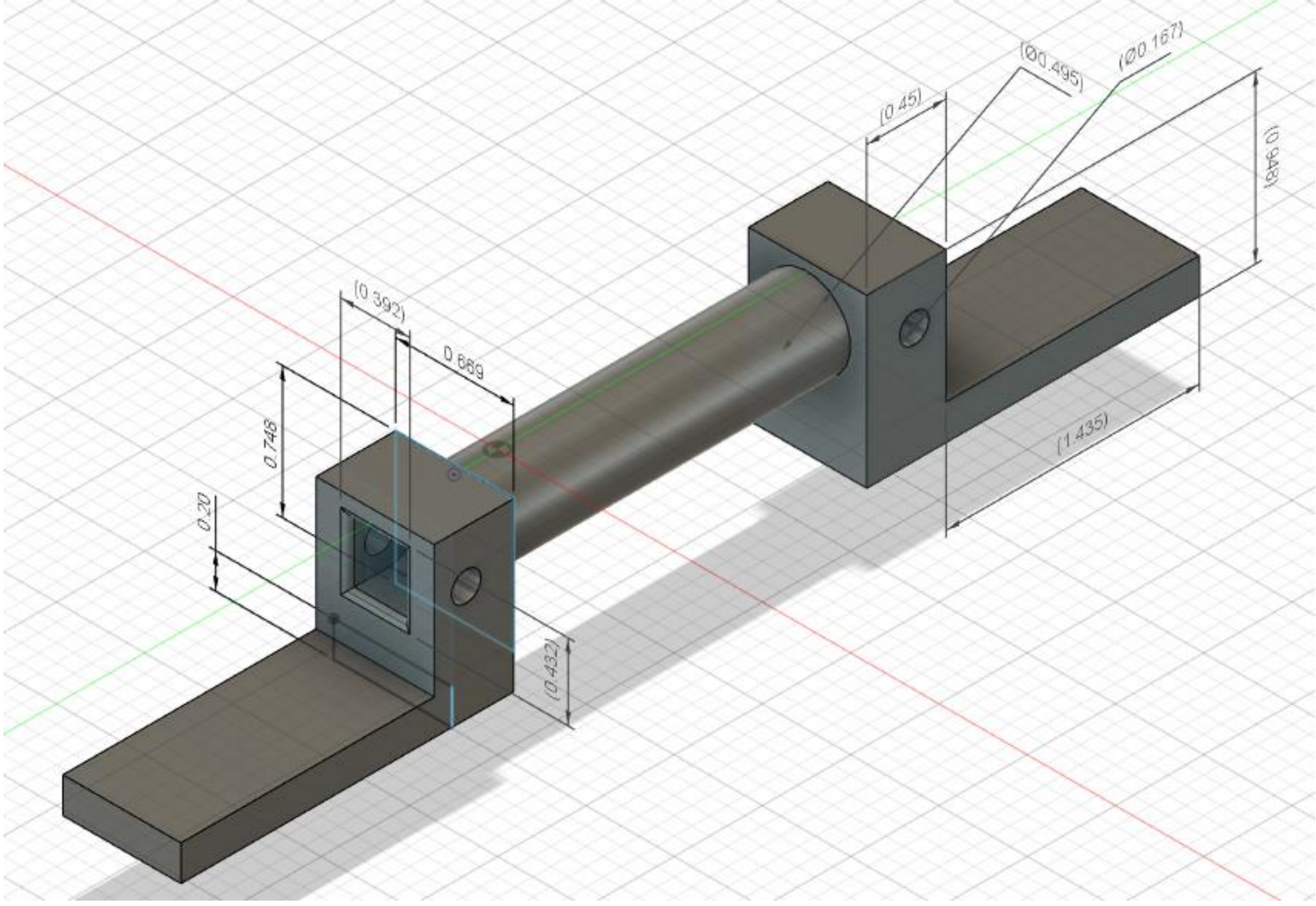

*Fig 5.17: Horizontal actuator holder design with dimensions*

Actuators of small size are selected for sideways actuation as these actuators need not carry any appreciable loads. The sideways actuation is only required to bear the minimum load of pipe and nozzle. This setup is designed in such a way that the Zero position of the actuators is in line with the edges of the chassis. Hence the actuation occurs outside the chassis area. This setup ensures that the pesticide is not sprayed on the chassis and the components inside it.

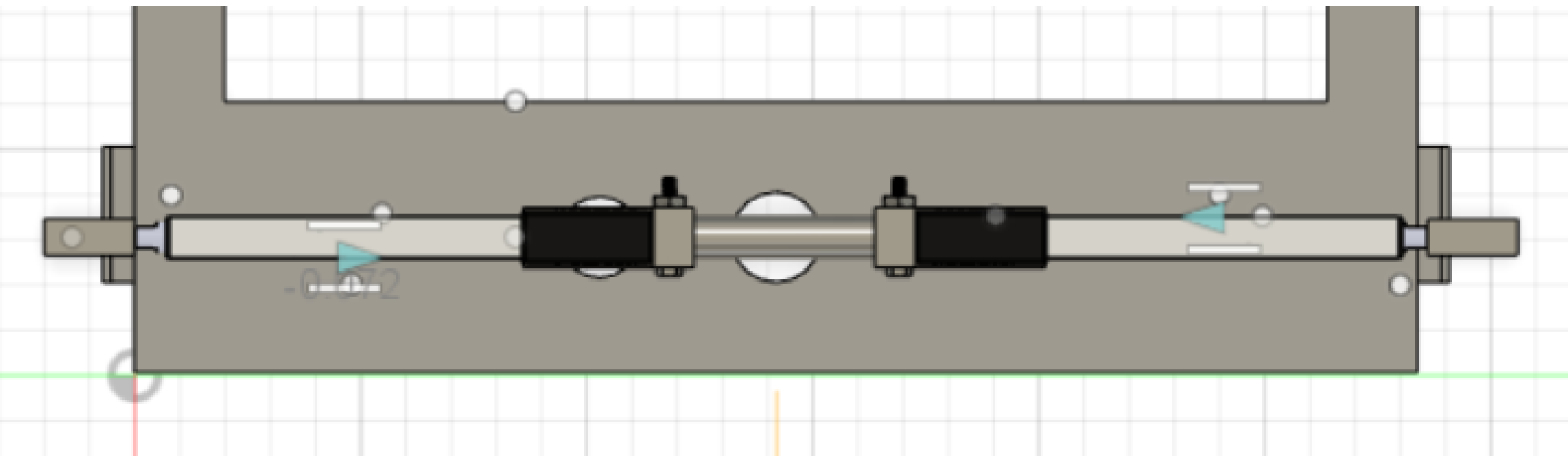

*Fig 5.18: Top view to represent the Zero position of Horizontal actuators*

### ***5.10: Vertical actuator***

Vertical Actuator is a linear actuator with an actuation of 20 inches, as shown in the figure.

*5.10.1 Specs of actuator*

| | | | |
|---|---|---|---|
| Input voltage | : 12 V | Force | : 35 lbs |
| Stroke | : 2 inches | Weight | : 2.1 lbs |

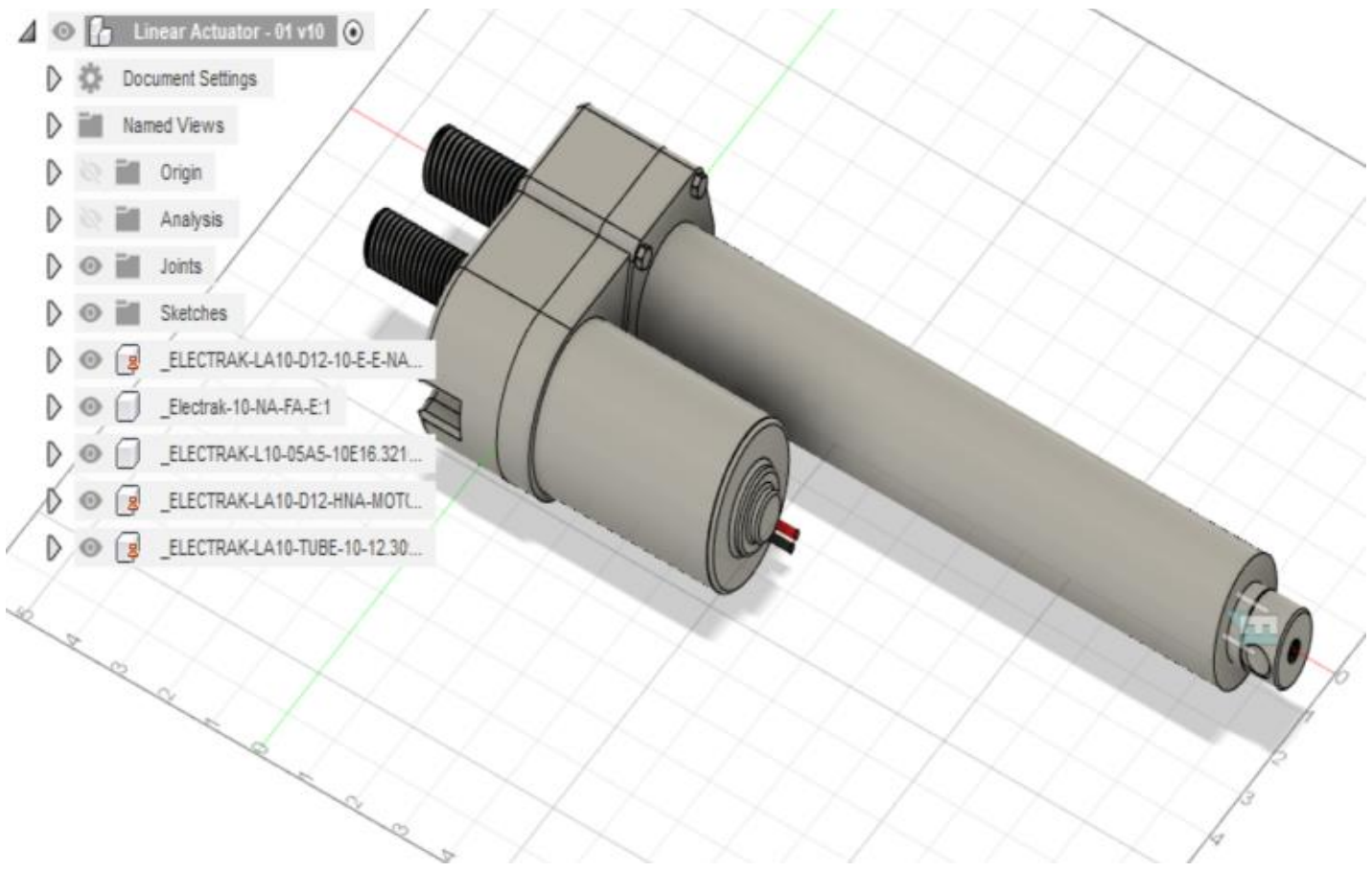

*Fig 5.19: Zero position of the actuator*

5.10.1 arrangement of actuator

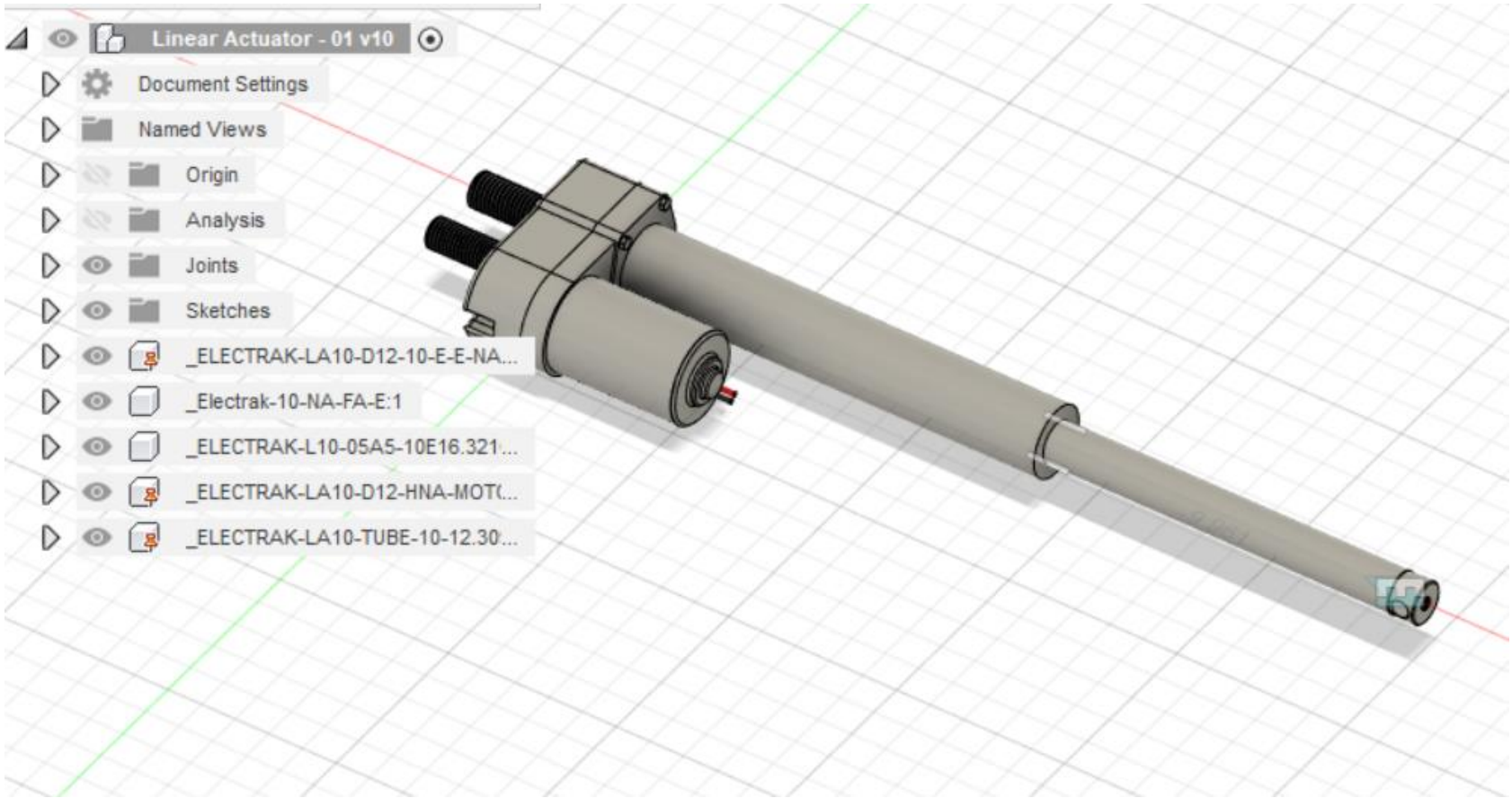

*Fig 5.20: Fully stretched actuator*

This actuator is placed at the center of the breadth which is in forward direction of the chassis. This is done by making two holes on the chassis as shown in the figure. Make sure that the holes are of the same dimension as the two screws fixed to the vertical actuator as shown in the above figure.

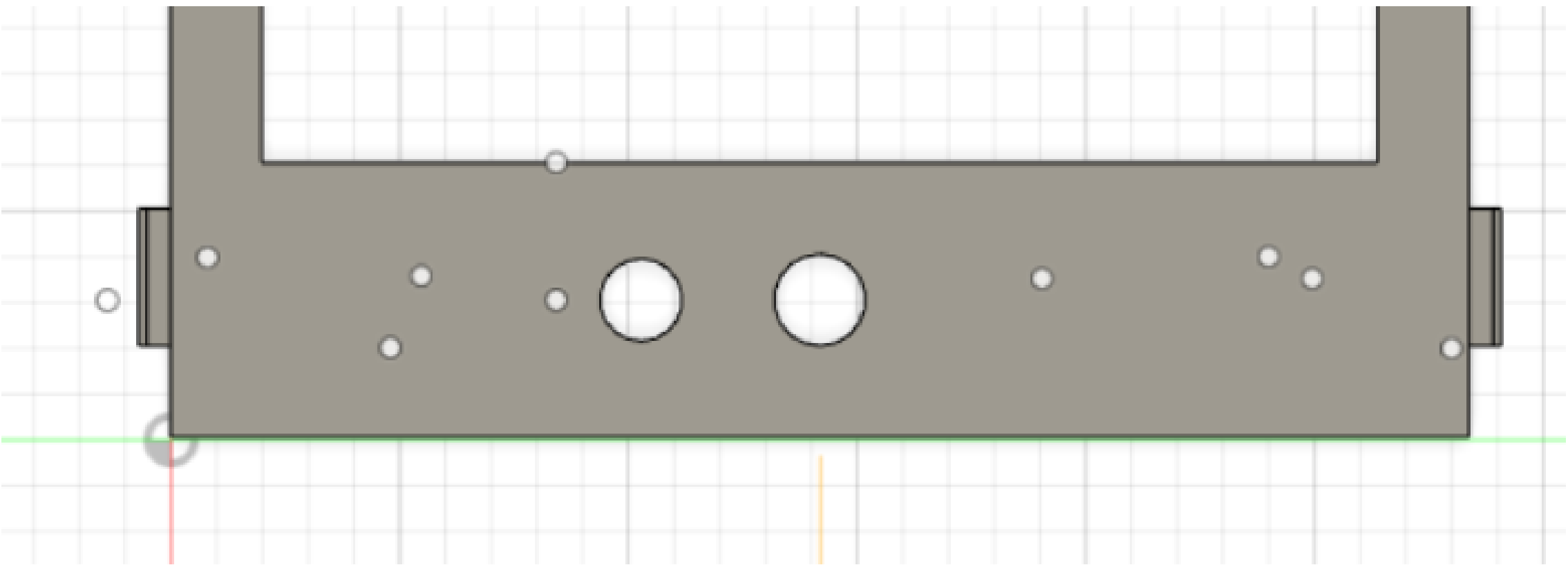

*Fig 5.21: Top view representing the holes required to hold the vertical actuator*

After placing the screws of the actuator in the above given holes. Four nuts that mate perfectly with the screws are chosen to fasten the screws and restrict any unnecessary movement of the Vertical actuator.

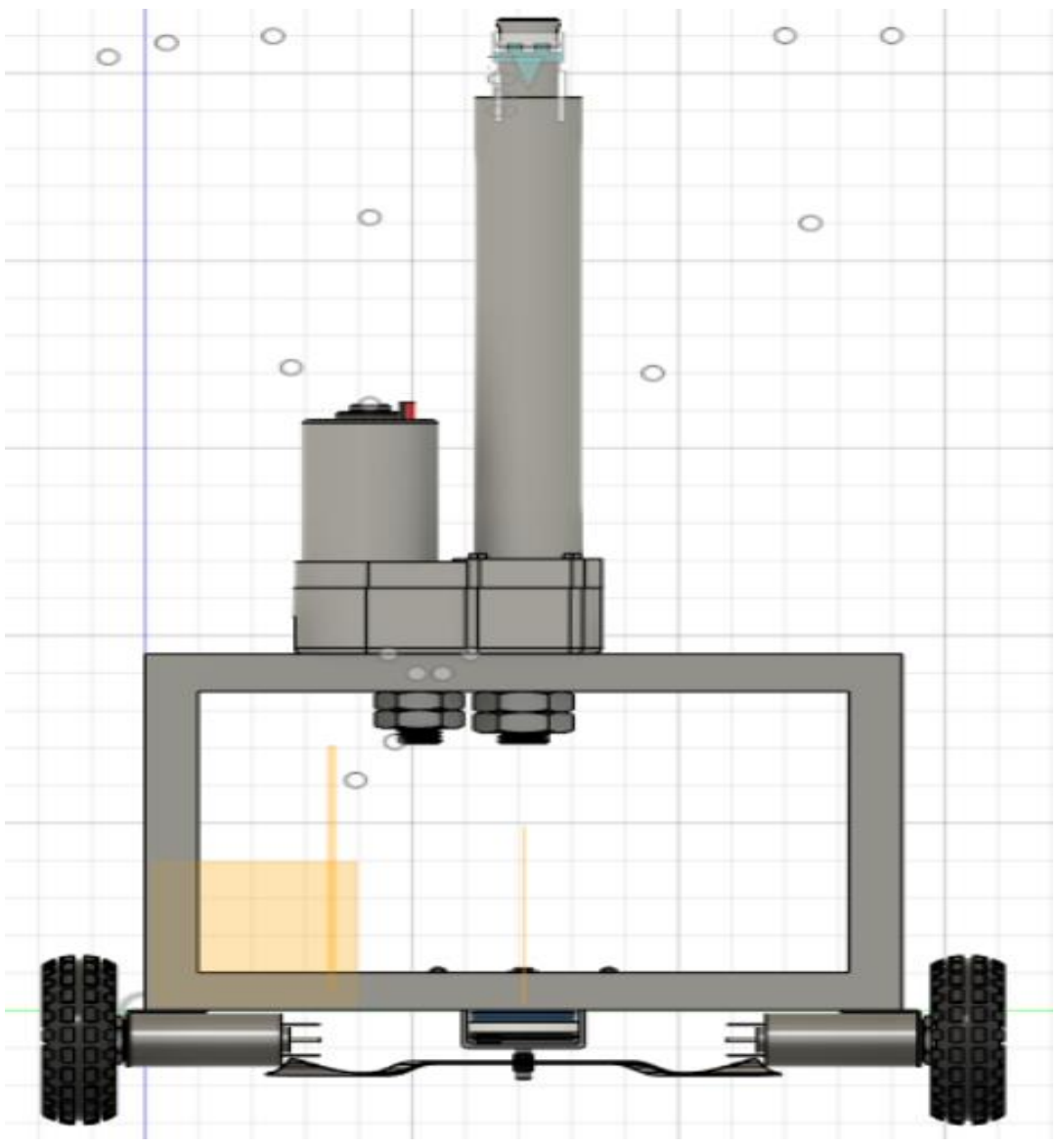

*Fig 5.22: Front view representing the height of vertical actuator*

Comparatively larger size actuator is chosen for this application as the slider of this actuator has to hold the entire horizontal actuator setup at its tip.

### *5.11: Making a joint between horizontal actuator setup and vertical actuator*

This is done by sliding the horizontal actuator setup inside the slot given at the tip of the slider of vertical actuator.

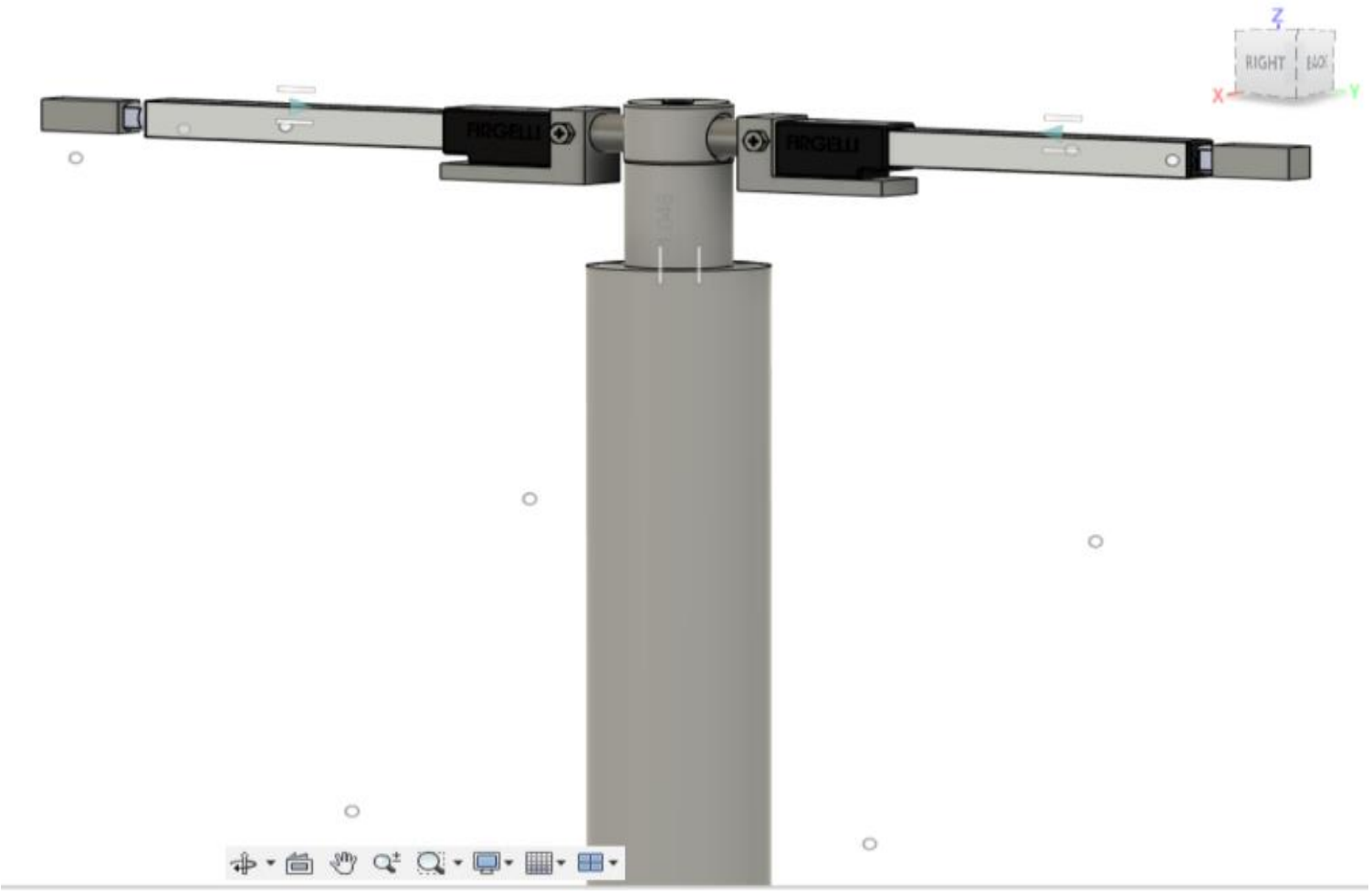

*Fig 5.23: T joint between actuators without holding screw*

To restrict the sliding motion of the horizontal actuator setup, make a threaded hole on the top of the slider and fasten a screw till its tight, with the horizontal actuator setup inside the slot of the slider of the vertical actuator as shown in the figure.

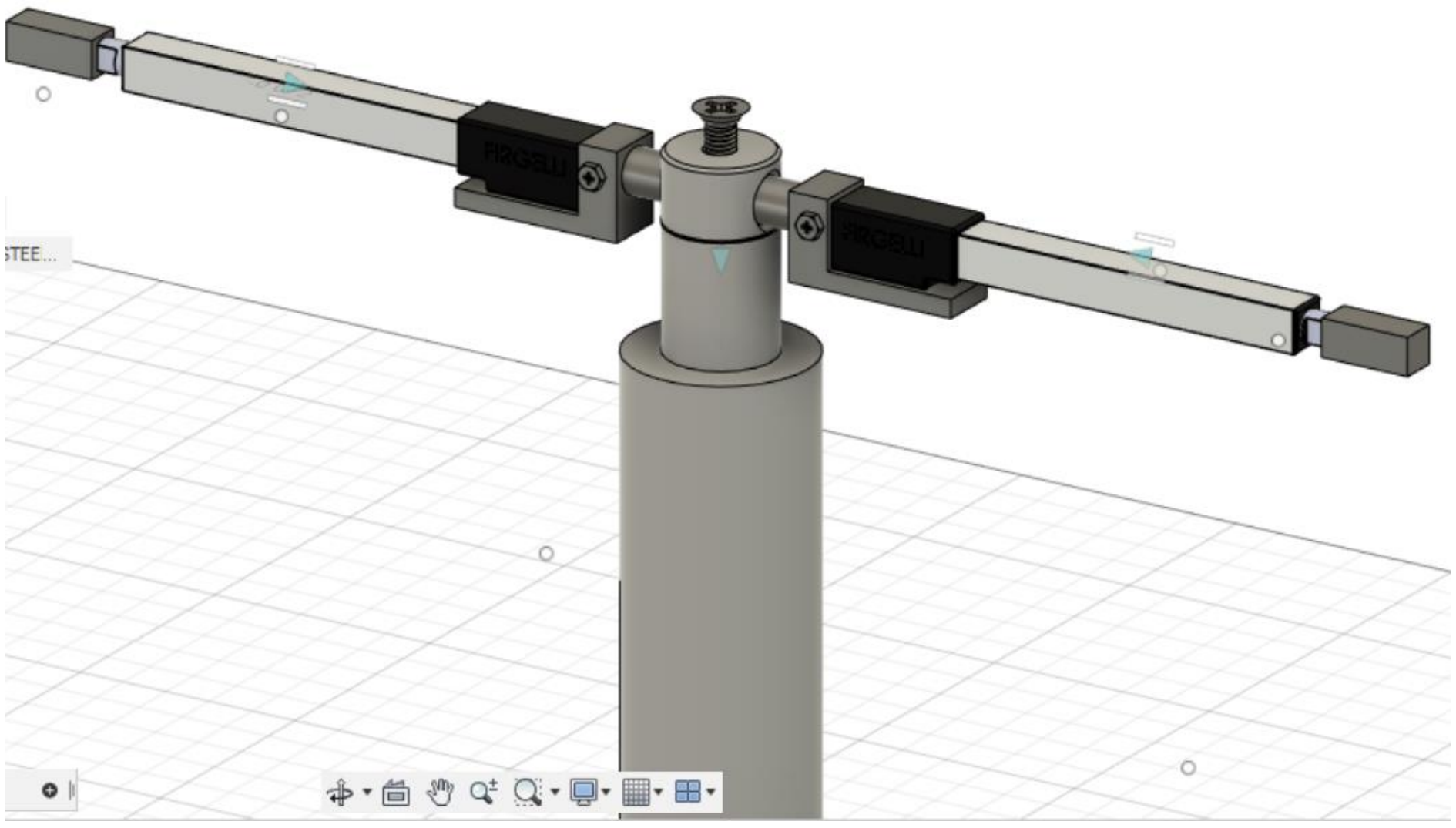

*Fig 5.24: T joint between actuators with the holding screw*

The above forms the complete Actuation setup with 2 Degrees of Freedom.(viz vertical sliding motion and sideways sliding motion)

### *5.12: Storage tank for pesticide*

Of the three spaces provided by the two partition rods placed at the bottom of the chassis, the one in the front (i.e., closest to the Actuation setup) is chosen as the place for placing the storage tank. By doing this we reduce the length of pipes to be used and prevent any leakages. The storage tank is designed in such a way that it fits exactly in the base. And the material chosen is plastic so as to reduce the weight of the setup. The storage tank is to have 2 holes, one for connecting pump to the pesticide in the tank and the other for filling the tank with pesticide. For ease of filling a funnel is placed above the filling hole. The storage tank is made by using the “shell” command and filleting the corners for proper stress distribution.

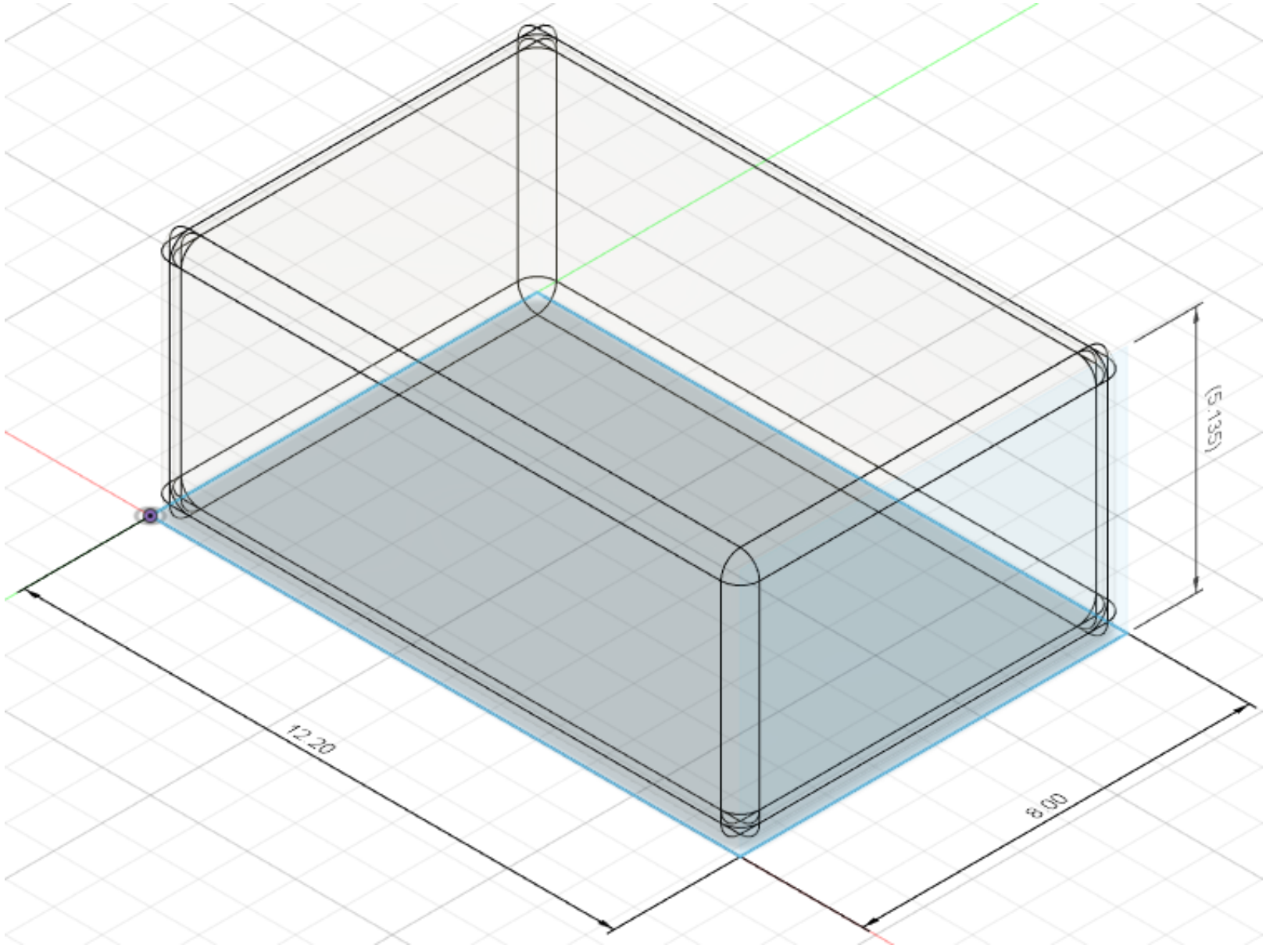

*Fig 5.25: Storage tank with dimensions*

***5.13: Pump***

A 12V DC pump is to be used for pumping the pesticide from storage tank to the Nozzles.

*5.13.1 The specs of pump :*

| | |
|---|---|
| Voltage rating | 12 V DC |
| Max current rating | 1000 mA |
| Flow rate | 350-700 L/H |
| Inlet dia | 0.354 in |
| Outlet dia | 0.295 in |

The pump is placed directly on the storage tank, the pump we have chosen has 3 suction cups which will stabilize the pump. The chosen pump is of centrifugal type as here the pesticide has to be pumped upwards (against gravity) at high speeds. The pump has to lift the pesticide to almost 30 inches.

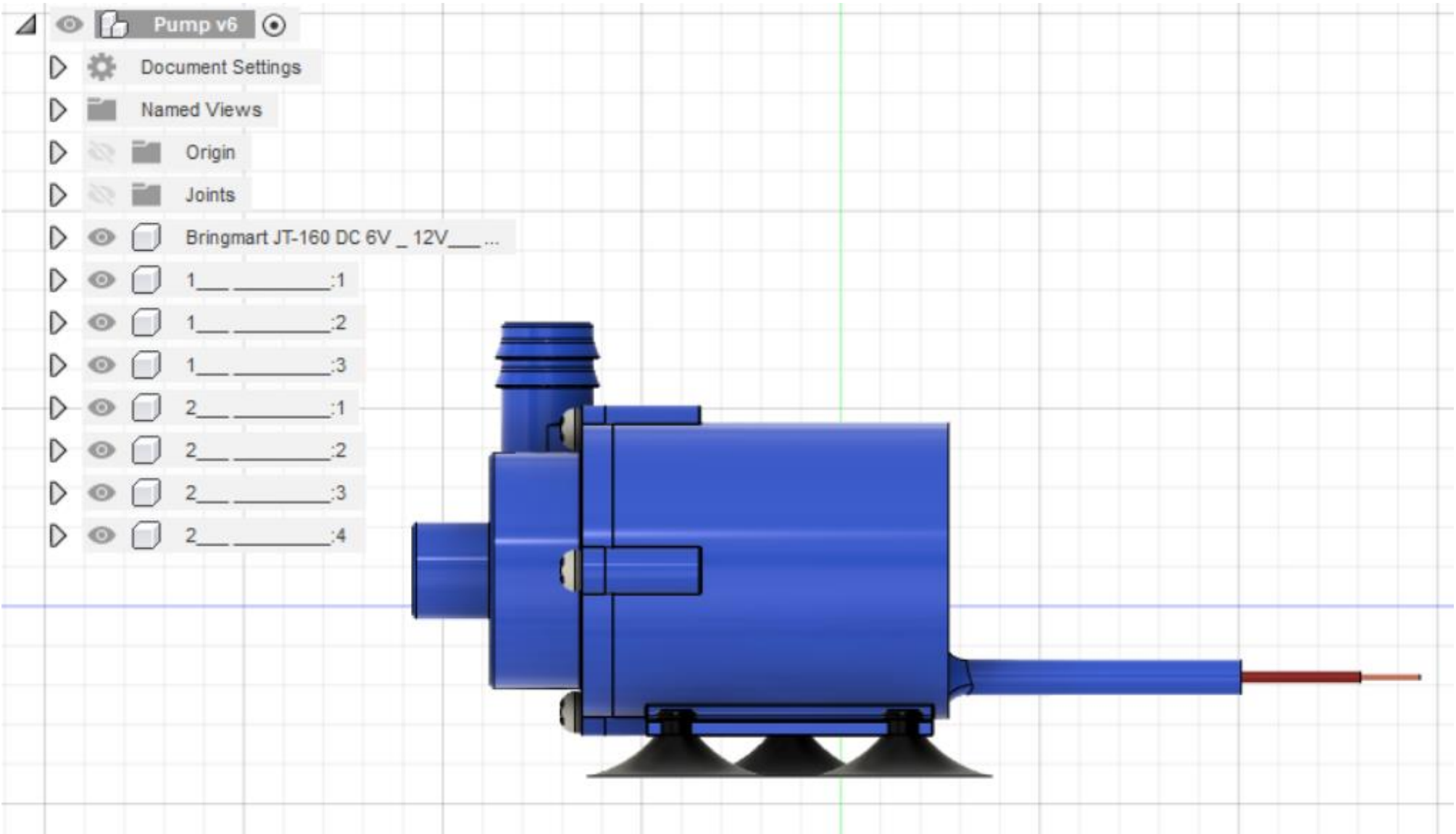

*Fig 5.26: 12 V DC Pump*

***5.14: Connecting pump and nozzles***

Since the Actuation setup is in the form of Alphabet T and the nozzles are at the end of the horizontal, we designed a Y joint to split the water flow to two Nozzles as shown in the figure.

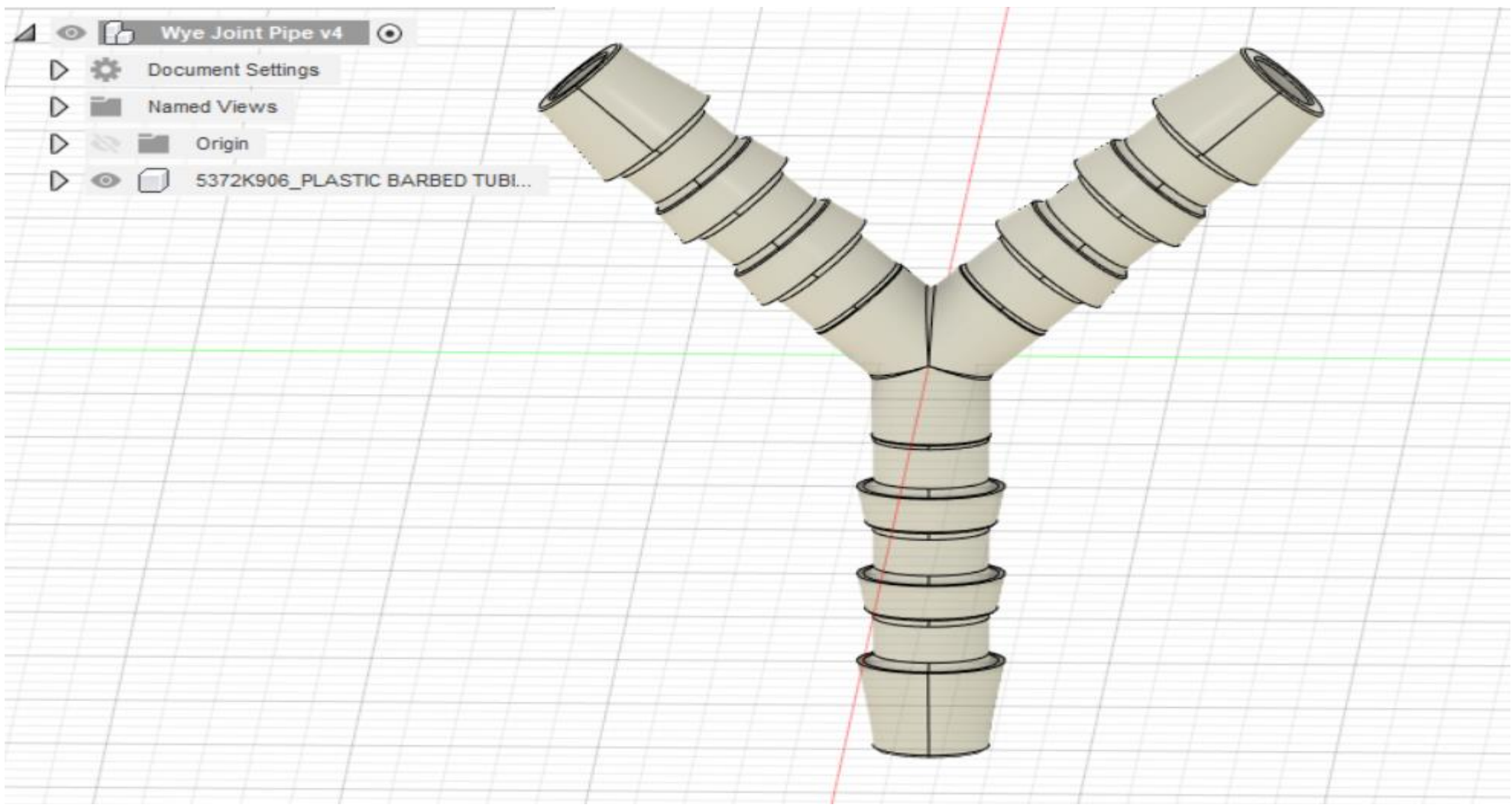

*Fig 5.27: Y Joint*

This Y joint is placed at the outlet of the pump as shown in the figure.

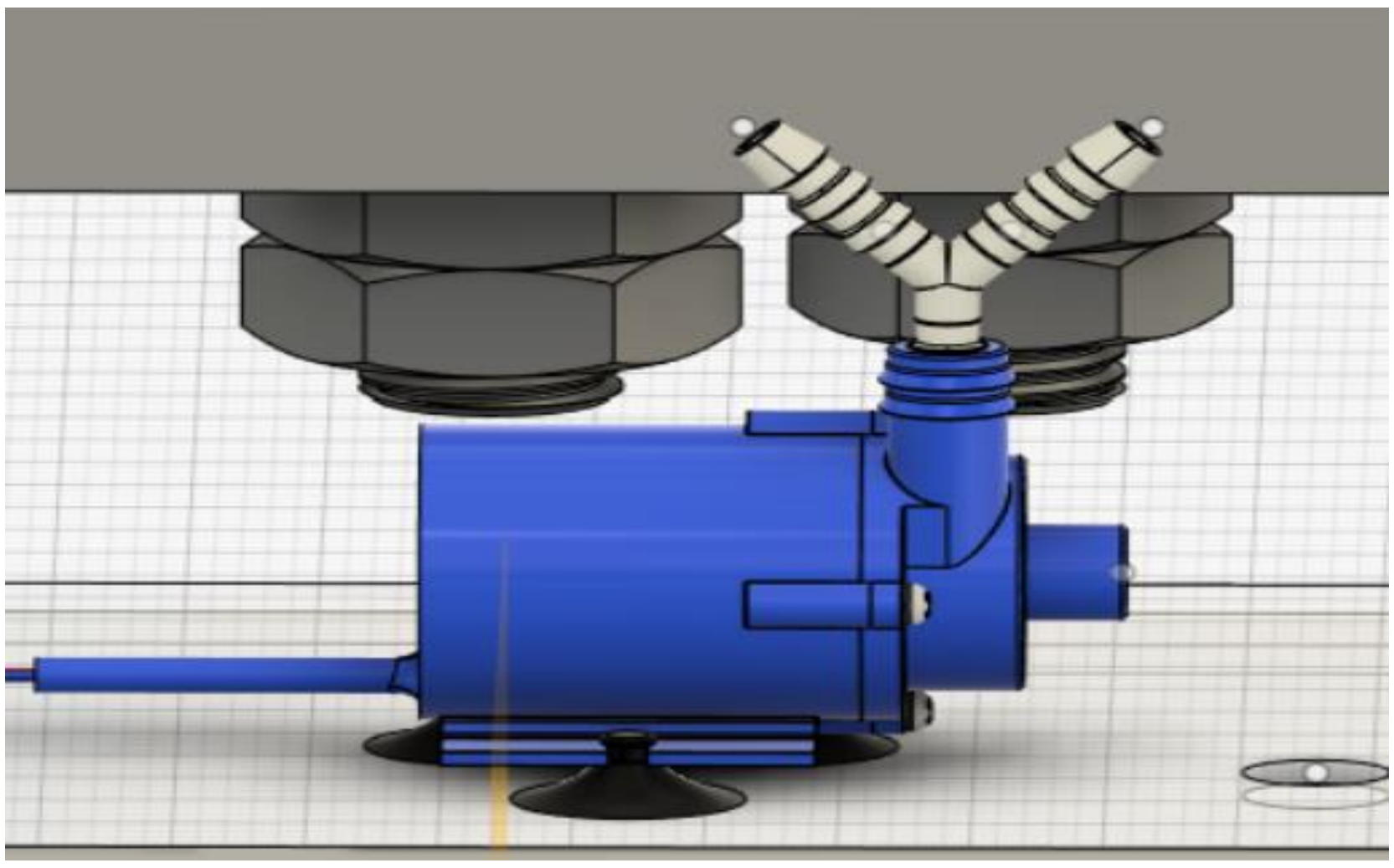

*Fig 5.28: Placing Y joint at the outlet of pump*

<u>*5.14.1 Nozzle placement:*</u> The design of the Nozzle is as shown in the figure below. Two of these are placed at the ends of the slider of horizontal actuators.

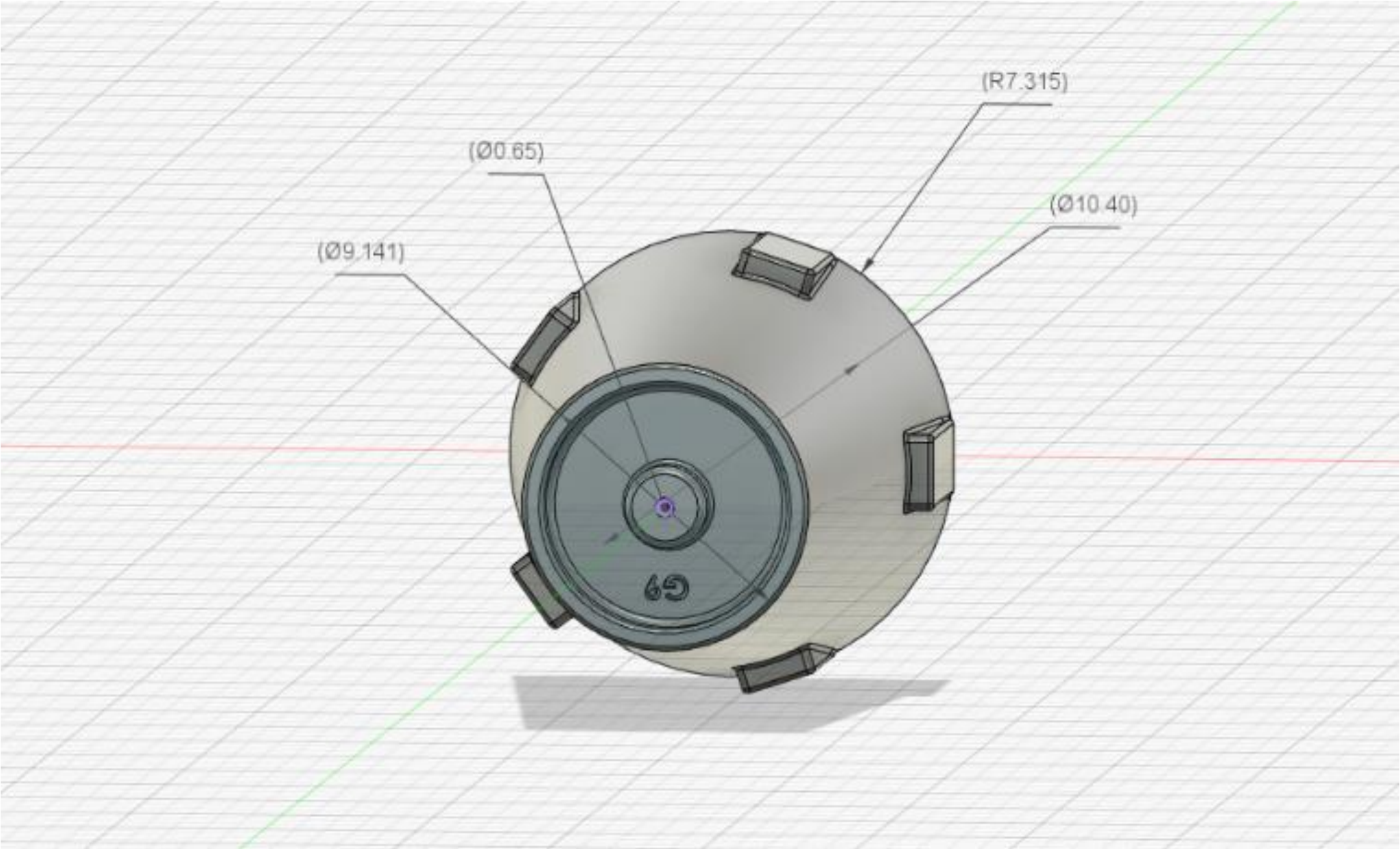

*Fig 5.29: Nozzle design with dimensions*

The nozzles are attached each at the ends of the two horizontal actuators and a rigid joint is made btw the end and the nozzle as shown in the figure.

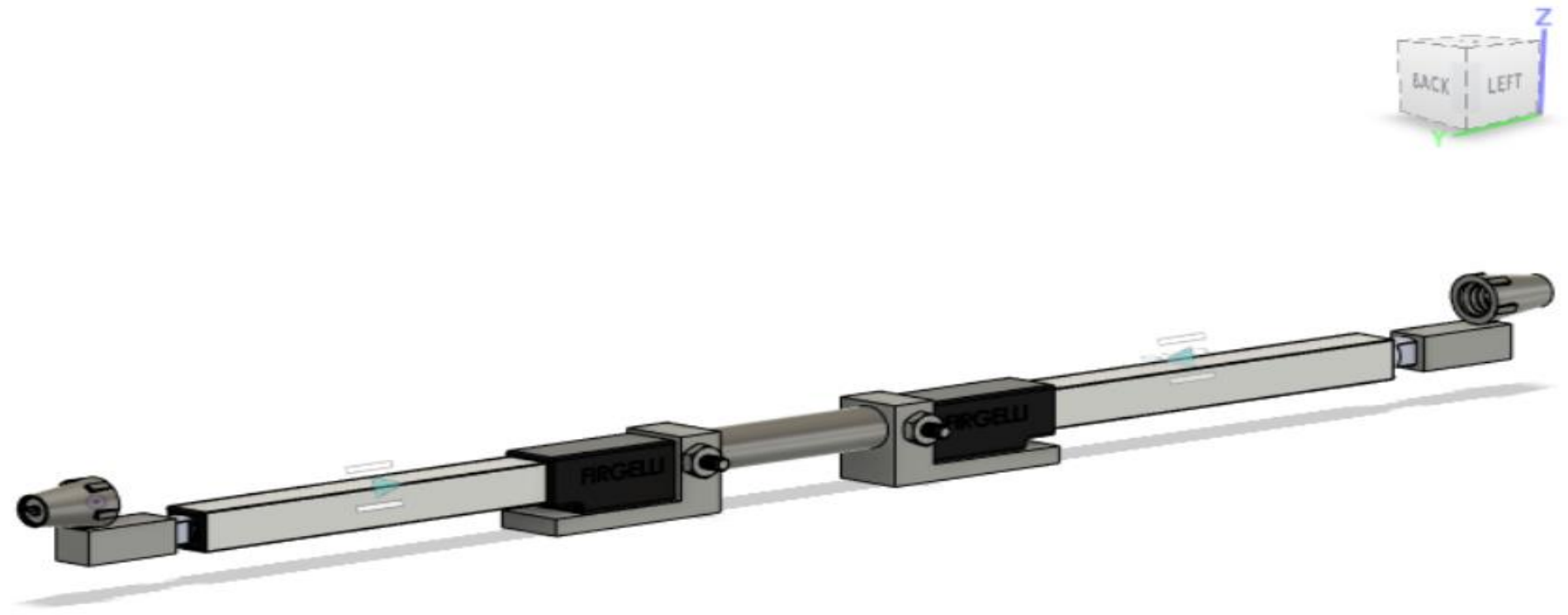

*Fig 5.30: Placing the nozzle at the tip of horizontal actuators*

Now to connect the Y joint to the Nozzles we designed 4 sets of pipes of which are in the form of spring coils to accommodate the vertical and horizontal actuations. Out of the 4 spring coil pipes, two are horizontal, which are placed on the either side of the horizontal actuator setup and two are vertical, which are placed on the either side of the vertical actuator as shown in the figure.

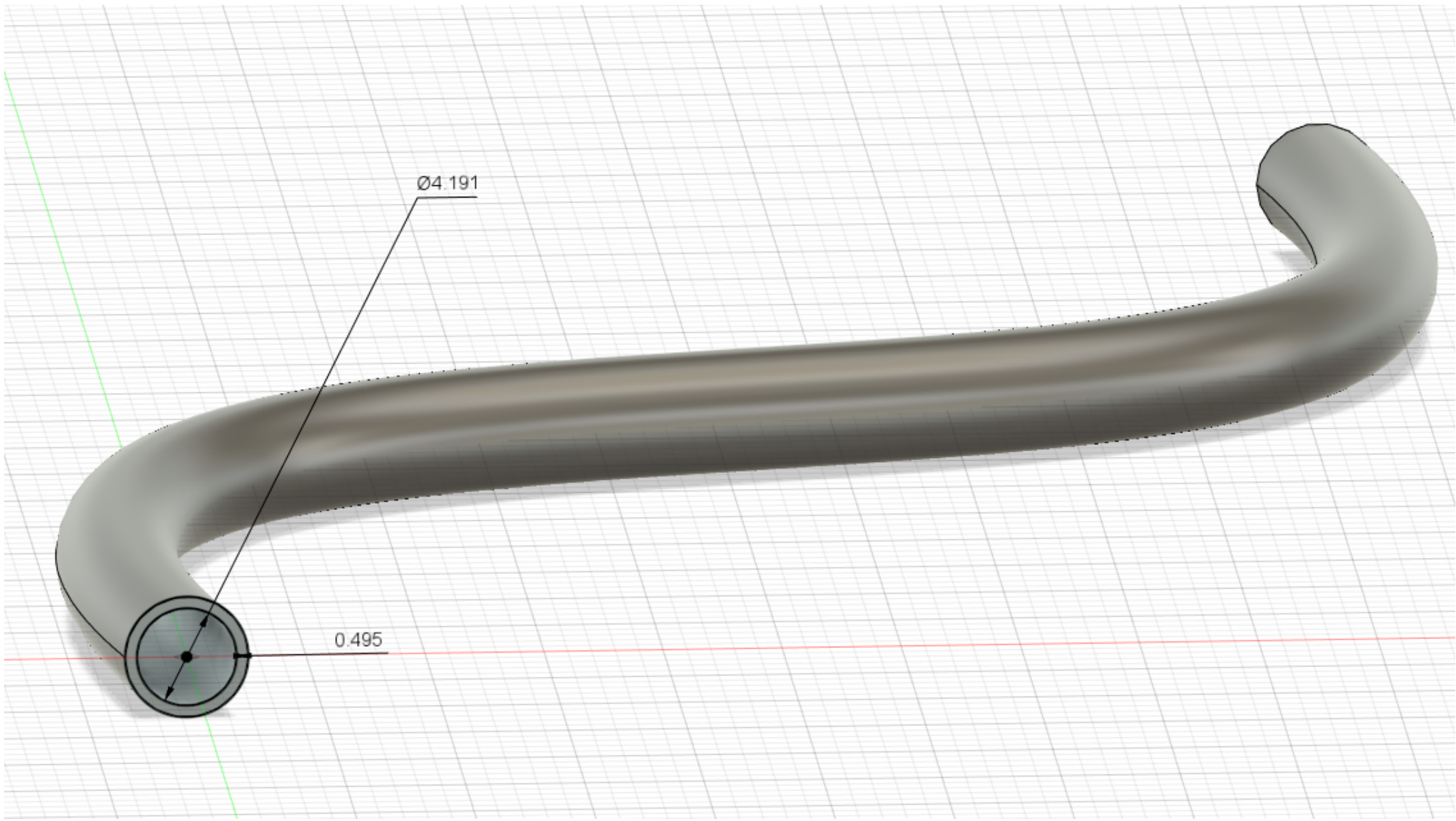

*Fig 5.31: Spring pipe coil design with dimensions*

The two free ends of the two vertical spring coiled pipes are connected to the Y joint.

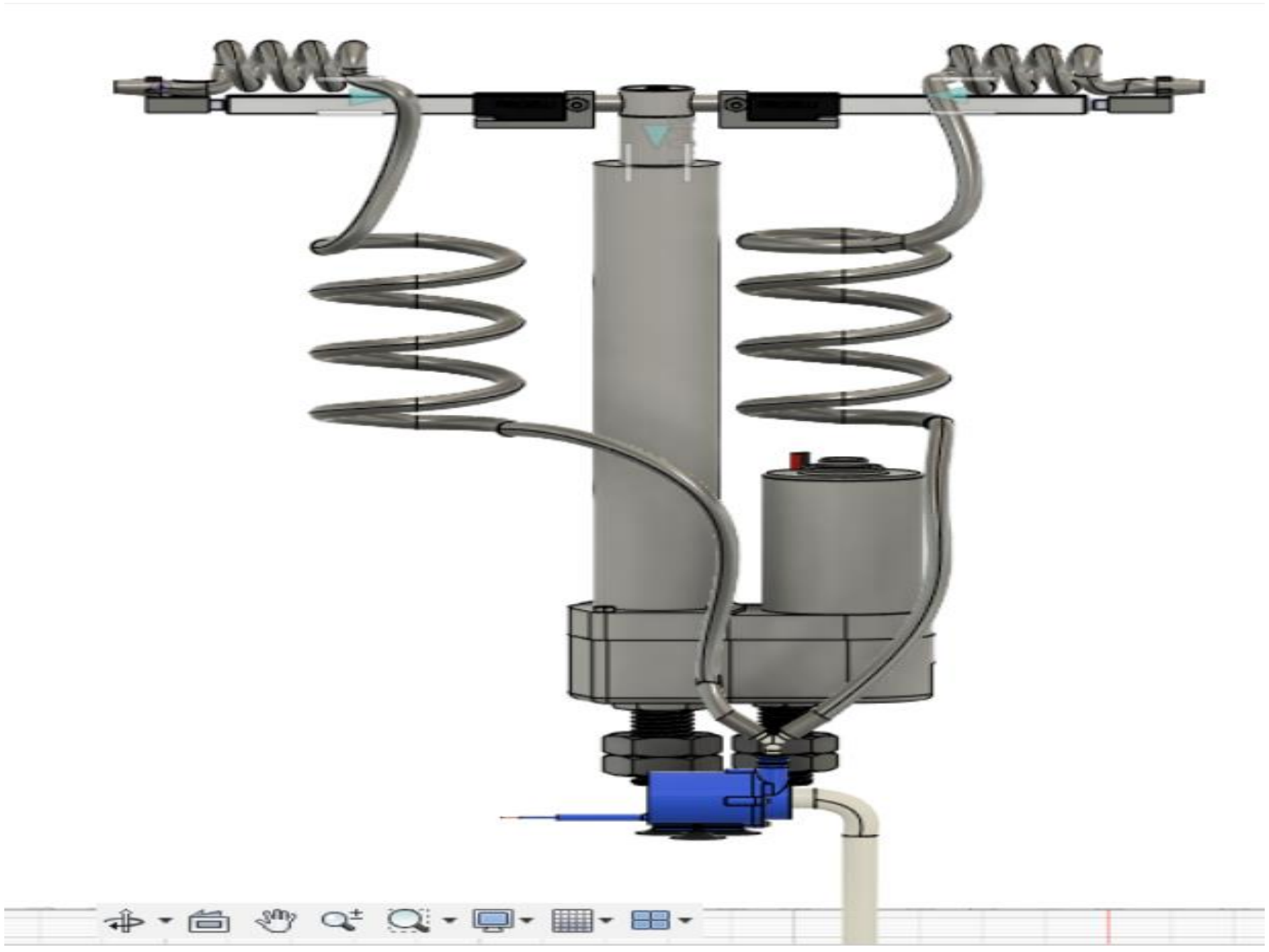

*Fig 5.32: Connection between pump and nozzles through pipes*

### *5.15: Battery*

The selection battery is of paramount importance. The battery has to be selected in such a way that it has the capacity to power all the electronic components viz 4 Drive motors, 1 Mower motor, 2 Horizontal actuators, 1 Vertical Actuator, 1 Pump, and the Arduino – Motor Driver setup. And it has to be rechargeable for uninterrupted low maintenance usage. Considering the above parameters we have selected sealed Lead-Acid Battery of 12V and Maximum current discharge of 1.1 Amps.

#### *5.15.1 Specs of battery:*

| | |
|---|---|
| Voltage | 12 V |
| Battery Capacity | 1.3 Ah |
| External height | 53 mm |
| External Width | 45 mm |
| External Depth | 96.5 mm |

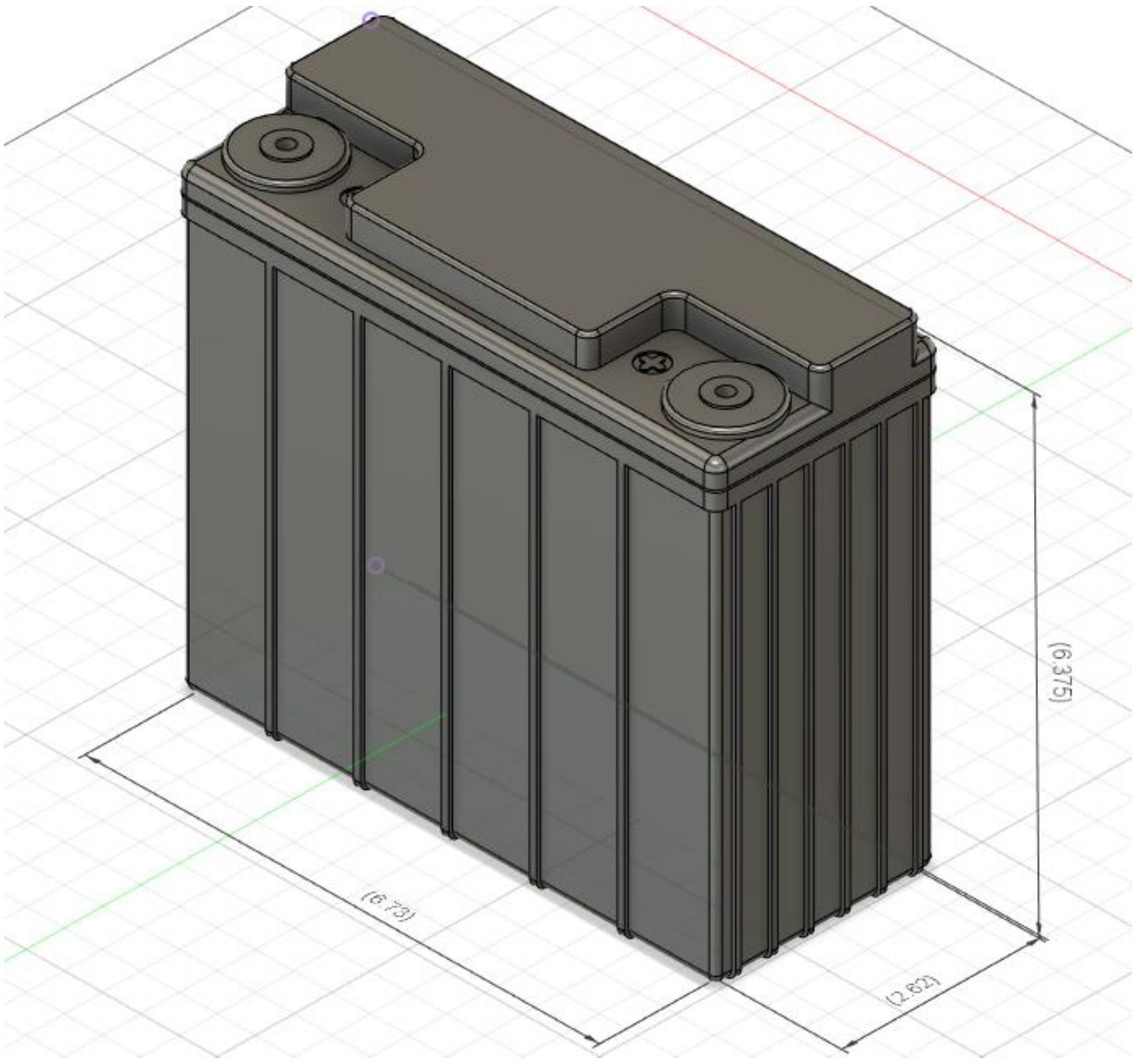

*Fig 5.33: Battery design with dimensions*

<u>*5.15.1 placement of battery:*</u>

The battery is placed in the center of the chassis so that it is accessible to all the electronic components. A Battery holder setup is created at the center of the chassis with a strap as shown in the figure.

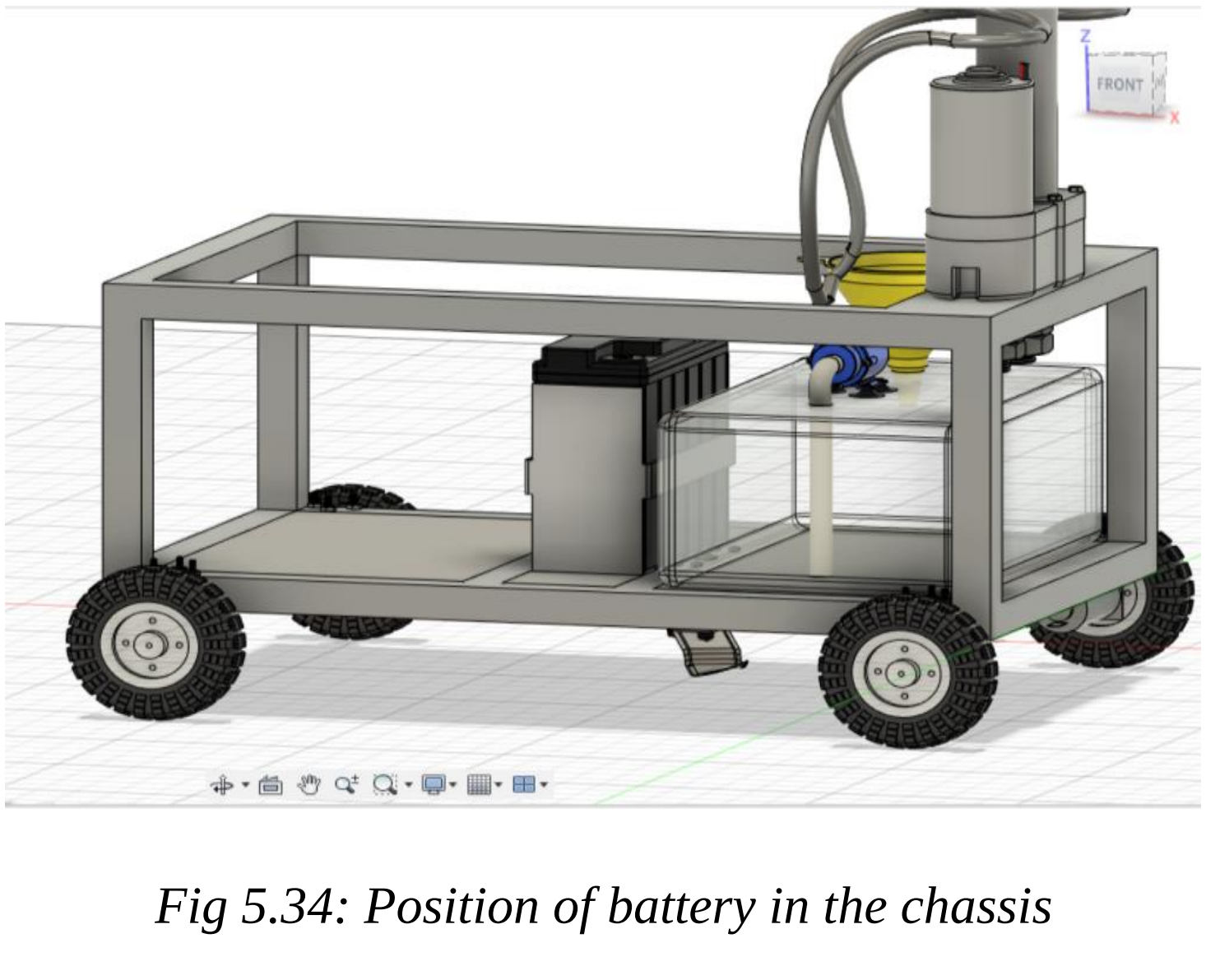

*Fig 5.34: Position of battery in the chassis*

### *5.16: Solar panel*

Solar panel is to be placed on the top frame of the chassis. It is so that the panel receives sunlight at an optimum angle. For places like India the angle at which the solar panel is to be placed is calculated and is placed accordingly.

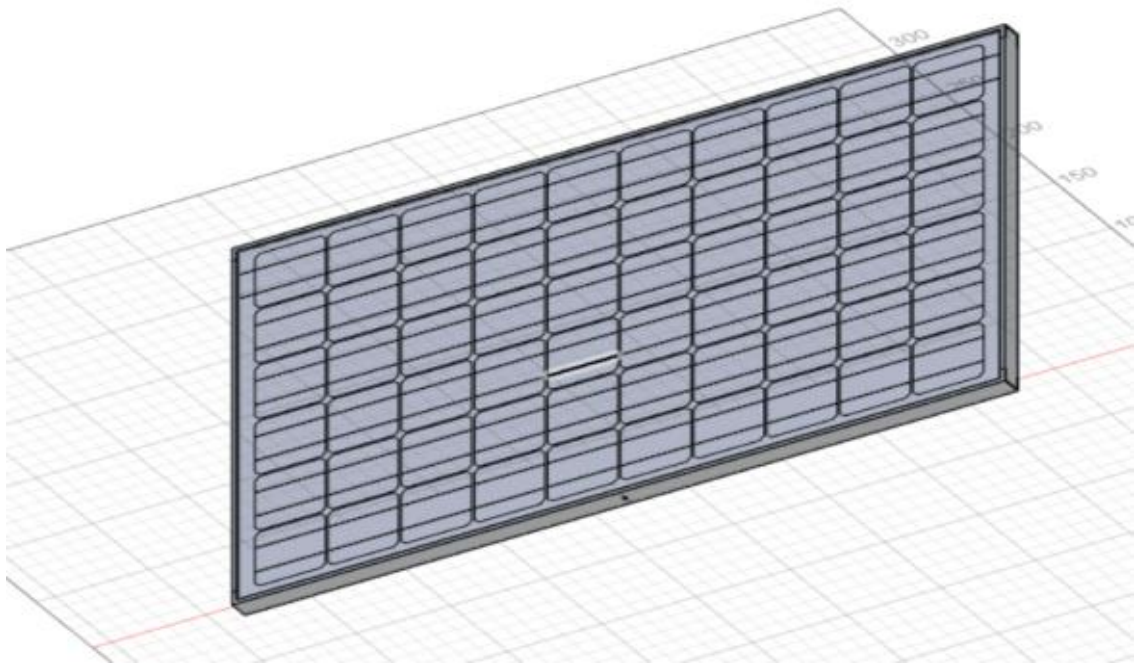

*Fig 5.35: Front – Solar Panel*

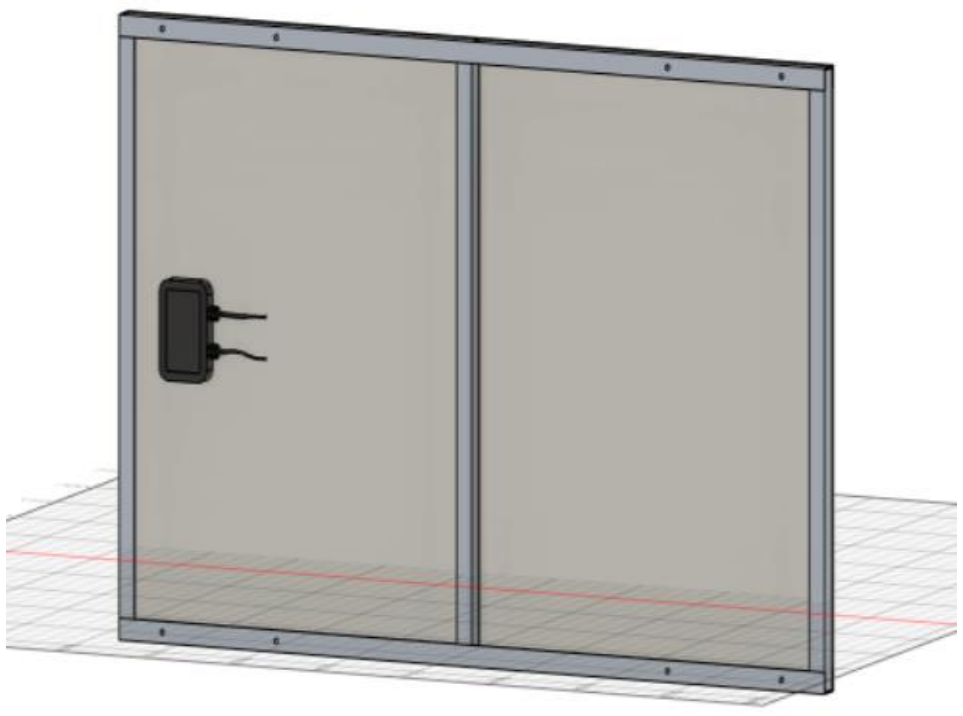

*Fig 5.36: Back – Solar Panel*

For better stability of the solar panel, two supports of length 10 inches are raised at the centre of the longitudinal rods of the top frame of the chassis. A rod perpendicular to the supports is placed to support one edge of the solar panel and the other edge of the solar panel is placed on the frame of the chassis as shown in the figure. The solar panel is hinged on the panel support rod so that the solar panel can be placed at any desired angle. This is because we know that the sun's position varies throughout the day.

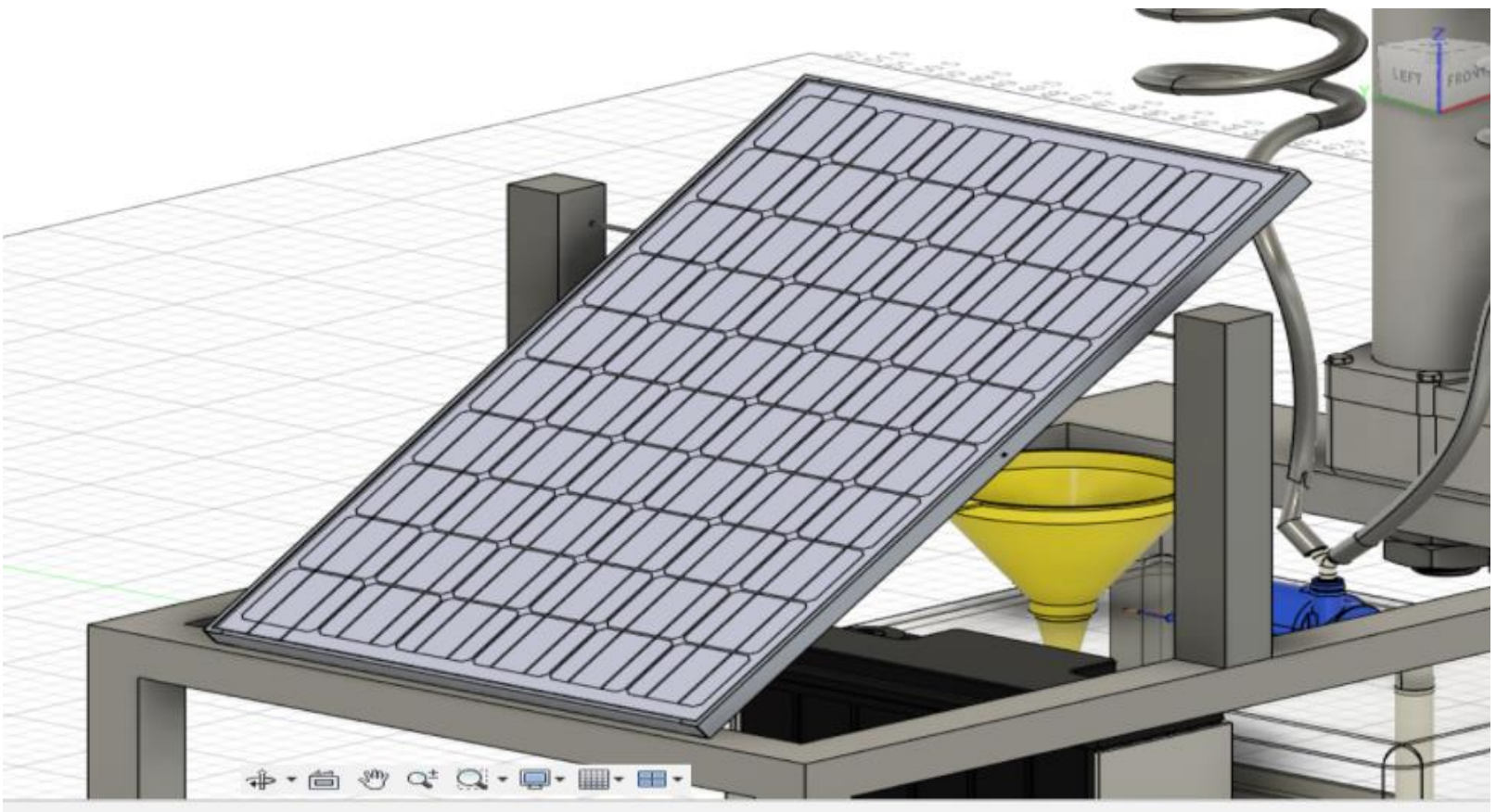

*Fig 5.37: Position of solar panel on the chassis*

Since the battery is rechargeable, it has to be seen that the solar panel recharges the battery whenever the battery discharges completely. Arrangements are made in such a way that the solar panel directly powers the battery when the optimum solar energy is available in the environment. And a Diode is to be placed between battery and Solar panel so that the solar panel doesn't overcharge the battery which may cause damage to battery.

### *5.17: Connecting battery and other electronic components through Arduino*

The connections are made from the battery through a programmed Arduino module to the all other electronic and electrical components. The programmed Arduino acts as the brain of the control system which executes the commands input by the user through a smart phone application using a Bluetooth module and make the electronic and electrical components function accordingly.

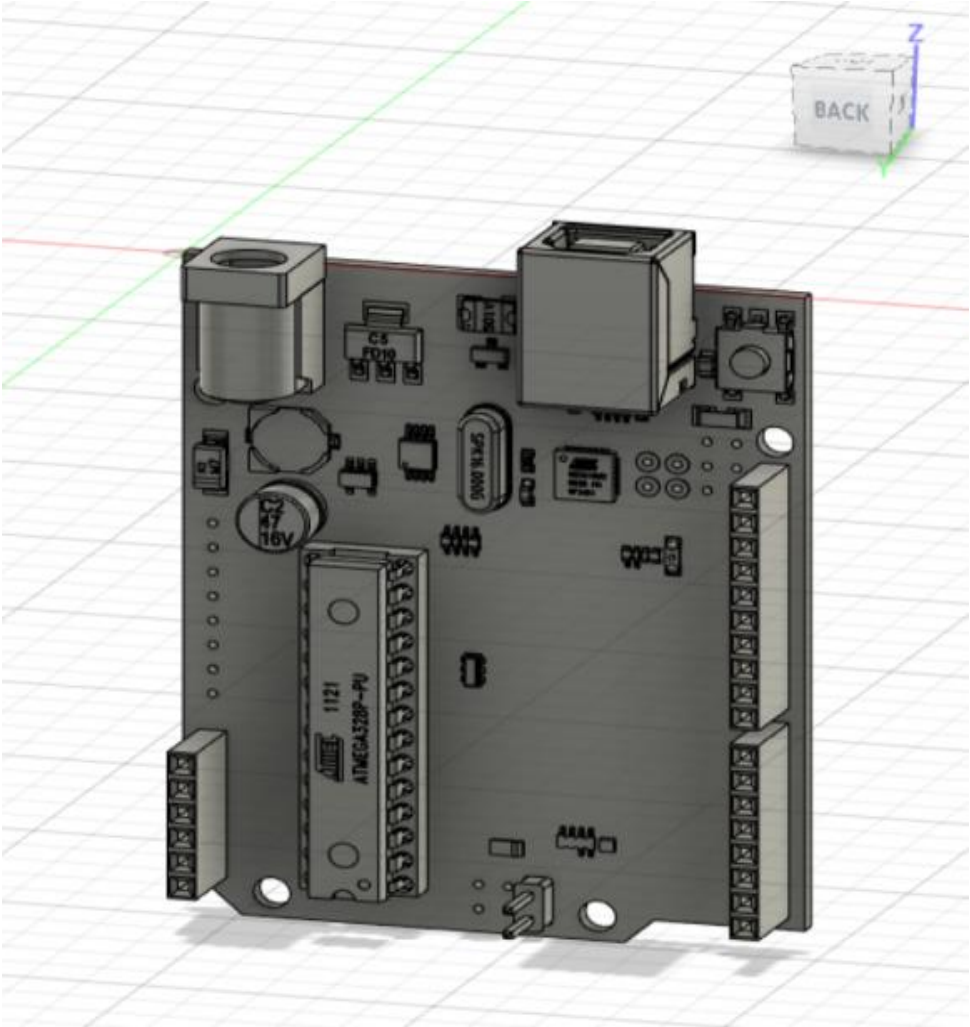

*Fig 5.38: Arduino UNO*

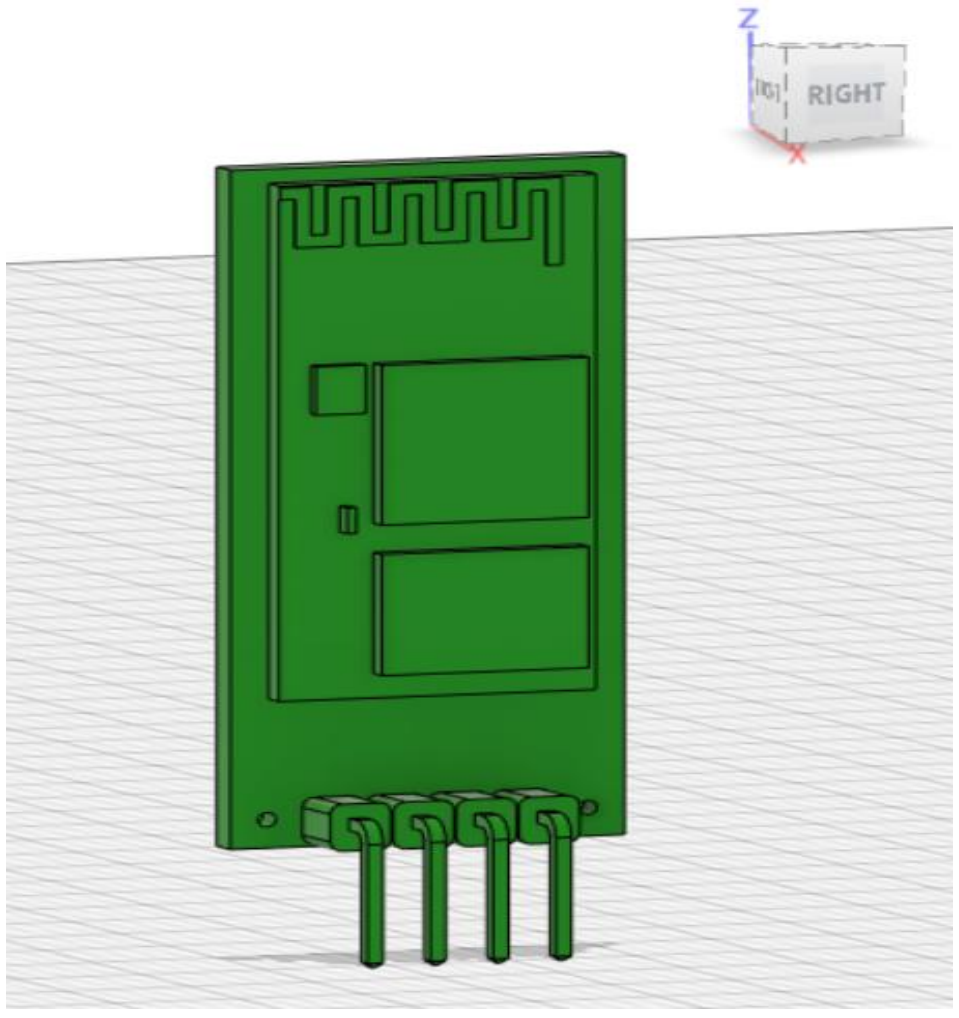

*Fig 5.39: HC-05 Bluetooth Module*

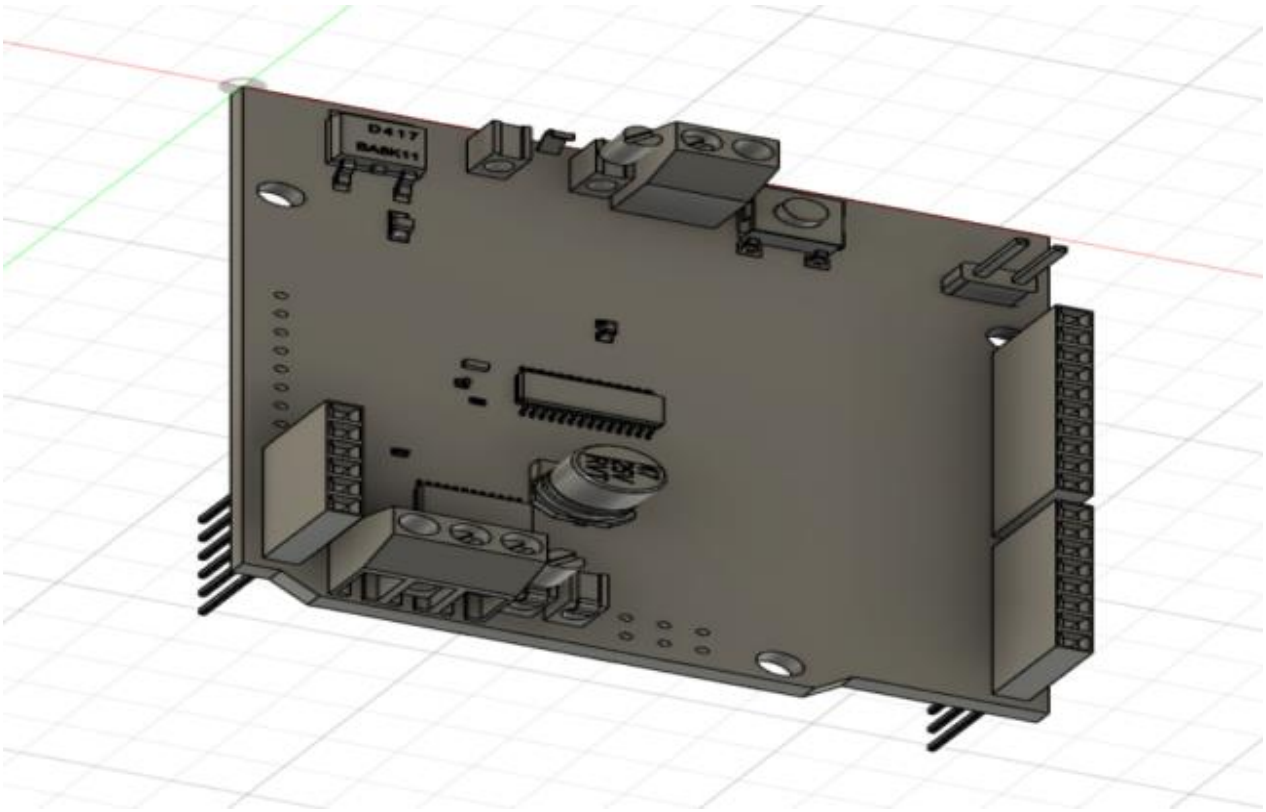

*Fig 5.40:  Motor Shield V2*

*5.18: Assembly*

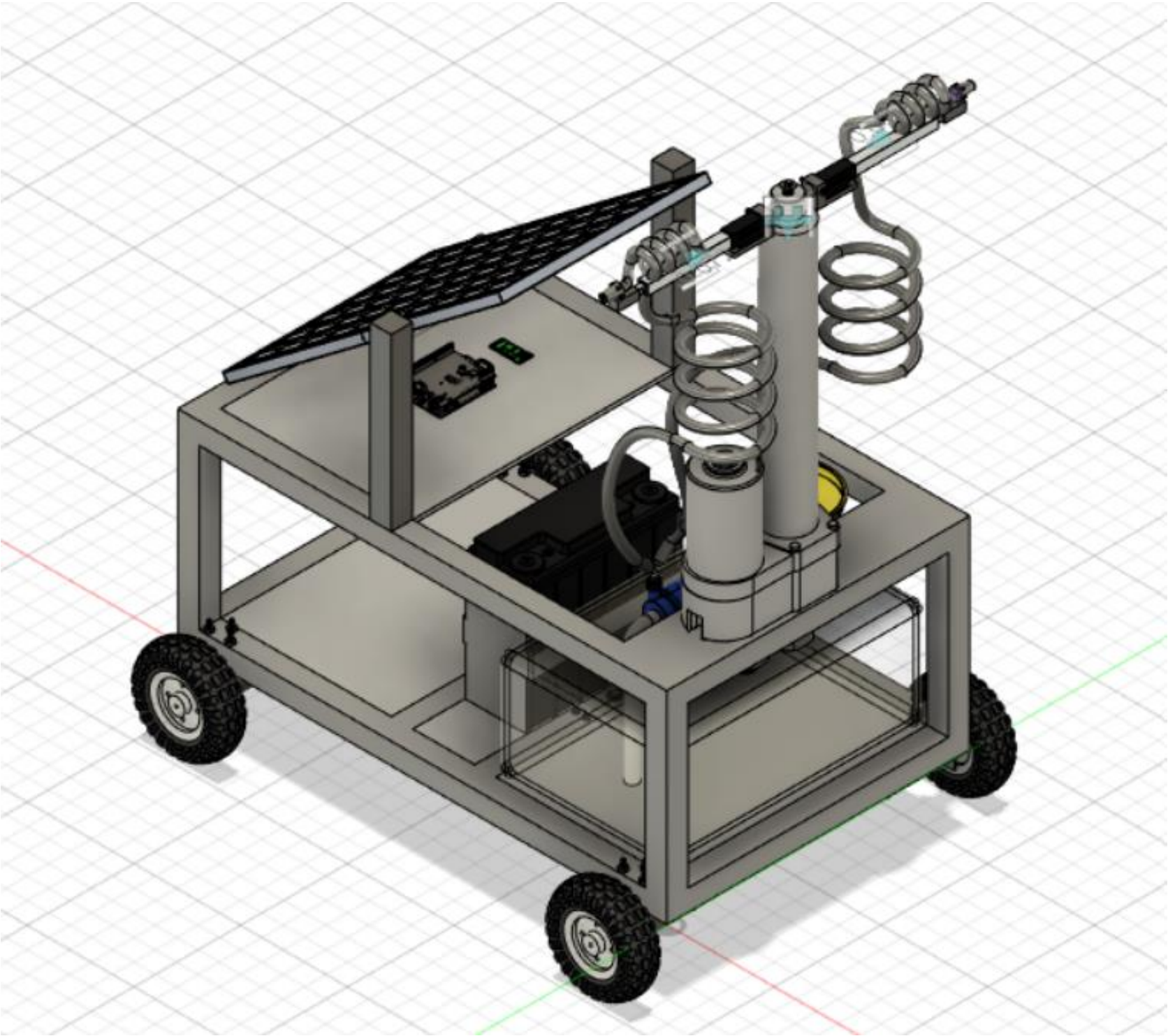

*Fig 5.41: Isometric view*

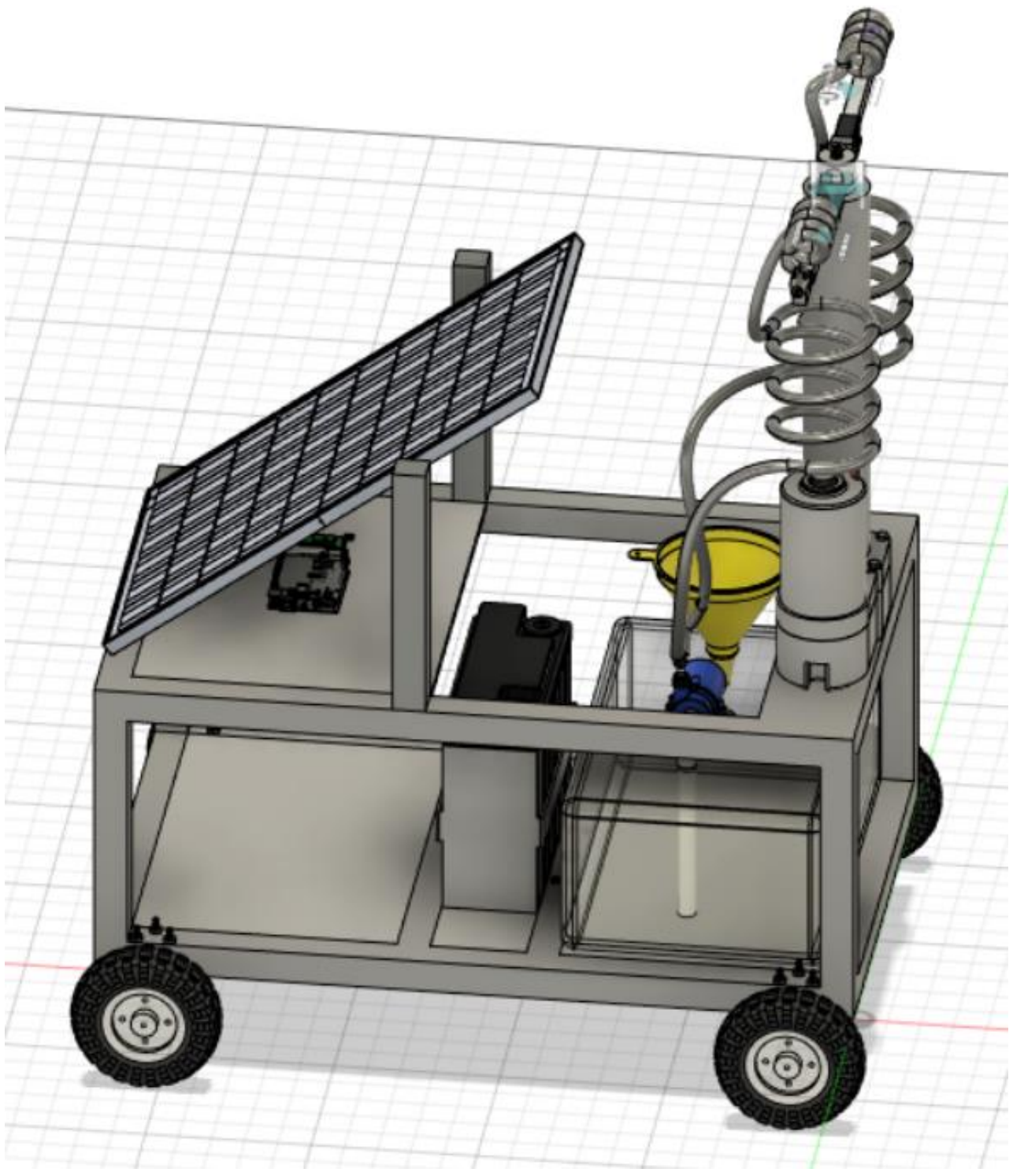

*Fig 5.42: Alternate view*

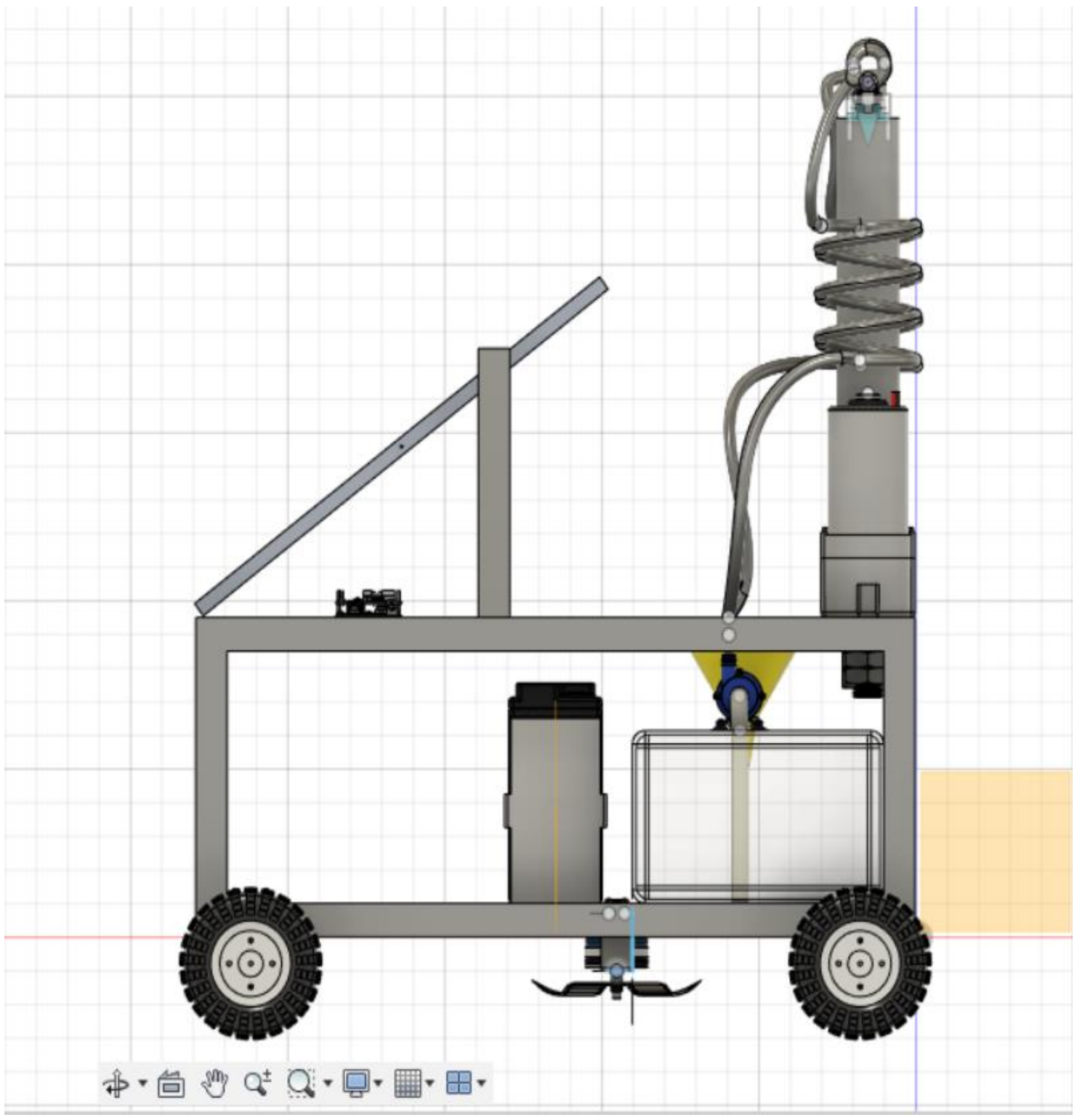

*Fig 5.43: Front view*

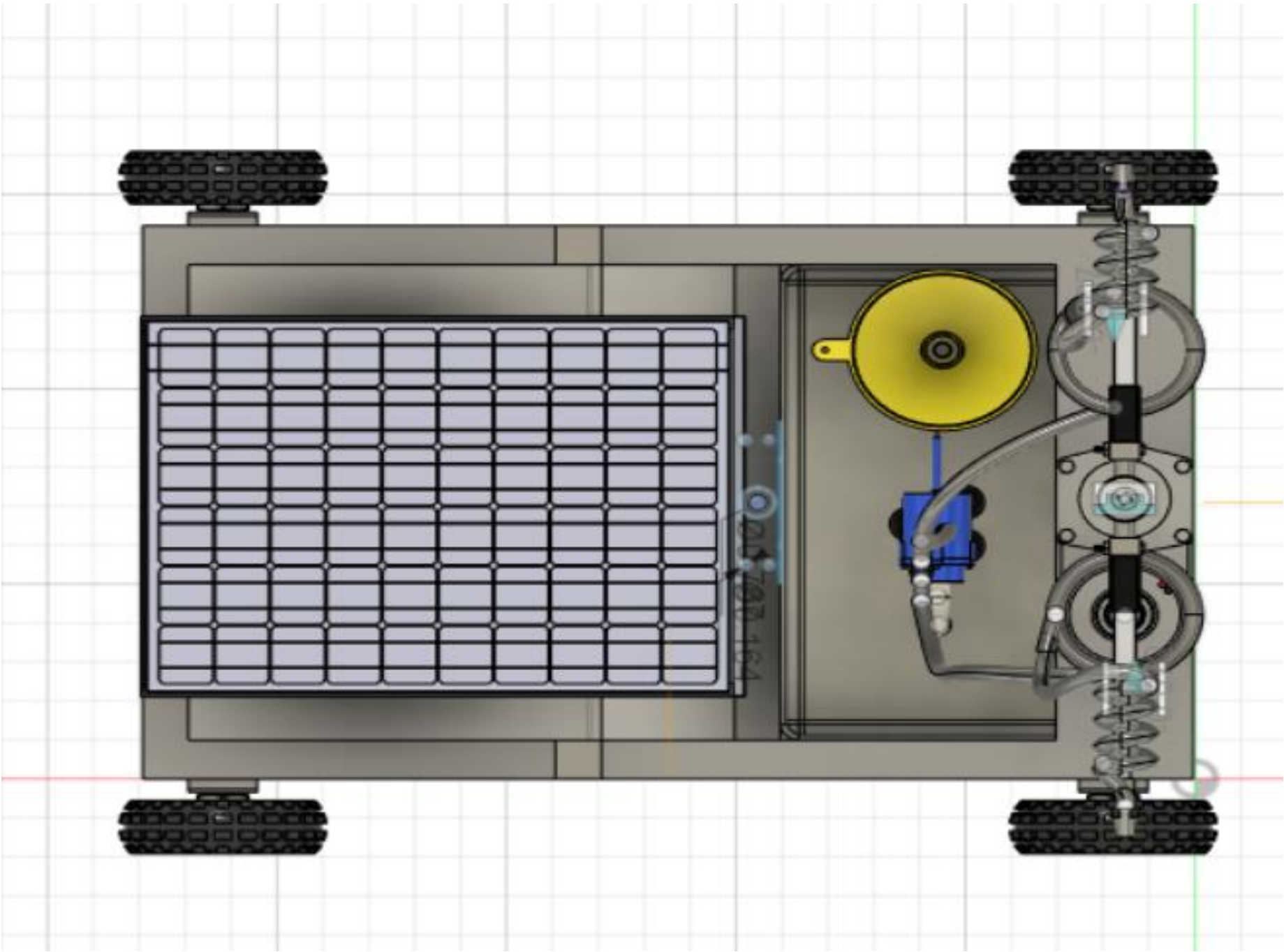

*Fig 5.44: Top view*

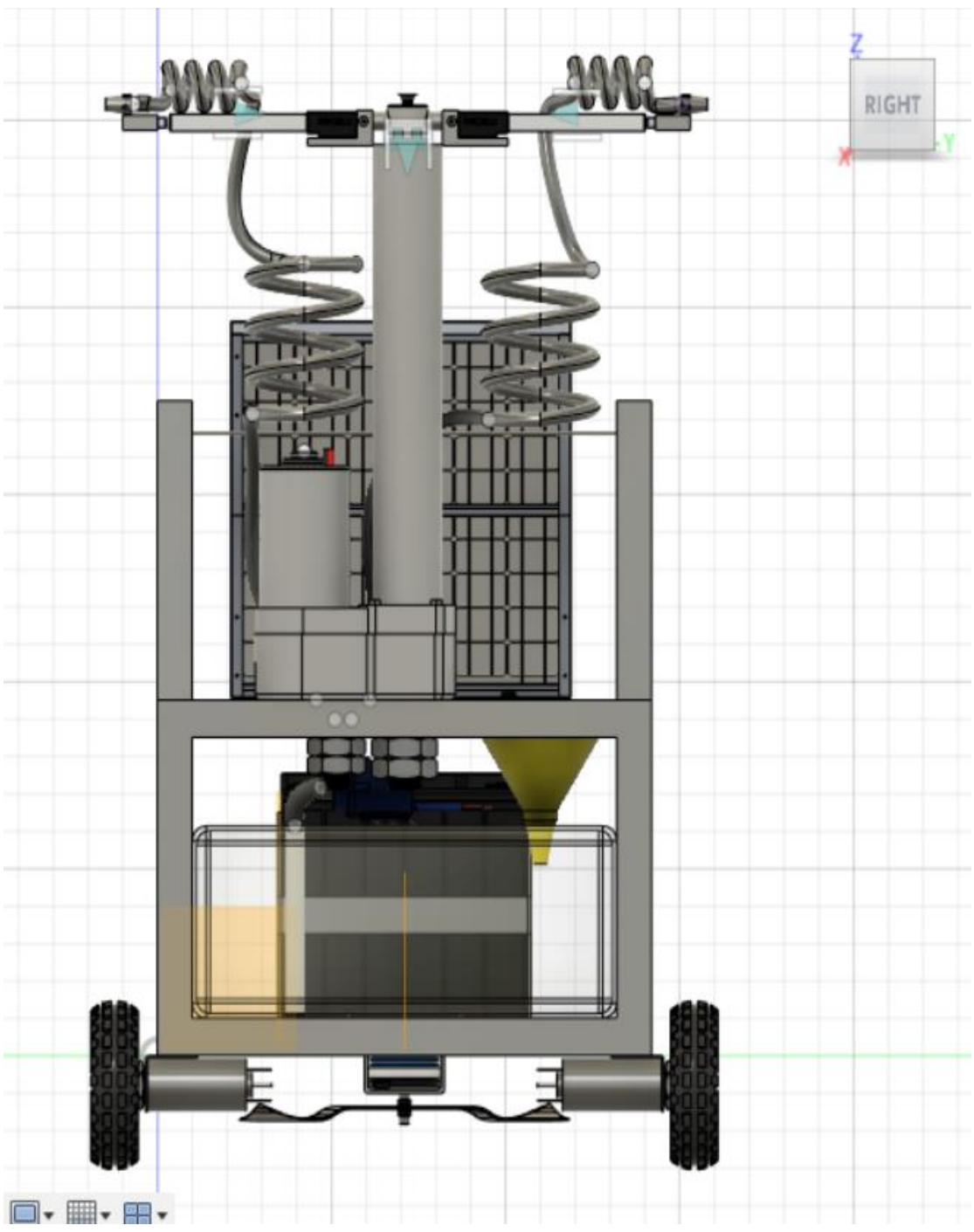

*Fig 5.45: Right side view*

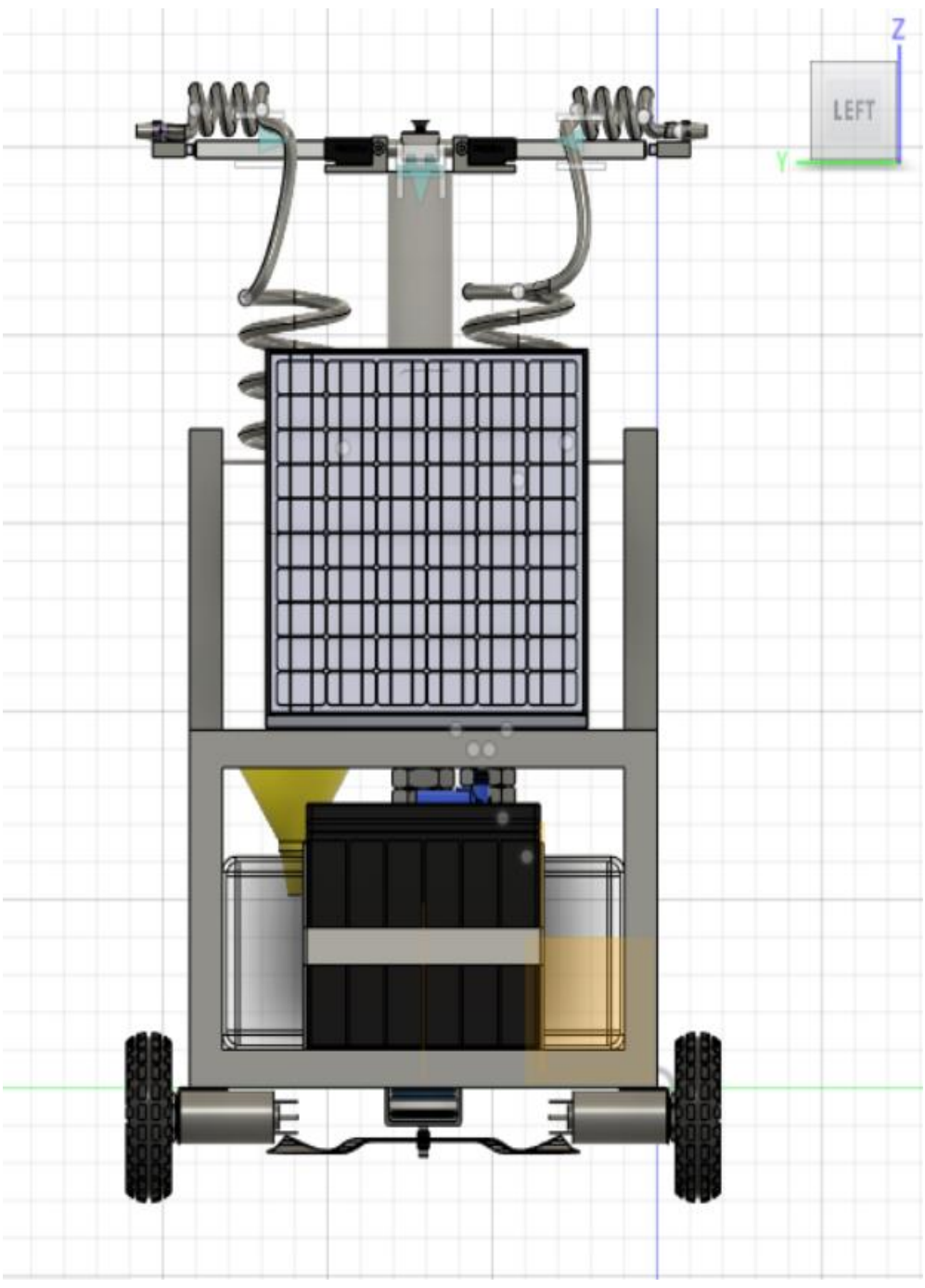

*Fig 5.46: Left side view*

# 6. PROPOSED SYSTEM

This system is a robot designed for agricultural purpose.

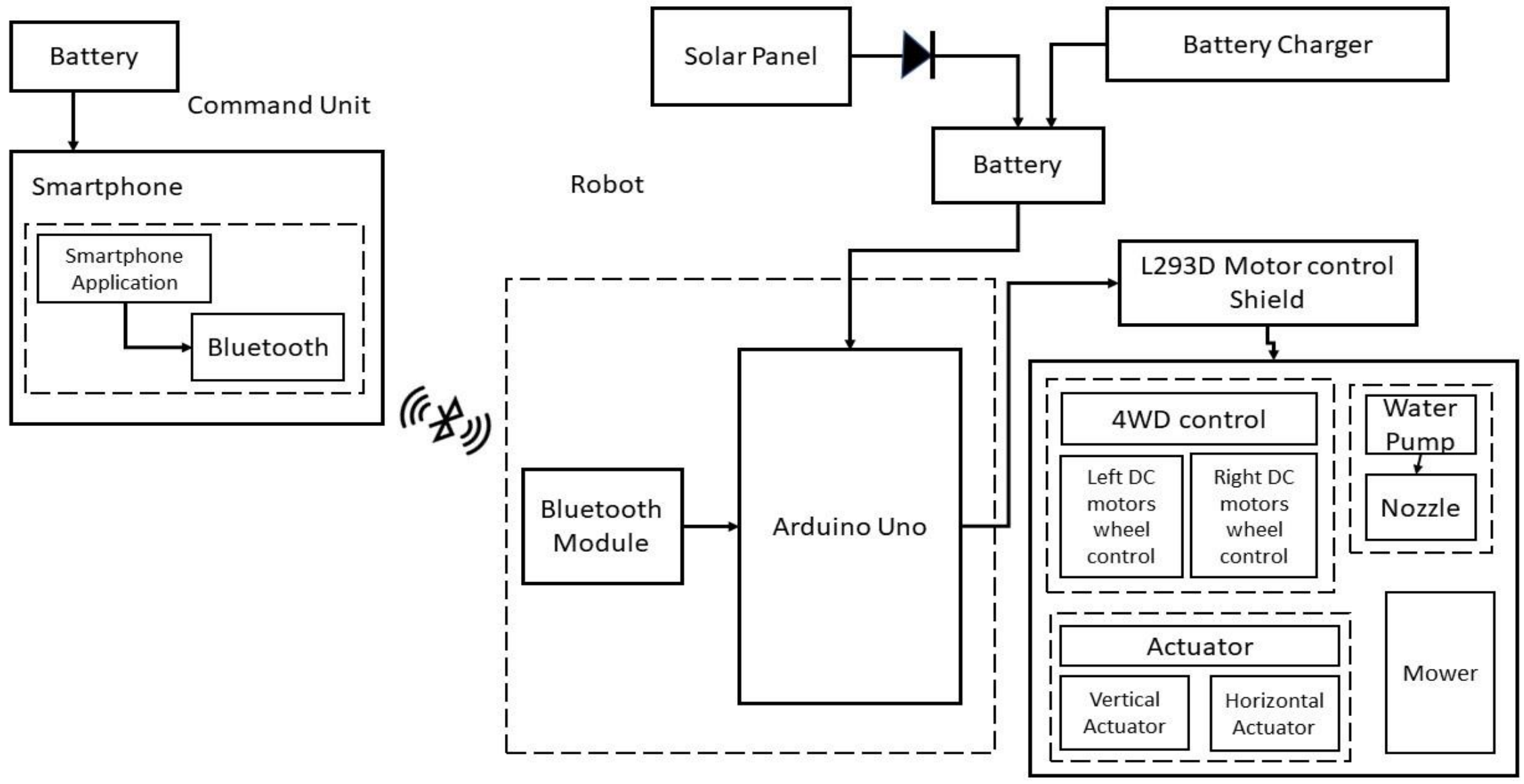

*Fig 6.1. Block diagram of proposed system*

This system is a robot designed for agricultural and sport fields maintenance purpose. Agro robot is developed with Arduino, L293D Motor control Shield, HC-05 Bluetooth Module, 4 Wheel drive with 4 DC Motors, Water pump, Nozzles, Mower setup, Battery and Solar Panel. This Agrobot with Manual Sun Tracking Solar Panel performs a number of concurrent operations. Today, the man power to do the farming is major concern, this machine removes the hurdles of agriculture laborers. Their efficiency and operating speed greatly affect the productivity. The different electrical components are connected to the combination of Arduino board and Motor Shield.

## *6.1 Fabrication*

- A long mild steel rod is square cross section with a side of 1 inch is taken and cut into 12 parts of the following dimensions.
  - 4 parts of length 22.9 inches.
  - 4 parts of length 14.2 inches.
  - 4 parts of length 9.5 inches.

- Now they are welded together to form a cuboid chassis of dimension (22.9*14.2*9.5 in3) as shown in the figure.

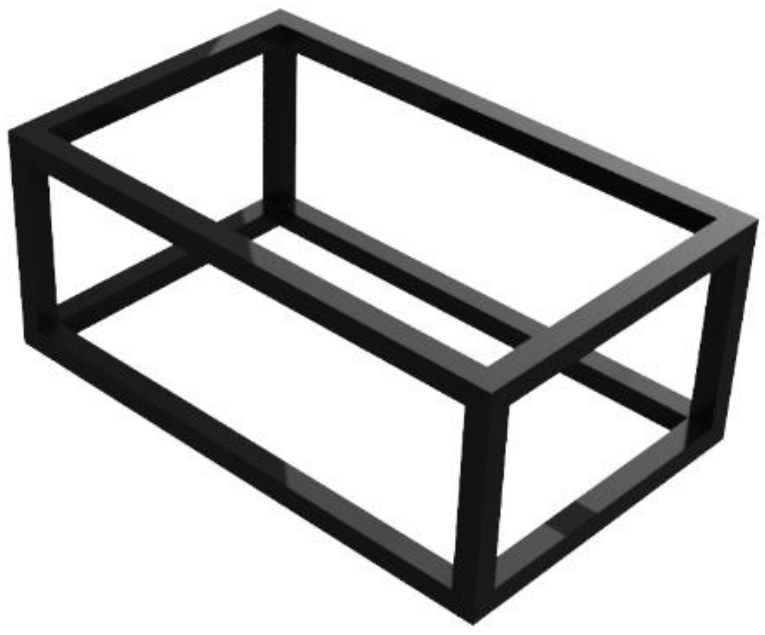

*Fig 6.2 The Chassis*

- Weld two L brackets at the back end of the chassis and two fibre glass pieces in the front to hold the DC Motors providing a 4-wheel drive as shown in the figure.

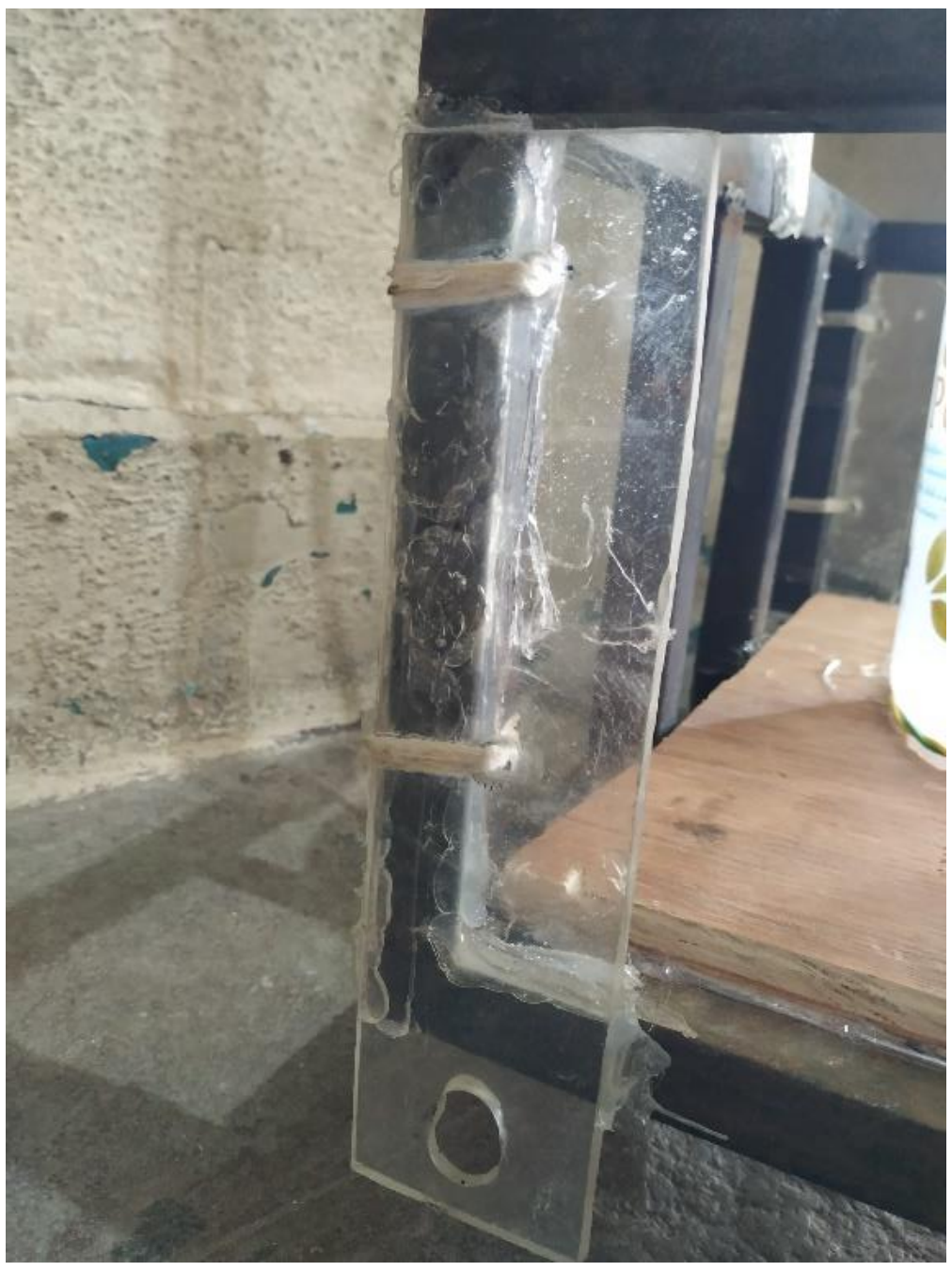

*Fig 6.3 Front wheels Motor Holder*

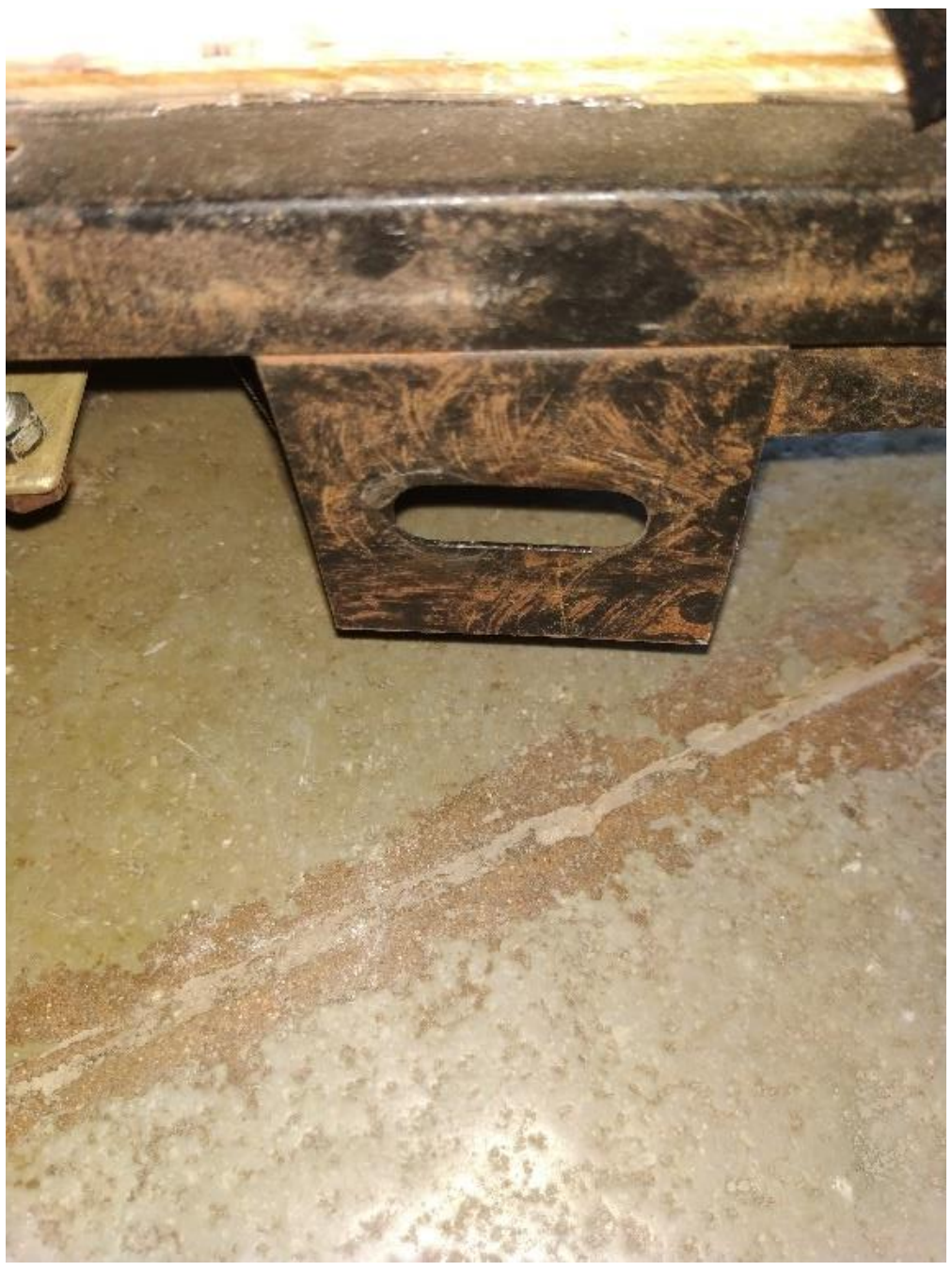

*Fig 6.4 Rear Wheels Motor Holder*

- Clamp the motors to the brackets and the fibre glass using the provided nut and secure the joint firmly. Now attach the front and rear wheels onto the motor shafts and tighten it with the screw provided to prevent the relative motion between the wheel and the shaft as shown in the figure.

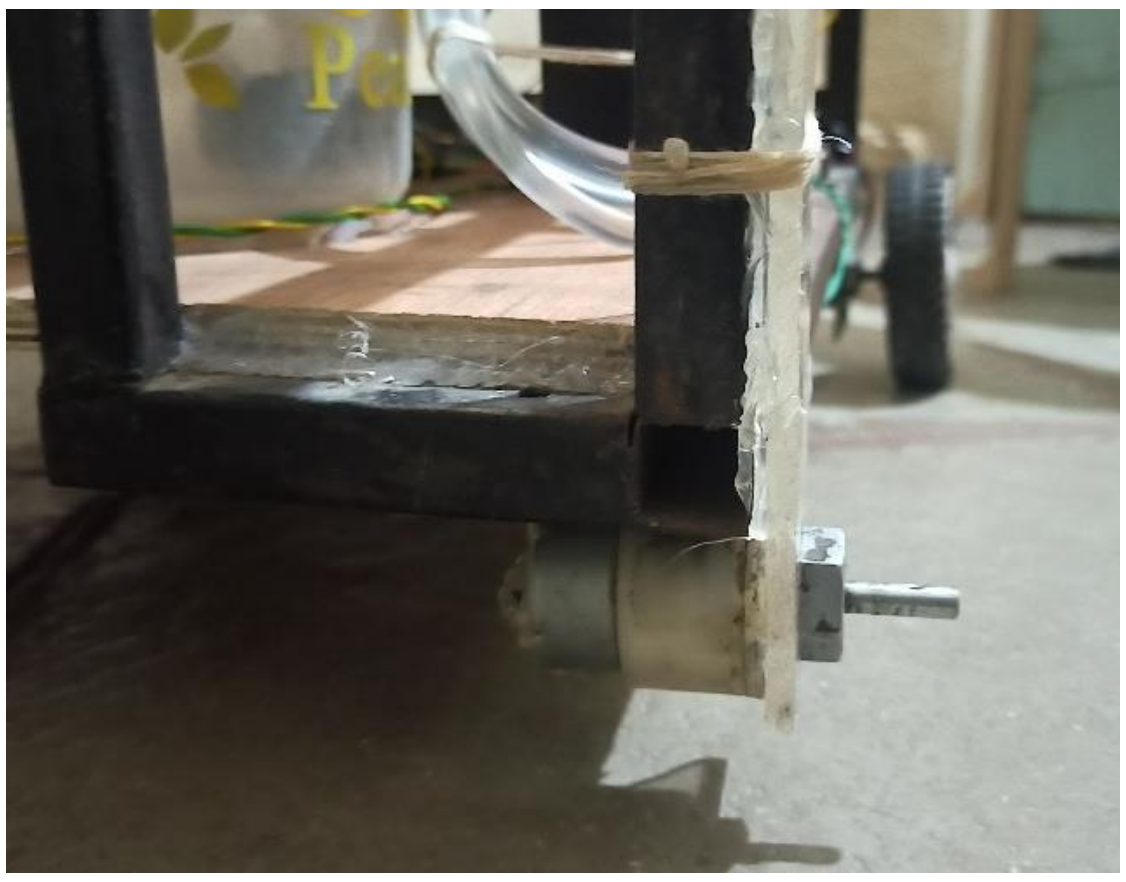

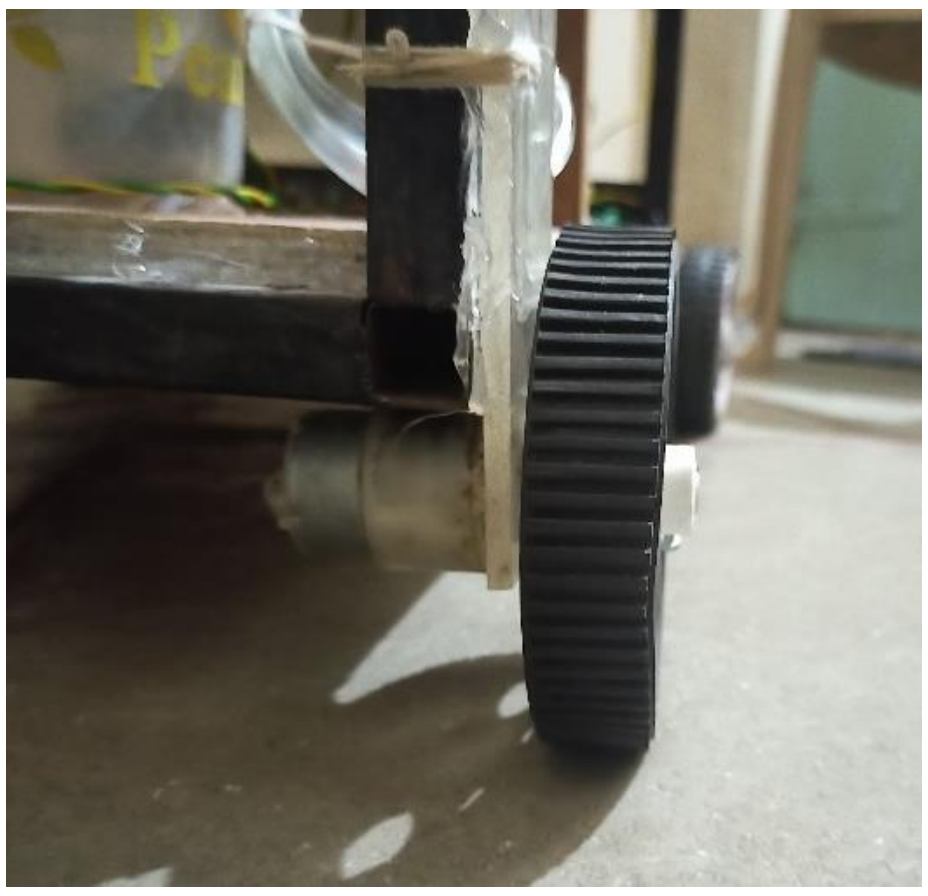

*Fig 6.5 Clamped Motor* *Fig 6.6 Wheel attached to motor shaft*

- Take a wood plank of dimensions 20.4*12.4 inch with a thickness of 0.5 inch and glue it on the base with hot glue. Cut a hole through the wooden plank to enable connections from the bottom parts to the Arduino.

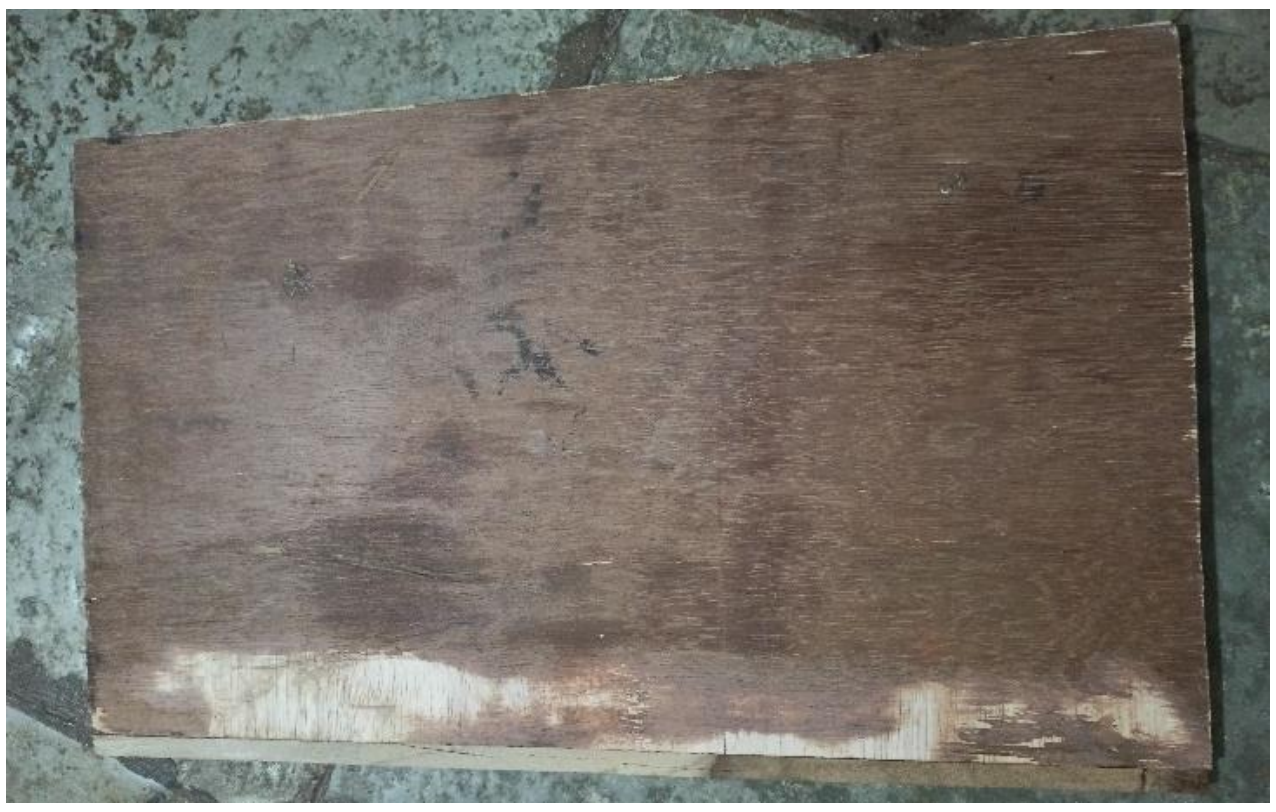

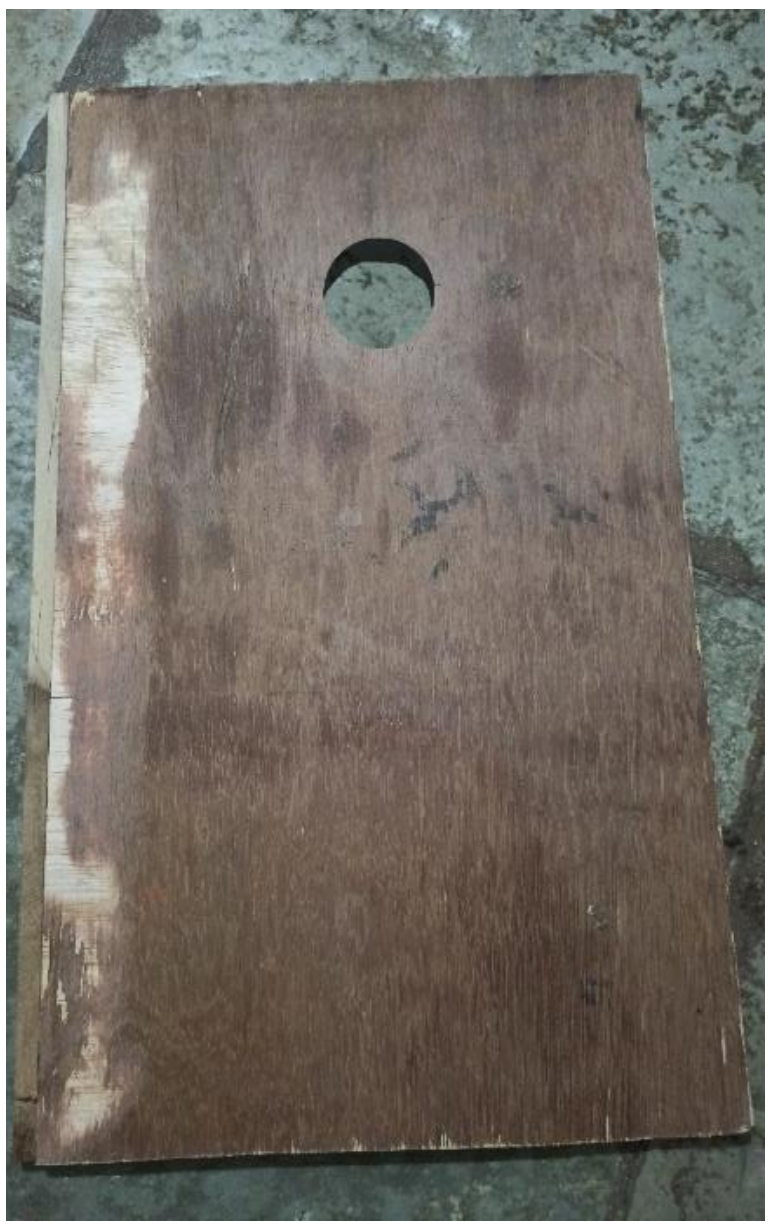

*Fig 6.7 Wooden plank* *Fig 6.8 Hole for connections*

- Get a specially made rod with a bracket in centre to accommodate the mower motor and clamp the motor to the bracket with the help of the nut. Secure the joint of the above rod to the chassis with help of nuts and bolt of 8mm dia. Attach a mower blade to the shaft of mower motor as shown in the figure.

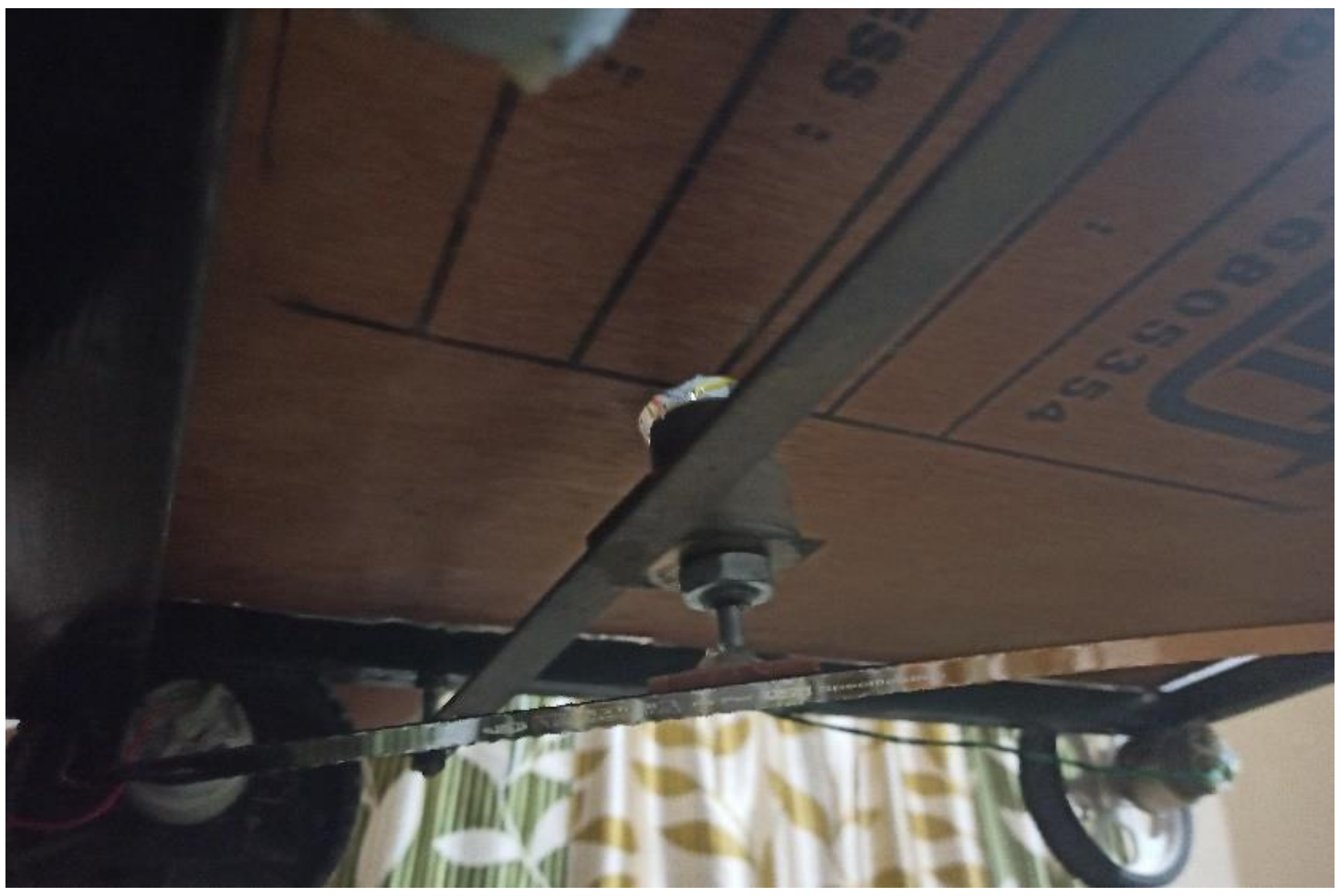

*Fig 6.9 Rod with bracket for Mower motor*

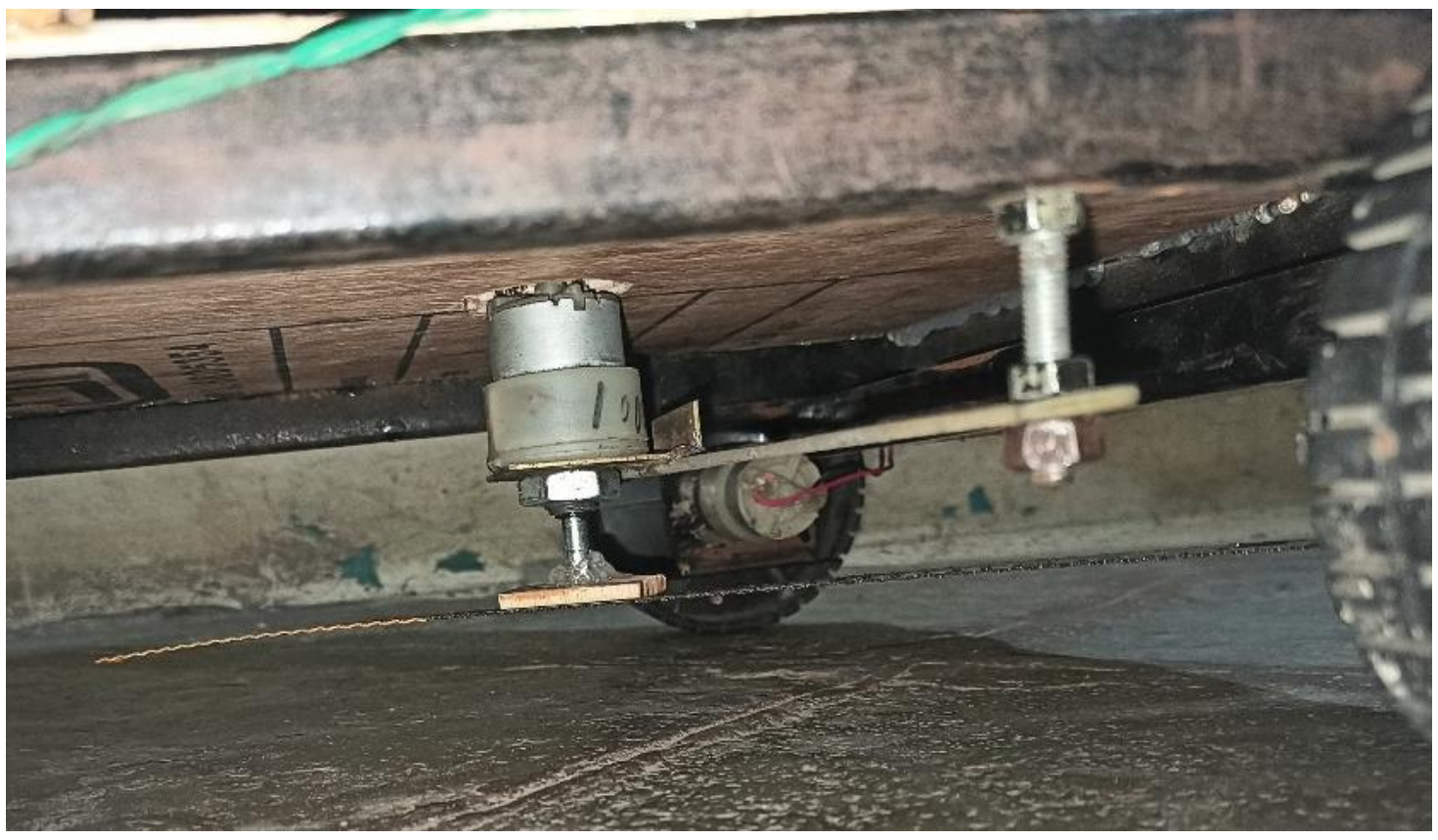

*Fig 6.10 Mower arrangement*

- Get two long PVC pipe of outer diameters 1 inch and 1.05 inch of 1 mm thickness, and cut them into 5 pieces as below
  - 1 inch dia. pipe – 2 parts of lengths 15.5 and 24 inches respectively.
  - 1.05 inch dia. pipe – 3 parts – 3.5, 3.5 and 15.5 inches respectively.
- Take the 1.05 inch dia. pipe of length 15.5 inches and make a slot of 1 inch of square cross-section at the bottom with a depth of 1 inch, sand it to smoothen the edges and attach it to the chassis. Drill a 8mm hole through one side along the diameter of the pipe and sand the edges to eliminate sharp and rough edges. Now stick a nut of 8mm dia. concentric with the through hole. Turn a bolt through the nut to complete the locking mechanism of the vertical actuation.

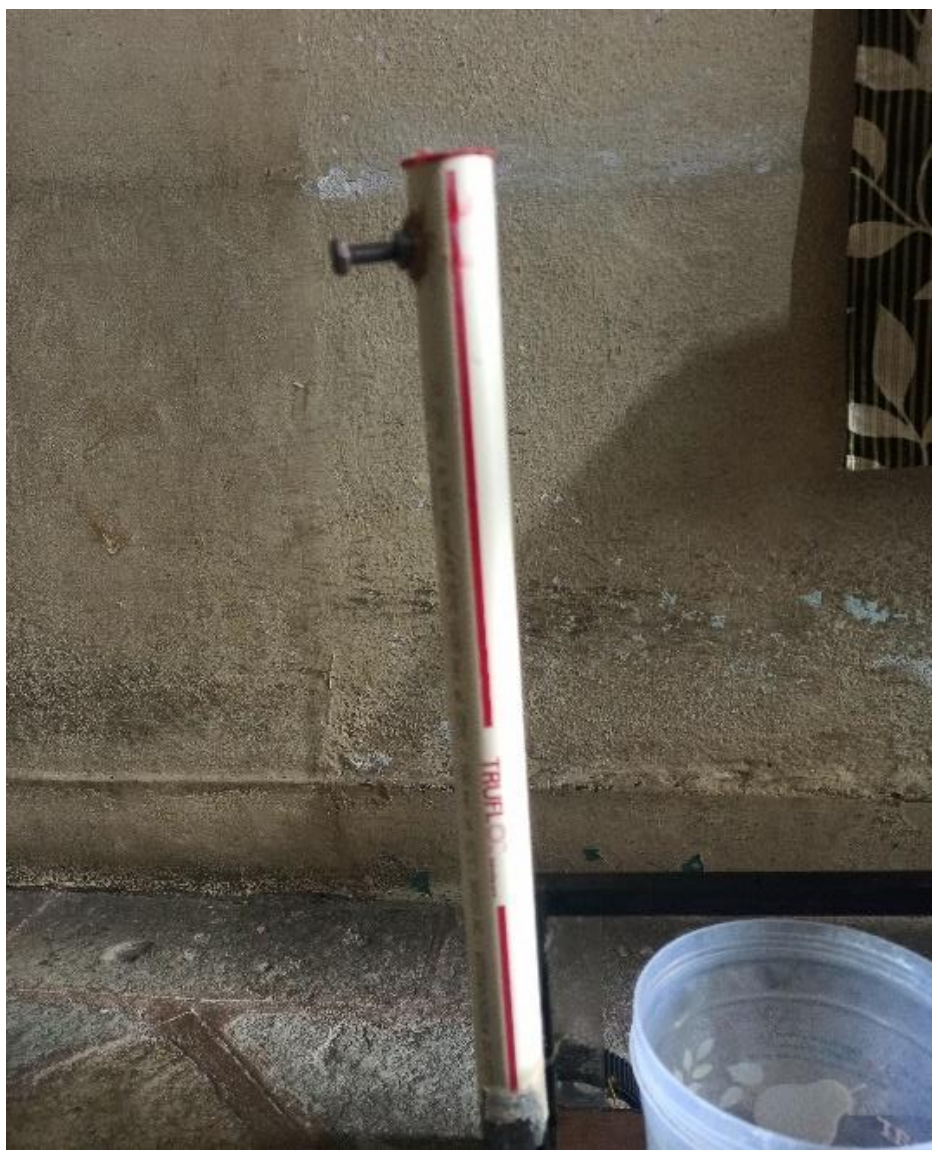

*Fig 6.11 Vertical Actuation*

- Take the 1 inch dia. pipe of length 15.5 inches and sand a curve on the top of radius 1 inch. Now take the other 1 inch dia. pipe of length 24 inches and stick it on the top of the curve previously made on 15.5 inch pipe and form a T-joint bisecting each other.
- Now take the two 1.05 inch dia pipe of lengths 3.5 inches (called sliders) and stick a custom made rubber washer of internal diameter of 0.99 inches on one end of each pipe. Now slide those pipes through the horizontal part (24 inches) of the previous T-joint on either side. Now make a custom-made stoppers on either side of the horizontal part of T joint by taking pieces 1 inch dia., cut it on one end and force it on the horizontal pipe for tight fit and glue it.
- The spray nozzles are fixed in ON position using a thread and glue joint . The nozzles are fixed at the centre of sliders on either side of T-joint with the axis of nozzle perpendicular to the sliding motion. Stick the water tubing to the Nozzle inlet as as shown in the figure.

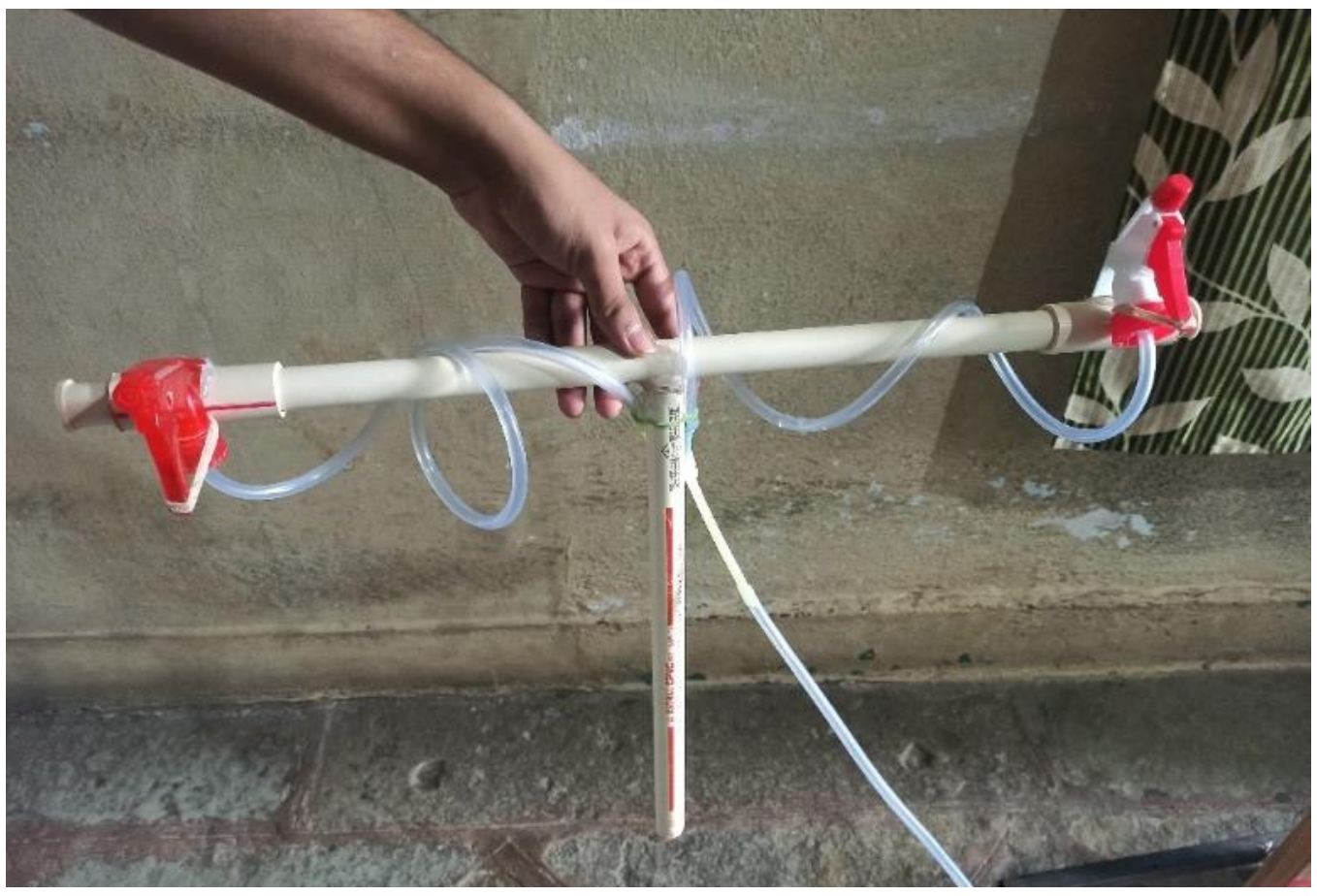

*Fig 6.12 T Joint*

- Now stick a custom-made rubber washer with internal dia. of 0.99 inches to the upper end of the 1.05 inch dia. pipe fixed to the chassis and push the vertical part of T-joint inside the vertical actuator outer contraption. Now lock the pipe at desired position using nut and bolt mechanism.

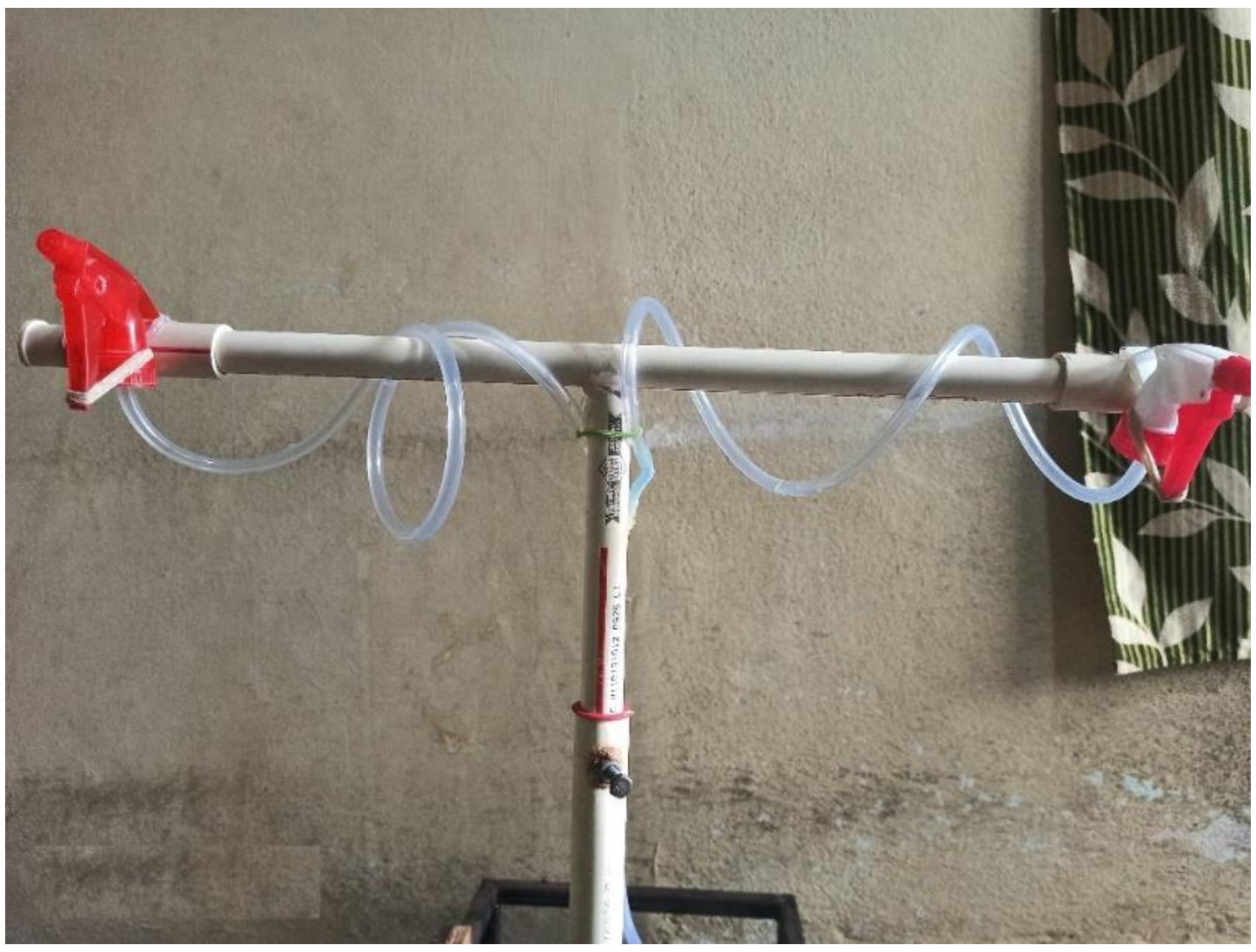

*Fig 6.13 Spraying arrangement*

- Take a plastic container make a hole through the side closer to the bottom of the container with a diameter of 0.8 inches to accommodate the pump inlet. Now seal the pump to the container by aligning the holes with M-seal to make a leakproof joint.

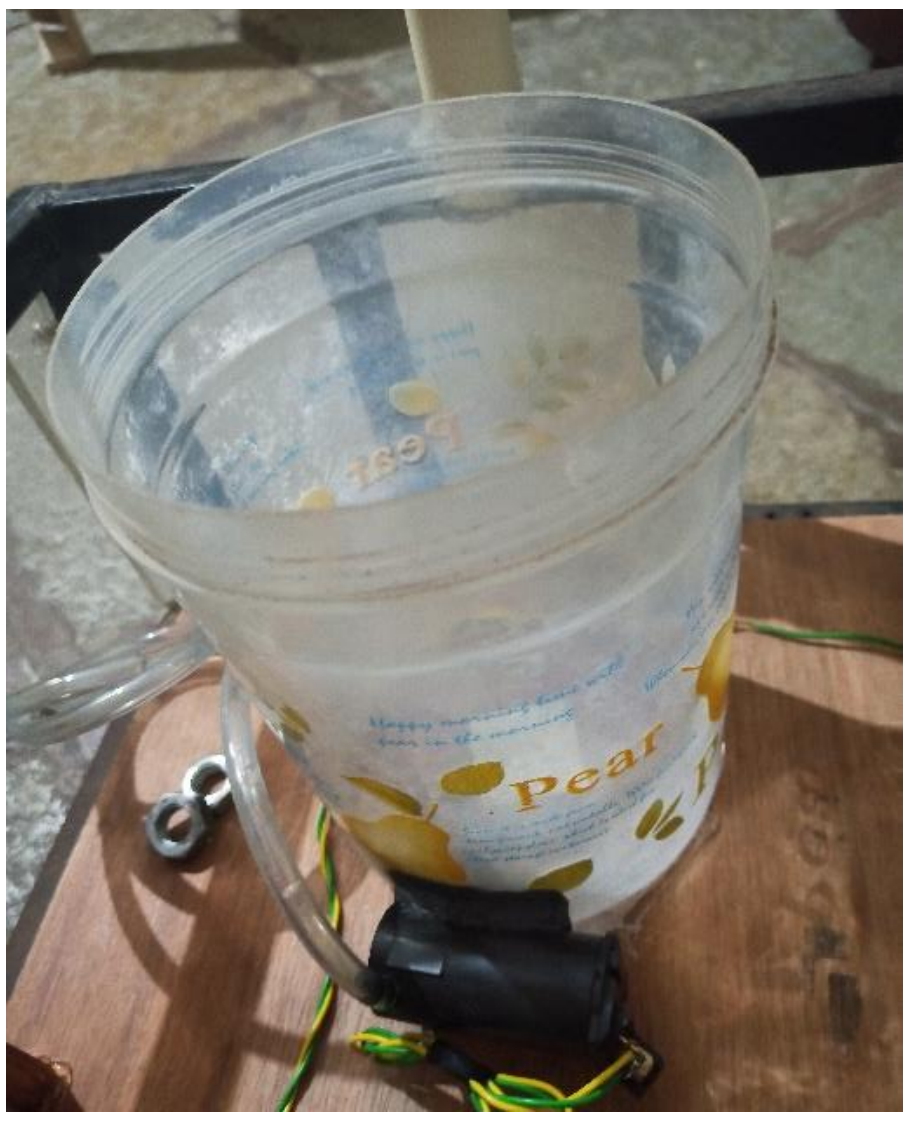

*Fig 6.14 Pump-Container arrangement*

- Make a Y joint using straws with Araldite Klear epoxy adhesive to make a leakproof joint and connect it to pump outlet and nozzle inlet water tubings.

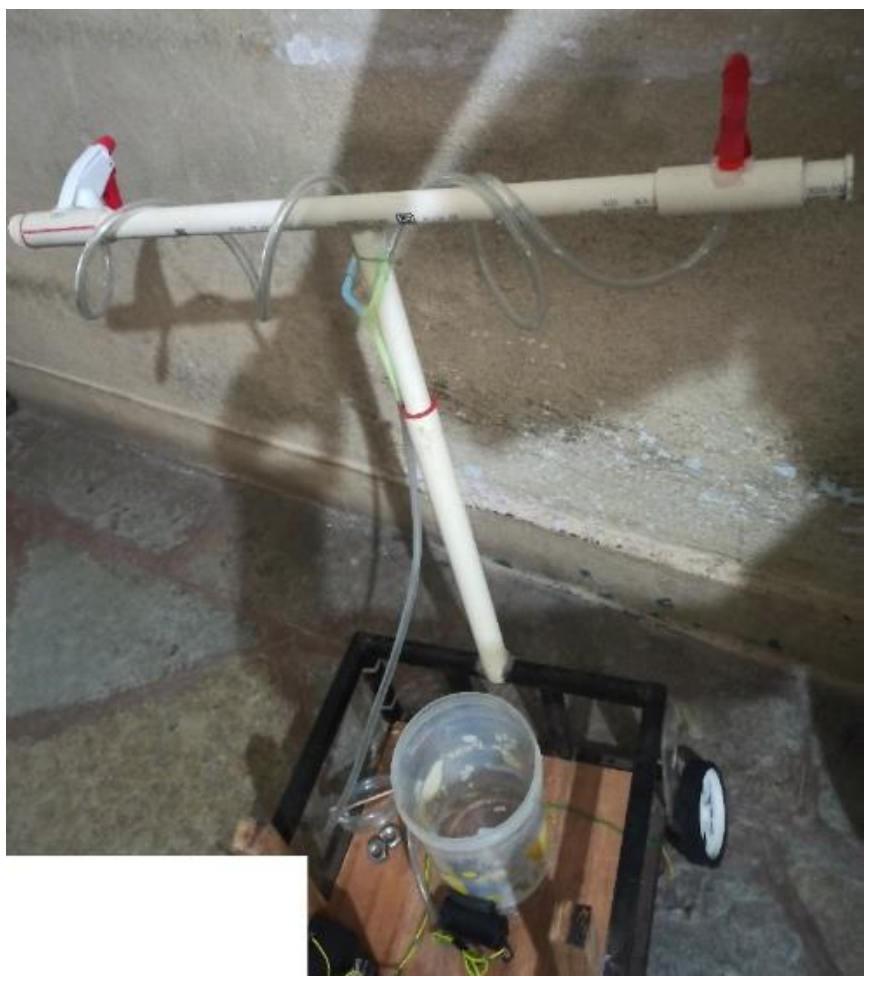

*Fig 6.15 Tubing between Nozzle and pump outlet through Y joint*

- Now stick 2 wooden pieces of 10*1.2*1.2 inches dimensions and stick it to the base wooden plank, at the left and right positions, in the centre of 20.4-inch dimension for supports to hold the solar panel in position as shown in the figure.

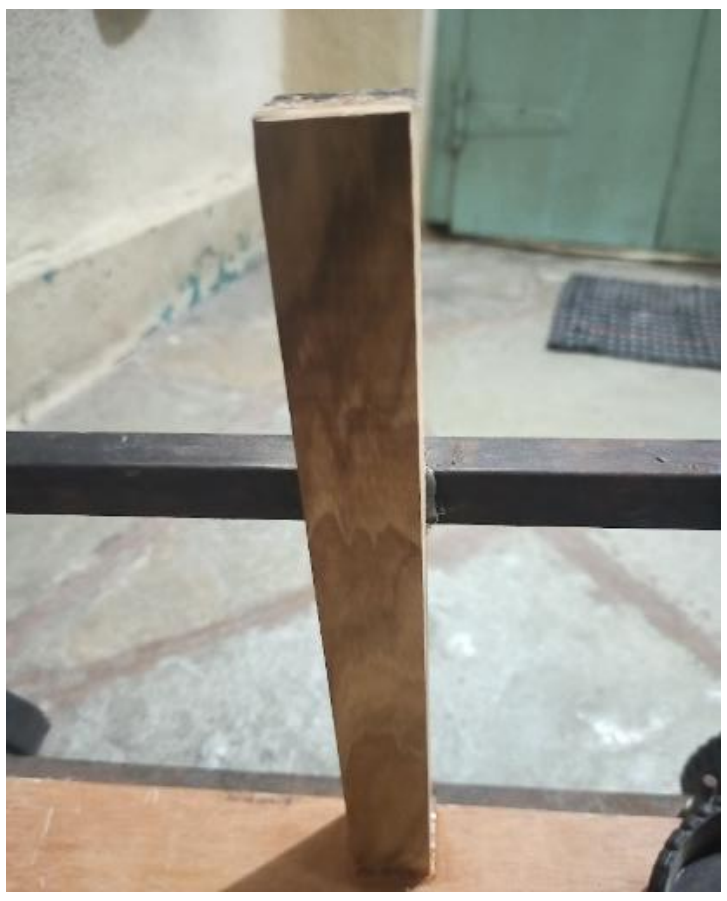

*Fig 6.16 Solar panel supports*

- Now put the circuit connections as shown in 'chap 6.2' to complete the prototype.

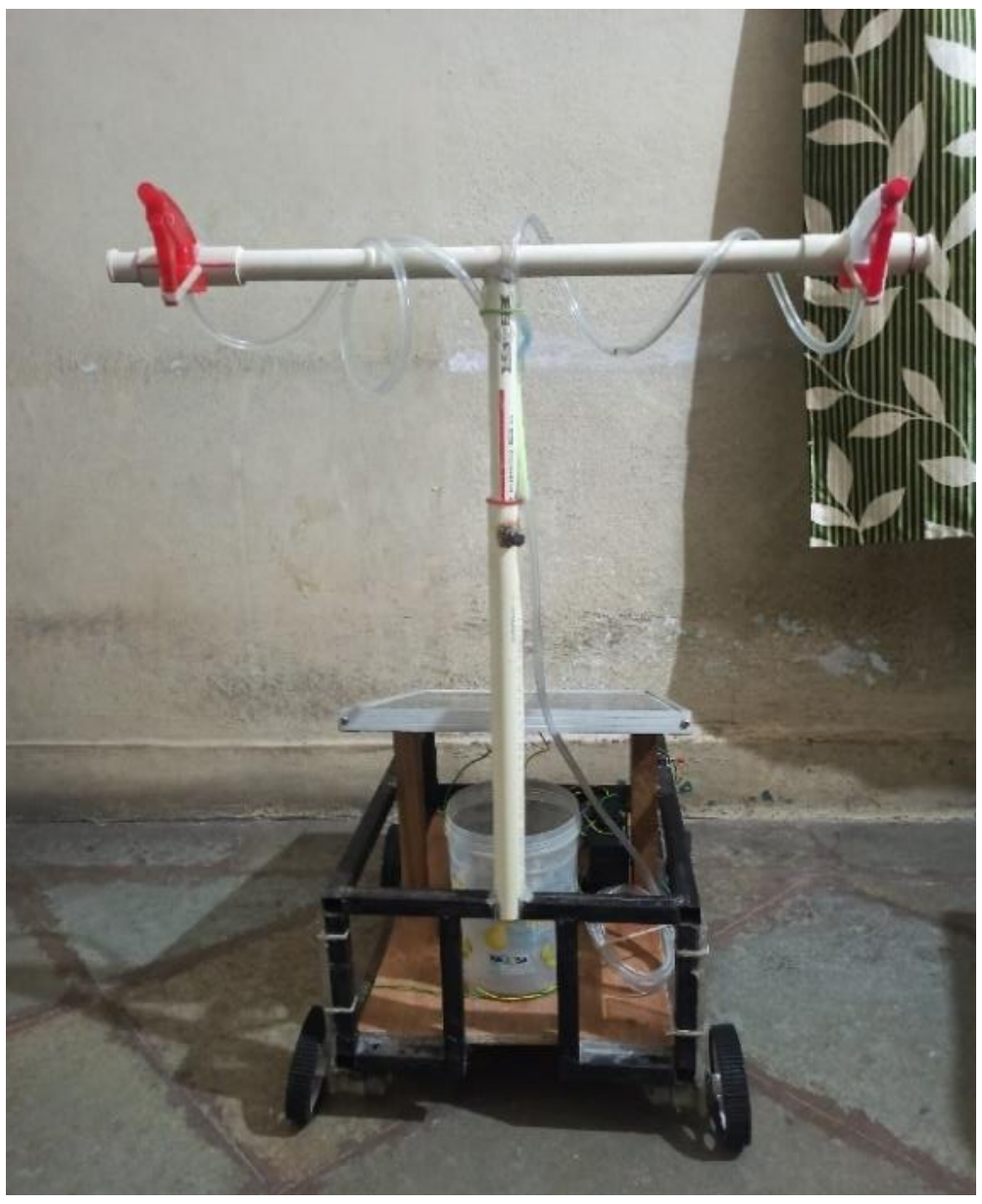

*Fig 6.17 Front View of the Prototype*

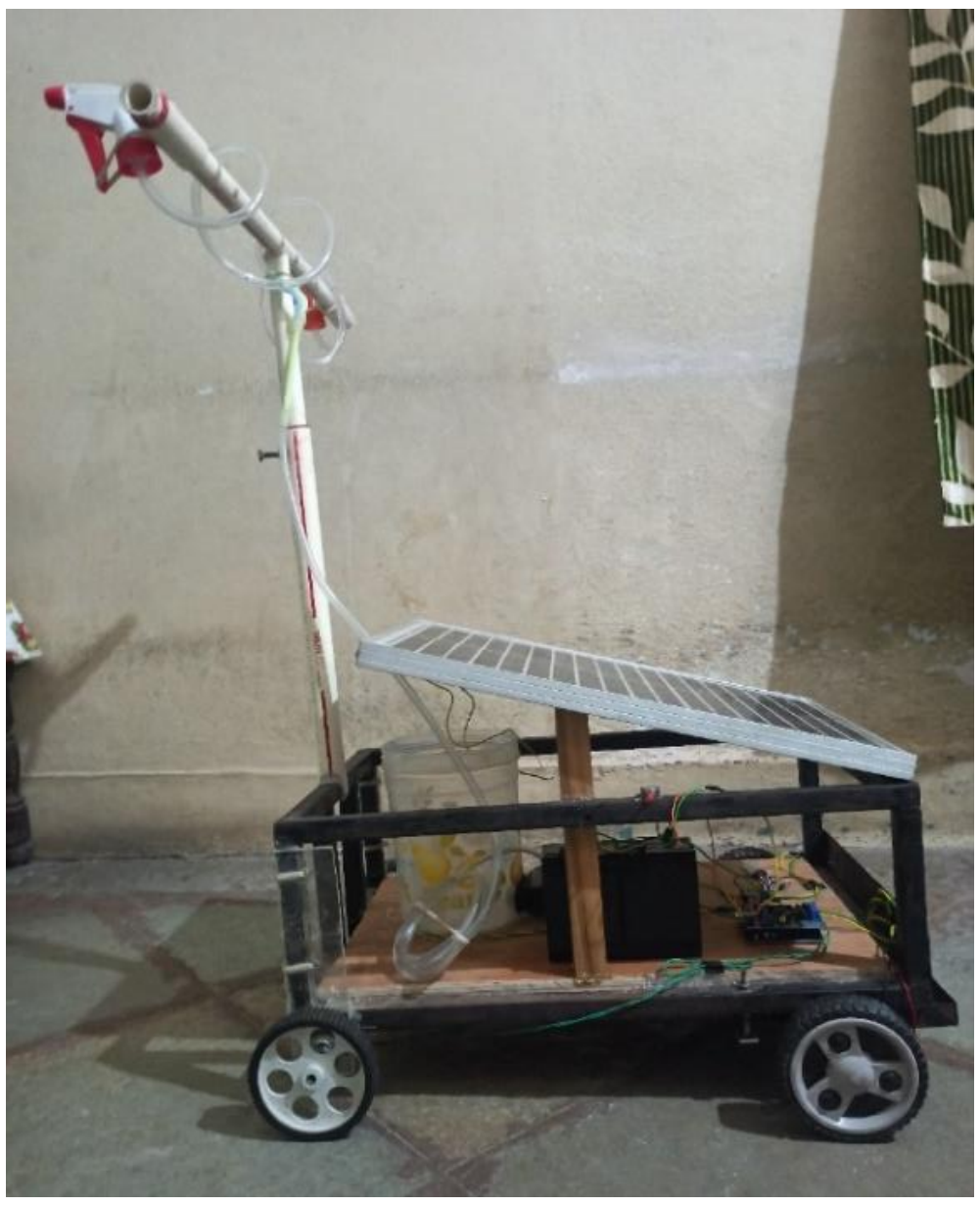

*Fig 6.18 Side View of the Prototype*

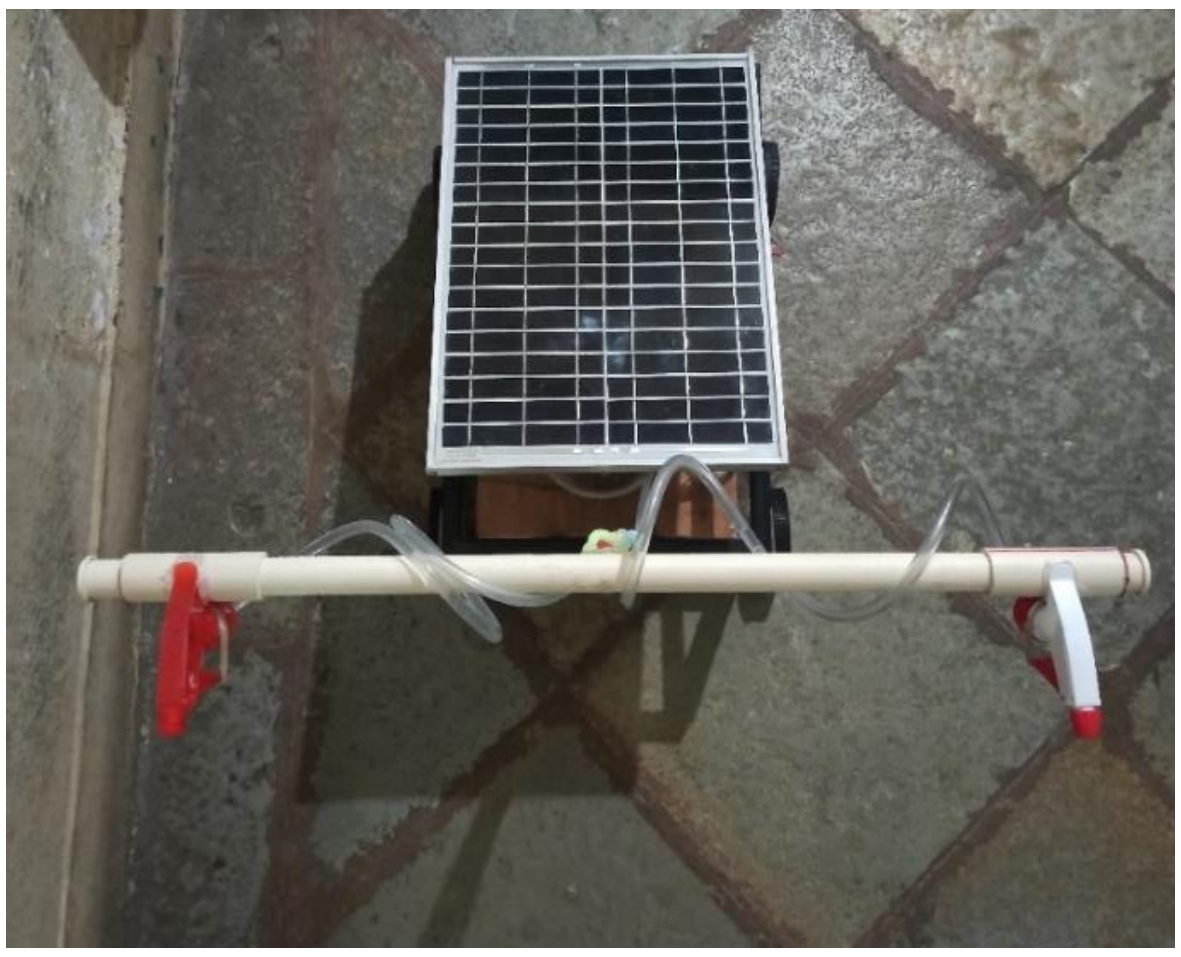

*Fig 6.19 Top View of the prototype*

### *6.2 Circuit connections between all electrical components*

The Bluetooth module Motor shield and the Arduino board are connected as shown in the figure.

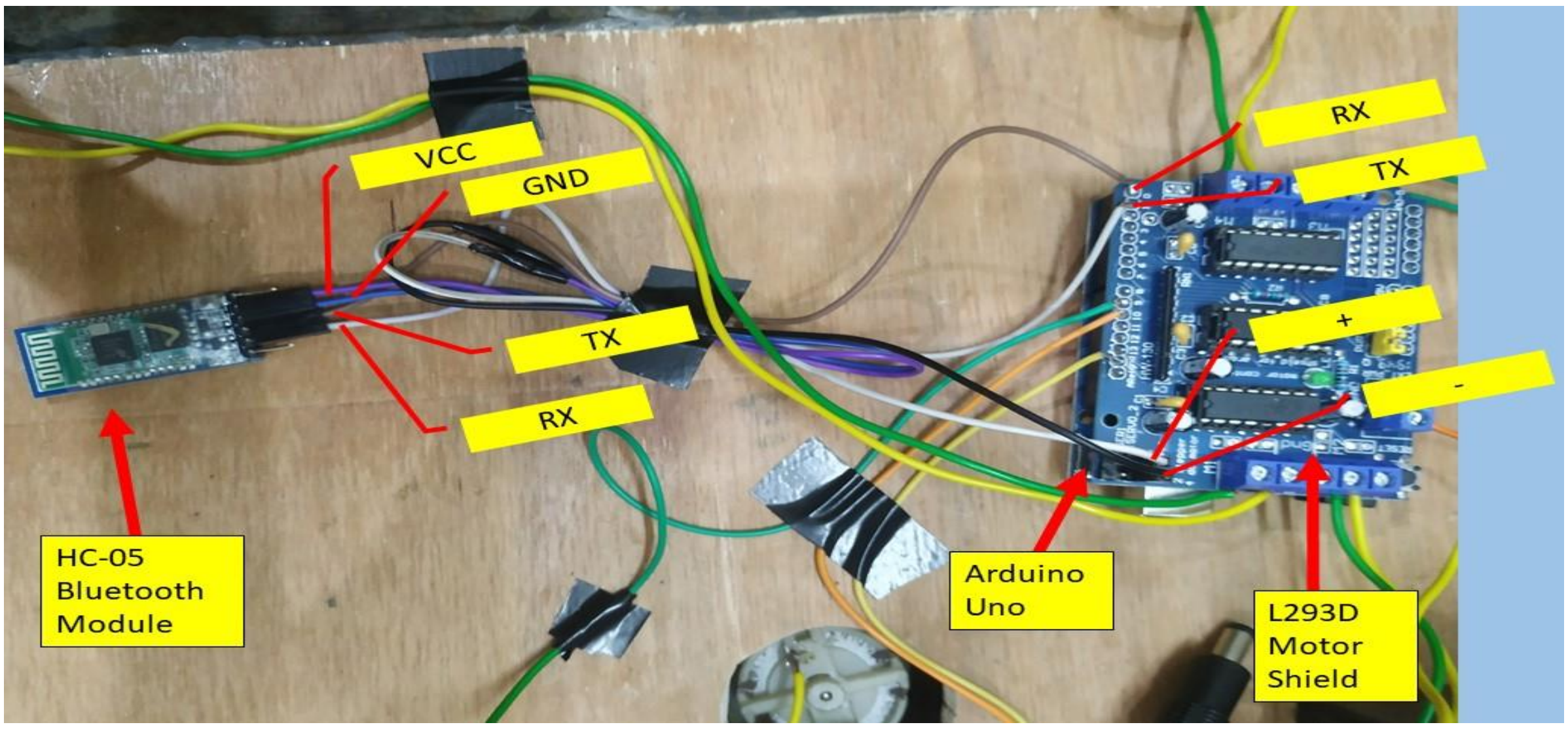

*Fig 6.20: Circuit 1: (Arduino, Motor shield and Bluetooth Module)*

After giving the above connections, give power supply to Motor shield and check if all the lights in Arduino, Motor shield and Bluetooth module are glowing.  To check the working of Bluetooth module

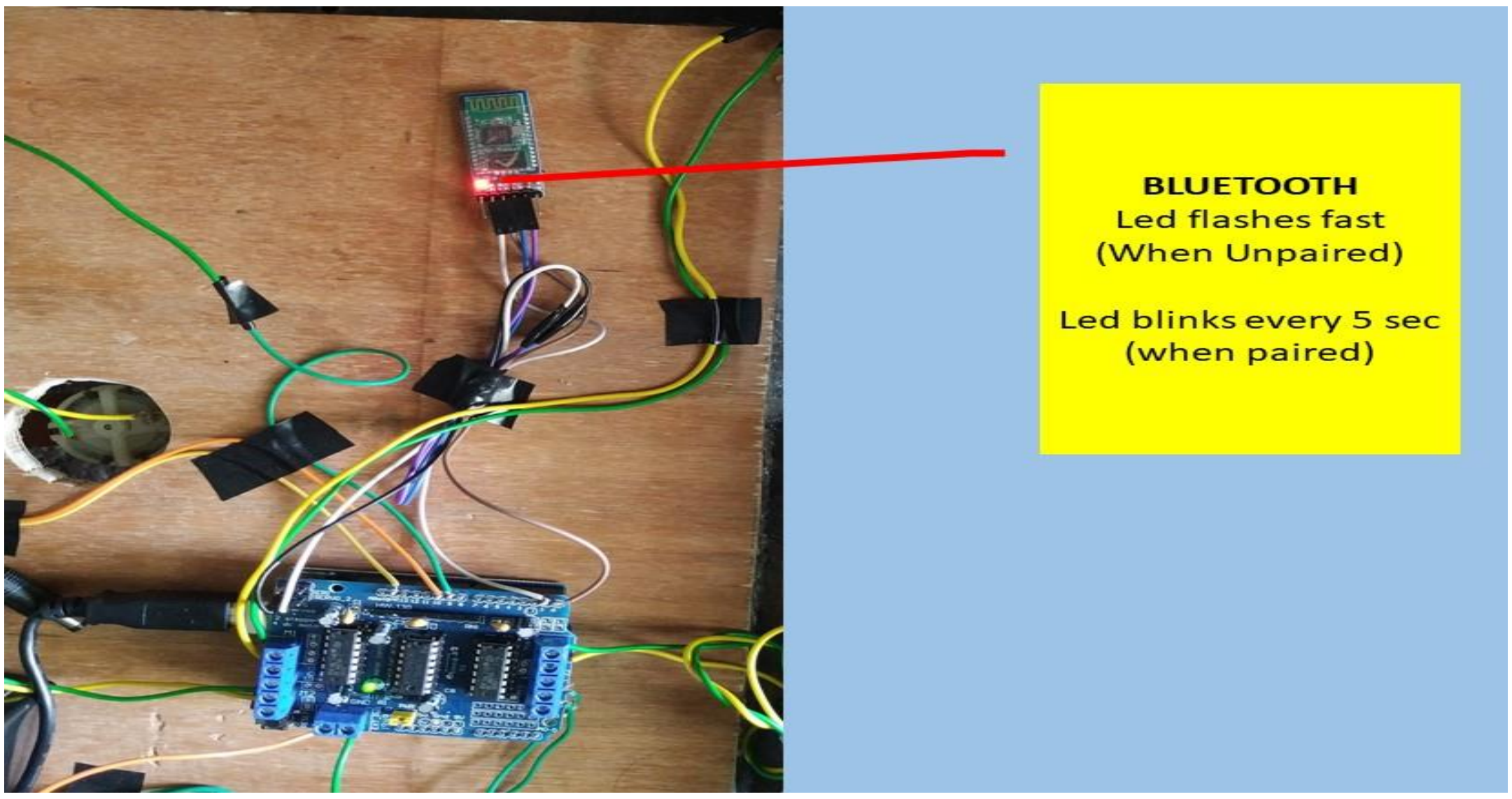

*Fig 6.21: Circuit 2: (Bluetooth)*

A typical motor shield can accommodate 4 DC motors, 2 servo motors and 2 stepper motors. For our application we only require 4 DC Motors to move in tandem, so the terminals of 4-wheel drive with DC motors are connected in M1, M2, M3 and M4 as shown in the figure. While connecting the wires to the positive and negative terminals of the motor shield make sure that they are connected in such a way that all the wheels rotate in the same direction either anti-clock wise or clockwise. The wires from the battery are screwed in the slots provided in the Motor Shield. A manual switch between Battery and Motor shield is provided so that we can switch ON and OFF the robot at our will. Otherwise, the circuit will always be in ON position which will waste a lot of power.

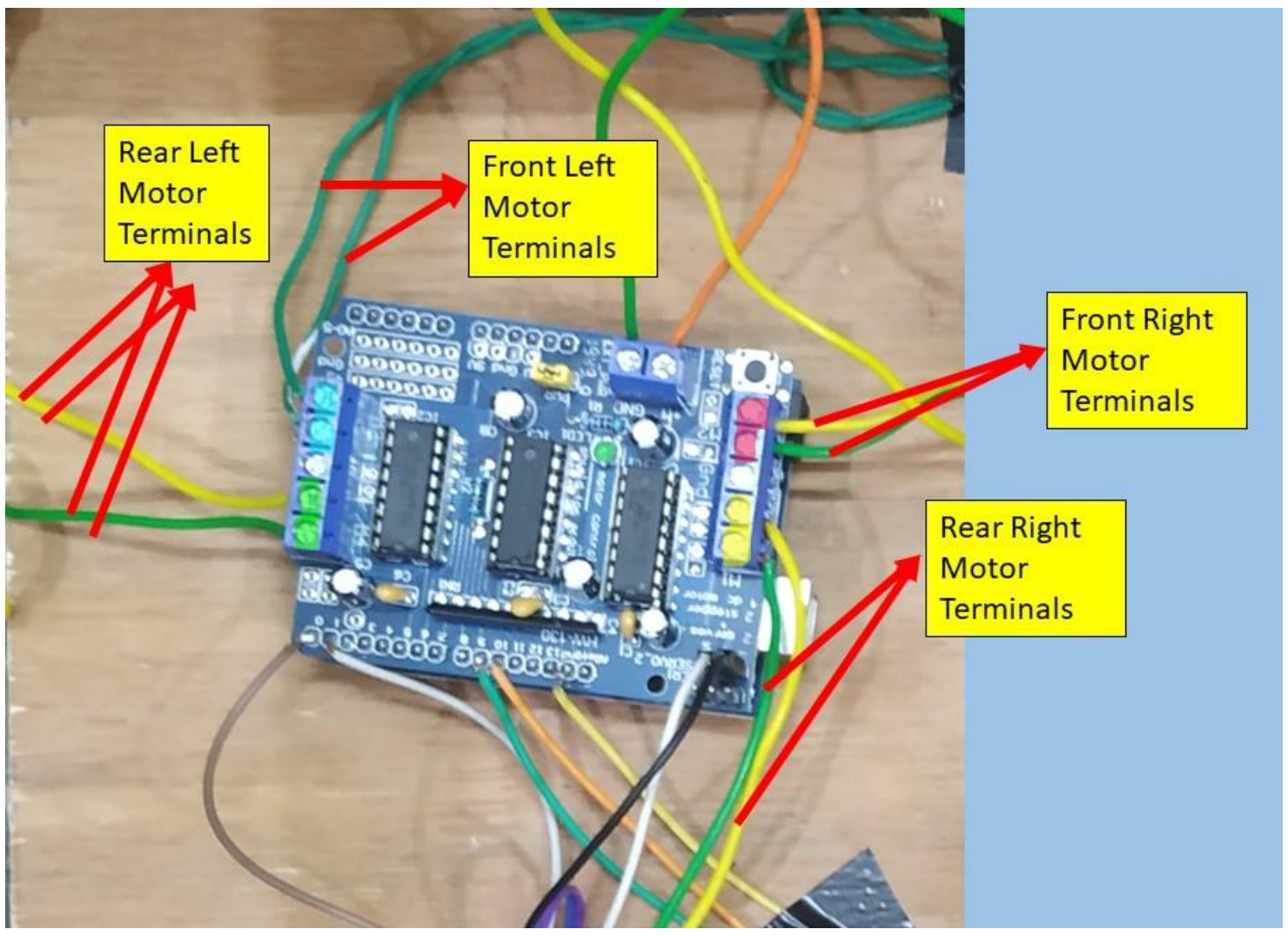

*Fig 6.22: Circuit 3: (Drive motor connections)*

Now the pump and the mower have to be operated independently but on command. This can be achieved by connecting positive terminals of Pump and Motor to any two of the D series (i.e. D9, D10, D11….so on) which are present on the Arduino Board, and the negative terminals of the pump and the Mower are to be connected to the ground of the Arduino board.

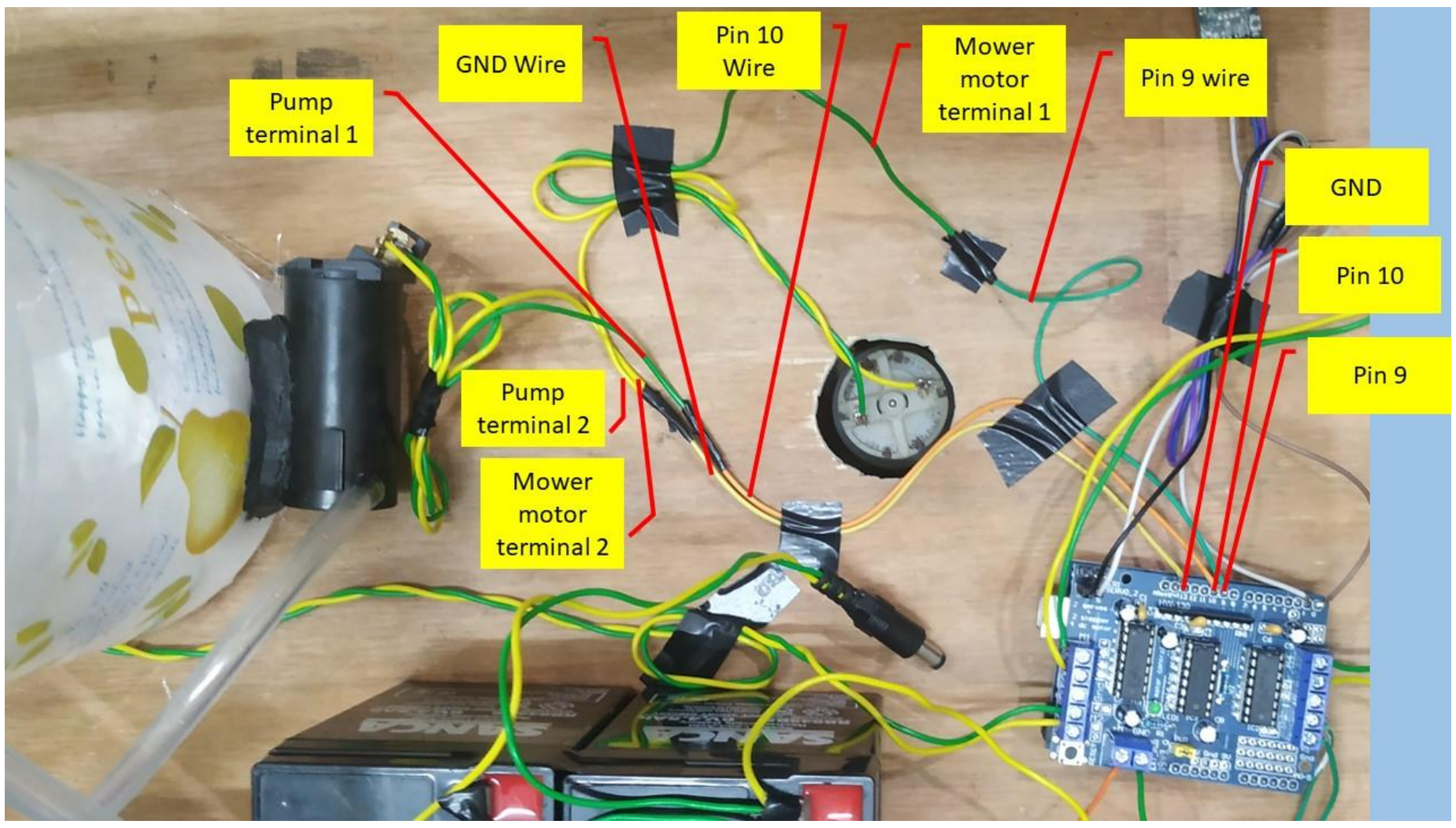

*Fig 6.23: Circuit 4: (Mower and Pump connections)*

For the chosen battery charging and discharging are from the same two terminals, so a diode is provided for the recharging of Battery through solar panel. It is placed on top of the positive terminal of the battery. This diode also makes sure that the battery does not overcharge. Now the power supply connections have to be made to Arduino and Motor shield so that they can distribute the power to all the electrical components and they run according to the input commands. A switch is provided to ON/OFF the circuit whenever needed.

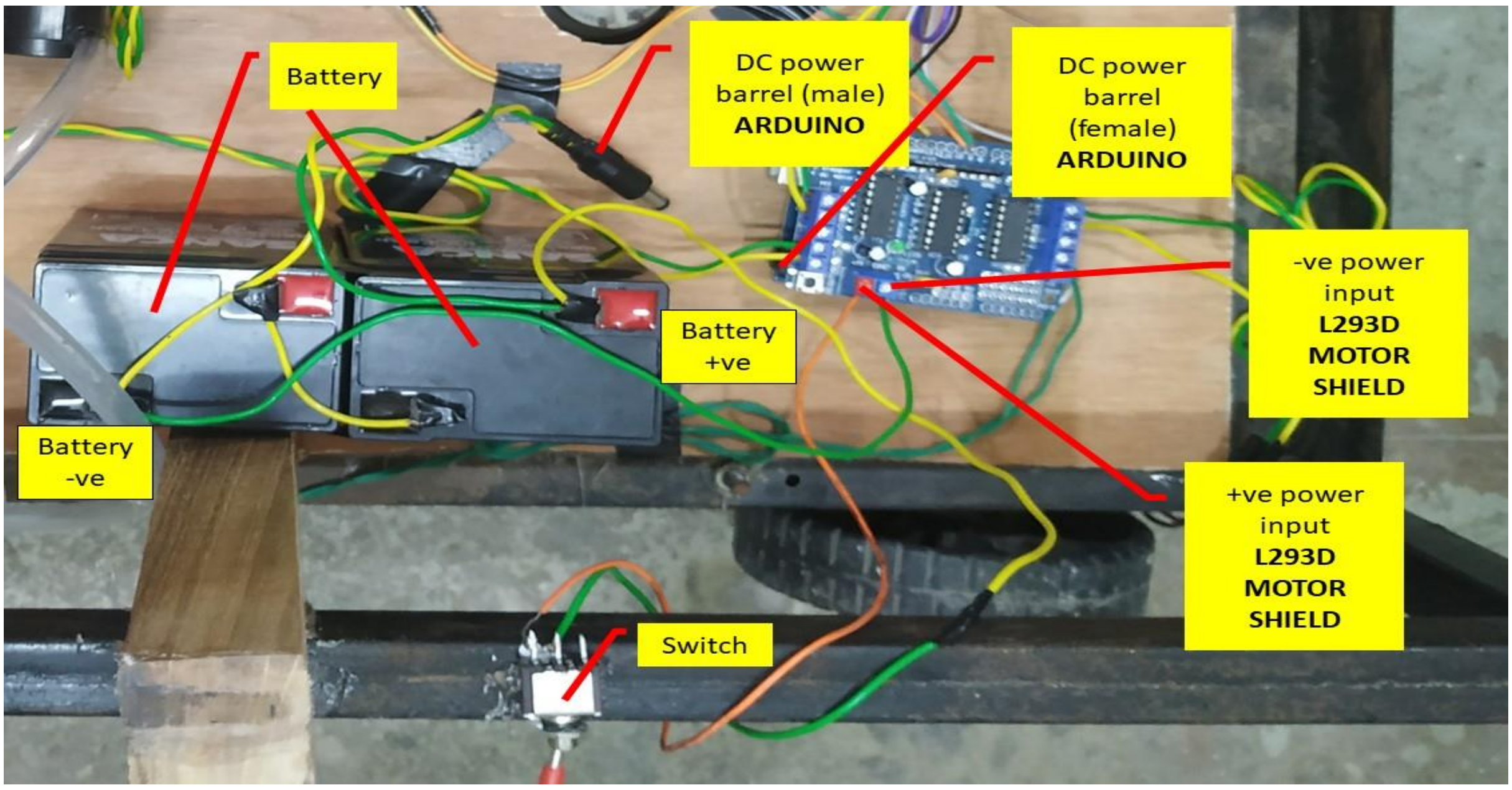

*Fig 6.24: Circuit 5: (Power supply connections)*

From the above circuit it is clear that both Arduino and Motor shield have separate terminals for Power supply input to them. So, both of them are powered individually. Generally, in small circuits the power supply to Motor shield alone would run the Arduino board as well but, in our application, there are a lot of components to be powered so it is necessary that the Motor shield and Arduino are powered individually and simultaneously.

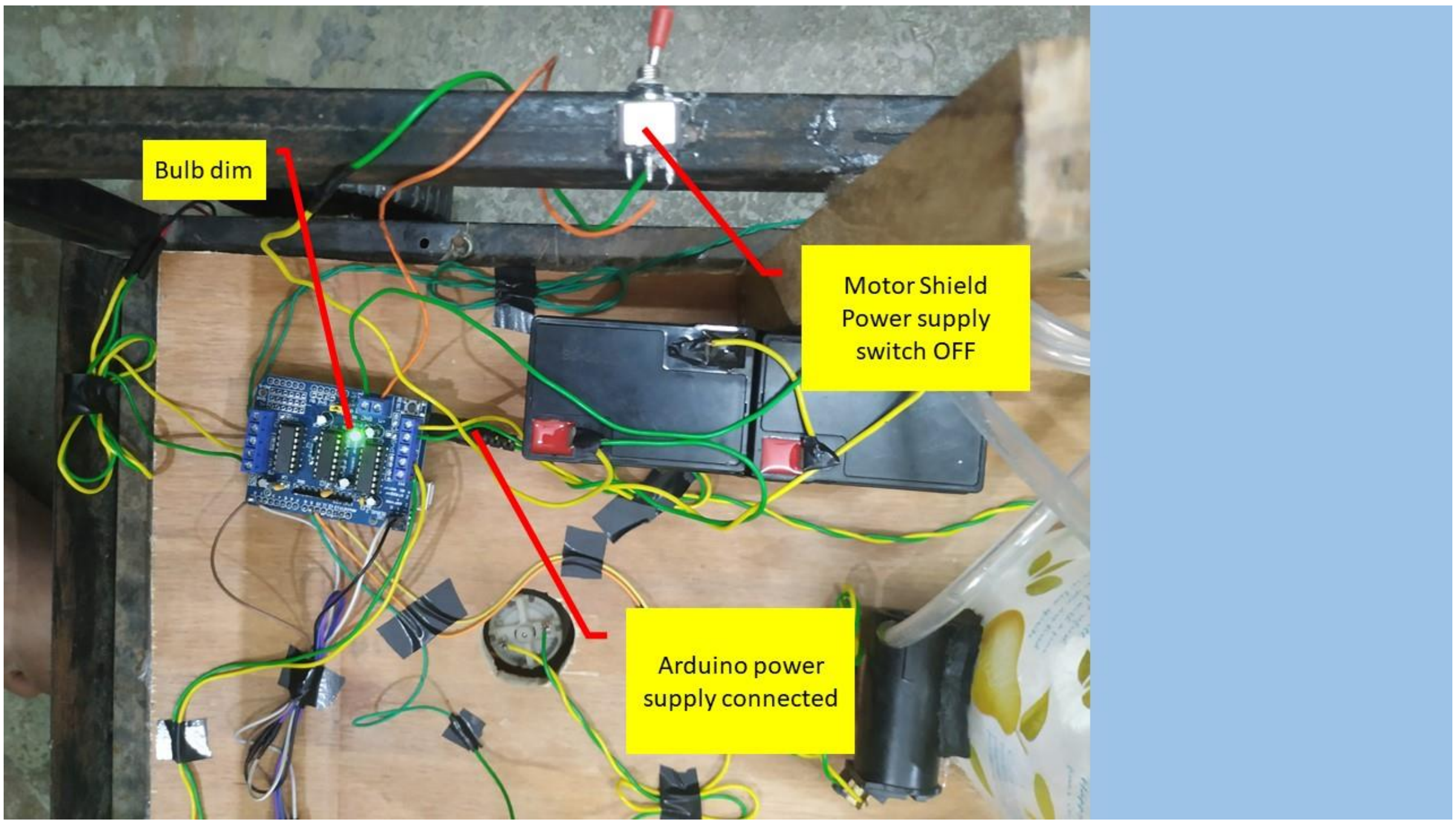

*Fig 6.25: Circuit 6: (Motor shield OFF and Arduino ON)*

From the above circuit we can see that only Arduino power supply is connected, so a dim light glows in the Motor shield. Motor shield power supply is cut off by the switch (OFF Position)

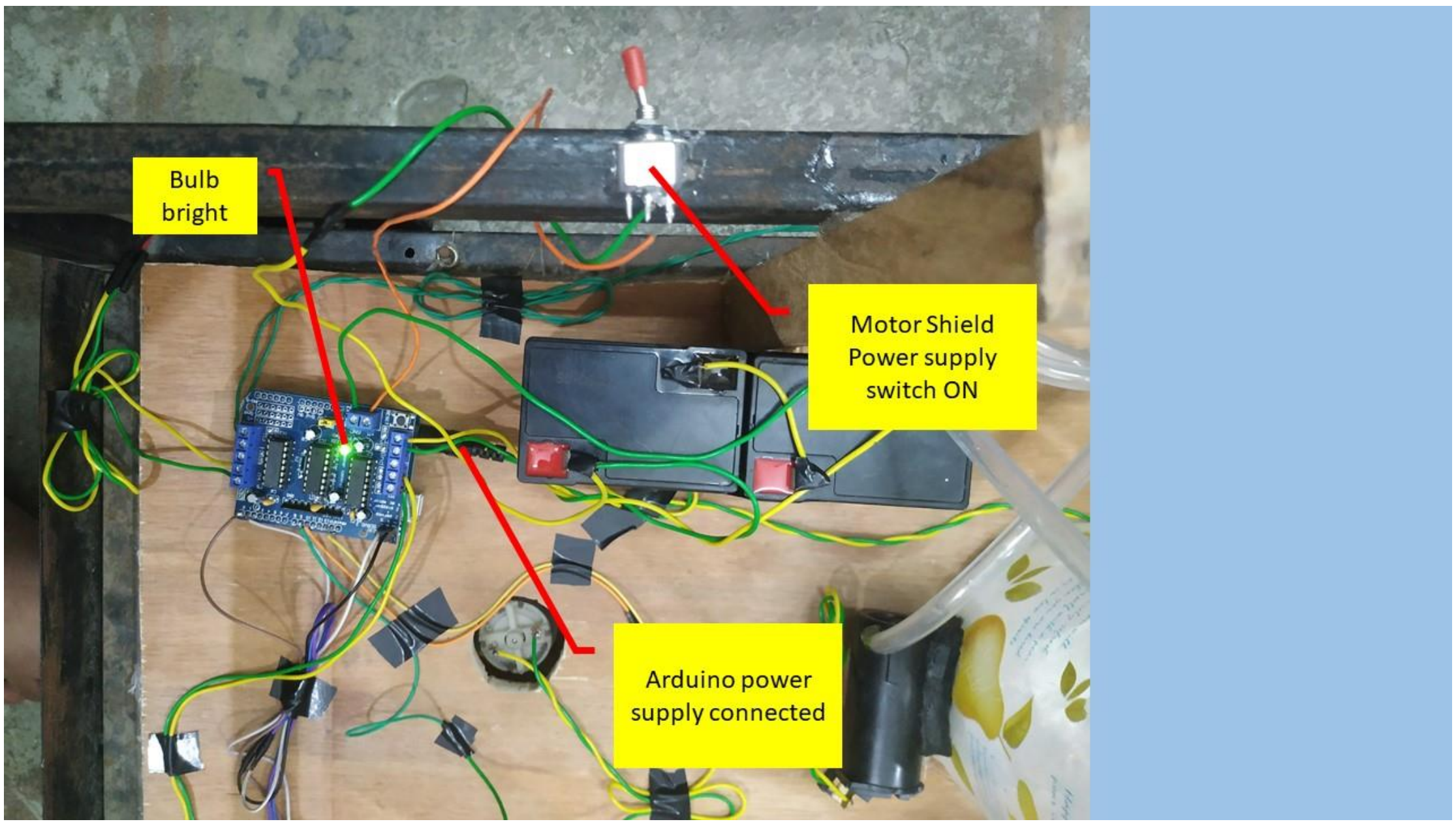

*Fig 6.26: Circuit 26: (Both Motor shield and Arduino ON)*

From the above circuit we can see that both Arduino and Motor power supply are connected, so a bright light glows in the Motor shield. Now the switch in in ON position.

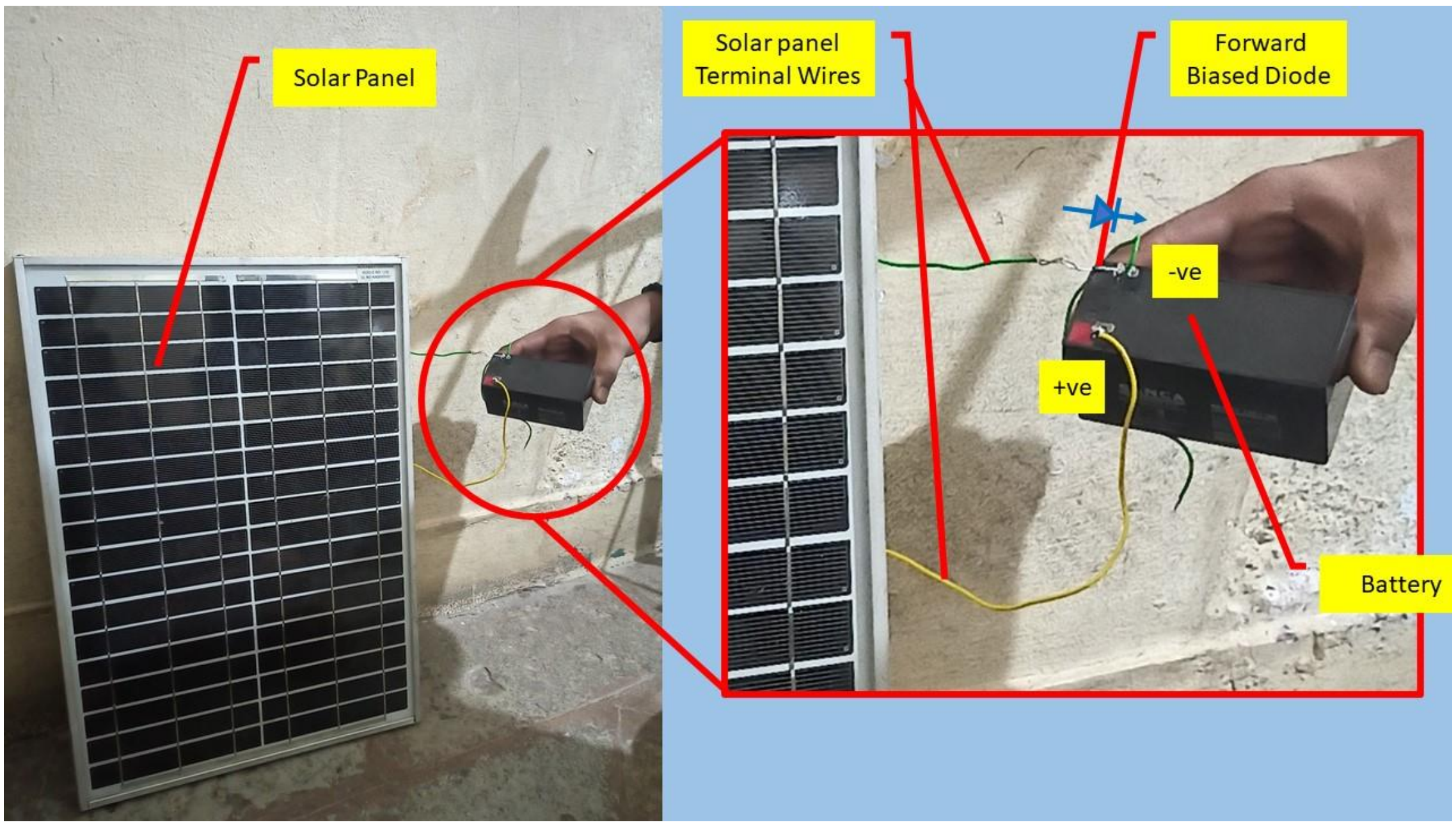

*Fig 6.27: Circuit 8: (Solar Panel and Battery)*

From the above diagram, we can see the connection of solar panel to the battery. A diode is provided in forward biased from solar panel to battery to prevent the back flow of current from battery to solar panel.

### *6.3 Program for the robot*

Before giving any connections to the Arduino and Motor shield, first compile a code in a PC in an app called ARDUINO IDE. Compile a code for the required application without any errors. Then plug in the Arduino board with the PC and upload the code.

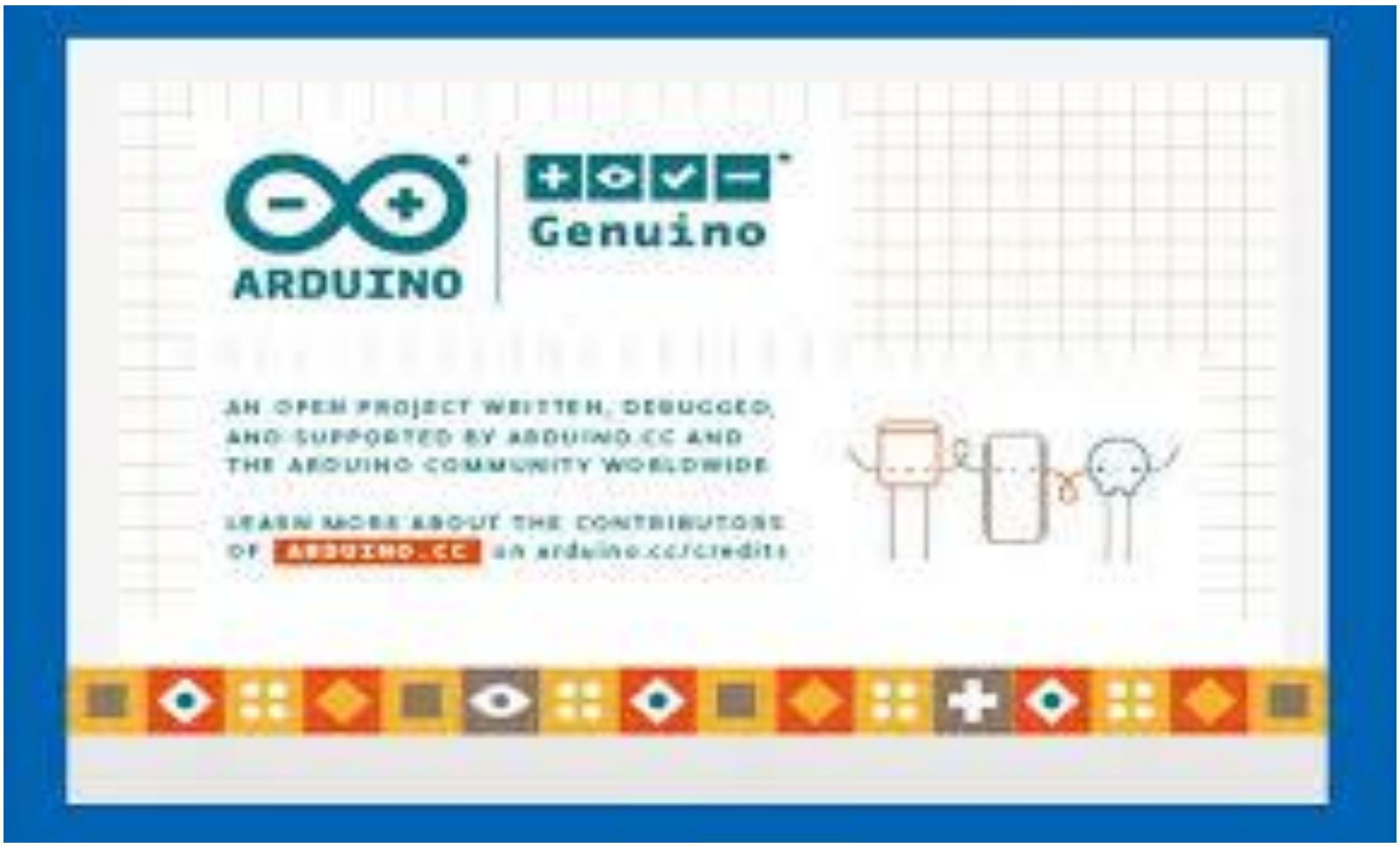

*Fig 6.28:  Arduino IDE Software*

Agrobot_sprayer_mower

```
#include <AFMotor.h>
#define mower_FR 9 //Mower Digital pin 9
#define pump_RE 10 //Pump Digital pin 10
boolean mower = false;
boolean pump = false;
//initial motors pin
AF_DCMotor motor1(1, MOTOR12_1KHZ);
AF_DCMotor motor2(2, MOTOR12_1KHZ);
AF_DCMotor motor3(3, MOTOR34_1KHZ);
AF_DCMotor motor4(4, MOTOR34_1KHZ);
char command;
void setup()
{
  pinMode(mower_FR, OUTPUT);
  pinMode(pump_RE, OUTPUT);
  Serial.begin(9600); //Set the baud rate to your Bluetooth module.
}
void loop()
{
  if (Serial.available() > 0) {
  command = Serial.read();
  Stop(); //initialize with motors stoped
  if (mower) {
  digitalWrite(mower_FR, HIGH);
  }
  if (!mower) {
  digitalWrite(mower_FR, LOW);
  }
  if (pump) {
  digitalWrite(pump_RE, HIGH);
  }
  if (!pump) {
  digitalWrite(pump_RE, LOW);
  }
```

```
//Change pin mode only if new command is different from previous.
//Serial.println(command);
  switch (command) {
    case 'F':
      forward();
      break;
    case 'B':
      back();
      break;
    case 'L':
      left();
      break;
    case 'R':
      right();
      break;
    case 'W':
      mower = true;
      break;
    case 'w':
      mower = false;
      break;
    case 'U':
      pump = true;
      break;
    case 'u':
      pump = false;
      break;
    }
  }
}
void forward()
{
  motor1.setSpeed(255); //Define maximum velocity
  motor1.run(FORWARD); //rotate the motorclockwise
  motor2.setSpeed(255); //Define maximum velocity
  motor2.run(FORWARD); //rotate the motorclockwise
  motor3.setSpeed(255);//Define maximum velocity
  motor3.run(FORWARD); //rotate the motorclockwise
  motor4.setSpeed(255);//Define maximum velocity
  motor4.run(FORWARD); //rotate the motorclockwise
}
void back()
{
  motor1.setSpeed(255); //Define maximum velocity
  motor1.run(BACKWARD); //rotate the motoranti-clockwise
  motor2.setSpeed(255); //Define maximum velocity
  motor2.run(BACKWARD); //rotate the motoranti-clockwise
  motor3.setSpeed(255); //Define maximum velocity
  motor3.run(BACKWARD); //rotate the motoranti-clockwise
  motor4.setSpeed(255); //Define maximum velocity
  motor4.run(BACKWARD); //rotate the motoranti-clockwise
}
void left()
{
  motor1.setSpeed(255); //Define maximum velocity
  motor1.run(BACKWARD); //rotate the motoranti-clockwise
  motor2.setSpeed(255); //Define maximum velocity
  motor2.run(BACKWARD); //rotate the motoranti-clockwise
  motor3.setSpeed(255); //Define maximum velocity
  motor3.run(FORWARD); //rotate the motorclockwise
  motor4.setSpeed(255); //Define maximum velocity
  motor4.run(FORWARD); //rotate the motorclockwise
}
```

```
void right()
{
  motor1.setSpeed(255); //Define maximum velocity
  motor1.run(FORWARD); //rotate the motorclockwise
  motor2.setSpeed(255); //Define maximum velocity
  motor2.run(FORWARD); //rotate the motorclockwise
  motor3.setSpeed(255); //Define maximum velocity
  motor3.run(BACKWARD); //rotate the motoranti-clockwise
  motor4.setSpeed(255); //Define maximum velocity
  motor4.run(BACKWARD); //rotate the motoranti-clockwise
}
void Stop()
{
  motor1.setSpeed(0); //Define minimum velocity
  motor1.run(RELEASE); //stop the motor whenrelease the button
  motor2.setSpeed(0); //Define minimum velocity
  motor2.run(RELEASE); //rotate the motor clockwise
  motor3.setSpeed(0); //Define minimum velocity
  motor3.run(RELEASE); //stop the motor whenrelease the button
  motor4.setSpeed(0); //Define minimum velocity
  motor4.run(RELEASE); //stop the motor whenrelease the button
}
```

Note: While uploading the program to the Arduino board make sure that RX and TX connections between Bluetooth module and Arduino board are removed.

### *6.4 Control of the robot*

The robot is controlled with the help of a smartphone application. The smartphone application comprises of many keys which help us to move the robot in any direction and also helps us to control other electrical equipment like Pump and Mower with just a tap on the keys. The keys present on the smartphone application are listed below.

| | | | |
|---|---|---|---|
| Forward | Reverse | Left | Right |
| Pump: ON/OFF | Mower: ON/OFF | Speed control slider | Stop |

The smartphone application is connected to the Arduino board through the Bluetooth module. So, when we tap on the keys, data from the smartphone is sent to the Arduino board via Bluetooth. Then, the Arduino on receiving the above data operates the robot through the motor driver circuit. The robot is charged with Battery which in turn is charged by solar power through the solar panel system.

| MATERIAL | QUANTITY |
| --- | --- |
| The chassis (Mild Steel) | 1 |
| Wooden plank (Base) | 1 |
| Wooden rods (Solar panel supports) | 2 |
| 12V DC Motors with Nuts | 5 |
| Front Motor Holders | 2 |
| Rear Motor Holders (Brackets) | 2 |
| Wheels | 4 |
| Long PVC Pipe (Dia 1 inch) | 1 |
| Long PVC pipe (Dia 1.05 inch) | 1 |
| Spray Nozzles | 2 |
| Tubing (3m long) | 1 |
| Y joint with straws | 1 |
| 12V DC Pump | 1 |
| Container (1 Litre capacity) | 1 |
| Cutting blade | 1 |
| Arduino Board | 1 |
| L293D Motor Control Shield | 1 |
| HC-05 Bluetooth Module | 1 |
| Electrical Wires (5m long) | 1 |
| Rubber washers (Internal dia 0.99 in) | 3 |
| Screws (M08) | 3 |
| Nuts (8mm) | 5 |
| Switch | 1 |
| Solar Panel | 1 |
| Forward Biased Diode | 1 |
| 6V Sealed Lead-Acid Battery (4.5Ah) | 2 |
| Custom Made Stoppers (PVC) | 2 |

*Table 6.1 Bill of Materials*

# 7. WORKING

## Manual Operation

First make sure that the program for the robot is successfully uploaded into the Arduino board and all the connections between electrical components and the Arduino are given exactly as mentioned in the above chapter.

Then install the RC Bluetooth application from the App store of the smartphone.

Switch ON the connection between Battery and the Arduino board. On doing this a bright green light will appear on the Arduino board and a slow blinking red light will appear on the Bluetooth module. This means that the Arduino board, Motor shield and the Bluetooth module are now active.

Now open the RC Bluetooth car app in the smartphone and open settings. Select the option that says "Connect to car". Then the Bluetooth module HC-05 will be visible. Click on that to pair the smartphone with the robot. When the robot is paired the blinking of the red light on the Bluetooth module will be faster.

Before giving any movement to the robot, adjust the vertical, horizontal actuators and nozzle according to the requirements of the field.

For the movement of robot keys are provided.

When the "Forward" key is pressed then all the 4 DC motors rotate in the same direction (assume clockwise) which in turn rotates the wheels and the robot moves forward.

When the "Reverse" key is pressed then all the 4 DC motors rotate in the same direction (but this time assume anti-clockwise) which in turn rotates the wheels and the robot moves reverse.

When the "Left" key is pressed then the front and rear DC Motors on the right side of the chassis move faster than the left ones and thus the robot moves Left.

When the "Right" key is pressed then the front and rear DC Motors on the left side of the chassis move faster than the right ones and thus the robot moves Right.

Using the keys mentioned above the robot is placed at such a position that when the pump is switched ON using the “Pump: ON/OFF” key, the nozzle sprays the pesticide directly onto the affected plants. By using the movement keys the whole field can be covered with pesticides.

When the spraying of pesticides is done, the pump is switched OFF using the same key.

The Mowing or Harvesting operation can also be performed. Using the movement keys mentioned above the robot is placed at such a position that when the Mower is switched ON using the “Mower: ON/OFF” key, the unwanted weeds or harvested crops are cut off by the blades of the mower. The clearance of the blade from the ground is adjustable. So, we can control the level of grass field to be maintained if needed. By using the movement keys the crops or weeds in the whole field can be Mowed.

When the Mowing operation is done, the Mower is switched OFF using the same key.

When all the operations are done, the robot is called back and the connection between the Battery and the Arduino board is switched OFF and the robot is preferably placed in an environment with sunlight so that the battery can be recharged.

# 8. EXPERIMENTAL ANALYSIS

In order to evaluate the performance of the system, several trials were performed on the robot. We have evaluated different functions that are to be performed by the robot. The functions are listed below.

1. Pesticide spraying
2. Mowing operation
3. Robot’s movement
4. Battery performance

The results of the tests for each function are given below in detail.

***8.1 Pesticide spraying***

Pesticide spraying has two aspects,

1. The distances and areas covered by the actuation setup
2. The distances and areas covered by the sprayer

*8.1.1The distances and areas covered by the actuation setup:* The position of the Nozzle is with the help of the combination of Horizontal and Vertical actuation, this combination is referred as the actuation setup. This is done so as to increase the work volume of the robot. The actuation setup is designed in such a way that it has 4 Degrees of Freedom, viz

1. Vertical actuation on Y Axis
2. Horizontal sliding motion provided for Nozzle on X Axis
3. Rotation of the Nozzle in the YZ Plane about X axis
4. Rotation of the Horizontal actuating setup in XZ plane about Y axis

The distances and areas covered in each degree of freedom is discussed below.

*8.1.1.a.) In vertical actuation:* The distance covered is only Y axis. The distances are given by us after the consideration of various crop lengths.

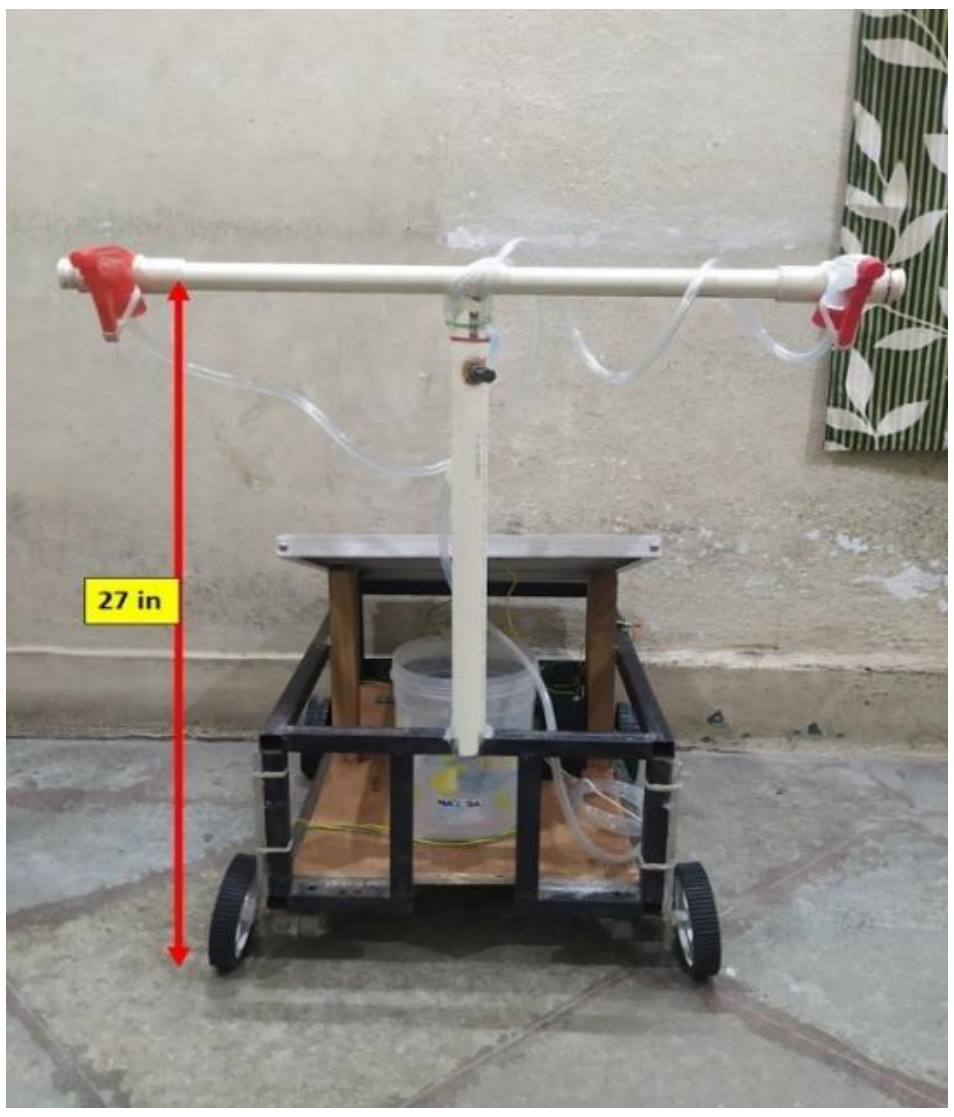

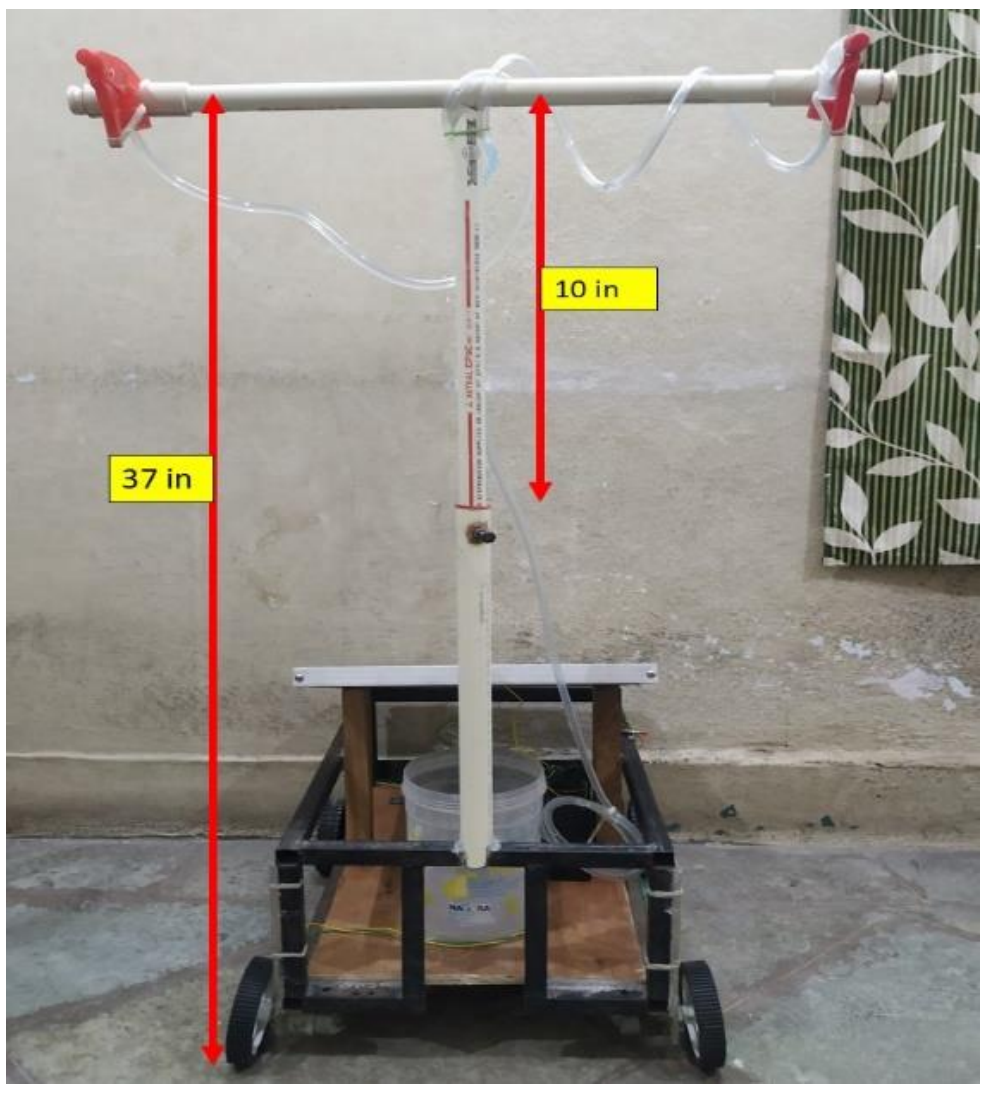

*Fig 8.1: Minimum Vertical Actuation*　　*Fig 8.2: Maximum Vertical Actuation*

The maximum height of the nozzle from the ground the nozzle spray can reach is 56.8 inches. Actuation height is 10 inches. The minimum height of the nozzle from the ground the nozzle spray can reach is 46.8 inches.

*8.1.1.b.) In Horizontal sliding motion provided for Nozzle on X Axis:* The distance covered is only X axis.

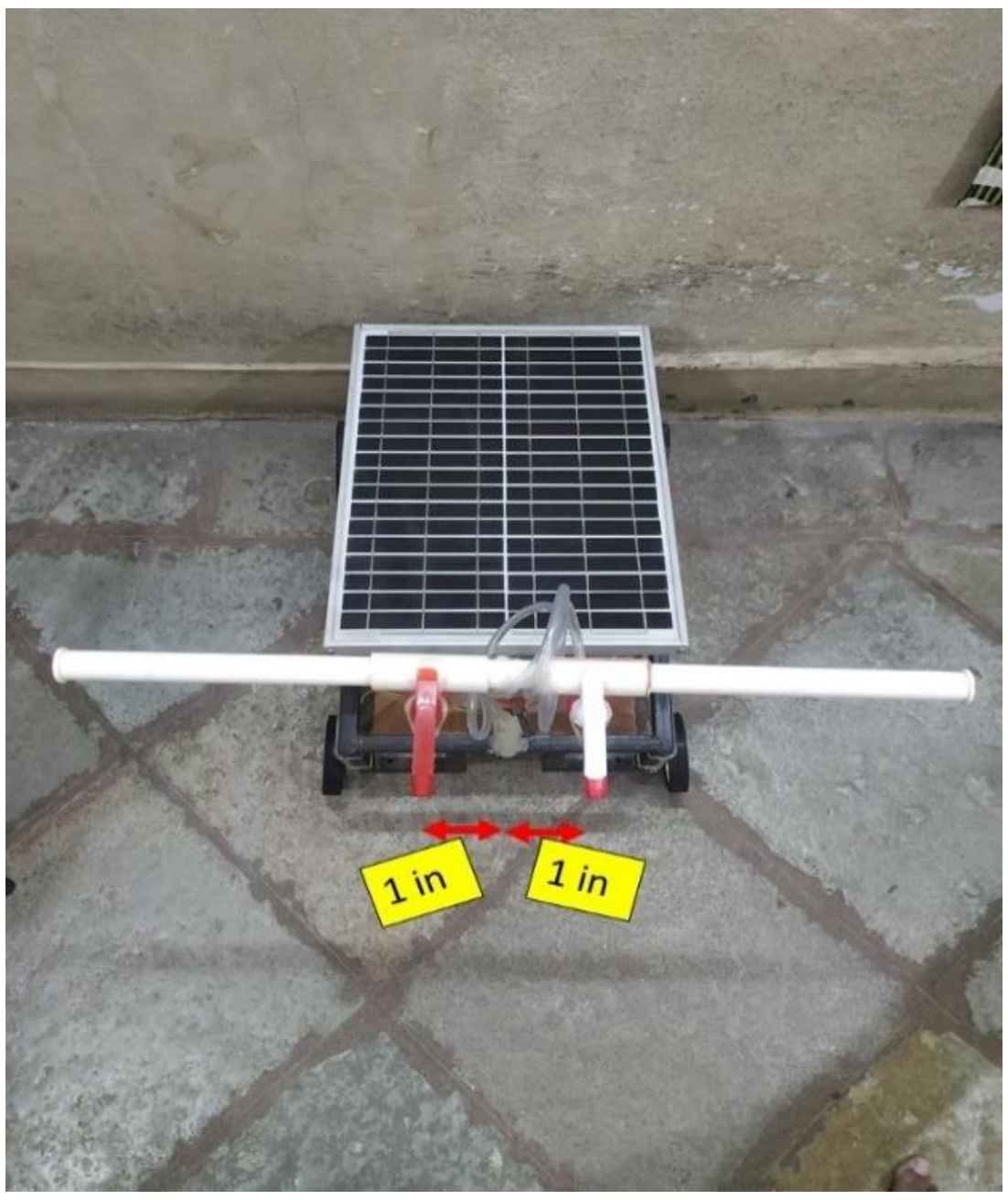

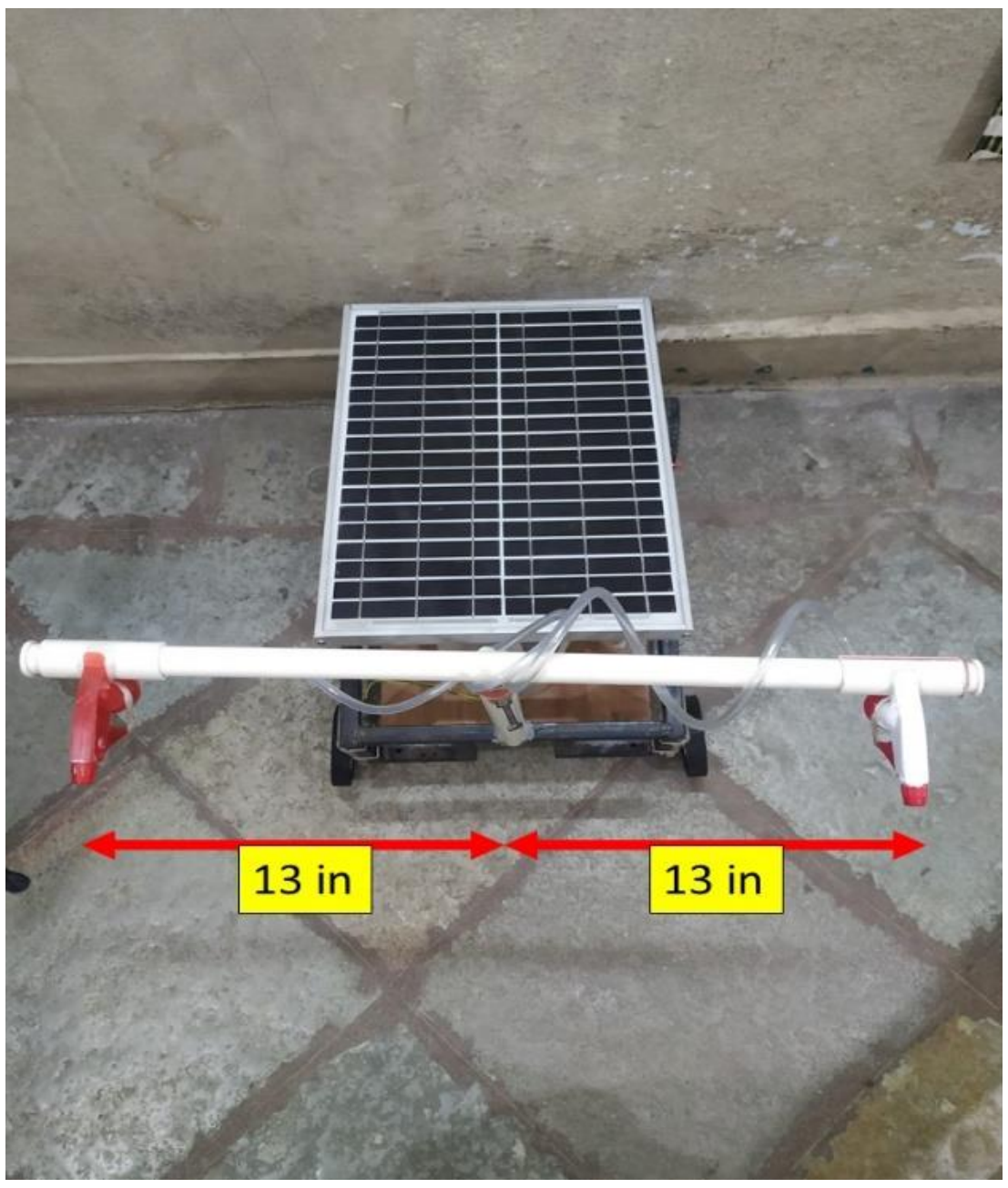

*Fig 8.3: Minimum Horizontal Actuation*　　*Fig 8.4: Maximum Horizontal Actuation*

The maximum length the nozzle can reach from the center of the chassis is 32.6 inches. Actuation length is 20.1 inches. The Zero position of the Nozzle is located at a distance of 12.5 inches from the center of the chassis.

*8.1.1.c.) Clearance of the robot:*

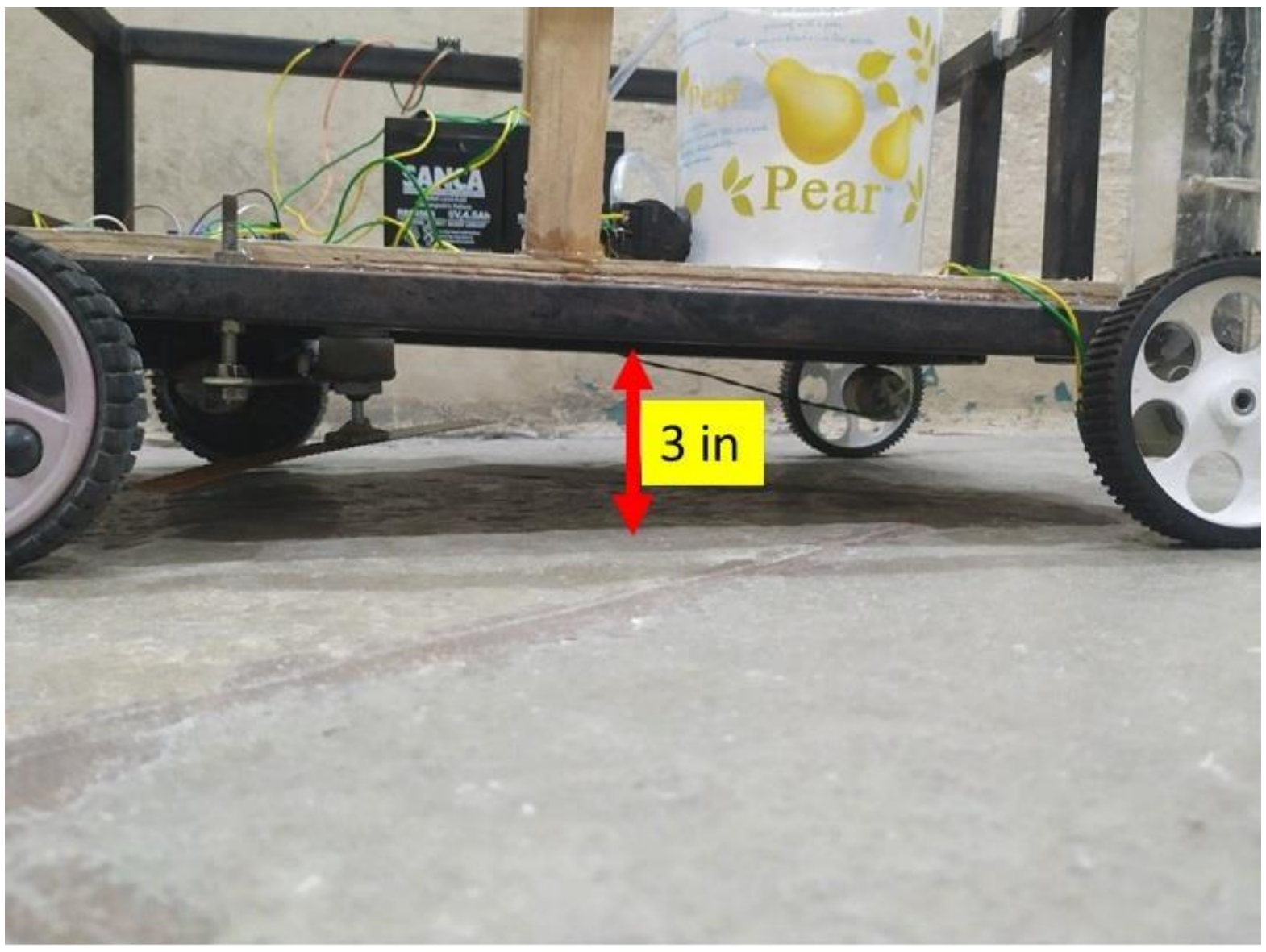

*Fig 8.5: ground clearance of chassis*

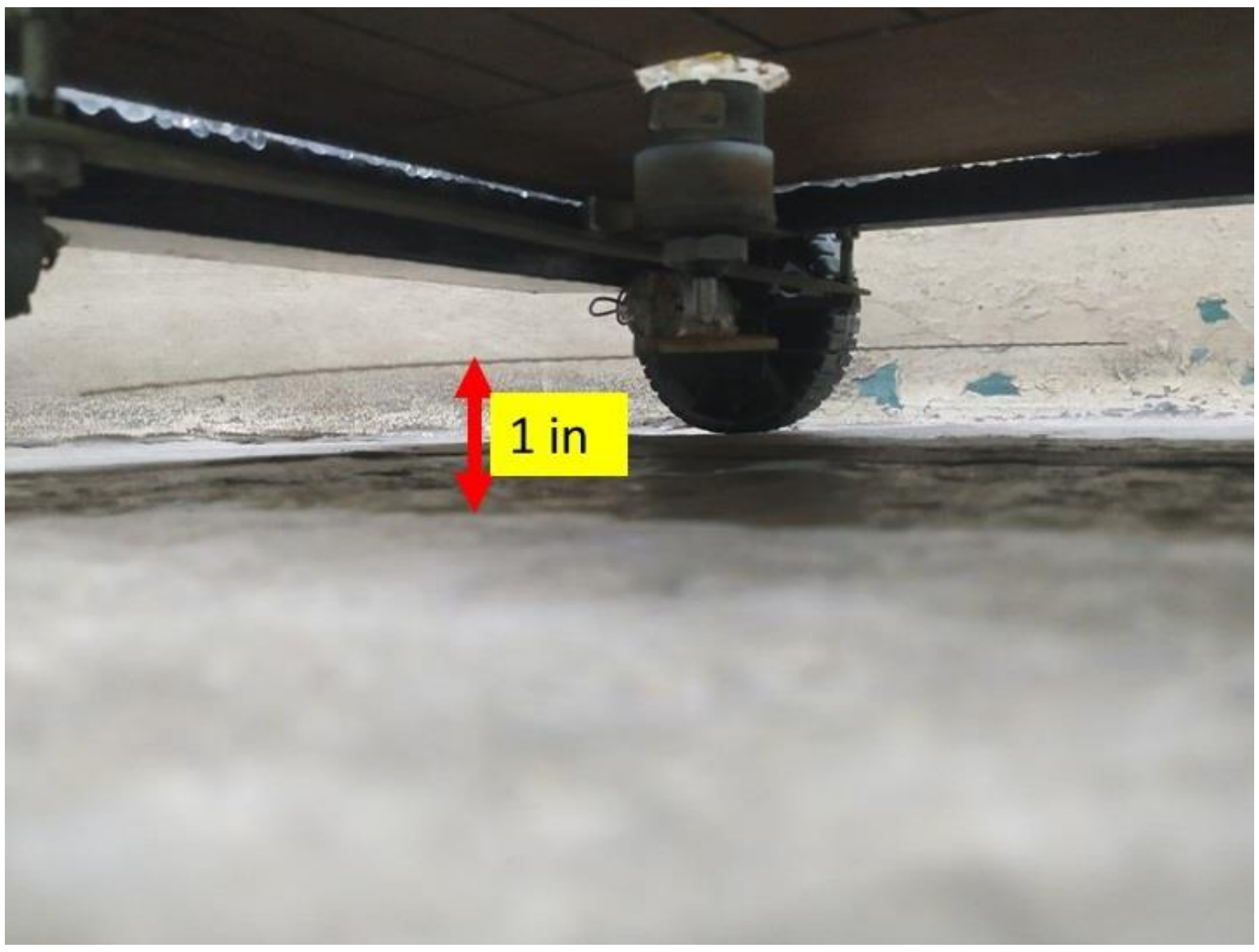

*Fig 8.6: ground clearance of mower blade*

8.1.1.d.) In Rotation of the Horizontal actuating setup in XZ plane about Y axis.

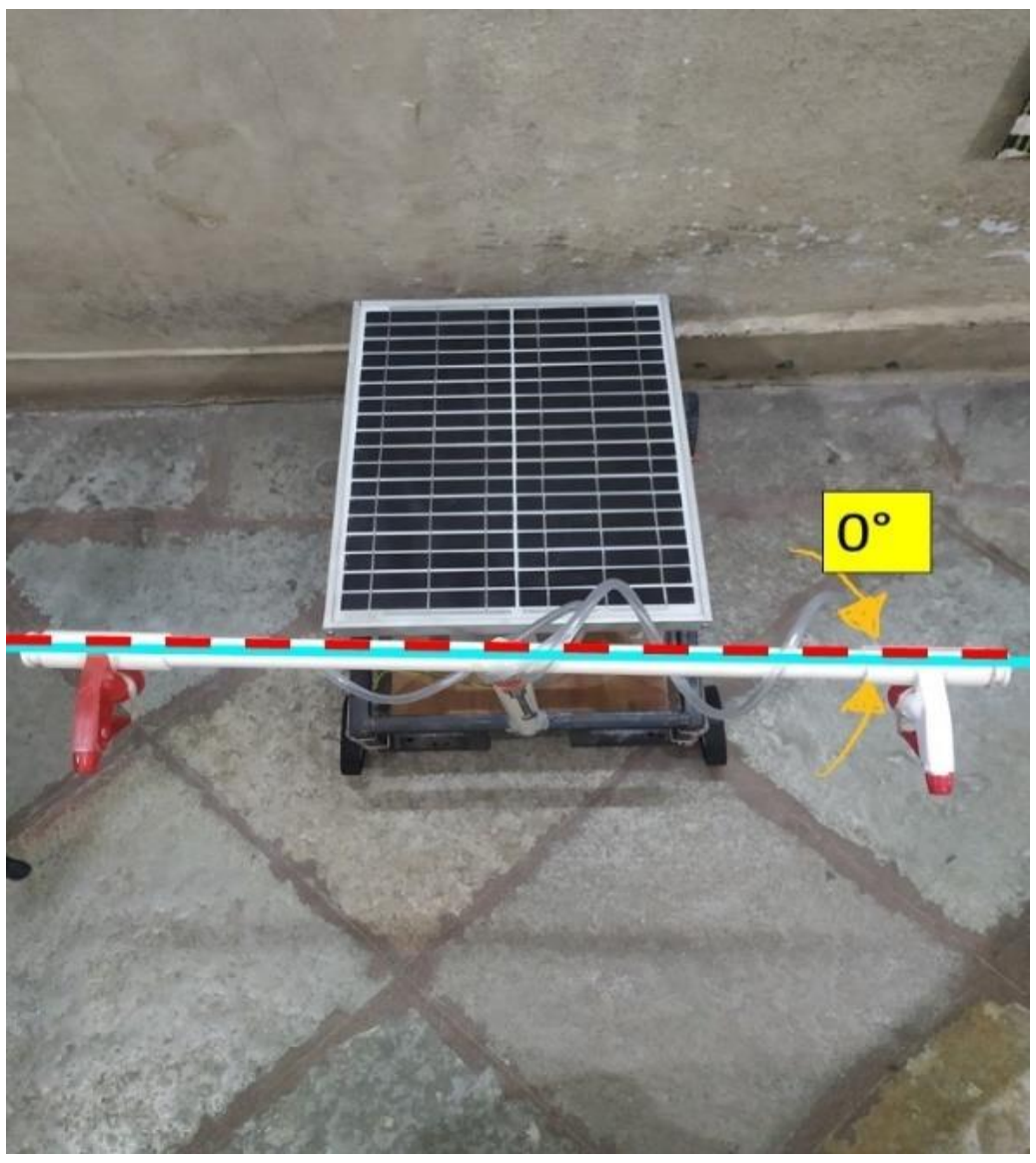

Fig 8.7: $0^o$ Rotation of Actuator in XZ Plane w.r.t Y axis

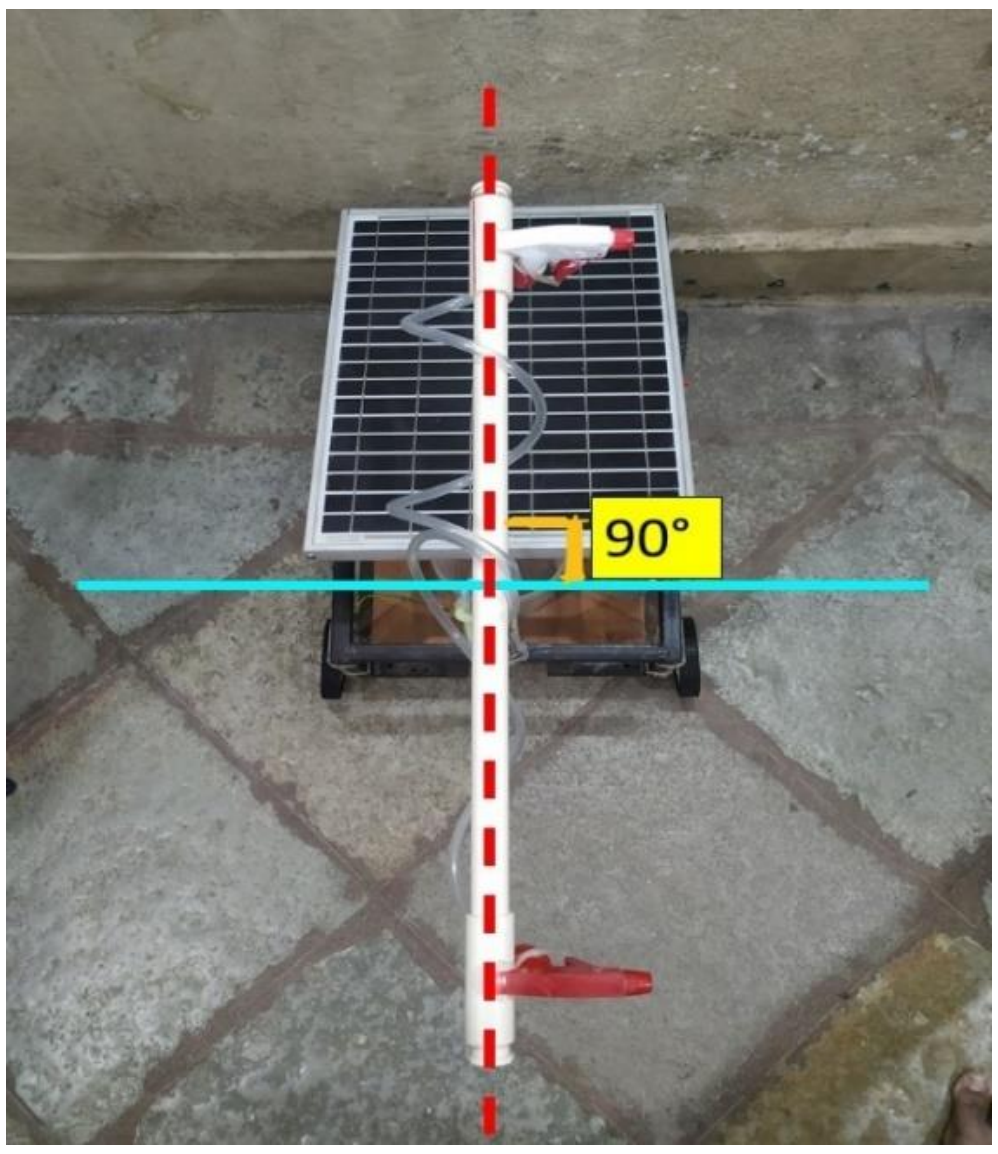

Fig 8.8: $90^o$ Rotation of Actuator in XZ plane w.r.t Y axis (Anti-clock wise)

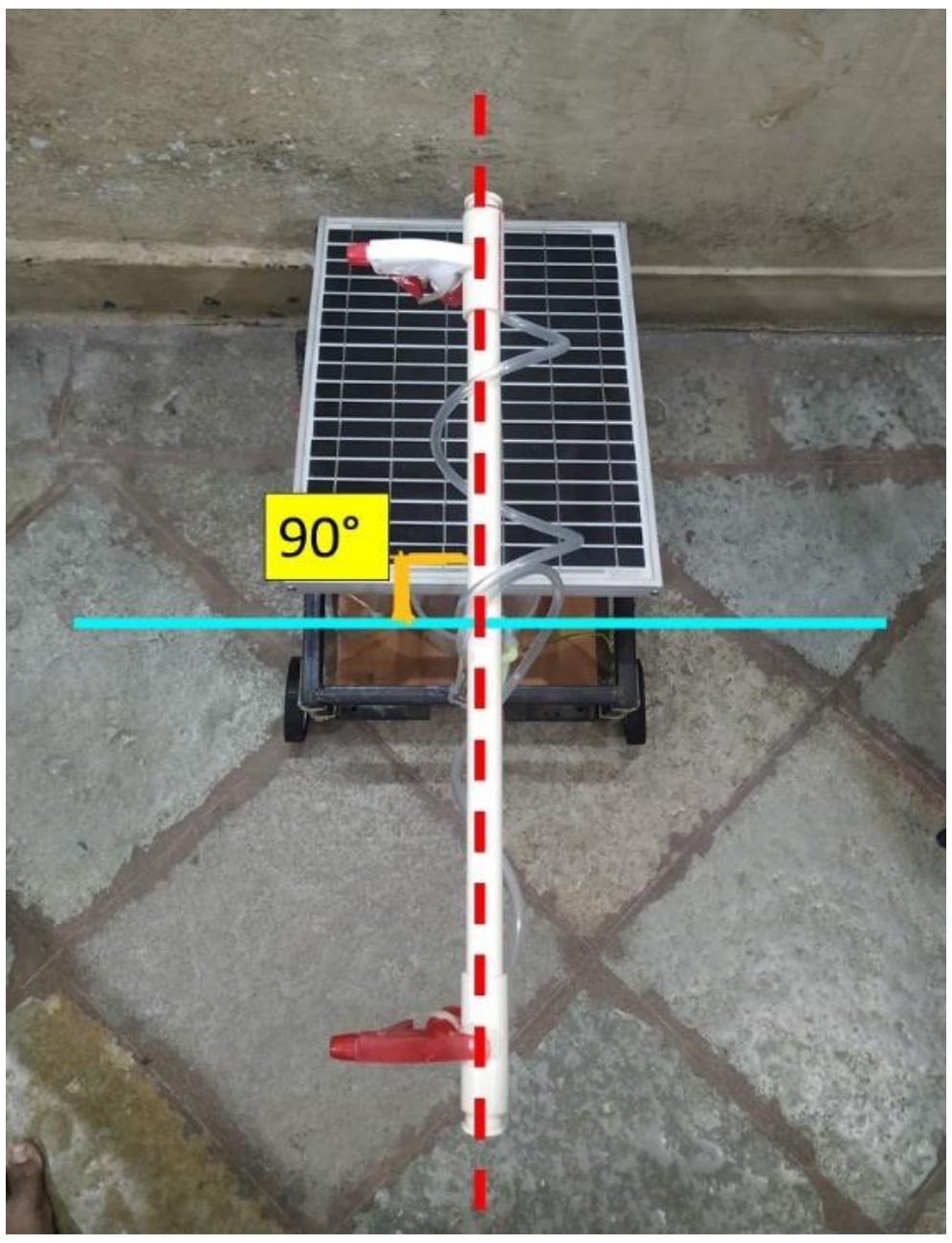

Fig 8.9: $90^o$ Rotation of Actuator in XZ Plane w.r.t Y axis (Clockwise)

In the figure shown below, the smaller circle represents the area that is not covered by the nozzle. Let's assume this to be A1. The two dots on the smaller circle represent the Zero position of the nozzle on the horizontal actuator.

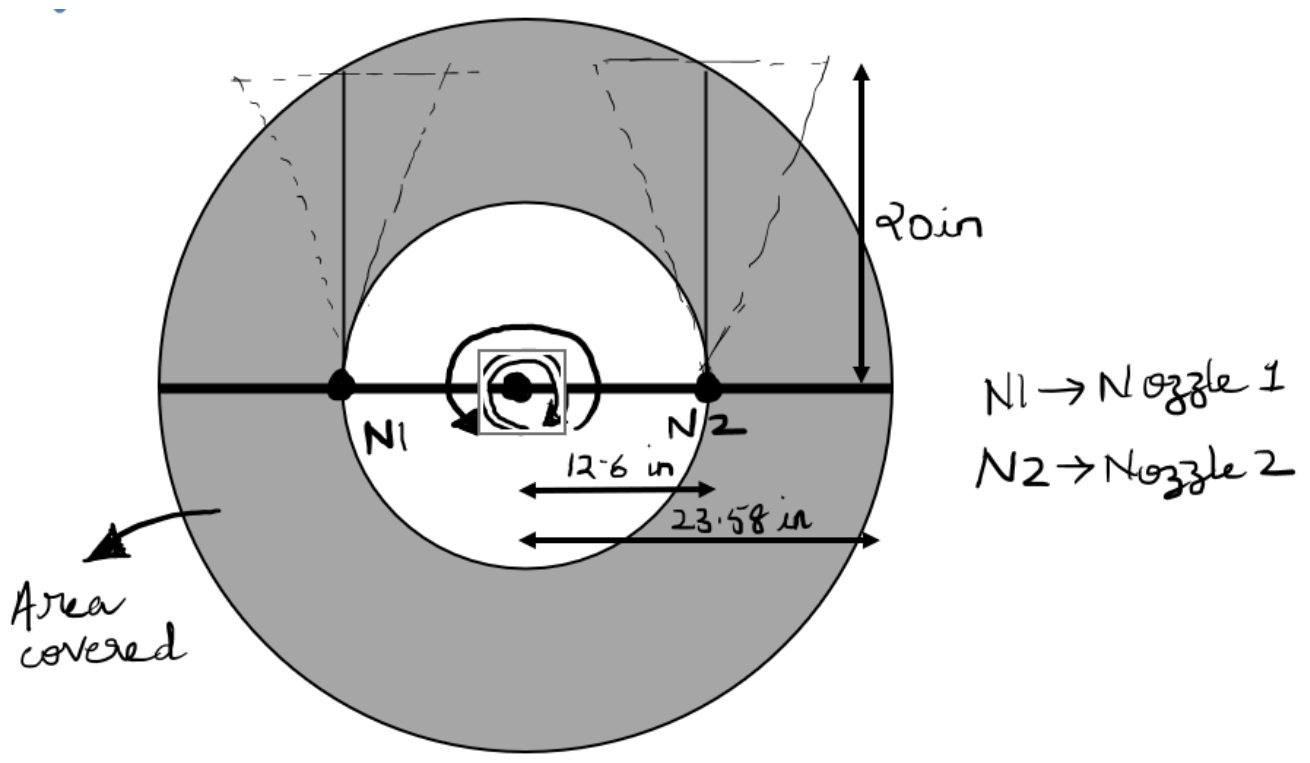

*Fig 8.10: Area Covered*

The larger circle represents the area covered by the nozzle when the nozzle is kept at the ends of Horizontal actuator. Let's assume this area to be A2. Therefore, the area covered by the nozzle during the rotation of the Horizontal actuating setup in XZ plane about Y axis is given by A2 – A1. Let's assume that to be A3.

Then A3 = A2-A1 = $\pi(32.6)^2$ - $\pi(12.5)^2$

Therefore A3 = 2840 sq. in

<u>*8.1.1.e.) In Rotation of the Nozzle in the YZ Plane about X axis:*</u> Maximum rotation of the spray nozzle allowed is 80 degrees in this scenario as any further rotation is obsolete.

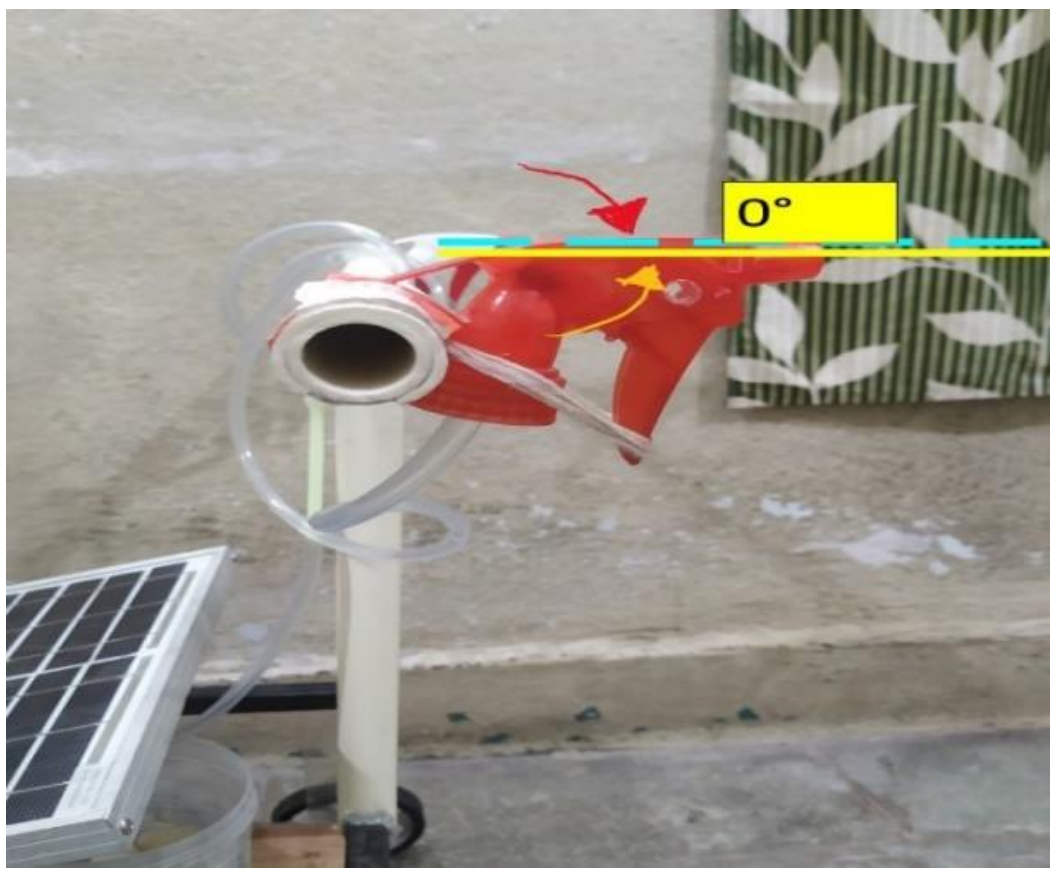

*Fig 8.11: $0^o$ Rotation of Nozzle in the YZ plane w.r.t X axis*

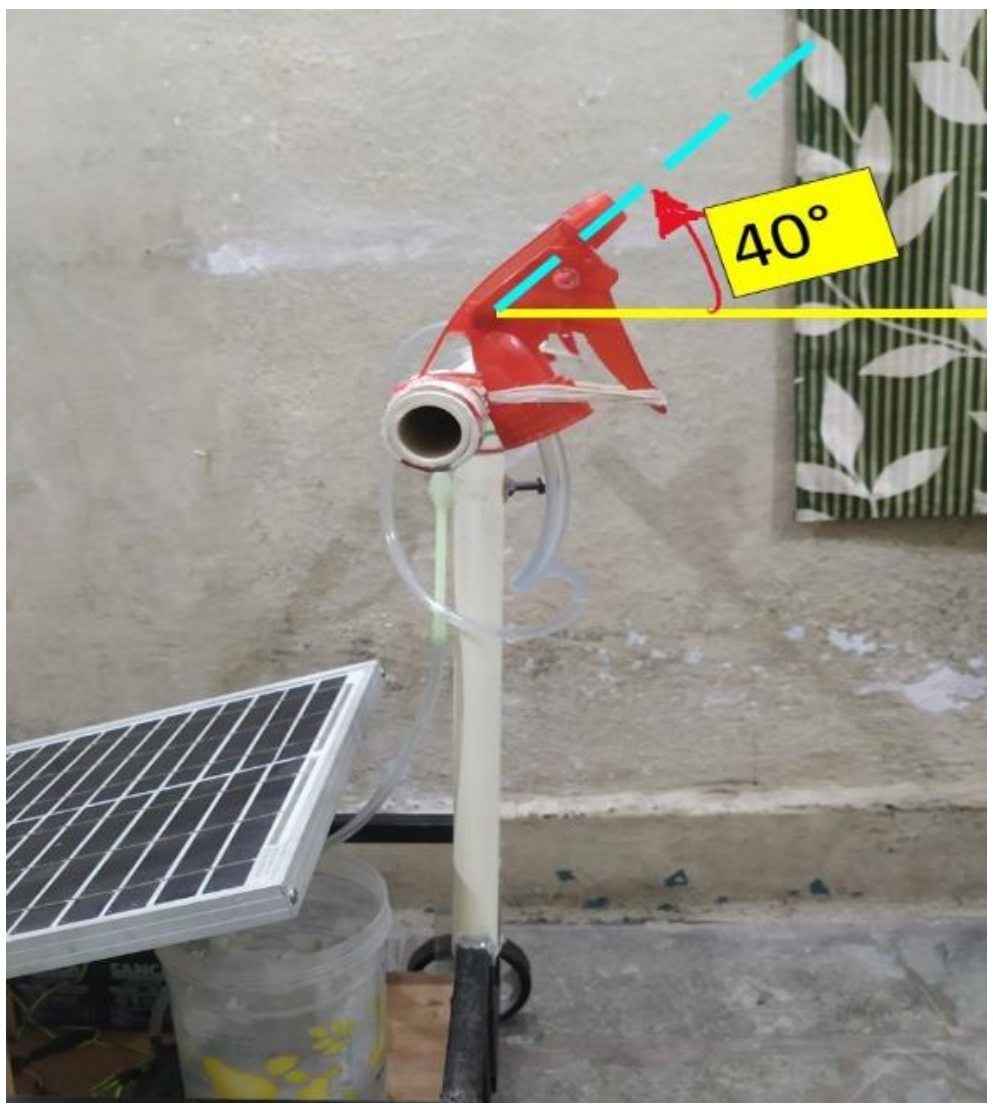

*Fig 8.12: 40° Upwards Rotation of Nozzle in the YZ Plane w.r.t X axis*

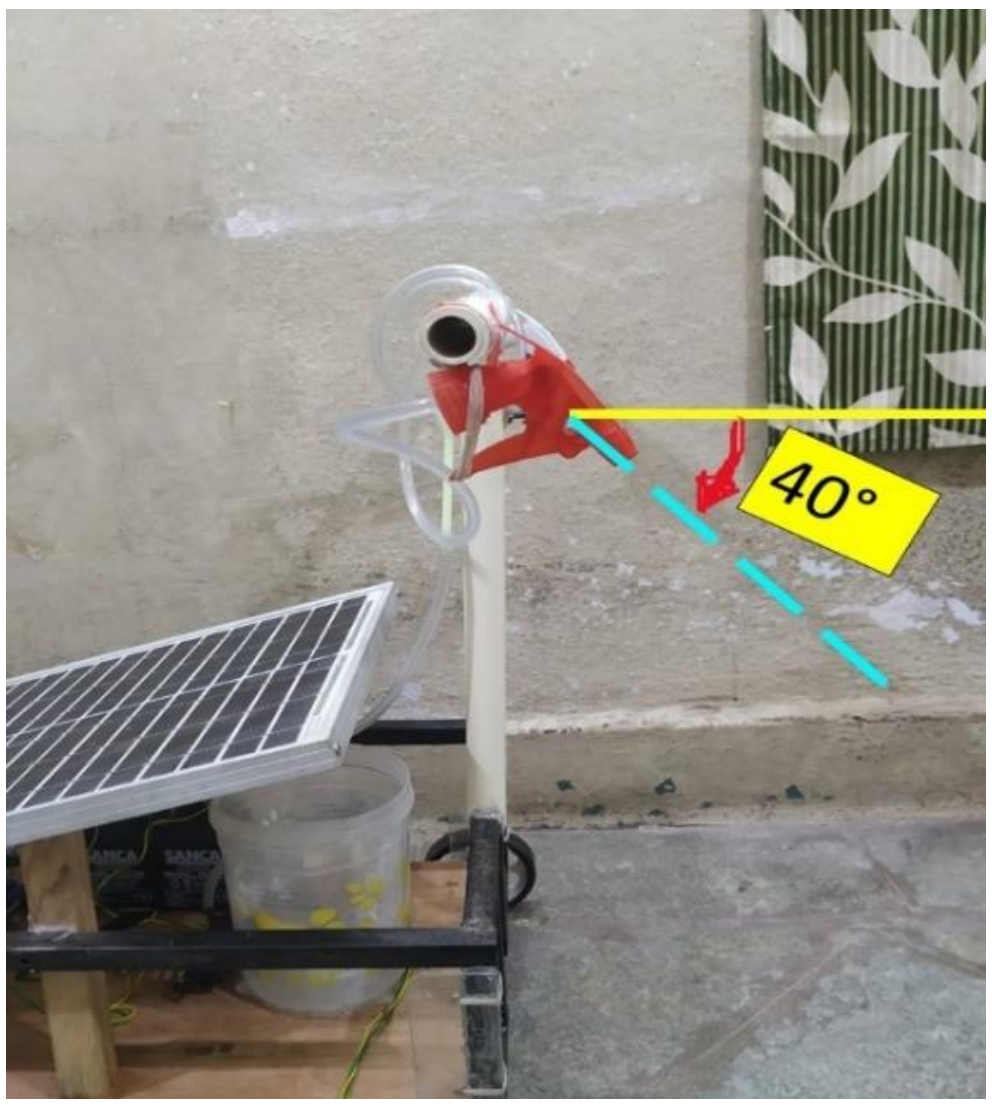

*Fig 8.13: 40° Downwards Rotation of Nozzle in the YZ Plane w.r.t X axis*

Additional Height gained Due to rotation in YZ plane.

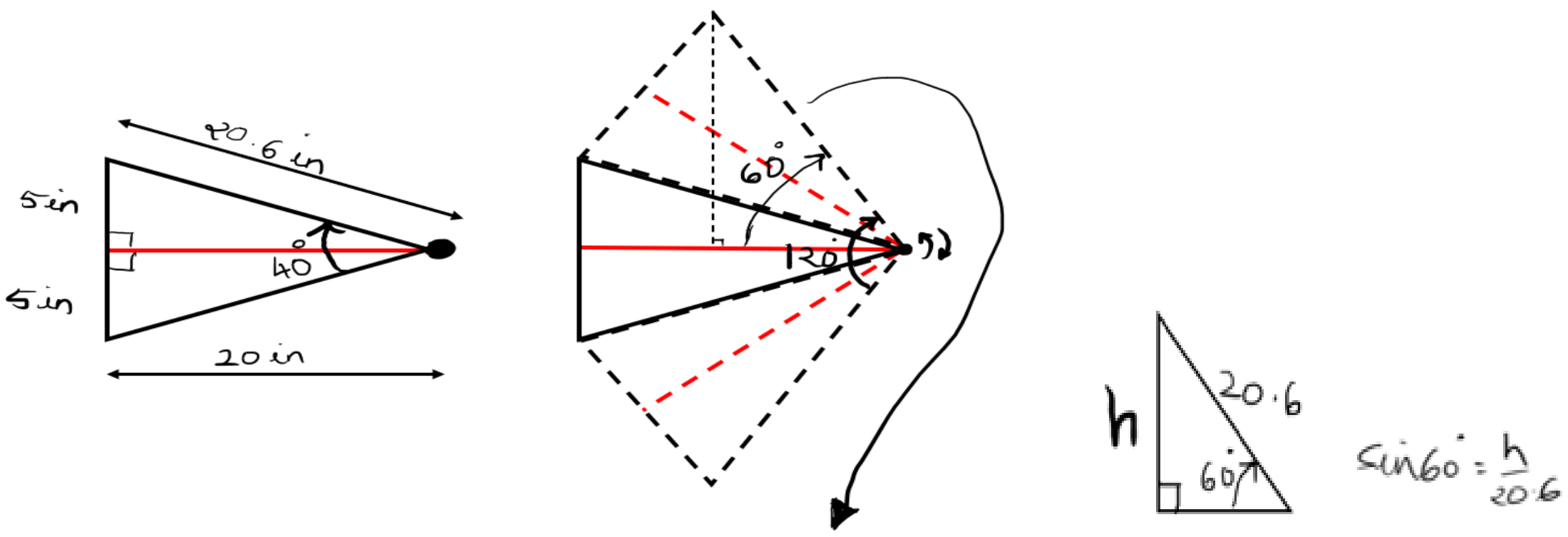

*Fig 8.14: Additional Height gained*

$X^2 = 5^2 + 20^2$

$= 25 + 400$

$X^2 = 425$

Therefore, X = 20.6 inches

Hence, h = 20.6* Sin(60)

h = 17.84 in

## *8.1.2. Distances and areas covered by the spray at different positions:*

Nozzle specs

| | |
|---|---|
| The cone angle: | 120 degrees |
| Outlet dia: | 0.65mm |
| Inlet dia: | 10.40mm |
| Nozzle type: | Full Cone Spray |

8.1.3 Literature survey on liquid drop specifications

| Droplet Diameter | Droplet type | Droplet coverage (in Drops Per Inch) | Nozzle cap Position |
|---|---|---|---|
| 100 (Very Fine hole) | Fine mist | 1,150 dpi | Loose |
| 1,000 (extra coarse) | Heavy rain | 1 dpi | Tight |
| 0 | None | 0 | Fully closed |

With the above table we can infer that the distance covered and the area of spray completely depends on the Nozzle cap position. The nozzle cap position can be tightened or loosened as it has internal threads. Therefore, by turning the nozzle cap we can control the type of spray we require for our application.

The distance covered by the spray varies with the no. of turns given to the nozzle cap. Hence the distances covered by the spray at different no. of turns given to the nozzle cap is tabulated along with the droplet diameter.

8.1.4 Number of turns vs Distance sprayed

| Droplet diameter | Distance sprayed in inches | Number of turns the nozzle cap is rotated |
|---|---|---|
| 100 (VF) | 9 | 1 turn |
| 150 (F) | 16 | 3 turns |
| 200 (F) | 26 | 5 turns |
| 1000 (XC) | 35 | 7 turns |

*Calculation of area covered by the Full cone spray:* From the side view a cone would look like a Triangle. A 2D representation of the Full cone spray is given below. The estimated dimensions of the cone from the tests at 4 turns are:
vertical cone dia - 10 inch = 2R (Therefore R = 5 in)

average spraying distance adjusted (ht. of vertical cone) - 20 inches

Flow rate through Pump of 1-2L/min

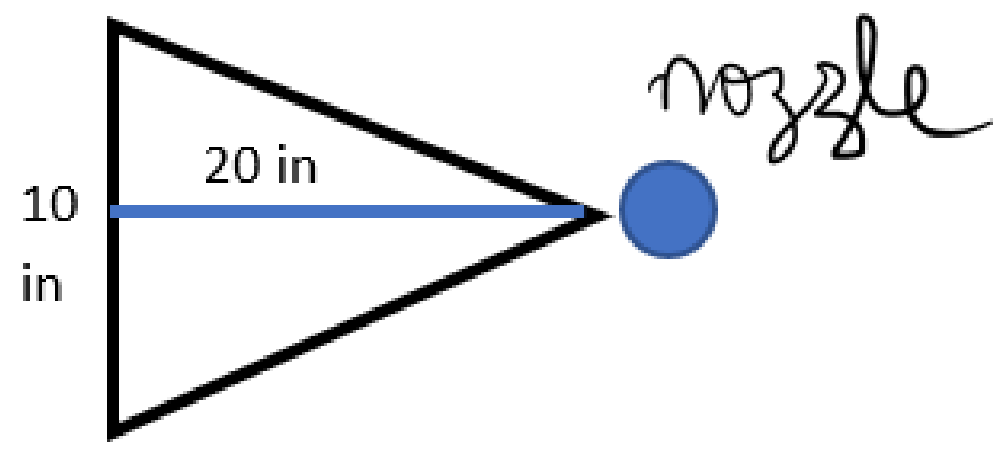

*Fig 8.15: Spray Area*

Let the slant height of the cone be L

The L is found out using Pythagoras theorem

$L^2 = 20^2 + 5^2$

Therefore L = 20.6 in

The area covered by the spray is equal to the Total Surface area of the cone.

The total surface area of the cone is given by

$TSA = \pi RL + \pi R^2$

Where R= Radius of the base = 5 in

L = Slant Height = 20.6 in

Hence TSA = 411.11 $in^2$

Therefore, the area covered by the spray is equal to 411.58 $in^2$

### *8.2. Mowing operation*

The Mower blade is set to a position with 3-inch Ground clearance. The Mower holder setup is joined to the chassis with the help of a screw and nut joint. So, the ground clearance of the mower is adjustable to meet different requirements. The mower blade length is 31cm. The Mower is connected to a high-speed DC Motor with speed rating 1000 RPM.

Effective cutting area when stationary = $\pi d^2/4$

Where d = Blade length,

Hence the effective cutting area comes out to be 468 sq. in. (i.e 0.3 sq. m.)

### *8.3. Robot's movement*

The robot is run by 4 DC Motors which are connected to Arduino which in turn is connected to a smartphone application via Bluetooth. The robot motion is controlled by the smartphone application.

The maximum speed of the Robot when fully loaded with all the components is 1.43 m/s. The High torque DC Motors are preferred which have a speed rating of 250 RPM.

### *8.4. Battery performance*

The difference between the conceptual design and the prototype is that the conceptual design has two Electrically operated Actuators. (One horizontal and One vertical) And so the power consumption varies between the two designs and hence the calculation of battery performance for the designs are done separately.

Battery backup is chosen as the parameter to evaluate the battery performance. Battery backup is the amount of time for which the battery can power the entire robot before getting fully discharged.

Battery backup is defined on the ratio of Battery Capacity to the Maximum current draw of the robot. Since all the electrical components are powered by the battery. The Maximum current draw will be the sum of the currents drawn by individual components. The capacity of the chosen battery is 4.5 Amps which is more than enough for powering all the components.

#### *8.4.1. Battery back-up of Prototype*

The individual current draw of components is given below:

DC Motor – 0.06A

No. of DC Motors = 5 (4 Drive motors for wheels + 1 Mower Motor)

Pump – 0.3A          Arduino board – 0.02 A

The total current draw is given as = (0.06*5) + 0.3 + 0.02 = 0.62 A

Therefore, Battery backup = Battery Capacity / Total current draw

= 4.5Ah/0.62 = 7.25hrs

Therefore, the battery can run this robot for 7.25 hours with a single charge

*8.4.2. Battery backup of Conceptual design*

Since the conceptual design has two components more (i.e One Horizontal Actuator and One Vertical Actuator) and rest all the components being same, the total current draw of the conceptual design can be found out by just adding the current draws of actuators to the total current draw of the Prototype. The current draw of the Actuators is given below.

Linear Actuator (Horizontal) – 0.3 A

No. of Horizontal actuators = 2

Linear Actuator (Vertical) – 0.5

Total current draw for CD = Total current draw for Prototype + (0.3*2) + 0.5

= 1.72 + 0.6 + 0.5

= 1.72 A

Battery backup = Battery Capacity / Total current draw

= 4.5Ah/1.72 = 2.61 hrs

Therefore, the battery can run the robot for 2.61 hours with a single charge

*8.5. Time taken to charge the battery with the Solar panel.*

The power rating of the chosen solar panel is 100 W. The voltage output given by the solar panel at 12:10 PM (IST) is found to be 21V. It is determined using the Multimeter. Hence, the amperage of the solar panel is found out using the equation

Power = Voltage * Current

Therefore, the current output of the solar panel is 4.5 Amps

The time taken by the solar panel to charge the battery would be the ratio of the battery capacity to the Current supplied by the panel.

Hence Time = Battery capacity / Current supplied

= 4.5 Ah / 4.5 A

= 1 Hour

Therefore, it would take 1 Hour for the solar panel to completely charge the Battery.

## 9. FUTURE SCOPE

Mechatronics is playing an enormous role in agricultural production and management. There is a desire for autonomous and timesaving technology in agriculture to possess efficient farm management. The researchers are now aiming towards different types of farming parameters to style autonomous multipurpose agricultural robots because of traditional farm machineries and topological dependent. Till date the multipurpose agricultural robots have been researched and developed mainly for harvesting, fertilizer spraying, picking fruits, sowing, solar energy and monitoring of crops. Robots like these are brilliant replacements for manpower to a better extent as they deploy unmanned sensors and machinery systems. The agricultural benefits of development of these autonomous and intelligent robots are to improve repetitive precision, efficiency, reliability and minimization of soil compaction and chemical utilization. The robots have the potential of multitasking, sensory measures, idle operation as well working in odd operating conditions. The study on multipurpose agricultural robot system had been done using model structure design along with various precision farming machineries. With fully automated farms in the future, robots can perform all the tasks like ploughing, seed sowing, pesticides spraying, monitoring of pests and diseases, harvesting, etc. This allows the farmers to just supervise the robots without the need of manual operation. In the future robots may also run on PLC and SCADA with automatic systems. In this paper, overview of mechatronics approach of our multipurpose agriculture robot for precision Agriculture in India and worldwide development is reviewed.

## 10. CONCLUSION

- The prototype gave a fairly good rate of area coverage with a reasonably low operating cost. The system addresses the issue of dearth of agricultural labour and ensures safe agricultural practices by completely eliminating, handling of harmful chemicals, cutting crops and extensive labour by the farmer as it can be operated remotely.

- The proposed spraying & mower robot is suitable for small and medium scale farmers. Large scale production of the spraying unit will reduce the cost significantly giving partial thrust to Indian agriculture practices.

- The unit can be scaled up based on the requirement. The developed system can not only be used for spraying fertilizer, pesticides, fungicides, lawn watering and crop cutting , weeding and lawn mowing but also for maintenance of sports fields like cricket ground.

- With the proposed design of the robot in this project, the above mentioned gaps can be eliminated completely. This project integrates two of the major activities in agriculture which are Pesticide spraying and Crop Cutting (or Weed Removal).

- Workload on the farmers is decreased and health problems also. Successful in constructing robot which can be travelled on rough, uneven surfaces also and weighing enough load of pump and other equipment. Successful in developing a robot whose construction is enough to withstand the challenges of the field.

## 11.REFERENCES

1. Adamides, G.; Katsanos, C.; Parmet, Y.; Christou, G.; Xenos, M.; Hadzilacos, T.; Edan, Y. HRI usability evaluation of interaction modes for a teleoperated agricultural robotic sprayer. Appl. Ergon. **2017**, 62, 237–246. [CrossRef] [PubMed]
2. Balloni, S.; Caruso, L.; Cerruto, E.; Emma, G.; Schillaci, G. A Prototype of Self-Propelled Sprayer to Reduce Operator Exposure in Greenhouse Treatment. In Proceedings of the Ragusa SHWA International Conference: Innovation Technology to Empower Safety, Health and Welfare in Agriculture and Agro-food Systems, Ragusa, Italy, 15–17 September 2008
3. Bonaccorso, F.; Muscato, G.; Baglio, S. Laser range data scan-matching algorithm for mobile robot indoor self-localization. In Proceedings of the World Automation Congress (WAC), Puerto Vallarta, Mexico, 24–28 June 2012; pp. 1–5.
4. Berenstein, R.; Shahar, O.B.; Shapiro, A.; Edan, Y. Grape clusters and foliage detection algorithms for autonomous selective vineyard sprayer. Intell. Serv. Robot. **2010**, 3, 233–243. [CrossRef]
5. Bergerman, M.; Singh, S.; Hamner, B. Results with autonomous vehicles operating in specialty crops. In Proceedings of the 2012 IEEE International Conference on Robotics and Automation (ICRA), St. Paul, MN, USA, 14–18 May 2012; pp. 1829–1835.
6. Bechar, A.; Vigneault, C. Agricultural robots for field operations. Part 2: Operations and systems. Biosyst. Eng. **2016**, 153, 110–128. [CrossRef]
7. Bechar, A.; Vigneault, C. Agricultural robots for field operations: Concepts and components. Biosyst. Eng. **2016**, 149, 94–111. [CrossRef]
8. Binod Poudel, Ritesh Sapkota, Ravi Bikram Shah, Navaraj Subedi, Anantha Krishna G.L, Design and fabrication of solar powered semi-automatic pesticide sprayer.
9. Cunha, M.; Carvalho, C.; Marcal, A.R.S. Assessing the ability of image processing software to analyse spray quality on water-sensitive papers used as artificial targets. Biosyst. Eng. **2012**, 111, 11–23. [CrossRef]
10. Damalas, C.A.; Koutroubas, S.D. Farmers' exposure to pesticides: Toxicity types and ways of prevention. Toxics **2016**, 4, 1. [CrossRef] [PubMed]
11. Flourish Project. Available online: (accessed on 21 June 2019).
12. González, R.; Rodríguez, F.; Sánchez-Hermosilla, J.; Donaire, J.G. Navigation techniques for mobile robots ingreenhouses. Appl. Eng. Agric. **2009**, 25, 153–165. [CrossRef]
13. Harshit Jain,Nikunj Gangrade, Sumit Paul, Harshal Gangrade, Jishnu Ghosh, Design and fabrication of Solar pesticide sprayer
14. Julian Senchez-Hermosilla, Francisco Rodriguez Ramon Gonzalez, Jose Luis Guzman2and Manuel Berenguel, A mechatronic description of an autonomous mobile robot for agricultural tasks in greenhouses
15. Kiran Kumar B M,M S Indira, S Nagaraja Rao Pranupa S, Design and development of Three DoF Solar powered smart spraying agricultural robot.

16. Navigation Inside a Greenhouse. Robotics **2018**, 7, 22. [CrossRef]

17. Philip J. Sammons, Tomonari Furukawa and Andrew Bulgin ,Autonomous pesticide spraying robot for use in a greenhouse.

18. Reis, R.; Mendes, J.; do Santos, F.N.; Morais, R.; Ferraz, N.; Santos, L.; Sousa, A. Redundant robot localization system based in wireless sensor network. In Proceedings of the IEEE International Conference on Autonomous Robot Systems and Competitions (ICARSC), Torres Vedras, Portugal, 25–27 April 2018; pp. 154–159.

19. Robots in Agriculture. Available online: (accessed on 21 June 2019).

20. Song, Y.; Sun, H.; Li, M.; Zhang, Q. Technology Application of Smart Spray in Agriculture: A Review. Intell. Autom. Soft Comput. **2015**, 21, 319–333. [CrossRef]

**21.** Salyani, M.; Zhu, H.; Sweeb, R.D.; Pai, N. Assessment of spray distribution with water-sensitive paper. Agric. Eng. Int. CIGR J. **2013**, 15, 101–111.

**22.** Siciliano, B.; Khatib, O. Springer Handbook of Robotics; Force Control; Springer: New York, NY, USA, 2008;

pp. 161–185.

23. Sánchez-Hermosilla, J.; González, R.; Rodríguez, F.; Donaire, J.G. Mechatronic description of a laser autoguided vehicle for greenhouse operations. Sensors **2013**, 13, 769–784. [CrossRef] [PubMed]

24. Tony E. Grift, Design and development of Autonomous robots for agricultural applications.

25. Vijaykumar N Chalwa, Shilpa S Gundagi, Mechatronics based remote controlled agricultural robot.

26. Vinerobot Project. Available online: (accessed on 21 June 2019).

27. Yan Li, Chunlei Xia, Jangmyung Lee, Vision based pest detection and automatic spray of greenhouse plant. **2018**, 152, 363–374. [CrossRef]